\documentclass{article}

\usepackage[final]{neurips_2022}

\usepackage[utf8]{inputenc} % allow utf-8 input
\usepackage[T1]{fontenc}    % use 8-bit T1 fonts
\usepackage{hyperref}       % hyperlinks
\usepackage{url}            % simple URL typesetting
\usepackage{booktabs}       % professional-quality tables
\usepackage{amsfonts}       % blackboard math symbols
\usepackage{nicefrac}       % compact symbols for 1/2, etc.
\usepackage{microtype}      % microtypography
\usepackage{xcolor}         % colors

\usepackage{tikz}
\usetikzlibrary{positioning,calc,bayesnet,external}
\RequirePackage[noend]{algorithmic}
\usepackage[ruled]{algorithm}

\usepackage{graphicx}
\usepackage{subcaption}
\usepackage{wrapfig}

\newcommand{\conf}{\mathbf{x}}
\newcommand{\Conf}{\mathbf{X}}
\newcommand{\confspace}{\mathcal{X}}
\newcommand{\kernel}{k}
\renewcommand{\algorithmicrequire}{\textbf{Input:}}
\renewcommand{\algorithmicensure}{\textbf{Output:}}
\newcommand{\algabbr}{\textsc{DyHPO}}

\newcommand{\comment}[1]{}

\usepackage{mathtools}
\usepackage{amsthm}
\usepackage{amsmath}
\usepackage{amssymb}
\DeclareMathOperator*{\argmax}{arg\,max}
\DeclareMathOperator*{\argmin}{arg\,min}

\title{Supervising the Multi-Fidelity Race \\ of Hyperparameter Configurations}

\author{%
  Martin Wistuba$\thanks{equal contribution}~~\thanks{work does not relate to position at Amazon}$  \\
  Amazon Web Services, Berlin, Germany \\
  \texttt{marwistu@amazon.com} \\
  \And
  Arlind Kadra$^*$ \\
  University of Freiburg, Freiburg, Germany \\
  \texttt{kadraa@cs.uni-freiburg.de} \\
  \And
  Josif Grabocka \\
  University of Freiburg, Freiburg, Germany \\
  \texttt{grabocka@cs.uni-freiburg.de} \\
}

\begin{document}

\maketitle

\begin{abstract}
% ---- Previous Abstract commented
%The automation of machine learning is an important step towards democratization of machine learning and its more efficient use.
%One important aspect is the automation of hyperparameter or neural architecture optimization.
%This work is devoted to hyperparameter optimization in the gray-box setting, in which the performance of a hyperparameter configuration  is measured with different fidelities while training a model.
%This allows for terminating machine learning jobs early and using the saved computational budget to explore other configurations.
%We propose a novel surrogate model which combines a Gaussian Process with a deep kernel which explicitly consumes the multi-fidelity data as an input.
%Furthermore, we propose a variation of the popular expected improvement acquisition function for the multi-fidelity setting.
%This allows us to derive \algabbr{} as a variation of Bayesian optimization.
%On different deep learning tasks, we demonstrate that its ability to dynamically choose from all configurations at any time grants more flexibility, which leads to better performance.

Multi-fidelity (gray-box) hyperparameter optimization techniques (HPO) have recently emerged as a promising direction for tuning Deep Learning methods. However, existing methods suffer from a sub-optimal allocation of the HPO budget to the hyperparameter configurations. In this work, we introduce DyHPO, a Bayesian Optimization method that learns to decide which hyperparameter configuration %'s neural network 
to train further in a dynamic race among all feasible configurations. %We propose both a new deep kernel for Gaussian Processes that embeds the learning curve dynamics, as well as a new acquisition function that incorporates multi-budget information.
We propose a new deep kernel for Gaussian Processes that embeds the learning curve dynamics, and an acquisition function that incorporates multi-budget information. We demonstrate the significant superiority of DyHPO against state-of-the-art hyperparameter optimization methods through large-scale experiments comprising 50 datasets (Tabular, Image, NLP) and diverse architectures (MLP, CNN/NAS, RNN).

\end{abstract}
\section{Introduction}

%% ---------- HERE is the previous introduction
%The right selection of machine learning algorithms, neural network architectures or hyperparameter settings are vital for every deployed machine learning system.
%Therefore, it is of no surprise that the research community has dedicated a lot of effort on mechanisms for finding good settings manually or automatically.
%Within the last decade, more and more algorithms have been introduced for automatically tackling this problem and therefore making machine learning more accessible and scalable.
%Often, these algorithms address the problem of hyperparameter optimization or algorithm selection as a \emph{black-box optimization problem}.
%In this case, the black-box function is a function which takes as an input the hyperparameter configuration and returns the validation score you would obtain by training the algorithm with this configuration.
%However, in practice this assumption is too strong and it can be relaxed.
%Many algorithms can provide intermediate feedback during training which can be considered to predict the final validation score.
%Ensembling techniques such as random forest or gradient boosting allow to measure the validation score when adding weak learners iteratively.
%Machine learning methods that learn incrementally from data, such as neural networks, allow to measure the validation score after each update step.
%In these cases, hyperparameter optimization can be considered a \emph{gray-box optimization problem}.

Hyperparameter Optimization (HPO) is arguably an acute open challenge for Deep Learning (DL), especially considering the crucial impact HPO has on achieving state-of-the-art empirical results. Unfortunately, HPO for DL is a relatively under-explored field and most DL researchers still optimize their hyperparameters via obscure trial-and-error practices. On the other hand, traditional Bayesian Optimization HPO methods~\citep{Snoek2012,Bergstra2011} are not directly applicable to deep networks, due to the infeasibility of evaluating a large number of hyperparameter configurations. In order to scale HPO for DL, three main directions of research have been recently explored. \textit{(i) Online HPO} methods search for hyperparameters during the optimization process via meta-level controllers~\citep{Chen2017,Parker-Holder2020}, however, this online adaptation can not accommodate all hyperparameters (e.g. related to architectural changes).  \textit{(ii) Gradient-based HPO} techniques, on the other hand, compute the derivative of the validation loss w.r.t. hyperparameters by reversing the training update steps~\citep{Maclaurin2015,Franceschi2017,Lorraine2020}, however, the reversion is not directly applicable to all cases (e.g. dropout rate). The last direction, \textit{(iii) Gray-box HPO} techniques discard sub-optimal configurations after evaluating them on lower budgets~\citep{Li2017,Falkner2018}.

In contrast to the online and gradient-based alternatives, gray-box approaches can be deployed in an off-the-shelf manner to all types of hyperparameters and architectures. The gray-box concept is based on the intuition that a poorly-performing hyperparameter configuration can be identified and terminated by inspecting the validation loss of the first few epochs, instead of waiting for the full convergence. The most prominent gray-box algorithm is Hyperband~\citep{Li2017}, which is based on successive halving. It runs random configurations at different budgets (e.g. number of epochs) and successively halves these configurations by keeping only the top performers. Follow-up works, such as BOHB~\citep{Falkner2018} or DEHB~\citep{Awad2021}, replace the random sampling of Hyperband with a sampling based on Bayesian optimization or differentiable evolution. 

Despite their great practical potential, gray-box methods suffer from a major issue. The low-budget (few epochs) performances are not always a good indicator for the full-budget (full convergence) performances. For example, a properly regularized network converges slower in the first few epochs, however, typically performs better than a non-regularized variant after the full convergence. In other words, there can be a poor rank correlation of the configurations' performances at different budgets.

\begin{wrapfigure}{r}{0.5\textwidth}
\begin{minipage}{0.5\textwidth}
  \centering
  \includegraphics[width=1.0\columnwidth]{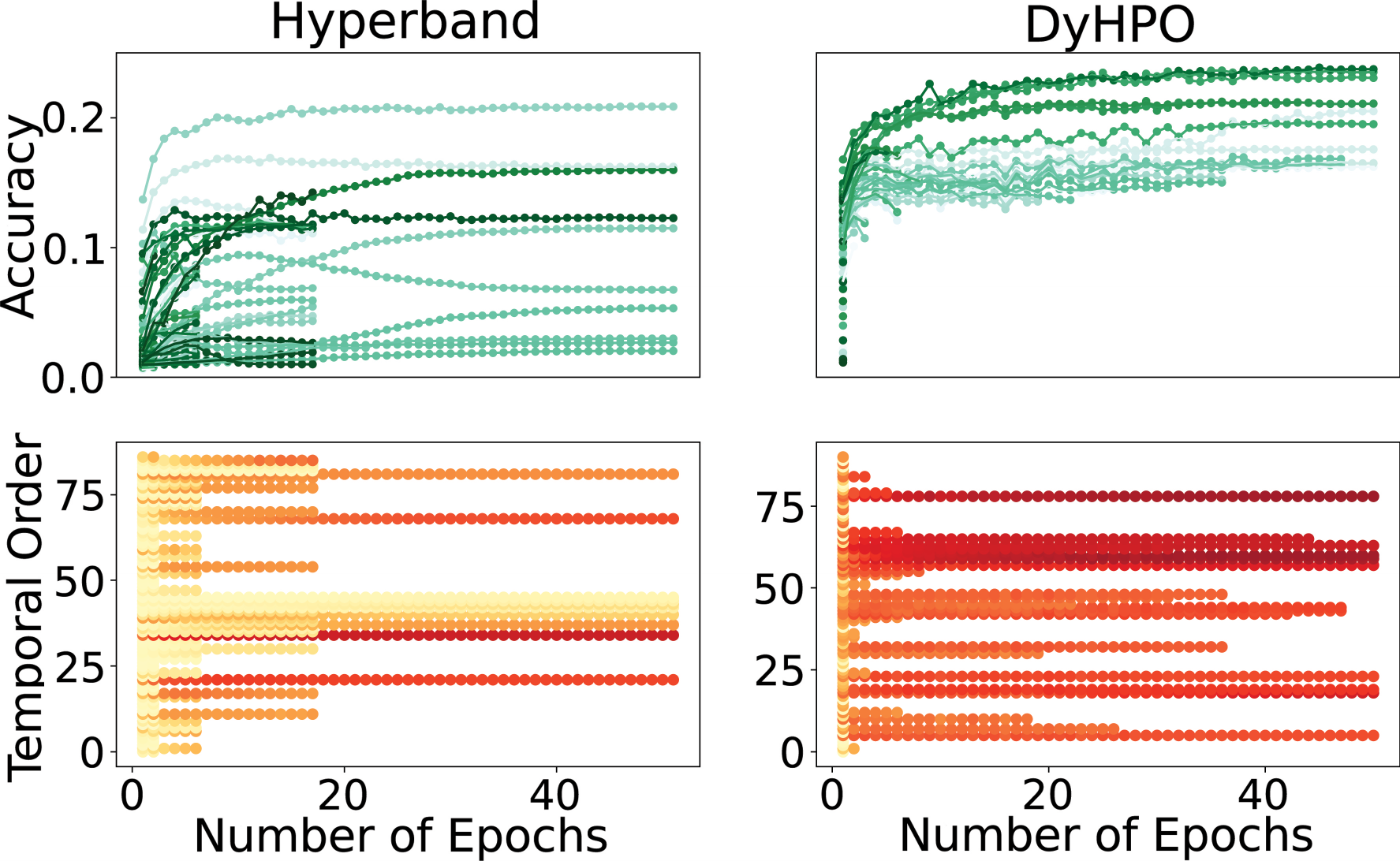}
  \caption{%Learning curves observed during the search. The darker the learning curve, the later it was evaluated during the search. 
  \textbf{Top:} The learning curve for different hyperparameter configurations. The darker the learning curve, the later it was evaluated during the search. \textbf{Bottom:} The hyperparameter indices in a temporal order as evaluated during the optimization and their corresponding curves.}  
%Bottom row: y-axis shows a sequence of learning curve evaluations (bottom to top). The color indicates accuracy. The darker red the higher the accuracy.}
  \label{fig:motivation}
\end{minipage}
\vspace{-0.4cm}
\end{wrapfigure}

%We introduce \algabbr{}, a gray-box method that \emph{dynamically} decides how many configurations to try and how much budget to spend on each configuration. \algabbr{} is a Bayesian Optimization (BO) approach based on Gaussian Processes (GP), 

We introduce \algabbr{}, a Bayesian Optimization (BO) approach based on Gaussian Processes (GP), that proposes a novel treatment to the multi-budget (a.k.a. multi-fidelity) setup. In this perspective, we propose a deep kernel GP that captures the learning dynamics. As a result, we train a kernel capable of capturing the similarity of a pair of hyperparameter configurations, even if the pair's configurations are evaluated at different budgets. Furthermore, we extend Expected Improvement~\citep{Jones1998_Efficient} to the multi-budget case, by introducing a new mechanism for the incumbent configuration of a budget. 

We illustrate the differences between our racing strategy and successive halving with the experiment of Figure~\ref{fig:motivation}, where, we showcase the HPO progress of two different methods on the "Helena" dataset from the LCBench benchmark~\citep{Zimmer2021}.
%Random search is an example of a black-box approach that trains each candidate until completion without considering the intermediate scores.
Hyperband~\citep{Li2017} is a gray-box approach that \textit{statically} pre-allocates the budget for a set of candidates (Hyperband bracket) according to a predefined policy.
However, \algabbr{} \textit{dynamically} adapts the allocation of budgets for configurations after every HPO step (a.k.a. a dynamic race). As a result, \algabbr{} invests only a small budget on configurations that show little promise as indicated by the intermediate scores.

The joint effect of modeling a GP kernel across budgets together with a dedicated acquisition function leads to \algabbr{} achieving a statistically significant empirical gain against state-of-the-art gray-box baselines~\citep{Falkner2018,Awad2021}, including prior work on multi-budget GPs~\citep{Kandasamy2017,Kandasamy2020} or neural networks~\citep{Li2020}. We demonstrate the performance of \algabbr{} in three diverse %popular types of 
deep learning architectures (MLP, CNN/NAS, RNN) and 50 datasets of three diverse modalities (tabular, image, natural language processing). We believe our method is a step forward toward making HPO for DL practical and feasible. Overall, our contributions can be summarized as follows:

\begin{itemize}
    \item We introduce a novel Bayesian surrogate for gray-box HPO optimization. Our novel surrogate model predicts the validation score of a machine learning model based on both the hyperparameter configuration, the budget information, and the learning curve.
    \item We derive a simple yet robust way to combine this surrogate model with Bayesian optimization, reusing most of the existing components currently used in traditional Bayesian optimization methods.
    \item Finally, we demonstrate the efficiency of our method for HPO and neural architecture search tasks compared to the current state-of-the-art methods in HPO, by outperforming seven strong HPO baselines with a statistically significant margin. As an overarching goal, we believe our method is an important step toward scaling HPO for DL. 
\end{itemize}
\section{Related Work on Gray-box HPO}
\paragraph*{Multi-Fidelity Bayesian Optimization and Bandits.}
Bayesian optimization is a black-box function optimization framework that has been successfully applied in optimizing hyperparameter and neural architectures alike~\citep{Snoek2012,Kandasamy2018, Bergstra2011}.
To further improve Bayesian optimization, several works propose low-fidelity data approximations of hyperparameter configurations by training on a subset of the data~\citep{Swersky2013, Klein2017}, or by terminating training early~\citep{Swersky2014}.
Additionally, several methods extend Bayesian optimization to multi-fidelity data by engineering new kernels suited for this problem~\citep{Swersky2013,Swersky2014,Poloczek2017}.
\cite{Kandasamy2016} extends GP-UCB~\citep{Srinivas2010} to the multi-fidelity setting by learning one Gaussian Process (GP) with a standard kernel for each fidelity.
Their later work improves upon this method by learning one GP for all fidelities that enables the use of continuous fidelities~\citep{Kandasamy2017}.
The work by \cite{Takeno2020} follows a similar idea but proposes to use an acquisition function based on information gain instead of UCB.
While most of the works rely on GPs to model the surrogate function, \cite{Li2020} use a Bayesian neural network that models the complex relationship between fidelities with stacked neural networks, one for each fidelity.

% Previous implementation
% \paragraph*{Multi-Fidelity Bandits and Bayesian Optimization}
%Hyperband~\citep{Li2017} is a multi-fidelity method for hyperparameter optimization which, due to its simplicity and strong performance, enjoys great popularity.
%The algorithm selects hyperparameter configurations at random and uses successive halving~\citep{Jamieson2016} with different settings to early-stop less promising training runs.
%Several improvements have been proposed with the aim to replace the random sampling of hyperparameter configurations with a more guided approach.
%Among those, the most notable ones are Bayesian optimization and differential evolution~\citep{Awad2021}.
Hyperband~\citep{Li2017} is a bandits-based multi-fidelity method for hyperparameter optimization that selects hyperparameter configurations at random and uses successive halving~\citep{Jamieson2016} with different settings to early-stop less promising training runs. 
Several improvements have been proposed to Hyperband with the aim to replace the random sampling of hyperparameter configurations with a more guided approach~\citep{Bertrand2017,Wang2018,Wistuba2017}.
BOHB~\citep{Falkner2018} uses TPE~\citep{Bergstra2011} and builds a surrogate model for every fidelity adhering to a fixed-fidelity selection scheme.
DEHB~\citep{Awad2021} samples candidates using differential evolution which handles large and discrete search spaces better than BOHB.
\cite{Mendes2021} propose a variant of Hyperband which allows to skip stages.

%Among those, the most notable ones are Bayesian optimization and differential evolution.
%From the domain of Bayesian optimization, BOHB uses TPE~\citep{Bergstra2011} and builds a surrogate model for every fidelity adhering to a fixed-fidelity selection scheme,  Lastly, from the domain of differential evolution, DEHB builds upon BOHB by incorporating differential evolution, thus, handling discrete and large hyperparameter search spaces better compared to BOHB.

\paragraph{Learning Curve Prediction}
A variety of methods attempt to extrapolate a partially observed learning curve in order to estimate the probability that a configuration will improve over the current best solution.
\cite{Domhan2015} propose to ensemble a set of parametric functions to extrapolate a partial learning curve. While this method is able to extrapolate with a single example, it requires a relatively long learning curve to do so.
The work by \cite{Klein2017a} build upon the idea of using a set of parametric functions.
The main difference is that they use a heteroscedastic Bayesian model to learn the ensemble weights.
\cite{Baker2018} propose to use support vector machines (SVM) as an auto-regressive model.
The SVM predicts the next value of a learning curve, the original learning curve is augmented by this value and we keep predicting further values.
The work by \cite{Gargiani2019} use a similar idea but makes prediction based on the last $K$ observations only and uses probabilistic models.
\cite{Wistuba2020} propose to learn a prediction model across learning curves from different tasks to avoid the costly learning curve collection.
In contrast to \algabbr{}, none of these methods selects configuration but is limited to deciding when to stop a running configuration.

\paragraph{Multi-Fidelity Acquisition Functions}
\cite{Klein2017} propose an acquisition function which allows for selecting hyperparameter configurations and the dataset subset size.
The idea is to reduce training time by considering only a smaller part of the training data.
In contrast to $\text{EI}_\text{MF}$, this acquisition function is designed to select arbitrary subset sizes whereas $\text{EI}_\text{MF}$ is intended to slowly increase the invested budget over time.
\cite{Mendes2020} extend the work of \cite{Klein2017} to take business constraints into account.

\paragraph*{Deep Kernel Learning with Bayesian Optimization.}
We are among the first to use deep kernel learning with Bayesian optimization and to the best of our knowledge the first to use it for multi-fidelity Bayesian optimization.
\cite{Rai2016} consider the use of a deep kernel instead of a manually designed kernel in the context of standard Bayesian optimization, but, limit their experimentation to synthetic data and do not consider its use for hyperparameter optimization.
\cite{Perrone2018,Wistuba2021} use a pre-trained deep kernel to warm start Bayesian optimization with meta-data from previous optimizations.
The aforementioned approaches are multi-task or transfer learning methods that require the availability of meta-data from related tasks.

%Our aim is to improve upon BOHB by developing a multi-fidelity Bayesian optimization approach that is not limited to the Hyperband bracket policy of selecting and discarding hyperparameter configurations.
%Furthermore, we use a deep kernel that allows for flexible kernel learning and overcomes the problem of manually engineering kernels.
%Finally, we propose the first surrogate model for Bayesian optimization method that explicitly takes the learning curve as an input.

In contrast to prior work, we propose a method that introduces deep learning to multi-fidelity HPO with Bayesian Optimization, and captures the learning dynamics across fidelities/budgets, combined with an acquisition function that is tailored for the gray-box setup. %Furthermore, our work represents an important step towards scaling HPO for Deep Learning (DL), by demonstrating a statistically significant reduction in terms of HPO time on a series of DL network architectures and a large set of diverse datasets.

\section{Dynamic Multi-Fidelity HPO}\label{sec:dyhpo}

%In this section, we will describe \algabbr{}, our proposed method for hyperparameter optimization in the gray-box setting.
%At first, we will describe the surrogate model which is a Gaussian Process with a deep convolutional kernel.
%Then, we describe a variation of the popular expected improvement acquisition function~\citep{Jones1998_Efficient}, modified to consider multiple fidelities, and conclude with the final algorithm.

\subsection{Preliminaries}

\paragraph{Gray-Box Optimization.}
%Since many machine learning algorithms allow to measure at various fidelities, a relaxation of the black-box to the gray-box optimization problem is in many cases logical and allows for significantly faster optimization.
The gray-box HPO setting allows querying configurations with a smaller budget compared to the total maximal budget $B$.
Thus, we can query from the response function $f:\confspace\times\mathbb{N}\rightarrow\mathbb{R}$ where $f_{i,j}=f(\conf_i, j)$ is the response after spending a budget of $j$ on configuration $\conf_i$.
As before, these observations are noisy and we observe $y_{i,j}=f(\conf_i,j)+\varepsilon_j$ where $\varepsilon_j\sim\mathcal{N}(0,\sigma_{j,n}^2)$.
Please note, we assume that the budget required to query $f_{i,j+b}$ after having queried $f_{i,j}$ is only $b$. Furthermore, we use the learning curve $\mathbf{Y}_{i,j-1}=(y_{i,1},\ldots,y_{i,j-1})$ when predicting $f_{i,j}$.

\paragraph{Gaussian Processes (GP).} Given a training data set $\mathcal{D}=\{(\conf_i,y_i)\}_{i=1}^n$, the Gaussian Process assumption is that $y_{i}$ is a random variable and the joint distribution of all $y_{i}$ is assumed to be multivariate Gaussian distributed as $\mathbf{y}\sim\mathcal{N}\left(m\left(\Conf\right),\kernel\left(\Conf,\Conf\right)\right)\enspace$.
Furthermore, $\mathbf{f}_{*}$ for test instances $\conf_{*}$ are jointly Gaussian with $\mathbf{y}$ as:
\begin{equation}
\left[\begin{array}{c}
\mathbf{y}\\
\mathbf{f}_{*}
\end{array}\right]\sim\mathcal{N}\left(m\left(\Conf,\conf_{*}\right),\left(\begin{array}{cc}
\mathbf{K}_{n} & \mathbf{K}_{*}\\
\mathbf{K}_{*}^{T} & \mathbf{K}_{**}
\end{array}\right)\right)\enspace.
\end{equation}
The mean function $m$ is often set to $\mathbf{0}$ and its covariance function $\kernel$ depends on parameters $\boldsymbol{\theta}$.
For notational convenience, we use $\mathbf{K}_{n} = \kernel\left(\Conf,\Conf|\boldsymbol{\theta}\right)+\sigma_{n}^{2}\mathbf{I}$, $\mathbf{K}_{*} =  \kernel\left(\Conf,\Conf_{*}|\boldsymbol{\theta}\right)$ and $\mathbf{K}_{**} =  \kernel\left(\Conf_{*},\Conf_{*}|\boldsymbol{\theta}\right)$ to define the kernel matrices. We can derive the posterior predictive distribution with mean and covariance as follows:
\begin{align}
\label{eq:gp-mean-cov}
\mathbb{E}\left[\mathbf{f}_{*}|\Conf,\mathbf{y},\Conf_*\right] = \mathbf{K}_{*}^{T}\mathbf{K}_{n}^{-1}\mathbf{y}, \; \text{cov}\left[\mathbf{f}_{*}|\Conf,\Conf_*\right] = \mathbf{K}_{**}-\mathbf{K}_{*}^{T}\mathbf{K}_{n}^{-1}\mathbf{K}_{*}
\end{align}

Often, the kernel function is manually engineered, one popular example is the squared exponential kernel.
However, in this work, we make use of the idea of deep kernel learning~\citep{Wilson2016}.
The idea is to model the kernel as a neural network $\varphi$ and learn the best kernel transformation $\mathbf{K}\left(\theta, w\right) :=\kernel(\varphi(\conf,w),\varphi(\conf';w)|\boldsymbol{\theta})$, which allows us to use convolutional operations in our kernel.

\subsection{Deep Multi-Fidelity Surrogate}\label{sub:surrogate}

We propose to use a Gaussian Process surrogate model that infers the value of $f_{i,j}$ based on the hyperparameter configuration $\conf_i$, the budget $j$ as well as the past learning curve $\mathbf{Y}_{i,j-1}$.
For this purpose, we use a deep kernel as:
\begin{align}
\label{eq:oursurrogate}
    \mathbf{K}\left(\theta, w\right) := k(&\varphi(\conf_i, \mathbf{Y}_{i,j-1}, j; w), \varphi(\conf_{i'}, \mathbf{Y}_{i',j'-1}, j'; w); \theta)
\end{align}

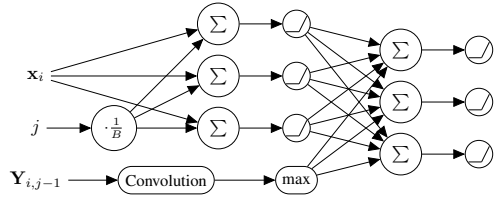
\begin{wrapfigure}{r}{0.5\textwidth}
  \centering
  \resizebox{0.47\textwidth}{!}{\begin{tikzpicture}
  % Styles
  \tikzstyle{boxstyle}=[rectangle,draw=black,minimum size=15,align=center, font=\fontsize{9}{0}\selectfont]
  \tikzstyle{circlestyle}=[circle,draw=black,minimum size=15,align=center, font=\fontsize{9}{0}\selectfont]
  \tikzstyle{roundedrect}=[rounded rectangle,draw=black,minimum size=15,align=center, font=\fontsize{9}{0}\selectfont]
  
  % Input
  %\node[rectangle,align=center](mutation_title) at (0.7, 1) {\bf Mutation};
  \node[rectangle](x_input) at (0, 2) {$\conf_i$};
  \node[rectangle](b_input) at (0, 1) {$j$};
  \node[rectangle](lc_input) at (0, 0) {$\mathbf{Y}_{i,j-1}$};
  
  \node[circlestyle](b_norm) at (1.5, 1) {$\cdot\frac{1}{B}$};
  \draw[->]   (b_input) -- (b_norm);
  
  % Layer 1
  \node[circlestyle](layer11) at (3.5, 3) {$\sum$};
  \node[circlestyle](layer12) at (3.5, 2) {$\sum$};
  \node[circlestyle](layer13) at (3.5, 1) {$\sum$};
  \draw[->]   (x_input) -- (layer11);
  \draw[->]   (x_input) -- (layer12);
  \draw[->]   (x_input) -- (layer13);
  \draw[->]   (b_norm) -- (layer11);
  \draw[->]   (b_norm) -- (layer12);
  \draw[->]   (b_norm) -- (layer13);
  \node[circlestyle](relu11) at (5, 3) {};
  \draw   (4.8, 2.85) -- (5.05,2.85);
  \draw   (5.05,2.85) -- (5.2,3.15);
  \node[circlestyle](relu12) at (5, 2) {};
  \draw   (4.8, 1.85) -- (5.05,1.85);
  \draw   (5.05,1.85) -- (5.2,2.15);
  \node[circlestyle](relu13) at (5, 1) {};
  \draw   (4.8, 0.85) -- (5.05,0.85);
  \draw   (5.05,0.85) -- (5.2,1.15);
  \draw[->]   (layer11) -- (relu11);
  \draw[->]   (layer12) -- (relu12);
  \draw[->]   (layer13) -- (relu13);
  
  % Conv
  \node[roundedrect](conv) at (2.5, 0) {Convolution};
  \node[roundedrect](max) at (5, 0) {max};
  \draw[->]   (lc_input) -- (conv);
  \draw[->]   (conv) -- (max);
  
  % Layer 2
  \node[circlestyle](layer21) at (7, 2.5) {$\sum$};
  \node[circlestyle](layer22) at (7, 1.5) {$\sum$};
  \node[circlestyle](layer23) at (7, 0.5) {$\sum$};
  \draw[->]   (relu11) -- (layer21);
  \draw[->]   (relu11) -- (layer22);
  \draw[->]   (relu11) -- (layer23);
  \draw[->]   (relu12) -- (layer21);
  \draw[->]   (relu12) -- (layer22);
  \draw[->]   (relu12) -- (layer23);
  \draw[->]   (relu13) -- (layer21);
  \draw[->]   (relu13) -- (layer22);
  \draw[->]   (relu13) -- (layer23);
  \draw[->]   (max) -- (layer21);
  \draw[->]   (max) -- (layer22);
  \draw[->]   (max) -- (layer23);
  \node[circlestyle](relu21) at (8.5, 2.5) {};
  \draw   (8.3, 2.35) -- (8.55,2.35);
  \draw   (8.55,2.35) -- (8.7,2.65);
  \node[circlestyle](relu22) at (8.5, 1.5) {};
  \draw   (8.3, 1.35) -- (8.55,1.35);
  \draw   (8.55,1.35) -- (8.7,1.65);
  \node[circlestyle](relu23) at (8.5, 0.5) {};
  \draw   (8.3, 0.35) -- (8.55,0.35);
  \draw   (8.55,0.35) -- (8.7,0.65);
  \draw[->]   (layer21) -- (relu21);
  \draw[->]   (layer22) -- (relu22);
  \draw[->]   (layer23) -- (relu23);
\end{tikzpicture}}
  \caption{The feature extractor $\varphi$ of our kernel.}
  \label{fig:deep-kernel}
\end{wrapfigure}

We use a squared exponential kernel for $k$ and the neural network $\varphi$ is composed of linear and convolutional layers as shown in Figure~\ref{fig:deep-kernel}. 
We normalize the budget $j$ to a range between $0$ and $1$ by dividing it by the maximum budget $B$.
Afterward, it is concatenated with the hyperparameter configuration $\conf_i$ and fed to a linear layer. The learning curve $\mathbf{Y}_{i,j-1}$ is transformed by a one-dimensional convolution followed by a global max pooling layer. Finally, both representations are fed to another linear layer.

Its output will be the input to the kernel function $k$. Both, the kernel $k$ and the neural network $\varphi$ consist of trainable parameters $\boldsymbol{\theta}$ and $\mathbf{w}$, respectively. We find their optimal values by computing the maximum likelihood estimates as:

%\begin{align}
%    \hat{\theta},\hat{w}&=\argmax_{\theta,w}p(\mathbf{y}|\Conf,\mathbf{Y},\theta,w) \\ \nonumber
%    &=\argmax_{\theta,w}\int p(\mathbf{y}|\Conf,\mathbf{Y},\theta,w)p(\mathbf{f}|\Conf,\mathbf{Y},\theta,w) d\mathbf{f}  \\ \nonumber
%    & \propto \argmin_{\theta,w}\mathbf{y}^{\mathrm{T}}\mathbf{K}\left(\theta, w\right)^{-1}\mathbf{y}+\log\left|\mathbf{K}\left(\theta, w\right)\right|
%\end{align}

\begin{align}
    \hat{\theta},\hat{w} \;=\;\argmax_{\theta,w}p(\mathbf{y}|\Conf,\mathbf{Y},\theta,w)  \propto \argmin_{\theta,w}\mathbf{y}^{\mathrm{T}}\mathbf{K}\left(\theta, w\right)^{-1}\mathbf{y}+\log\left|\mathbf{K}\left(\theta, w\right)\right|
\end{align}

In order to solve this optimization problem, we use gradient descent and Adam~\citep{Kingma2015} with a learning rate of $0.1$. Given the maximum likelihood estimates, we can approximate the predictive posterior through $p\left(f_{i,j}|\conf_{i},\mathbf{Y}_{i,j-1},j,\mathcal{D},\hat{\boldsymbol{\theta}},\hat{\mathbf{w}}\right)$, 
and ultimately compute the mean and covariance of this Gaussian using Equation~\ref{eq:gp-mean-cov}.

%\begin{equation}
%\label{eq:posteriorapprox}
%    p\left(f_{i,j}|\conf_{i},\mathbf{Y}_{i,j-1},j,\mathcal{D}\right)\approx
%    p\left(f_{i,j}|\conf_{i},\mathbf{Y}_{i,j-1},j,\mathcal{D},\hat{\boldsymbol{\theta}},\hat{\mathbf{w}}\right)
%\end{equation}

\subsection{Multi-Fidelity Expected Improvement}\label{sub:acq-fct}
Expected improvement~\citep{Jones1998_Efficient} is a commonly used acquisition function %for the black-box setting 
and is defined as:
\begin{equation}
\operatorname{EI}(\conf|\mathcal{D})=\mathbb{E}\left[\max\left\{f(\conf) - y^{\text{max}}, 0\right\}\right]\ ,
\end{equation}
where $y^{\text{max}}$ is the largest observed value of $f$. We propose a multi-fidelity version of it as:
\begin{equation}
\operatorname{EI_{MF}}(\conf,j|\mathcal{D})=\mathbb{E}\left[\max\left\{f(\conf,j) - y_j^{\text{max}}, 0\right\}\right]\ ,
\end{equation}
where:
\begin{equation}
    y_j^{\text{max}}=\begin{cases}
        \max\left\{y\ |\ ((\conf,\cdot,j), y)\in\mathcal{D}\right\} &  \text{if}\, ((\conf,\cdot,j),y)\in\mathcal{D}\\
        \max\left\{y\ |\ (\cdot,y)\in\mathcal{D}\right\} & \text{otherwise}
    \end{cases}
\end{equation}
Simply put, $y_j^{\text{max}}$ is the largest observed value of $f$ for a budget of $j$ if it exists already, otherwise, it is the largest observed value for any budget.
If there is only one possible budget, the multi-fidelity expected improvement is identical to expected improvement.

\subsection{The \algabbr{} Algorithm}

The \algabbr{} algorithm looks very similar to many black-box Bayesian optimization algorithms as shown in Algorithm~\ref{alg:gb-bo}.
The big difference is that at each step we dynamically decide which candidate configuration to train \emph{for a small additional budget}.

\begin{wrapfigure}{r}{0.6\textwidth}
%\vspace{-2cm}
\begin{minipage}{0.6\textwidth}
      \begin{algorithm}[H]
        \caption{\algabbr{} Algorithm}
        \label{alg:gb-bo}
        \begin{algorithmic}[1]
        \STATE $b(\conf) = 0\ \forall\conf\in\confspace$
        \WHILE{not converged}
        \STATE $\conf_{i}\leftarrow\argmax_{\conf\in\confspace}\operatorname{EI_{MF}}\left(\conf,b(\conf)+1\right)$ (Sec.~\ref{sub:acq-fct})
        \STATE Observe $y_{i,b(\conf_i)+1}$.
        \STATE $b(\conf_i)\leftarrow b(\conf_i)+1$
        \STATE $\mathcal{D}\leftarrow\mathcal{D}\cup\left\{((\conf_i, \mathbf{Y}_{i,b(\conf_i)-1},b(\conf_i)),y_{i,b(\conf_i)})\right\}$
        \STATE Update the surrogate on $\mathcal{D}$. (Sec.~\ref{sub:surrogate})
        \ENDWHILE
        \textbf{return} $\conf_{i}$ with largest $y_{i,\cdot}$.
        \end{algorithmic}
    \end{algorithm}
\end{minipage}
\end{wrapfigure}
Possible candidates are previously unconsidered configurations as well as configurations that did not reach the maximum budget.
In Line 2, the most promising candidate is chosen using the acquisition function introduced in Section~\ref{sub:acq-fct} and the surrogate model's predictions.
It is important to highlight that we do not maximize the acquisition function along the budget dimensionality.
Instead, we set the budget $b$ such that it is by exactly one higher than the budget used to evaluate $\conf_i$ before.
%If $\conf_i$ has not been evaluated for any budget yet ($b(\conf_i)=0$), $j$ is set to $1$.
This ensures that we explore configurations by slowly increasing the budget.
After the candidate and the corresponding budget are selected, the function $f$ is evaluated and we observe $y_{i,j}$ (Line 3).
This additional data point is added to $\mathcal{D}$ in Line 4.
Then in Line 5, the surrogate model is updated according to the training scheme described in Section~\ref{sub:surrogate}.
\section{Experimental Protocol}
\label{sec:experiments}

\subsection{Experimental Setup}
\label{subsec:expsetup}
We evaluate \algabbr{} in three different settings on hyperparameter optimization for tabular, text, and image classification against several competitor methods, the details of which are provided in the following subsections. We ran all of our experiments on an Amazon EC2 M5 Instance (m5.xlarge).
%These include Hyperband~\citep{Li2017}, BOHB~\citep{Falkner2018}, DEHB~\citep{Awad2021}, and Dragonfly~\citep{Metz2020}.
%We use Dragonfly's multi-fidelity optimizer~\citep{Kandasamy2017}.
%For a sanity check, we also compare against random search~\citep{Bergstra2012}.
%We use the publicly available implementations whenever available and we implemented Hyperband and random search ourselves.
In our experiments, we report the mean of ten repetitions and we report two common metrics, the regret and the average rank. The regret refers to the absolute difference between the score of the solution found by an optimizer compared to the best possible score.
If we report the regret as an aggregate result over multiple datasets, we report the mean over all regrets.
The average rank is the metric we use to aggregate rank results over different datasets.
We provide further implementation and training details in Appendix~\ref{app:implementation-details}.
Our implementation of \algabbr{} is publicly available.\footnote{\url{https://github.com/releaunifreiburg/DyHPO}}

%For each dataset, the best performing method obtains a rank of 1.
%Ties are broken by using the average rank, e.g., if the methods have scores 0.9, 0.8, 0.8, 0.7, the ranks are 1, 2.5, 2.5, and 4.
%For both metrics, smaller is better.

\subsection{Benchmarks}
\label{subsec:benchmarks}

In our experiments, we use the following benchmarks.
We provide more details in Appendix~\ref{app:benchmarks}.

%\begin{description}
\textbf{LCBench:} A learning curve benchmark~\citep{Zimmer2021} that evaluates neural network architectures for tabular datasets. LCBench contains learning curves for 35 different datasets, where 2,000 neural networks per dataset are trained for 50 epochs with Auto-PyTorch.

\textbf{TaskSet:} A benchmark that features diverse tasks~\cite{Metz2020} from different domains and includes 5 search spaces with different degrees of freedom, where, every search space includes 1000 hyperparameter configurations. In this work, we focus on a subset of NLP tasks (12 tasks) and we use the Adam8p search space with 8 continuous hyperparameters.

\textbf{NAS-Bench-201:} A benchmark consisting of 15625 hyperparameter configurations representing different architectures on the CIFAR-10, CIFAR-100 and ImageNet datasets~\cite{Dong2020}. NAS-Bench-201 features a search space of 6 categorical hyperparameters and each architecture is trained for 200 epochs.

\subsection{Baselines}
\label{subsec:baselines}

%In our experiments, we use the following baselines:

\textbf{Random Search:} A random/stochastic black-box search method for HPO.

\textbf{HyperBand:} A multi-arm bandit method that extends successive halving by multiple brackets with different combinations of the initial number of configurations, and their initial budget~\citep{Li2017}.

\textbf{BOHB:} An extension of Hyperband that replaces the random sampling of the initial configurations for each bracket with recommended configurations from a model-based approach~\citep{Falkner2018}. BOHB builds a model for every fidelity that is considered.

\textbf{DEHB:} A method that builds upon Hyperband by exploiting differential evolution to sample the initial candidates of a Hyperband bracket~\citep{Awad2021}.

\textbf{ASHA:} An asynchronous version of successive halving (or an asynchronous version of Hyperband if multiple brackets are run). ASHA~\cite{Li2020a} does not wait for all configurations to finish inside a successive halving bracket, but, instead promotes configurations to the next successive halving bracket in real-time.

\textbf{MF-DNN:} A multi-fidelity Bayesian optimization method that uses deep neural networks to capture the relationships between different fidelities~\cite{Li2020}.

\textbf{Dragonfly:} We compare against BOCA~\citep{Kandasamy2017} by using the Dragonfly library~\cite{Kandasamy2020}. This method suggests the next hyperparameter configuration as well as the budget it should be evaluated for.

\subsection{Research Hypotheses and Associated Experiments}

\textbf{Hypothesis 1:} \algabbr{} achieves state-of-the-art results in multi-fidelity HPO.

\textbf{Experiment 1:} We compare \algabbr{} against the baselines of Section~\ref{subsec:baselines} on the benchmarks of Section~\ref{subsec:benchmarks} with the experimental setup of Section~\ref{subsec:expsetup}. For TaskSet we follow the authors' recommendation and report the number of steps (every 200 iterations).

\textbf{Hypothesis 2:} \algabbr{}'s runtime overhead has a negligible impact on the quality of results.

\textbf{Experiment 2:} We compare \algabbr{} against the baselines of Section~\ref{subsec:baselines} over the wallclock time. The wallclock time includes both \textit{(i)} the optimizer's runtime overhead for recommending the next hyperparameter configuration, plus \textit{(ii)} the time needed to evaluate the recommended configuration. In this experiment, we consider all datasets where the average training time per epoch is at least 10 seconds, because, for tasks where the training time is short, there is no practical justification for complex solutions and their overhead. In these cases, we recommend using a random search. We don't report results for TaskSet because the benchmark lacks training times.

\textbf{Hypothesis 3:} \algabbr{} uses the computational budget more efficiently than baselines.

\textbf{Experiment 3:} To further verify that \algabbr{} is efficient compared to the baselines, we investigate whether competing methods spend their budgets on qualitative candidates. Concretely we: i) calculate the precision of the top (w.r.t. ground truth) performing configurations that were selected by each method across different budgets, ii) compute the average regret of the selected configurations across budget, and iii) we compare the fraction of top-performing configurations at a given budget that were not top performers at lower budgets, i.e. measure the ability to handle the poor correlation of performances across budgets.

%\subsection{Feedforward Neural Networks}

\begin{figure}[t]
  \centering
    \includegraphics[width=0.265\textwidth]{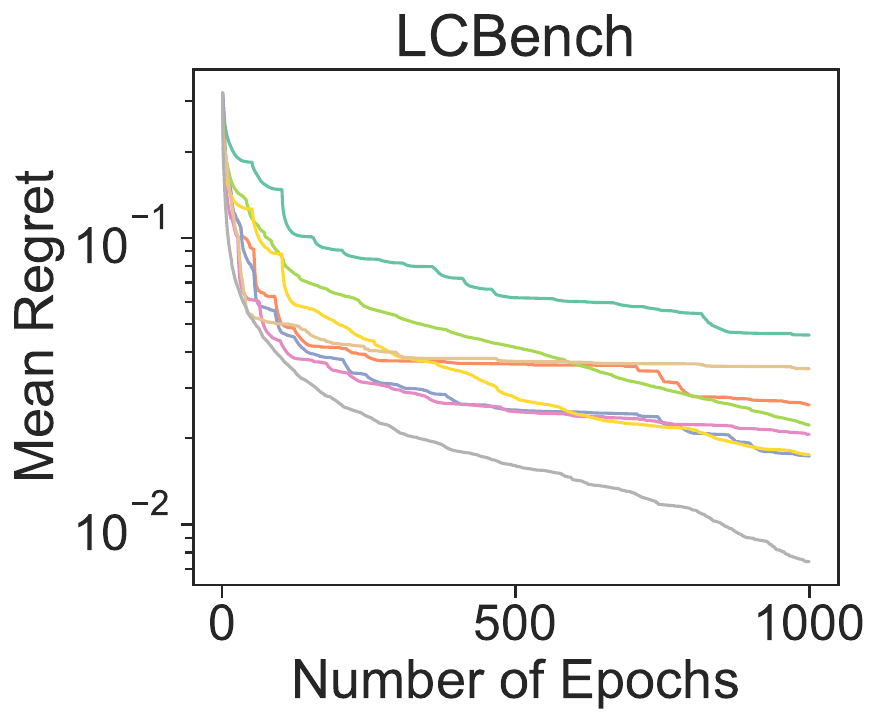}
    \includegraphics[width=0.265\textwidth]{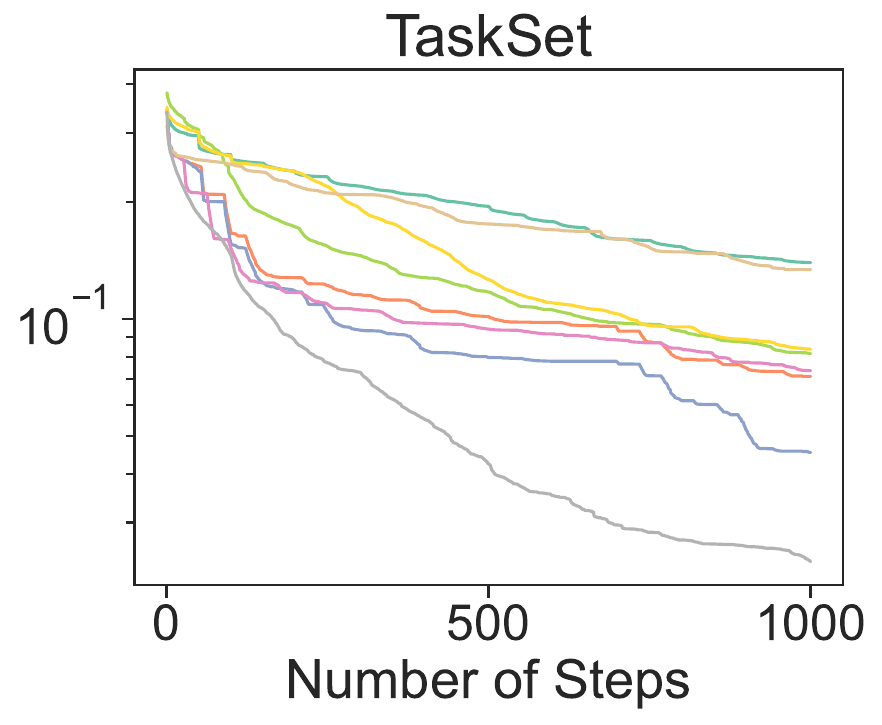}
    \includegraphics[width=0.3925\textwidth]{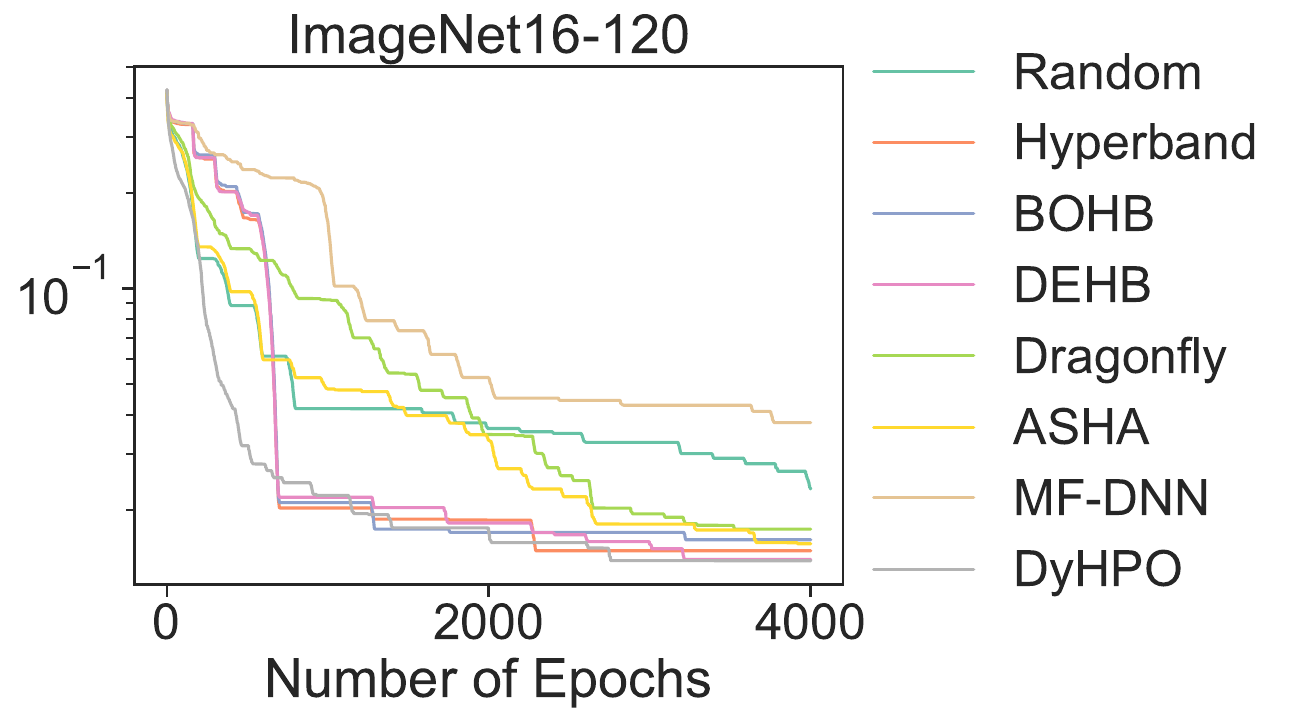}
  \caption{The mean regret for the different benchmarks over the number of epochs or steps (every 200 iterations). The results are aggregated over 35 different datasets for LCBench and aggregated over 12 different NLP tasks for TaskSet.}
  %Again, \algabbr{} shows the best performance among all methods for both evaluation metrics.  \algabbr{} achieves the best performance among all methods for both metrics.}
  \label{fig:regret_aggregated_results_steps}
\end{figure}

\section{Results} 
\label{main:results}

\paragraph{Experiment 1: \algabbr{} achieves state-of-the-art results.} In our first experiment, we evaluate the various methods on the benchmarks listed in Section \ref{subsec:benchmarks}.
We show the aggregated results in Figure~\ref{fig:regret_aggregated_results_steps}, the results show that \algabbr{} manages to outperform competitor methods over the set of considered benchmarks by achieving a better mean regret across datasets. Not only does \algabbr{} achieve a better final performance, it also achieves strong anytime results by converging faster than the competitor methods.
% In Figure~\ref{fig:regret_aggregated_results_steps}, we aggregate the normalized wallclock time by dividing the actual wallclock time of baselines by the total wallclock time of our method \algabbr{} including the overhead incurred by fitting the deep GP. In that manner, we can aggregate wallclock times across datasets.
For the extended results, related to the performance of all methods on a dataset level, we refer the reader to Appendix~\ref{app:additional_plots}.

\begin{figure}[ht]
  \centering
    \includegraphics[width=0.245\columnwidth]{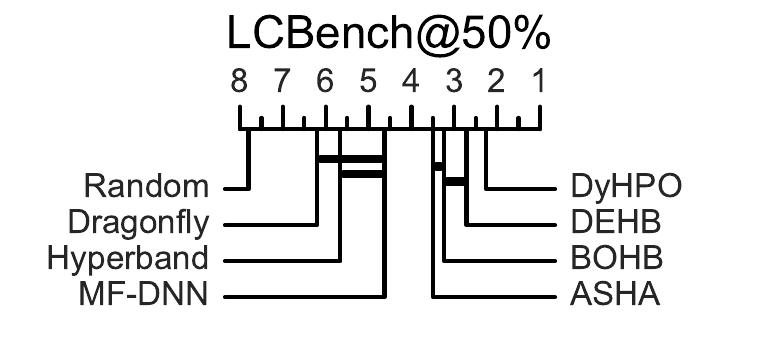}
    \includegraphics[width=0.245\columnwidth]{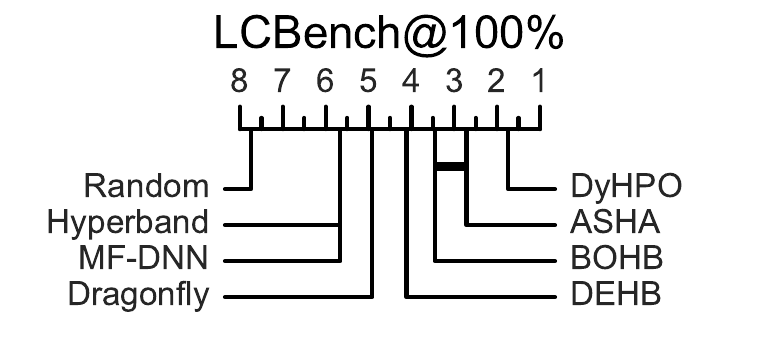}
    \includegraphics[width=0.245\columnwidth]{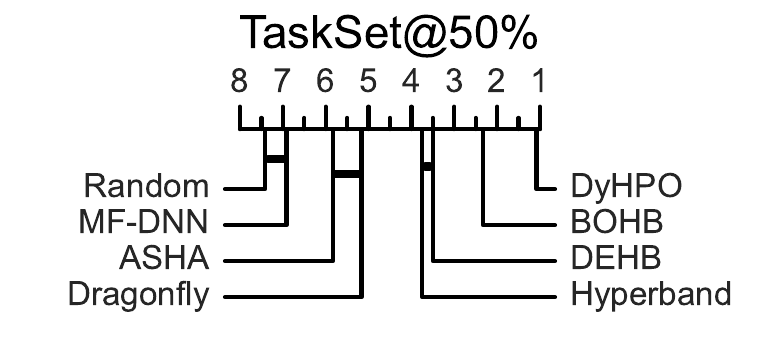}
    \includegraphics[width=0.245\columnwidth]{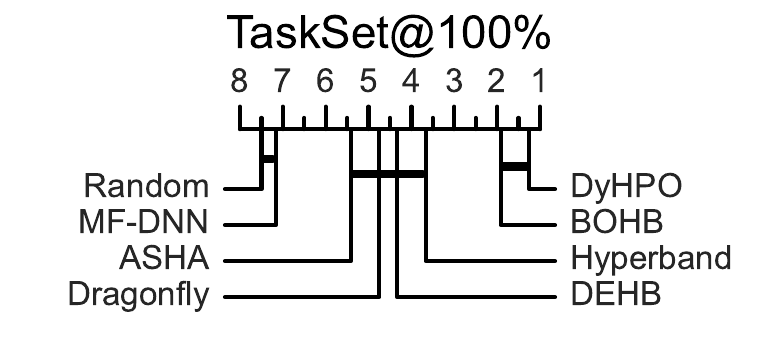}

  \caption{Critical difference diagram for LCBench and TaskSet in terms of the number of HPO steps. The results correspond to results after 500 and 1,000 epochs. Connected ranks via a bold bar indicate that performances are not significantly different ($p > 0.05$).}%\algabbr{}'s improvement is statistically significant in the majority of cases.}
  \label{fig:lcbench-cr}
\end{figure}

In Figure~\ref{fig:lcbench-cr}, we provide further evidence that \algabbr{}'s improvement over the baselines is statistically significant. The critical difference diagram presents the ranks of all methods and provides information on the pairwise statistical difference between all methods for two fractions of the number of HPO steps (50\% and 100\%). We included the LCBench and TaskSet benchmarks in our significance plots. NAS-Bench-201 was omitted because it has only 3 datasets and the statistical test cannot be applied. Horizontal lines indicate groupings of methods that are not significantly different.
As suggested by the best published practices~\cite{Demsar2006}, we use the Friedman test to reject the null hypothesis followed by a pairwise post-hoc analysis based on the Wilcoxon signed-rank test ($\alpha=0.05$). 

For LCBench, \algabbr{} already outperforms the baselines significantly after 50\% of the search budget, with a statistically significant margin. As the optimization procedure continues, \algabbr{} manages to extend its gain in performance and is the only method that has a statistically significant improvement against all the other competitor methods. Similarly, for TaskSet, \algabbr{} manages to outperform all methods with a statistically significant margin only halfway through the optimization procedure and achieves the best rank over all methods. However, as the optimization procedure continues, BOHB manages to decrease the performance gap with \algabbr{}, although, it still achieves a worse rank across all datasets. Considering the empirical results, we conclude that \textbf{Hypothesis 1 is validated and that \algabbr{} achieves state-of-the-art results on multi-fidelity HPO}.

\begin{figure*}[h]
  \centering
    \includegraphics[width=0.28\textwidth]{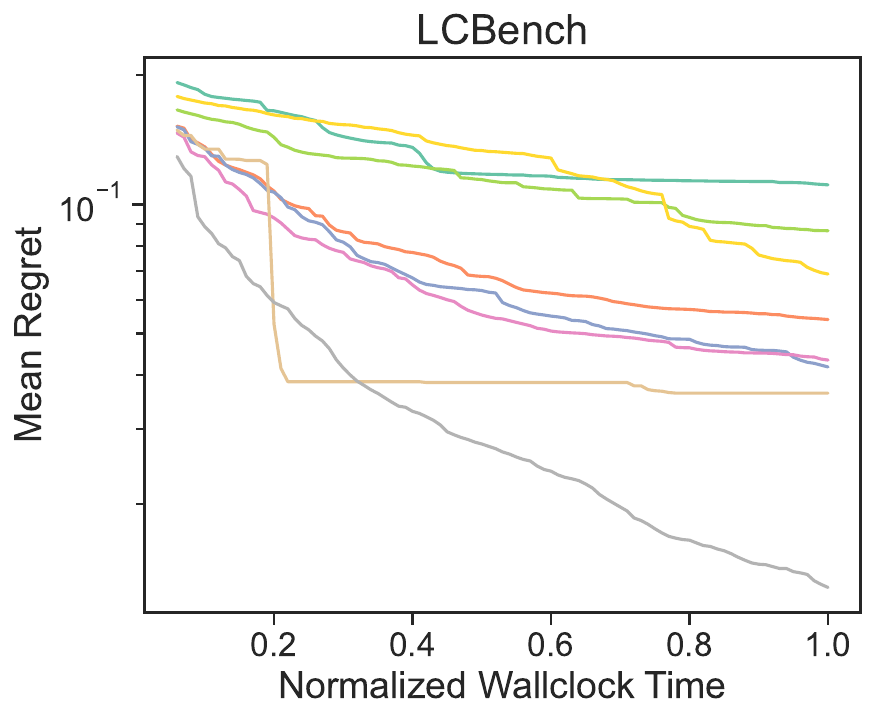}
    \includegraphics[width=0.41\textwidth]{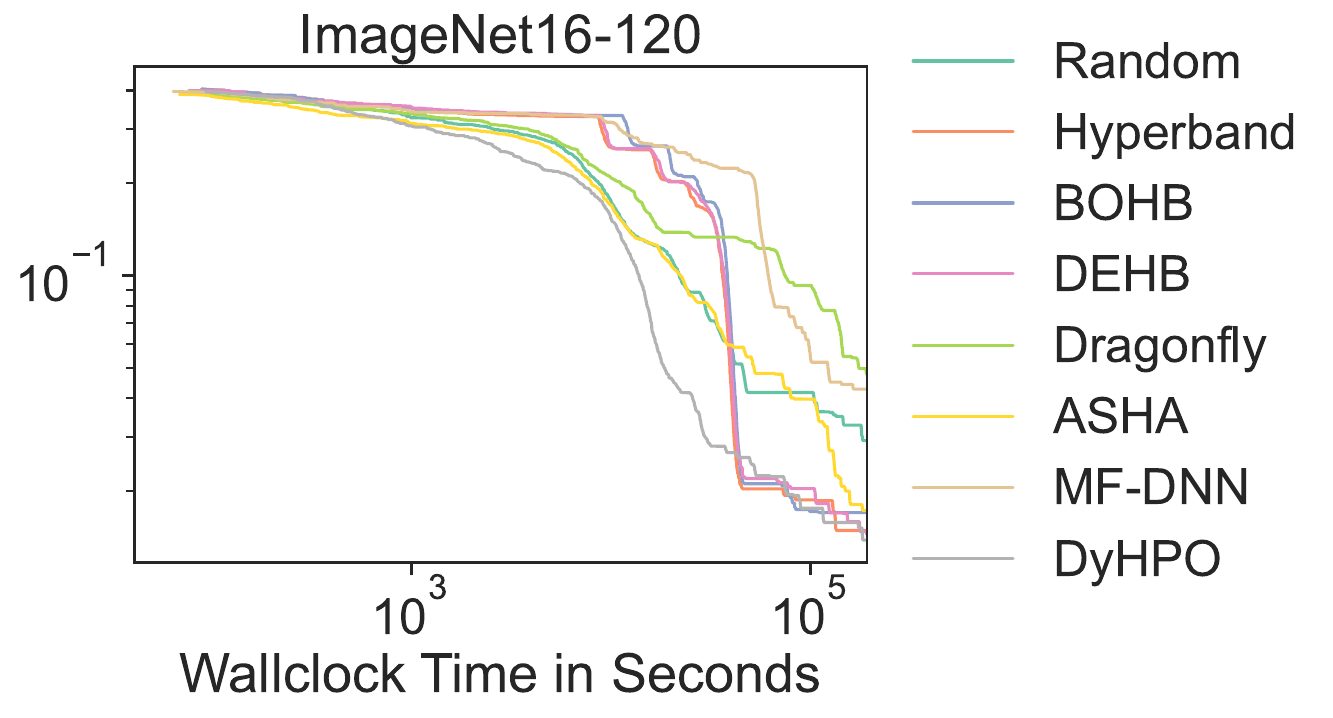}
    \hfill
    \begin{minipage}{0.28\textwidth}
    \vspace{-3.5cm}
    \includegraphics[width=\textwidth]{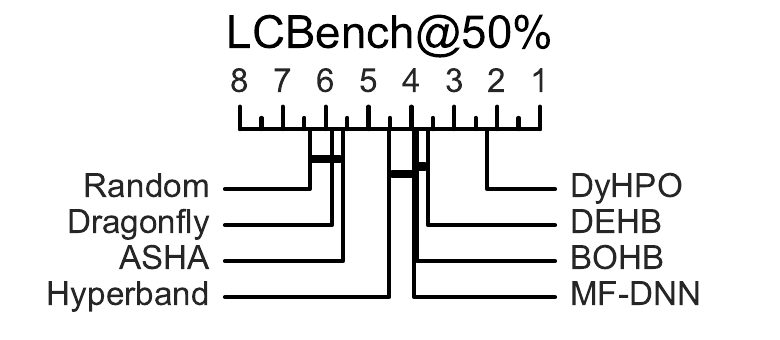}\\
    \includegraphics[width=\textwidth]{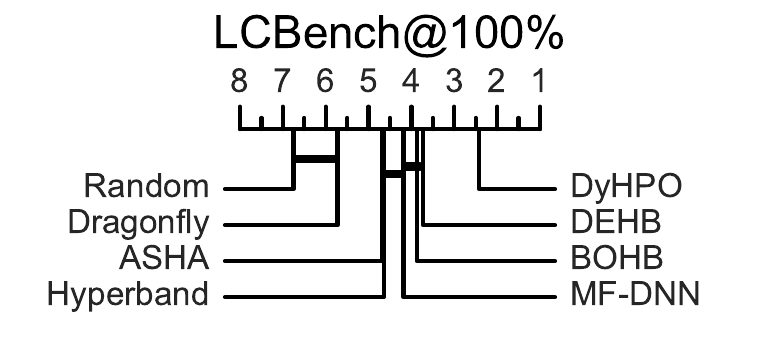}
    \end{minipage}
  \caption{\textbf{Left:} The regret over time for all methods during the optimization procedure for the LCBench benchmark and the ImageNet dataset from the NAS-Bench-201 benchmark. The normalized wall clock time represents the actual run time divided by the total wall clock time of \algabbr{} including the overhead of fitting the deep GP. \textbf{Right:} The critical difference diagram for LCBench halfway through the HPO wall-clock time, and in the end. Connected ranks via a bold bar indicate that differences are not significant ($p > 0.05$).}
  \label{fig:dyhpo_time_overhead}
\end{figure*}

\textbf{Experiment 2: On the impact of \algabbr{}'s overhead on the results.} We present the results of our second experiment in Figure~\ref{fig:dyhpo_time_overhead} (left), where, as it can be seen, \algabbr{} still outperforms the other methods when its overhead is considered. For LCBench, \algabbr{} manages to get an advantage fairly quickly and it only increases the gap in performance with the other methods as the optimization process progresses. Similarly, in the case of ImageNet from NAS-Bench-201, \algabbr{} manages to gain an advantage earlier than other methods during the optimization procedure. Although in the end \algabbr{} still performs better than all the other methods, we believe most of the methods converge to a good solution and the differences in the final performance are negligible. For the extended results, related to the performance of all methods on a dataset level over time, we refer the reader to the plots in Appendix~\ref{app:additional_plots}. Additionally, in Figure~\ref{fig:dyhpo_time_overhead} (right), we provide the critical difference diagrams for LCBench that present the ranks and the statistical difference of all methods halfway through the optimization procedure, and in the end. As it can be seen, \algabbr{} has a better rank with a significant margin with only half of the budget used and it retains the advantage until the end.
%We would like to note that we filtered the tasks considered for the critical difference diagrams by considering only datasets where the average training time per epoch is larger than 10 seconds. The reason being, that in any task where the training time is less than 10 seconds per epoch, any smart model based method will not be able to outperform random search. Since, the setup time would have a significant impact in a cloud environment. Based on the presented results, we conclude that \textbf{Hypothesis 2 is validated and \algabbr{}'s overhead does not impact the achieved state-of-the-art results}.

\begin{figure*}[t]
  \centering
    \includegraphics[width=0.32\textwidth]{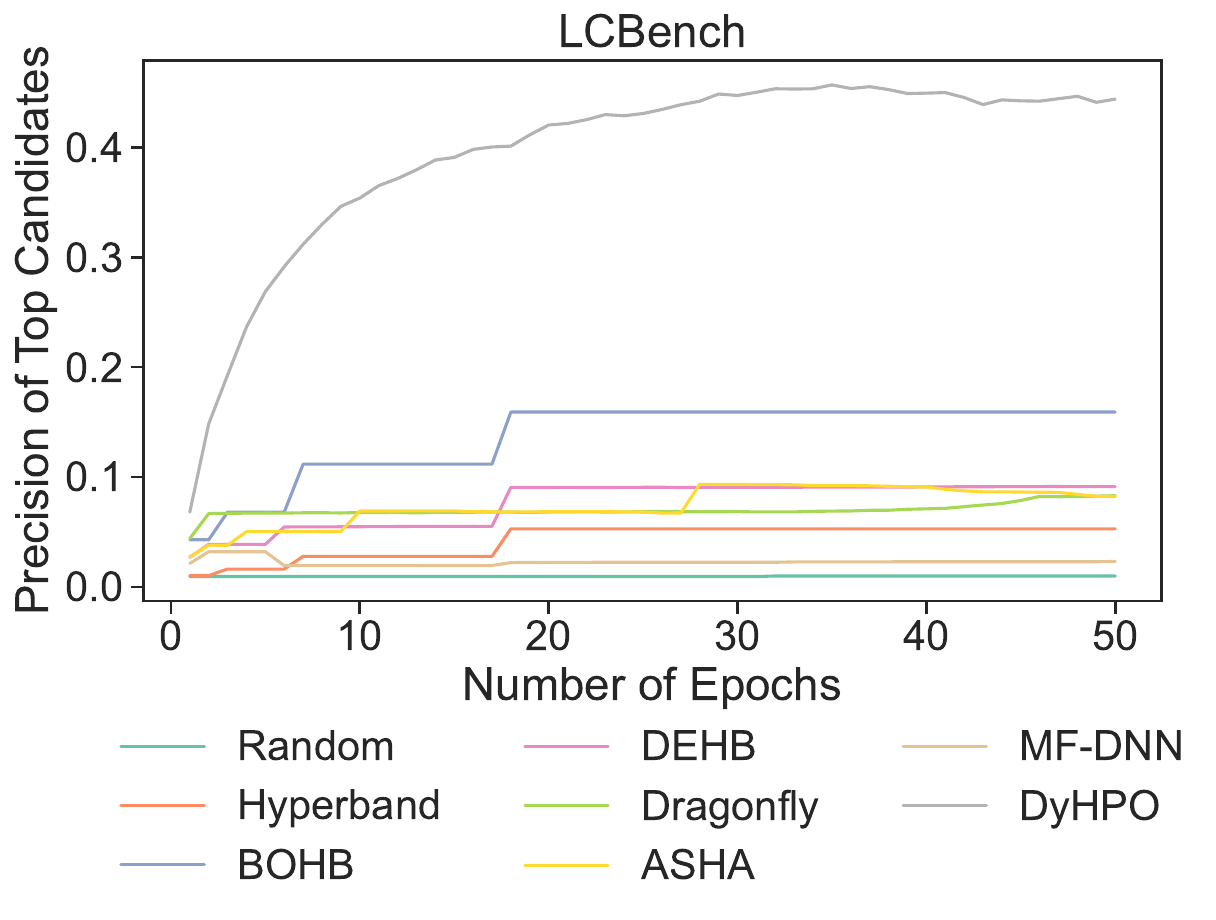}
    \includegraphics[width=0.32\textwidth]{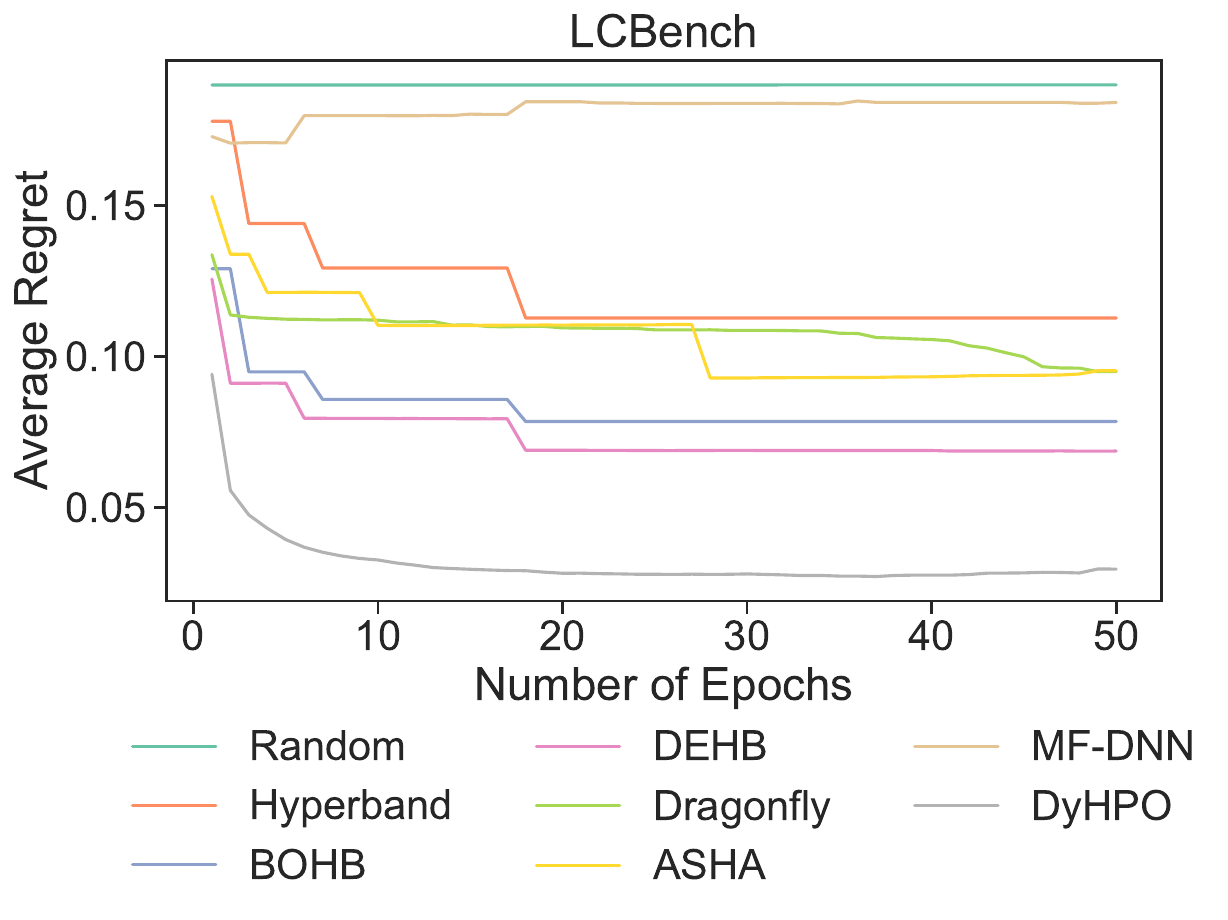}
    \includegraphics[width=0.32\textwidth]{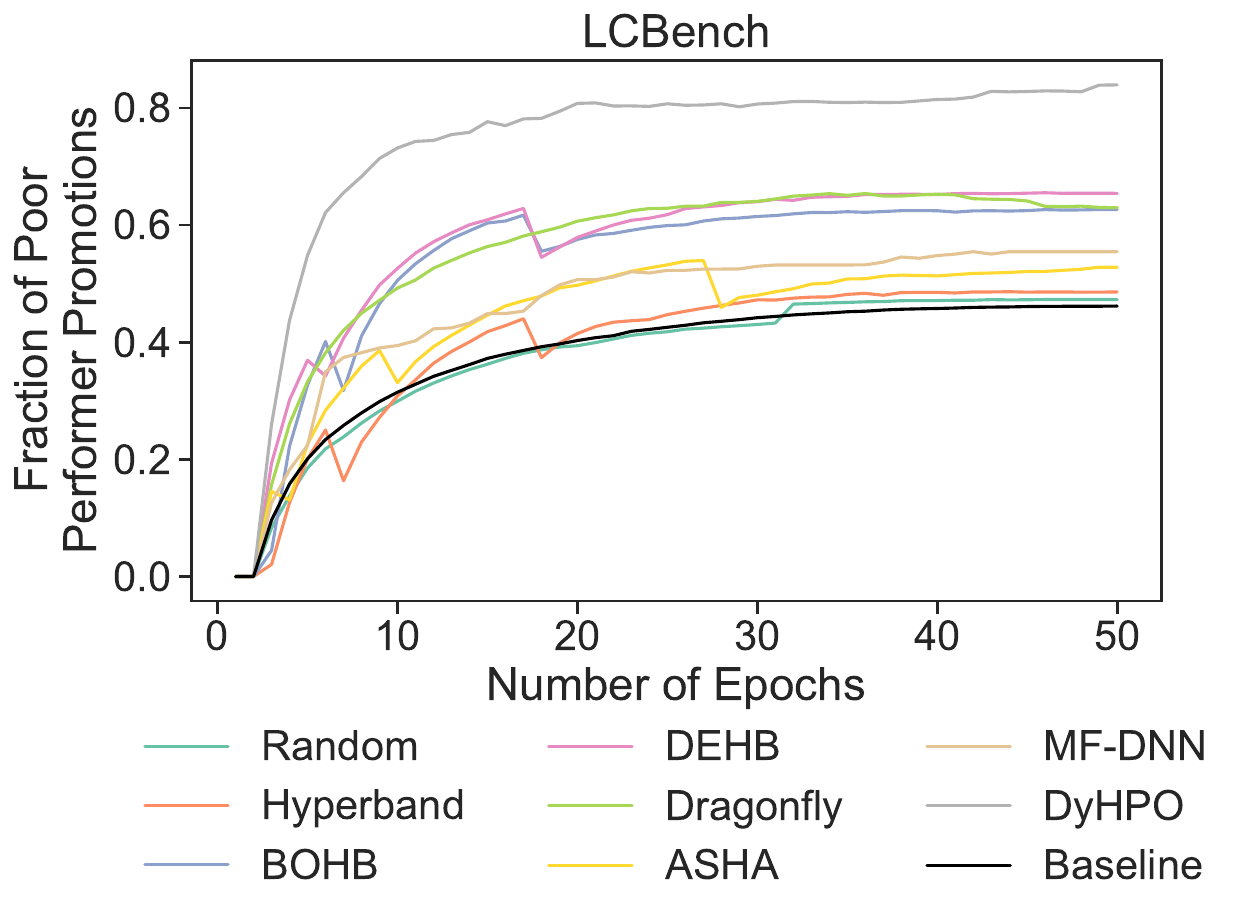}
    \includegraphics[width=0.32\textwidth]{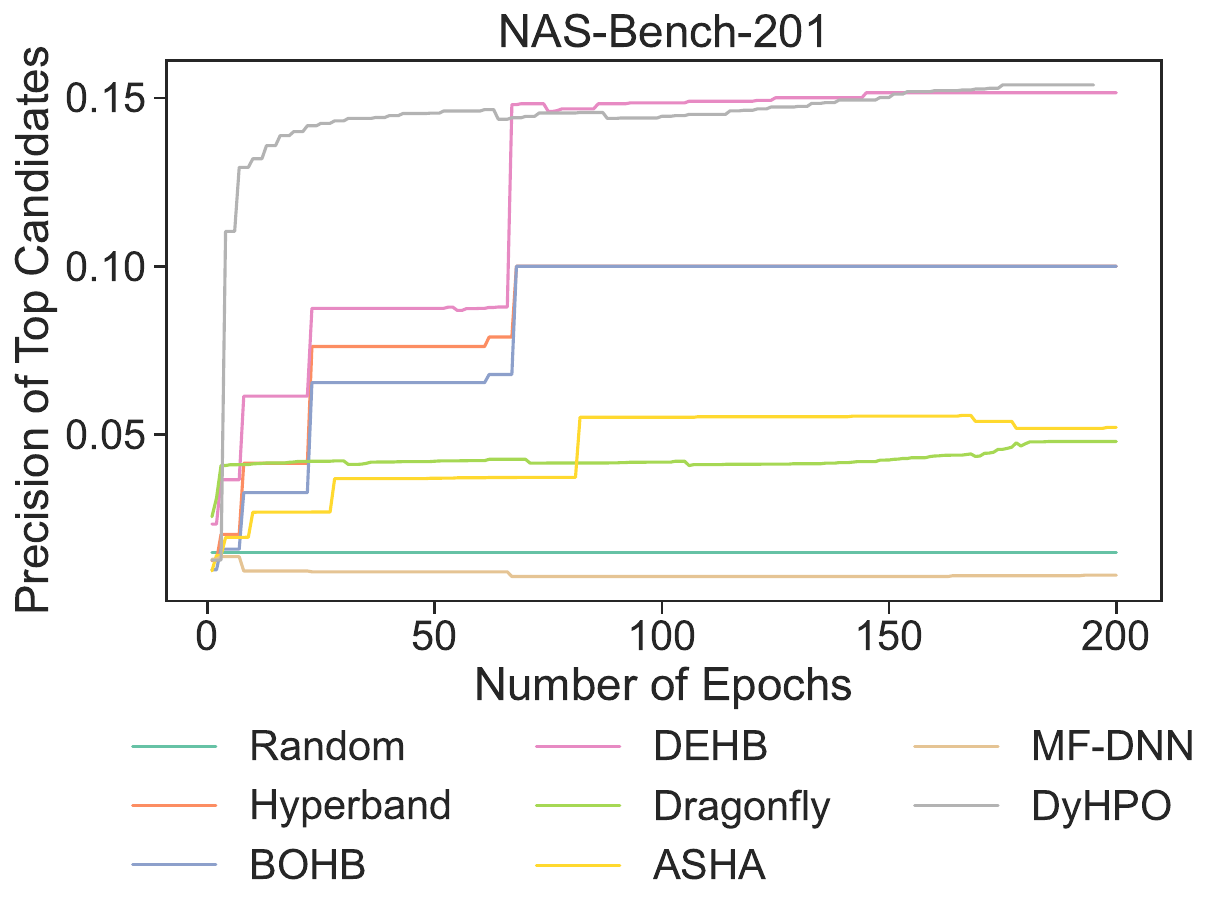}
    \includegraphics[width=0.32\textwidth]{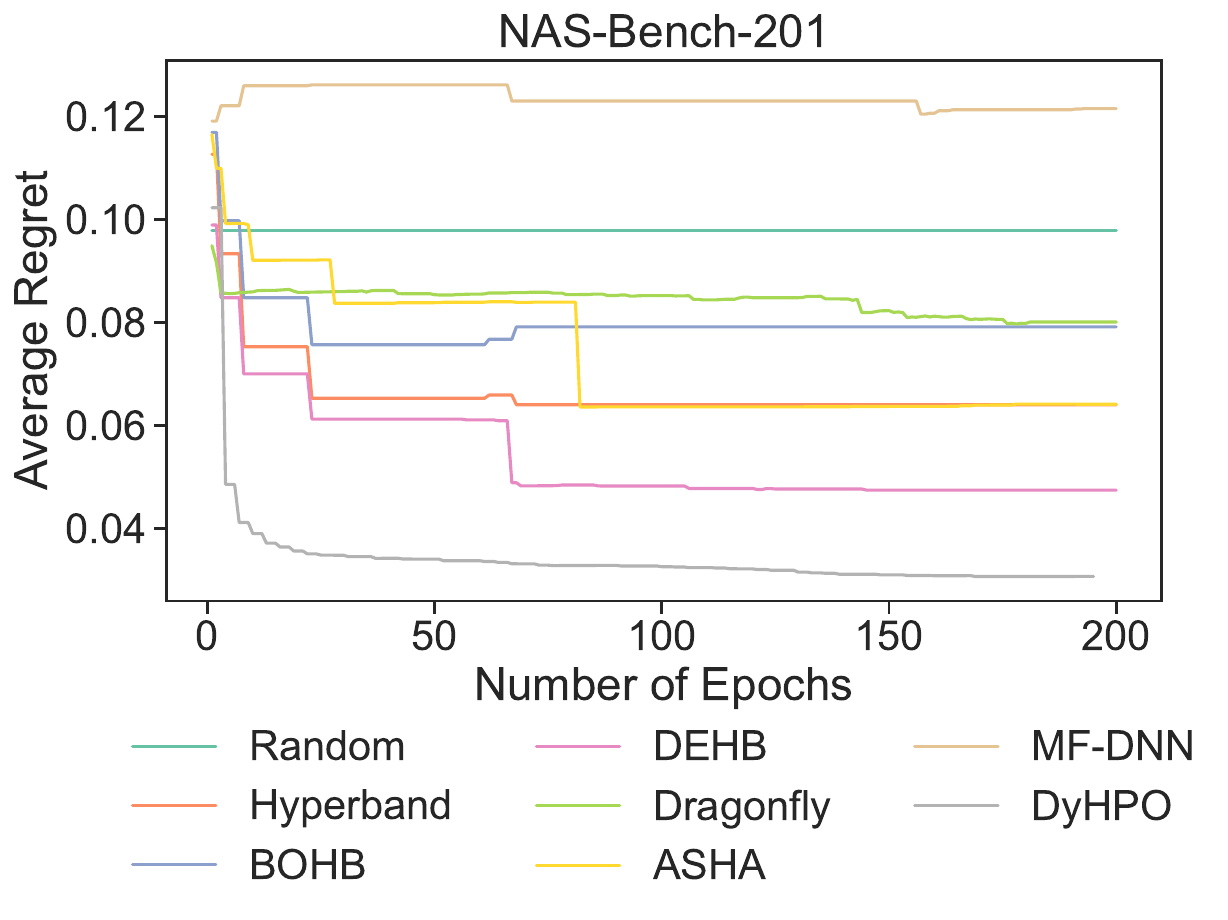}
    \includegraphics[width=0.32\textwidth]{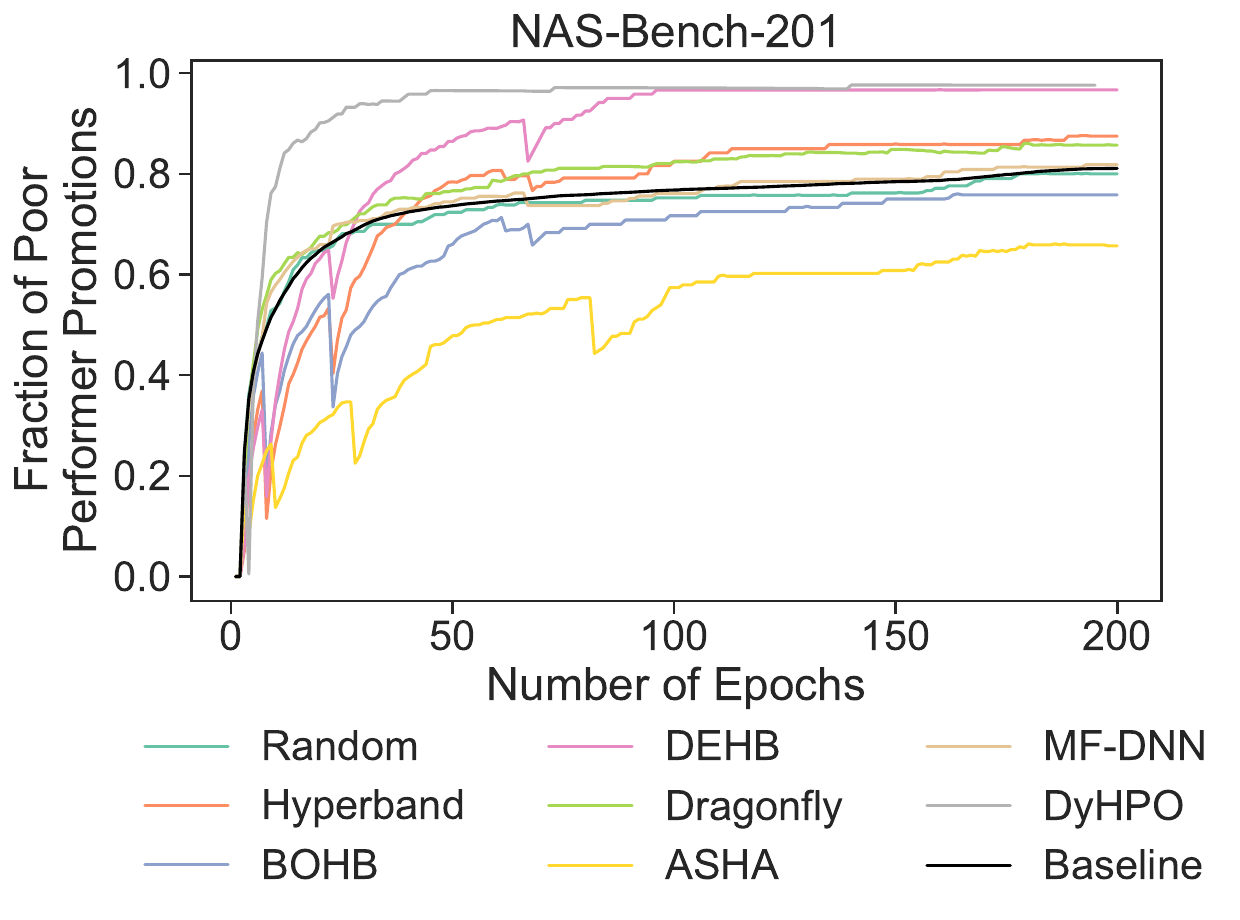}
  \caption{The efficiency of \algabbr{} as the optimization progresses. \textbf{Left:} The fraction of top-performing candidates from all candidates that were selected to be trained. \textbf{Middle:} The average regret for the configurations that were selected to be trained at a given budget. \textbf{Right:} The percentage of configurations that belong to the top 1/3 configurations at a given budget and that were in the top bottom 2/3 of the configurations at a previous budget. All of the results are from the LCBench and NAS-Bench-201 benchmark.}
  \label{fig:dyhpo_efficiency}
\end{figure*}

\textbf{Experiment 3: On the efficiency of \algabbr{}.} In Figure~\ref{fig:dyhpo_efficiency} (left), we plot the precision of every method for different budgets during the optimization procedure, which demonstrates that \algabbr{} effectively explores the search space and identifies promising candidates.
The precision at an epoch $i$ is defined as the number of top 1\% candidates that are trained, divided by the number of all candidates trained, both trained for at least $i$ epochs.
The higher the precision, the more relevant candidates were considered and the less computational resources were wasted.
For small budgets, the precision is low since \algabbr{} spends budget to consider various candidates, but then, promising candidates are successfully identified and the precision quickly increases. %On the contrary, the other methods dedicate significantly more resources to irrelevant candidates which explains why \algabbr{} finds good candidates faster. 
This argument is further supported in Figure~\ref{fig:dyhpo_efficiency} (middle), where we visualize the average regret of all the candidates trained for at least the specified number of epochs on the x-axis. In contrast to the regret plots, here we do not show the regret of the best configuration, but the mean regret of all the selected configurations. The analysis deduces a similar finding, our method \algabbr{} selects more qualitative hyperparameter configurations than all the baselines.

An interesting property of multi-fidelity HPO is the phenomenon of poor rank correlations among the validation performance of candidates at different budgets. In other words, a configuration that achieves a poor performance at a small budget can perform better at a larger budget. To analyze this phenomenon, we measure the percentage of "good"  configurations at a particular budget, that were "bad" performers in at least one of the smaller budgets. We define a "good" performance at a budget B when a configuration achieves a validation accuracy ranked among the top 1/3 of the validation accuracies belonging to all the other configurations that were run until that budget B. %Similarly, a "bad" performance at a budget B represents a configuration whose validation accuracy belongs to the bottom 2/3 of all configurations run at that budget B. 

In Figure~\ref{fig:dyhpo_efficiency} (right), we analyze the percentage of "good" configurations at each budget denoted by the x-axis, that were "bad" performers in at least one of the lower budgets. Such a metric is a proxy for the degree of the promotion of "bad" configurations towards higher budgets. We present the analysis for all the competing methods of our experimental protocol from Section~\ref{sec:experiments}. We have additionally included the ground-truth line annotated as "Baseline", which represents the fraction of past poor performers among all the feasible configurations in the search space. In contrast, the respective methods compute the fraction of promotions only among the configurations that those methods have considered (i.e. selected within their HPO trials) until the budget indicated by the x-axis. We see that there is a high degree of "good" configurations that were "bad" at a previous budget, with fractions of the ground-truth "Baseline" going up to 40\% for the LCBench benchmark and up to 80\% for the NAS-Bench-201 benchmark.

On the other hand, the analysis demonstrates that our method \algabbr{} has promoted more "good" configurations that were "bad" in a lower budget, compared to all the rival methods. In particular, more than 80\% of selected configurations from the datasets belonging to either benchmark were "bad" performers at a lower budget. The empirical evidence validates \textbf{Hypothesis 3 and demonstrates that \algabbr{} efficiently explores qualitative candidates.} We provide the results of our analysis for \algabbr{}'s efficiency on the additional benchmarks (Taskset) in Appendix~\ref{app:additional_plots}.

\textbf{Ablating the impact of the learning curve}

\begin{wrapfigure}{r}{0.5\textwidth}
\begin{minipage}{0.5\textwidth}
\vspace{-0.5cm}
%\begin{figure}[]
  \centering
    \includegraphics[width=0.85\columnwidth]{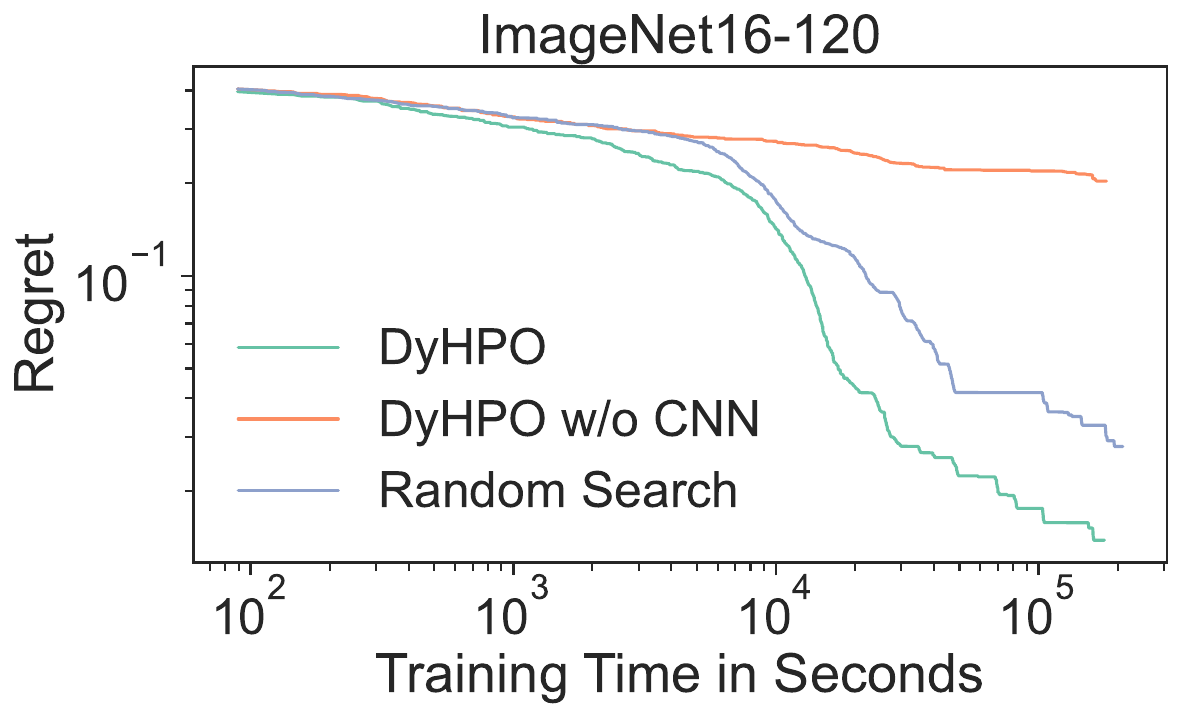}
  \caption{Ablating the impact of the learning curve on \algabbr{}.%, (i) with the learning curve as an input to the kernel and, (ii) without the learning curve.
  }
  \label{fig:ablation_main}
%\end{figure}
\end{minipage}
\vspace{-0.5cm}
\end{wrapfigure}

One of the main differences between \algabbr{} and similar methods~\cite{Kandasamy2017}, is that the learning curve is an input to the kernel function.
For this reason, we investigate the impact of this design choice.
We consider a variation of \algabbr{} w/o CNN, which is simply \algabbr{} without the learning curve.

It is worth emphasizing that both variants (with and without the learning curve) are multi-fidelity surrogates and both receive the budget information through the inputted index $j$ in Equation~\ref{eq:oursurrogate}. The only difference is that \algabbr{} additionally incorporates the pattern of the learning curve. 

We run the ablation on the NAS-Bench-201 benchmark and report the results for ImageNet, the largest dataset in our collection. The ablation results are shown in Figure~\ref{fig:ablation_main}, while the remaining results on the other datasets are shown in Figure~\ref{fig:results_nasbench201_ablation} of the appendix. Based on the results from our learning curve ablation, we conclude that the use of an explicit learning curve representation leads to significantly better results.

% One of the striking observations is that the learning curve has only a small impact for LCBench in contrast to NAS-Bench-201 where it is significant. Our hypothesis is that for LCBench the hyperparameter representation is a valuable feature such that the additional use of the learning curve (which in fact is already implicitly considered by the Gaussian Process) is not required.
 
%As can be seen in Figure~\ref{fig:ablation_main}, \algabbr{} the learning curve has a significant impact on the results in NAS-Bench-201. %We believe one of the reasons to be the poor architecture representation that does not allow for good predictions on its own.
%This hypothesis aligns well with our previous experiments, where BOHB did better on LCBench compared to Hyperband whereas we have not seen any difference on NAS-Bench 201.
%As a reminder, BOHB is a variation of Hyperband that considers the hyperparameter representation to sample candidates.

\section{Limitations of Our Method}
\label{main:limitations}

Although \algabbr{} shows a convincing and statistically significant reduction of the HPO time on diverse Deep Learning (DL) experiments, we cautiously characterized our method only as a "step towards" scaling HPO for DL. The reason for our restrain is the lack of tabular benchmarks for HPO on very large deep learning models, such as Transformers-based architectures~\citep{Devlin2019}. Additionally, the pause and resume part of our training procedure can only be applied when tuning the hyperparameters of parametric models, otherwise, the training of a hyperparameter configuration would have to be restarted. Lastly, for small datasets that can be trained fast, the overhead of model-based techniques would make an approach like random search more appealing.
\section{Conclusions}
In this work, we present \algabbr{}, a new Bayesian optimization (BO) algorithm for the gray-box setting.
We introduced a new surrogate model for BO that uses a learnable deep kernel and takes the learning curve as an explicit input.
Furthermore, we motivated a variation of expected improvement for the multi-fidelity setting.
Finally, we compared our approach on diverse benchmarks on a total of 50 different tasks against the current state-of-the-art methods on gray-box hyperparameter optimization (HPO).
Our method shows significant gains and has the potential to become the de facto standard for HPO in Deep Learning.

%\vspace{-0.2cm}
\section*{Acknowledgments}
Josif Grabocka and Arlind Kadra would like to acknowledge the grant awarded by the Eva-Mayr-Stihl Stiftung. In addition, this research was funded by the Deutsche Forschungsgemeinschaft (DFG, German Research Foundation) under grant number 417962828 and grant INST 39/963-1 FUGG (bwForCluster NEMO). In addition, Josif Grabocka acknowledges the support of the BrainLinks-BrainTools center of excellence.

\bibliography{references}
\bibliographystyle{plainnat}

%%%%%%%%%%%%%%%%%%%%%%%%%%%%%%%%%%%%%%%%%%%%%%%%%%%%%%%%%%%%
\section*{Checklist}

\begin{enumerate}

\item For all authors...
\begin{enumerate}
  \item Do the main claims made in the abstract and introduction accurately reflect the paper's contributions and scope? \answerYes{}
  \item Did you describe the limitations of your work?
    \answerYes{See Section \ref{main:limitations}.}
  \item Did you discuss any potential negative societal impacts of your work?
    \answerYes{See Section ``Societal Implications'' in the Appendix.}
  \item Have you read the ethics review guidelines and ensured that your paper conforms to them?
    \answerYes{}
\end{enumerate}

\item If you are including theoretical results...
\begin{enumerate}
  \item Did you state the full set of assumptions of all theoretical results?
    \answerNA{}
        \item Did you include complete proofs of all theoretical results?
    \answerNA{}
\end{enumerate}

\item If you ran experiments...
\begin{enumerate}
  \item Did you include the code, data, and instructions needed to reproduce the main experimental results (either in the supplemental material or as a URL)?
    \answerNo{We provide our main algorithm in Section \ref{sec:dyhpo} and we additionally provide the detailed implementation details in Appendix \ref{app:ex_setup} for all methods and benchmarks. We will release the code for the camera-ready version of our work.}
  \item Did you specify all the training details (e.g., data splits, hyperparameters, how they were chosen)?
    \answerYes{Please see Appendix \ref{app:ex_setup}.}
        \item Did you report error bars (e.g., with respect to the random seed after running experiments multiple times)?
    \answerYes{We report the statistical significance of the performance difference between our method and the baselines in Section \ref{main:results}}
        \item Did you include the total amount of compute and the type of resources used (e.g., type of GPUs, internal cluster, or cloud provider)?
    \answerYes{See Section \ref{subsec:expsetup}.}
\end{enumerate}

\item If you are using existing assets (e.g., code, data, models) or curating/releasing new assets...
\begin{enumerate}
  \item If your work uses existing assets, did you cite the creators?
    \answerYes{See Section \ref{subsec:benchmarks} and Section \ref{subsec:baselines}.}
  \item Did you mention the license of the assets?
    \answerYes{See Appendix \ref{app:benchmarks} and \ref{app:baselines} where we provide references to the assets where the license is included.}
  \item Did you include any new assets either in the supplemental material or as a URL?
    \answerNo{}
  \item Did you discuss whether and how consent was obtained from people whose data you're using/curating?
    \answerNA{The benchmarks and baselines are open-sourced.}
  \item Did you discuss whether the data you are using/curating contains personally identifiable information or offensive content?
    \answerNA{}{The data does not contain personally identifiable information or offensive content.}
\end{enumerate}

\item If you used crowdsourcing or conducted research with human subjects...
\begin{enumerate}
  \item Did you include the full text of instructions given to participants and screenshots, if applicable?
    \answerNA{}
  \item Did you describe any potential participant risks, with links to Institutional Review Board (IRB) approvals, if applicable?
    \answerNA{}
  \item Did you include the estimated hourly wage paid to participants and the total amount spent on participant compensation?
    \answerNA{}
\end{enumerate}

\end{enumerate}

%%%%%%%%%%%%%%%%%%%%%%%%%%%%%%%%%%%%%%%%%%%%%%%%%%%%%%%%%%%%

%%%%%%%%%%%%%%%%%%%%%%%%%%%%%%%%%%%%%%%%%%%%%%%%%%%%%%%%%%%%%%%%%%%%%%%%%%%%%%%
%%%%%%%%%%%%%%%%%%%%%%%%%%%%%%%%%%%%%%%%%%%%%%%%%%%%%%%%%%%%%%%%%%%%%%%%%%%%%%%
% APPENDIX
%%%%%%%%%%%%%%%%%%%%%%%%%%%%%%%%%%%%%%%%%%%%%%%%%%%%%%%%%%%%%%%%%%%%%%%%%%%%%%%
%%%%%%%%%%%%%%%%%%%%%%%%%%%%%%%%%%%%%%%%%%%%%%%%%%%%%%%%%%%%%%%%%%%%%%%%%%%%%%%
\newpage
\appendix
\onecolumn

\section*{Societal Implications}
\label{app:soc_implications}

In our work, we use only publicly available data with no privacy concerns. Furthermore, our algorithm reduces the overall time for fitting deep networks, therefore, saving computational resources and yielding a positive impact on the environment. Moreover, our method can help smaller research organizations with limited access to resources to be competitive in the deep learning domain, which reduces the investment costs on hardware. Although our method significantly reduces the time taken for optimizing a machine learning algorithm that achieves peak performance, we warn against running our method for an extended time only to achieve marginal gains in performance, unless it is mission-critical. Last but not least, in order to save energy, we invite the community to create sparse benchmarks with surrogates, instead of dense tabular ones.

\section{Experimental Setup}
\label{app:ex_setup}

\subsection{Benchmarks}
\label{app:benchmarks}

%In our experiments, we make use of three different benchmarks.
\paragraph*{LCBench.}
LCBench\footnote{\url{https://github.com/automl/LCBench}} is a feedforward neural network benchmark on tabular data which consists of 2000 configuration settings for each of the 35 datasets.
The configurations were evaluated during HPO runs with AutoPyTorch.
LCBench features a search space of 7 numerical hyperparameters, where every hyperparameter configuration is trained for 50 epochs. The objective is to optimize seven different hyperparameters of funnel-shaped neural networks, i.e., batch size, learning rate, momentum, weight decay, dropout, number of layers, and maximum number of units per layer.

\paragraph*{TaskSet.}
TaskSet\footnote{\url{https://github.com/google-research/google-research/tree/master/task_set}} is a benchmark that features over 1162 diverse tasks from different domains and includes 5 search spaces.
In this work, we focus on NLP tasks and we use the Adam8p search space with 8 continuous hyperparameters. %The search spaces include only numerical hyperparameters and we chose the Adam8p search space because it is the search space with the highest dimensionality that was not sampled heuristically.
We refer to Figure~\ref{fig:results_per_dataset_taskset} for the exact task names considered in our experiments.
The learning curves provided in TaskSet report scores after every 200 iterations.
We refer to those as "steps". The objective is to optimize eight hyperparameters for a set of different recurrent neural networks (RNN) that differ in embedding size, RNN cell, and other architectural features.
The set of hyperparameters consists of optimizer-specific hyperparameters, such as the learning rate, the exponential decay rate of the first and second momentum of Adam, $\beta_1$ and $\beta_2$, and Adam's constant for numerical stability $\varepsilon$.
Furthermore, there are two hyperparameters controlling linear and exponential learning rate decays, as well as L1 and L2 regularization terms.
%TaskSet features the learning curves for different hyperparameter configurations. In particular, for every hyperparameter configuration it features 5 repetitions with different seeds. We aggregate over the 5 repetitions to get the mean curves corresponding to every hyperparameter configuration. For every learning curve, a step/sample constitutes of 200 iterations, for a total of 10000 iterations, or 50 steps.

\paragraph*{NAS-Bench-201.}
NAS-Bench-201\footnote{\url{https://github.com/D-X-Y/NAS-Bench-201}} is a benchmark that has precomputed about 15,600 architectures trained for 200 epochs for the image classification datasets CIFAR-10, CIFAR-100, and ImageNet. The objective is to select for each of the six operations within the cell of the macro architecture one of five different operations. All other hyperparameters such as learning rate and batch size are kept fixed. NAS-Bench-201 features a search space of 6 categorical hyperparameters and each architecture is trained for 200 epochs. 

\subsection{Preprocessing}
\label{app:preprocessing}
In the following, we describe the preprocessing applied to the hyperparameter representation.
For LCBench, we apply a log-transform to batch size, learning rate, and weight decay.
For TaskSet, we apply it on the learning rate, L1 and L2 regularization terms, epsilon, linear and exponential decay of the learning rate.
All continuous hyperparameters are scaled to the range between 0 and 1 using sklearn's MinMaxScaler.
If not mentioned otherwise, we use one-hot encoding for the categorical hyperparameters.
As detailed in subsection~\ref{app:baselines}, some baselines have a specific way of dealing with them.
In that case, we use the method recommended by the authors.

\subsection{Framework}

The framework contains the evaluated hyperparameters and their corresponding validation curves.
The list of candidate hyperparameters is passed to the baseline-specific interface, which in turn, optimizes and queries the framework for the hyperparameter configuration that maximizes utility. Our framework in turn responds with the validation curve and the cost of the evaluation. In case a hyperparameter configuration has been evaluated previously up to a budget $b$ and a baseline requires the response for budget $b + 1$, the cost is calculated accordingly only for the extra budget requested.

\subsection{Implementation Details}
\label{app:implementation-details}
We implement the Deep Kernel Gaussian Process using GPyTorch 1.5~\citep{Gardner2018}.
We use an RBF kernel and the dense layers of the transformation function $\varphi$ have 128 and 256 units.
We used a convolutional layer with a kernel size of three and four filters.
All parameters of the Deep Kernel are estimated by maximizing the marginal likelihood.
We achieve this by using gradient ascent and Adam~\citep{Kingma2015} with a learning rate of $0.1$ and batch size of 64.
We stop training as soon as the training likelihood is not improving for 10 epochs in a row or we completed 1,000 epochs.
For every new data point, we start training the GP with its old parameters to reduce the required effort for training.

\subsection{Baselines}
\label{app:baselines}

\paragraph{Random Search \& Hyperband.}

Random search and Hyperband sample hyperparameter configurations at random and therefore the preprocessing is irrelevant.
We have implemented both from scratch and use the recommended hyperparameters for Hyperband, i.e. $\eta=3$.
%Our random search and Hyperband implementations query an index from the evaluated hyperparameters given by the framework, without performing any additional preprocessing to the hyperparameters. In the case of random search, the index is random, while, for Hyperband, the indices sampled are random in the beginning of a Hyperband bracket, while, in the upcoming successive halving (SH) brackets it retains the top $\eta$ performing configurations of the previous SH brackets.

\paragraph{BOHB.} For our experiments with BOHB, we use version 0.7.4 of the officially-released code\footnote{\url{https://github.com/automl/HpBandSter}}.
%In addition to the initial preprocessing common for all methods, in our BOHB interface, we encode categorical hyperparameters with a one-hot encoder. In our experiments, we use version 0.7.4 of the library.

\paragraph{DEHB.} For our experiments with DEHB, we use the official public implementation\footnote{\url{https://github.com/automl/DEHB/}}. We developed an interface that communicates between our framework and DEHB. In addition to the initial preprocessing common for all methods, we encode categorical hyperparameters with a numerical value in the interval [0, 1]. For a categorical hyperparameter $\conf_i$, we take $N_i$ equal-sized intervals, where $N_i$ represents the number of unique categorical values for hyperparameter $\conf_i$ and we assign the value for a categorical value $n \in N_i$ to the middle of the interval $[n, n + 1]$ as suggested by the authors. For configuring the DEHB algorithm we used the default values from the library.

\paragraph{Dragonfly.} We use the publicly available code of Dragonfly\footnote{\url{https://github.com/dragonfly/dragonfly}}.
No special treatment of categorical hyperparameters is required since Dragonfly has its own way to deal with them.
We use version $0.1.6$ with default settings.

\paragraph{MF-DNN.} We use the official implementation of MF-DNN by the authors\footnote{\url{https://github.com/shib0li/DNN-MFBO}}. Initially, we tried to use multiple incremental fidelity levels like for \algabbr{}, however, the method runtime was too high and it could not achieve competitive results. For that reason, we use only a few fidelity levels like the authors do in their work~\cite{Li2020}. We use the same fidelity levels as for Hyperband, DEHB, and BOHB to have a fair comparison between the baselines. We also use the same number of initial points as for the other methods to have the same maximal resource allocated for every fidelity level.

\paragraph{ASHA-HB.} We use the public implementation from the well-known optuna library (version $2.10.0$). We used the same eta, minimum and maximal budget as for HB, DEHB, and BOHB in our experiments, to have a fair comparison.

\section{Additional Plots}
\label{app:additional_plots}

\begin{figure}
  \centering
    \includegraphics[width=0.48\textwidth]{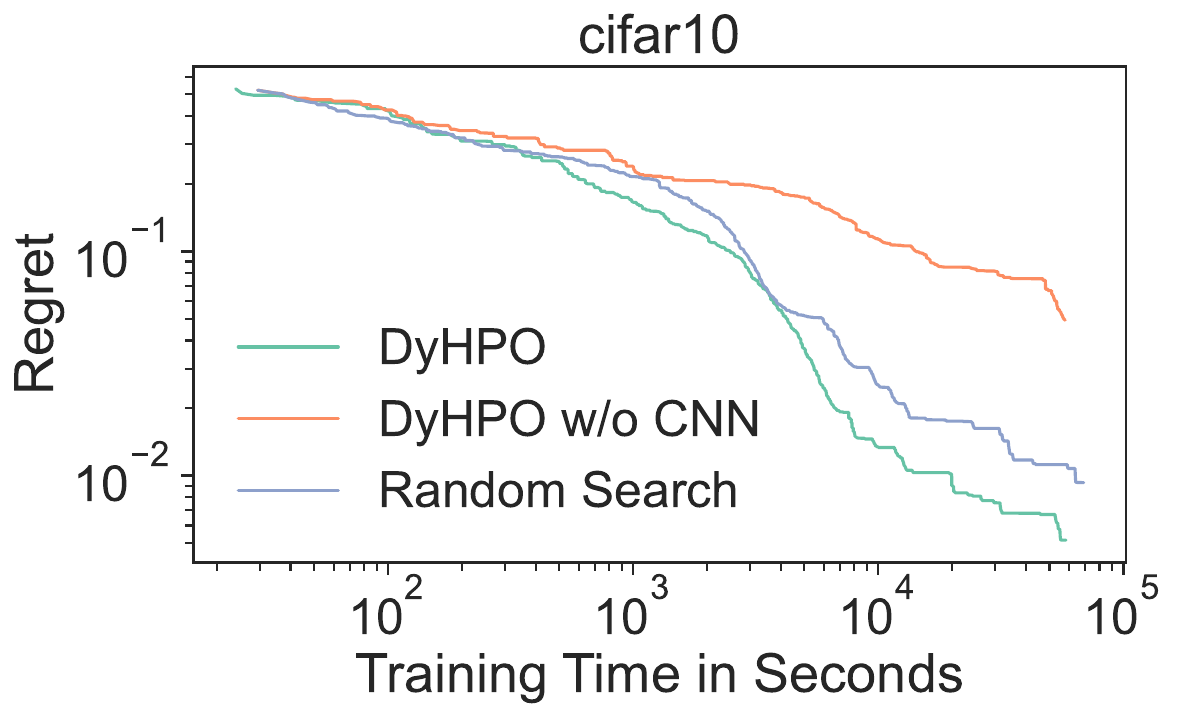}
    \includegraphics[width=0.48\textwidth]{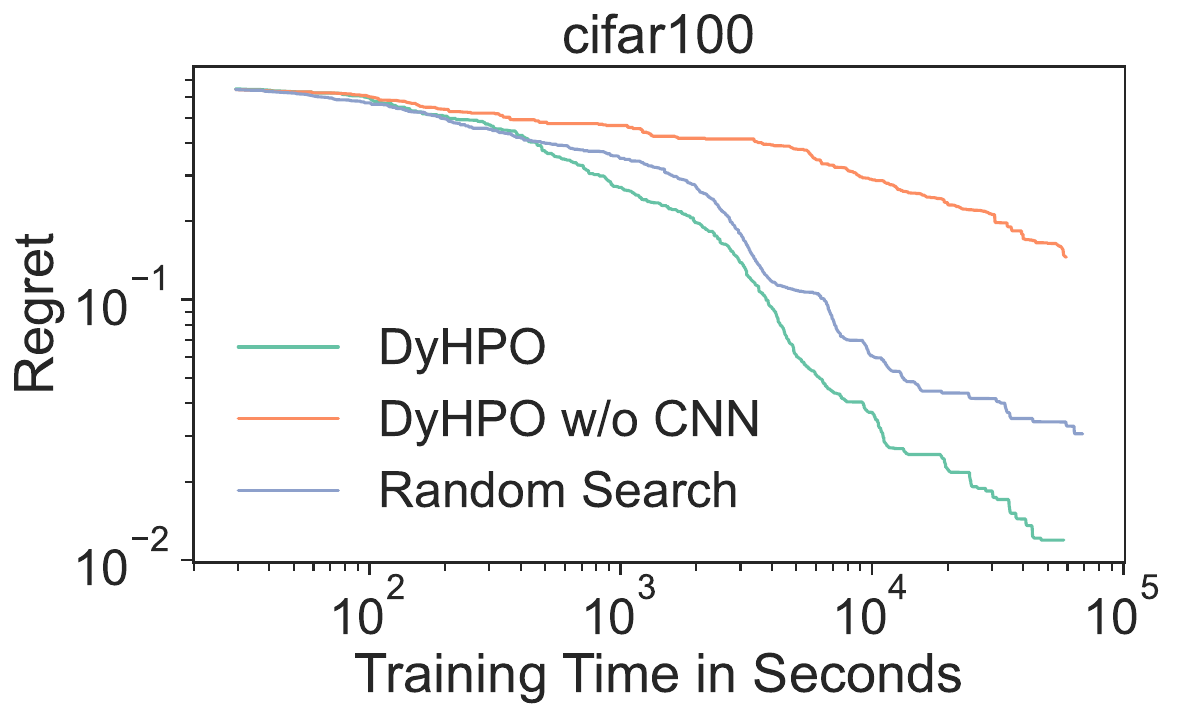}
  \caption{The learning curve ablation for the CIFAR-10 and CIFAR-100 tasks of NAS-Bench-201.}
  \label{fig:results_nasbench201_ablation}
\end{figure}

In Figure~\ref{fig:results_nasbench201_ablation}, we ablate the learning curve input in our kernel, to see the effect it has on performance for the CIFAR-10 and CIFAR-100 datasets from the NAS-Bench-201 benchmark. The results indicate that the learning curve plays an important role in achieving better results by allowing faster convergence and a better anytime performance. 

\begin{figure}[t]
  \centering
    \includegraphics[width=0.48\textwidth]{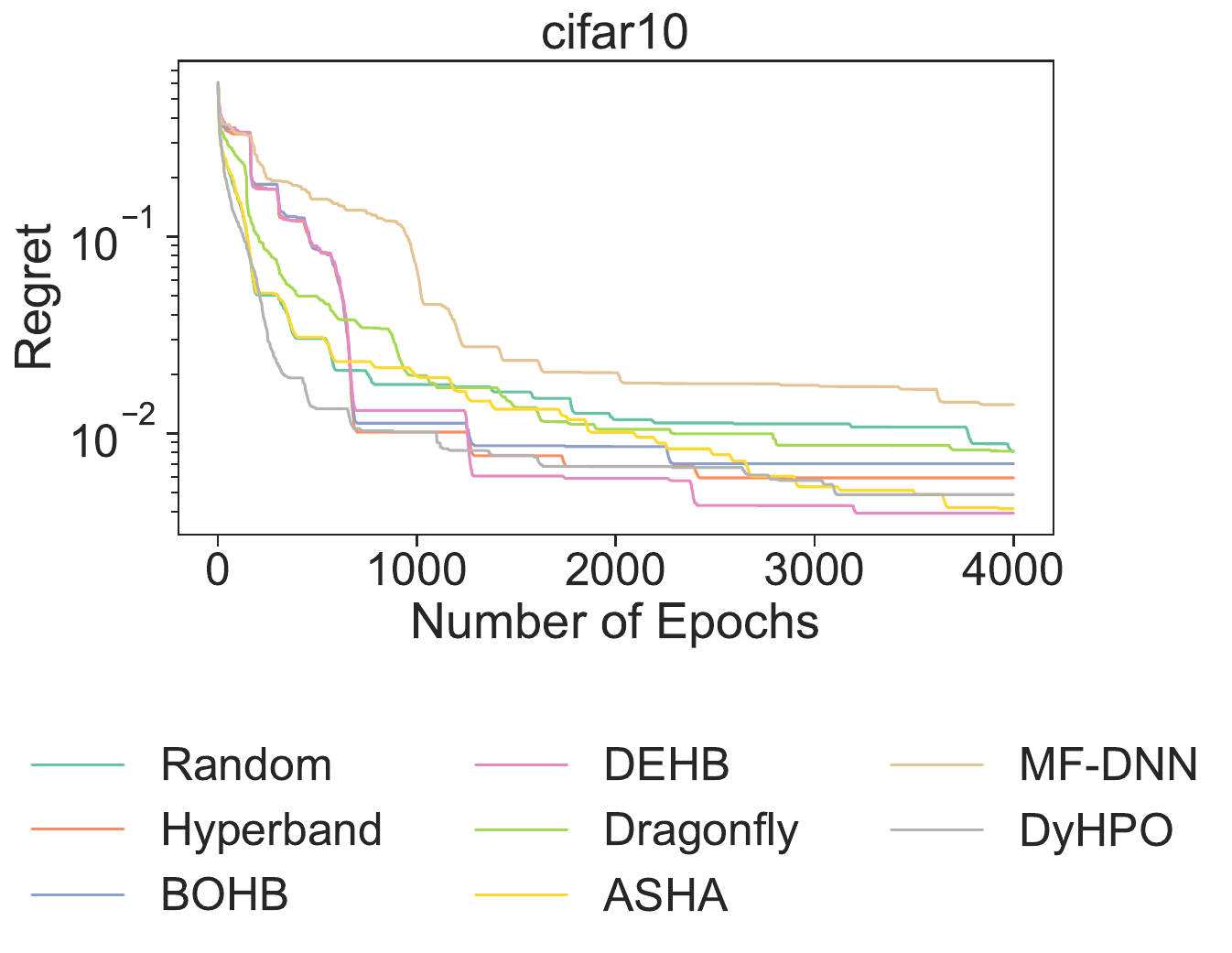}
    \includegraphics[width=0.48\textwidth]{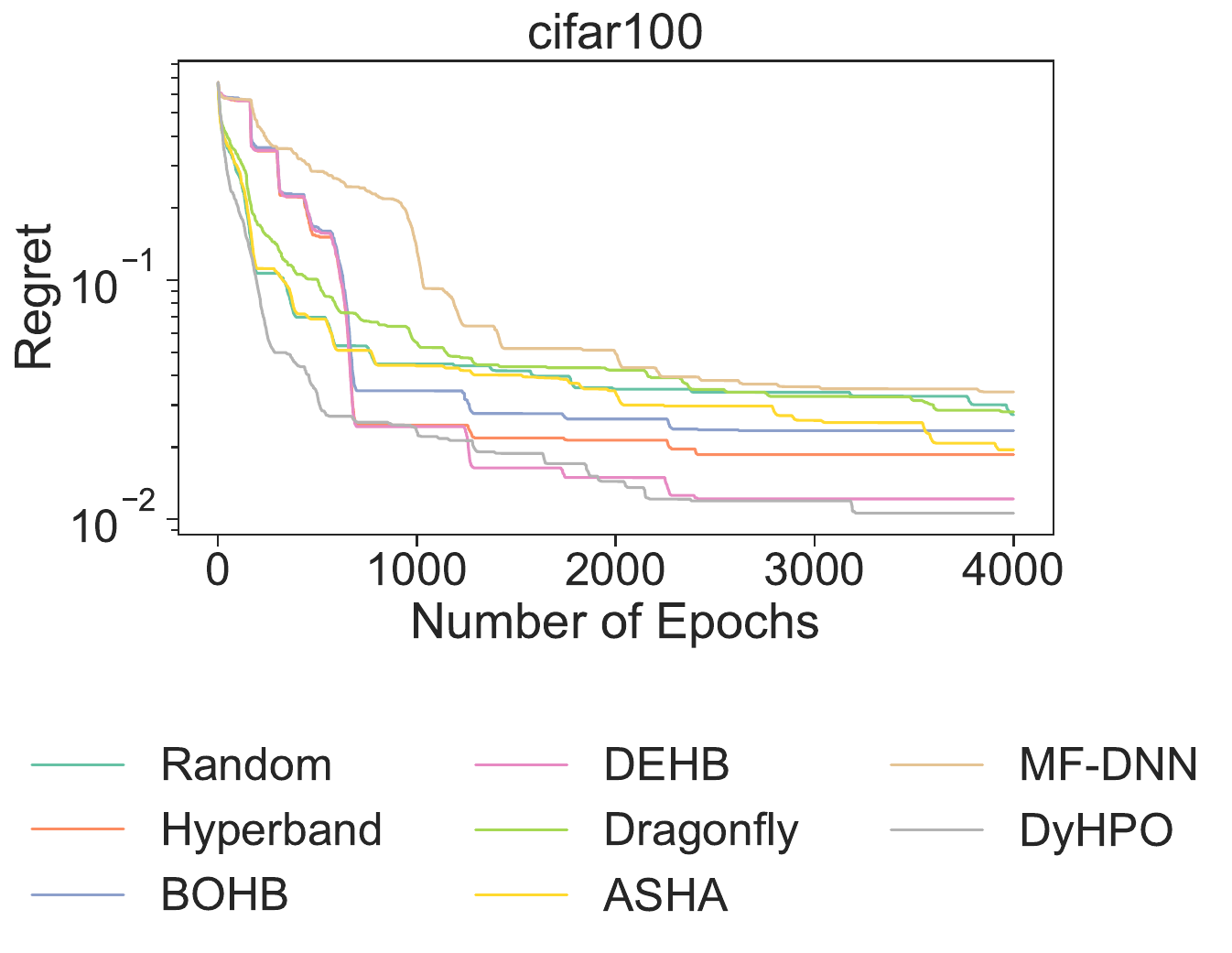}
  \caption{NAS-Bench-201 regret results over the number of epochs spent during the optimization.}
  \label{fig:results_nasbench201_time}
\end{figure}

\begin{figure}[t]
  \centering
    \includegraphics[width=0.48\textwidth]{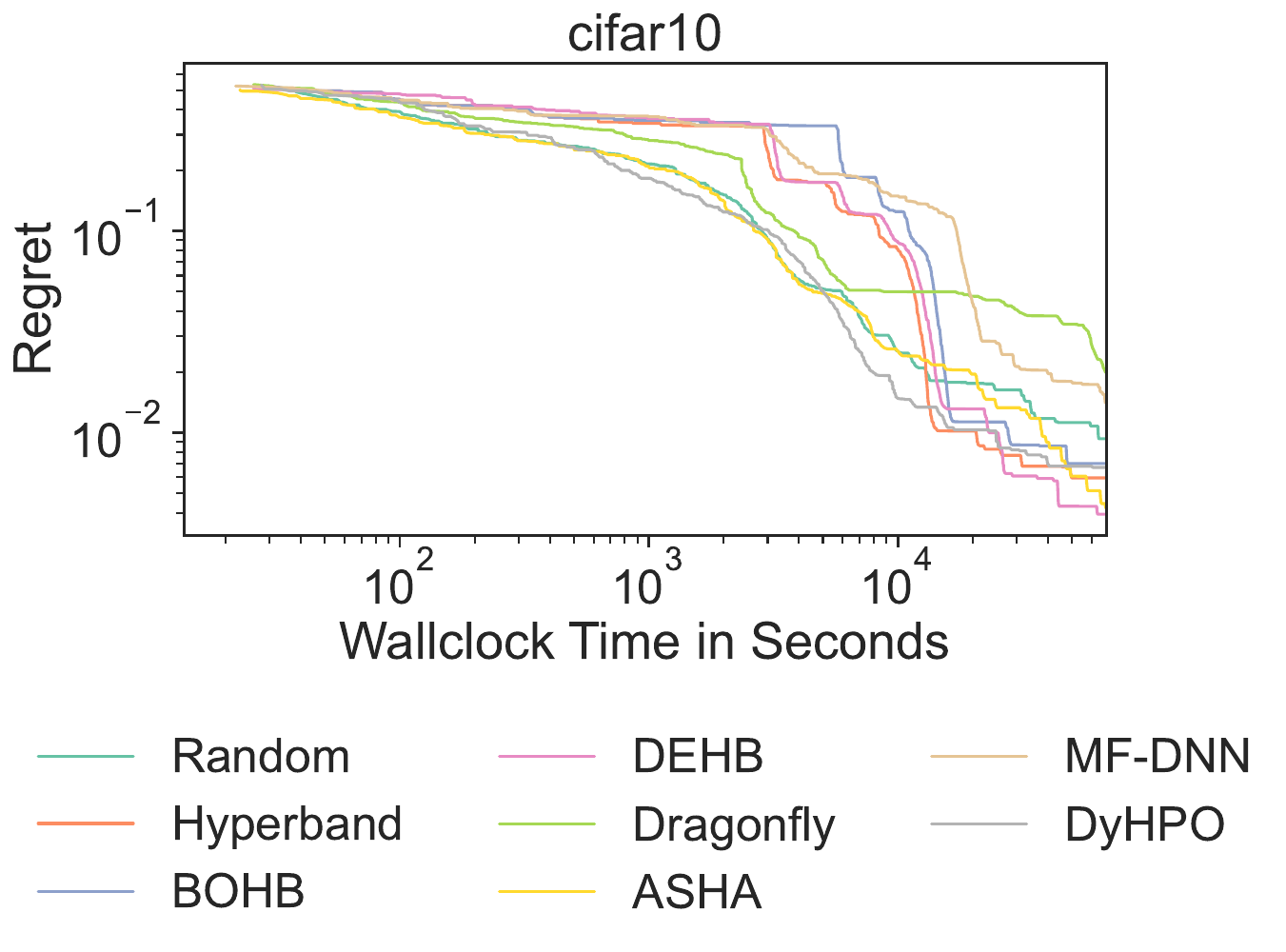}
    \includegraphics[width=0.48\textwidth]{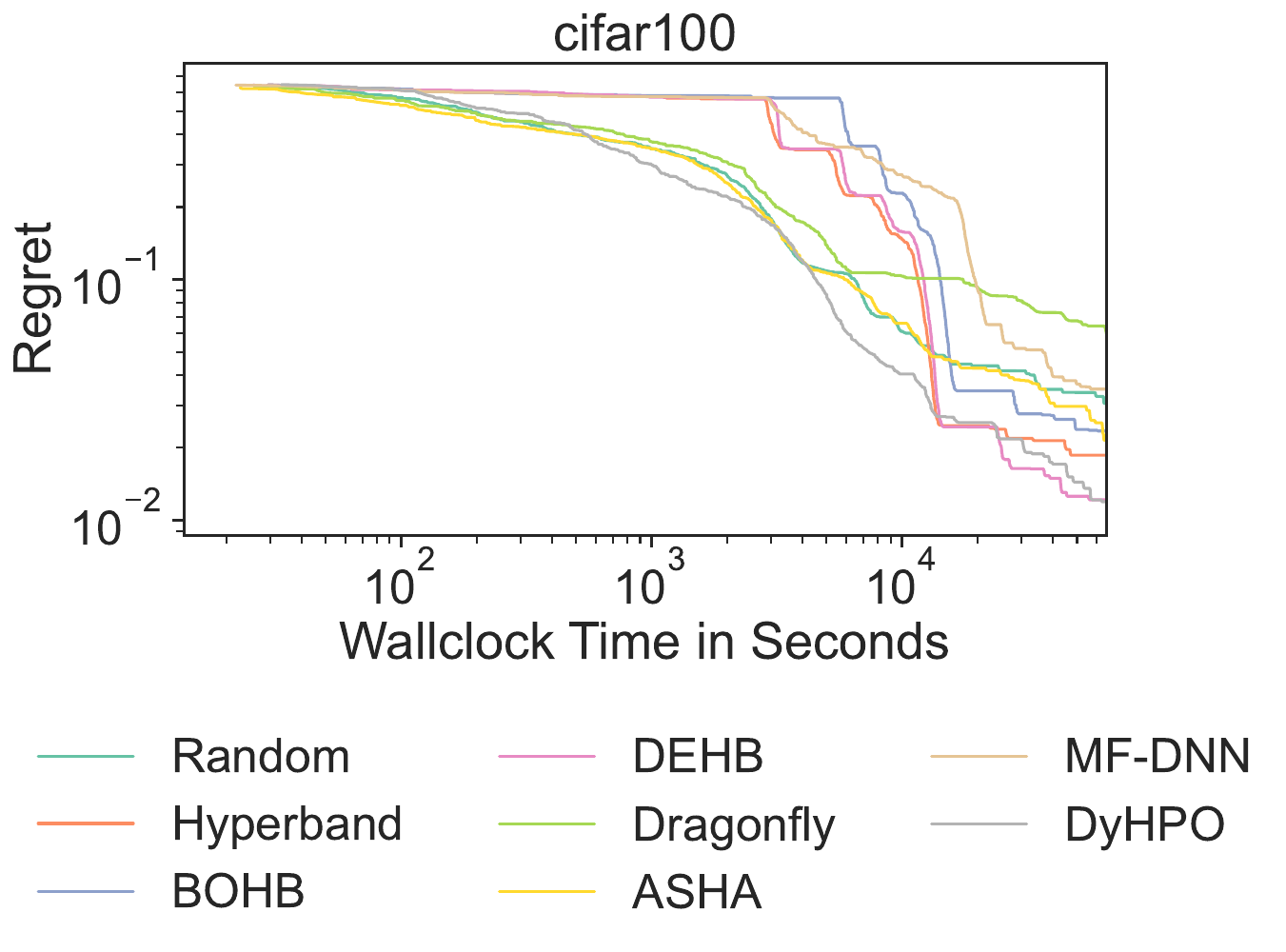}
  \caption{NAS-Bench-201 regret results over the total optimization time. The total time includes the method overhead time and the hyperparameter configuration evaluation time.}
  \label{fig:results_nasbench201_time_overhead}
\end{figure}

Additionally, in Figure~\ref{fig:results_nasbench201_time}, we show the performance comparison over the number of epochs of every method for the CIFAR-10 and CIFAR-100 datasets in the NAS-Bench-201 benchmark. While, in Figure~\ref{fig:results_nasbench201_time_overhead}, we present the performance comparison over time. As can be seen, \algabbr{} converges faster and has a better performance compared to the other methods over the majority of the time or steps, however, towards the end although it is the optimal method or close to the optimal method, the difference in regret is not significant anymore.

\begin{figure*}[t]
  \centering
    \includegraphics[width=0.32\textwidth]{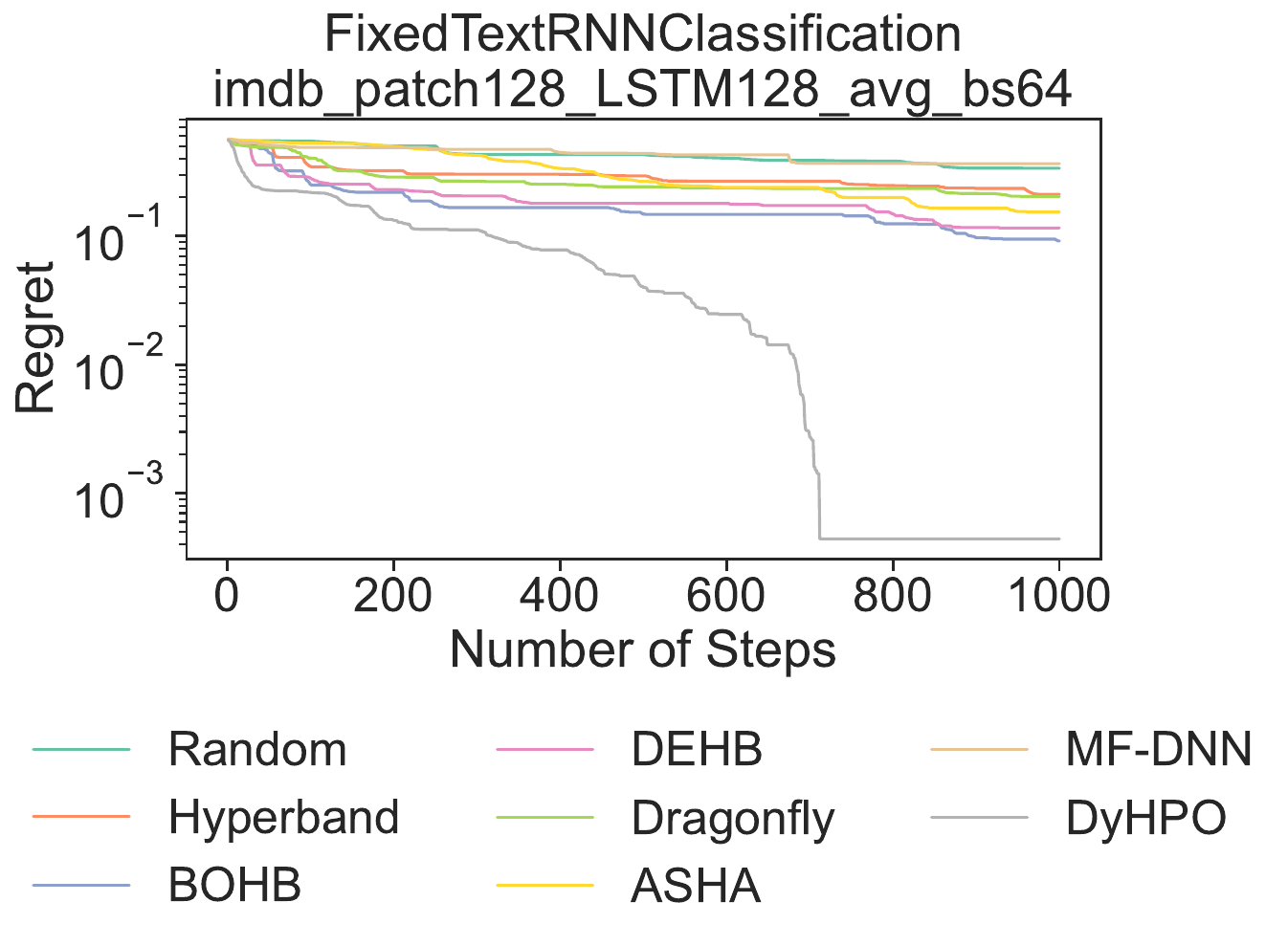}
    \includegraphics[width=0.32\textwidth]{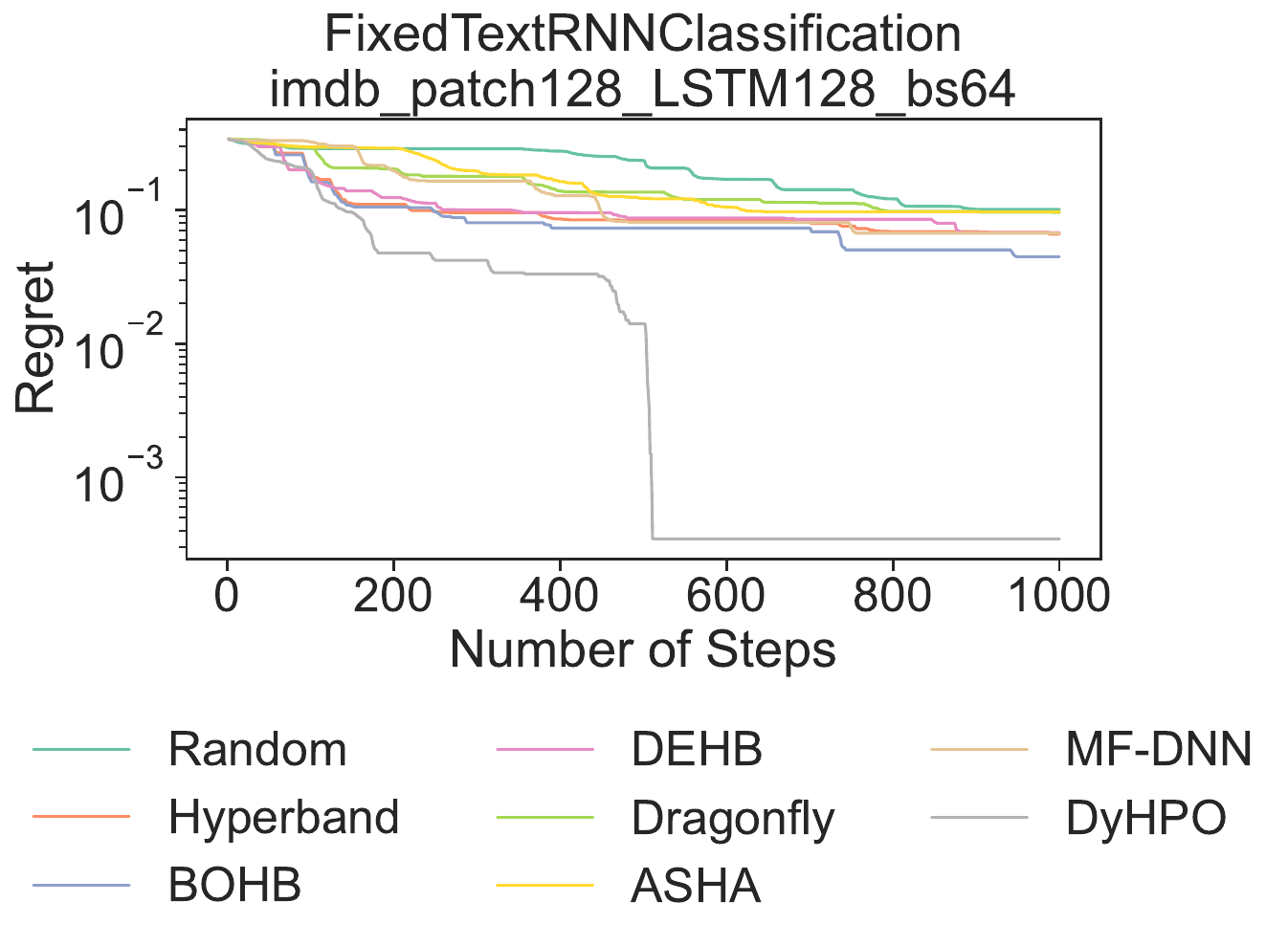}
    \includegraphics[width=0.32\textwidth]{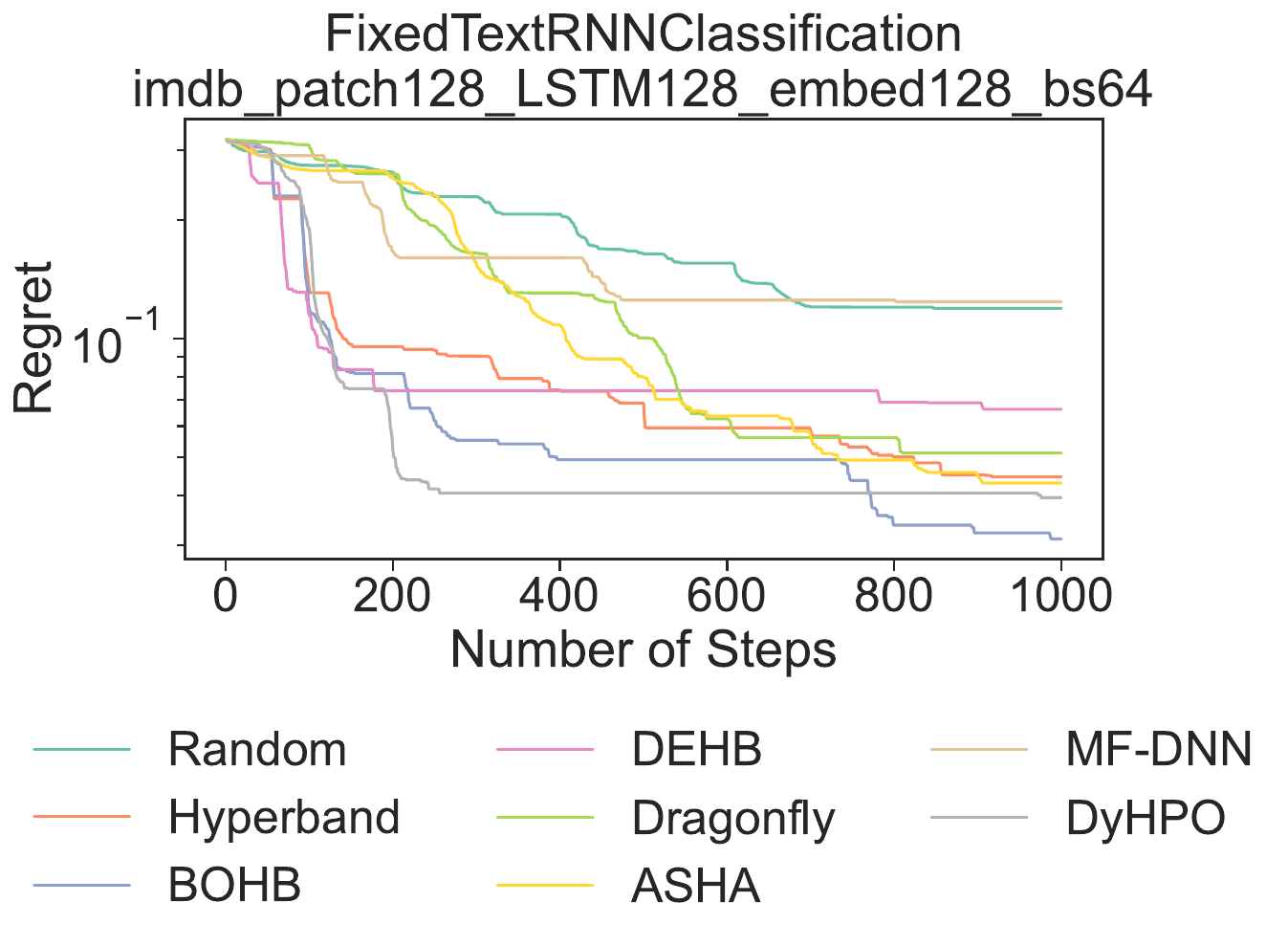}
    \includegraphics[width=0.32\textwidth]{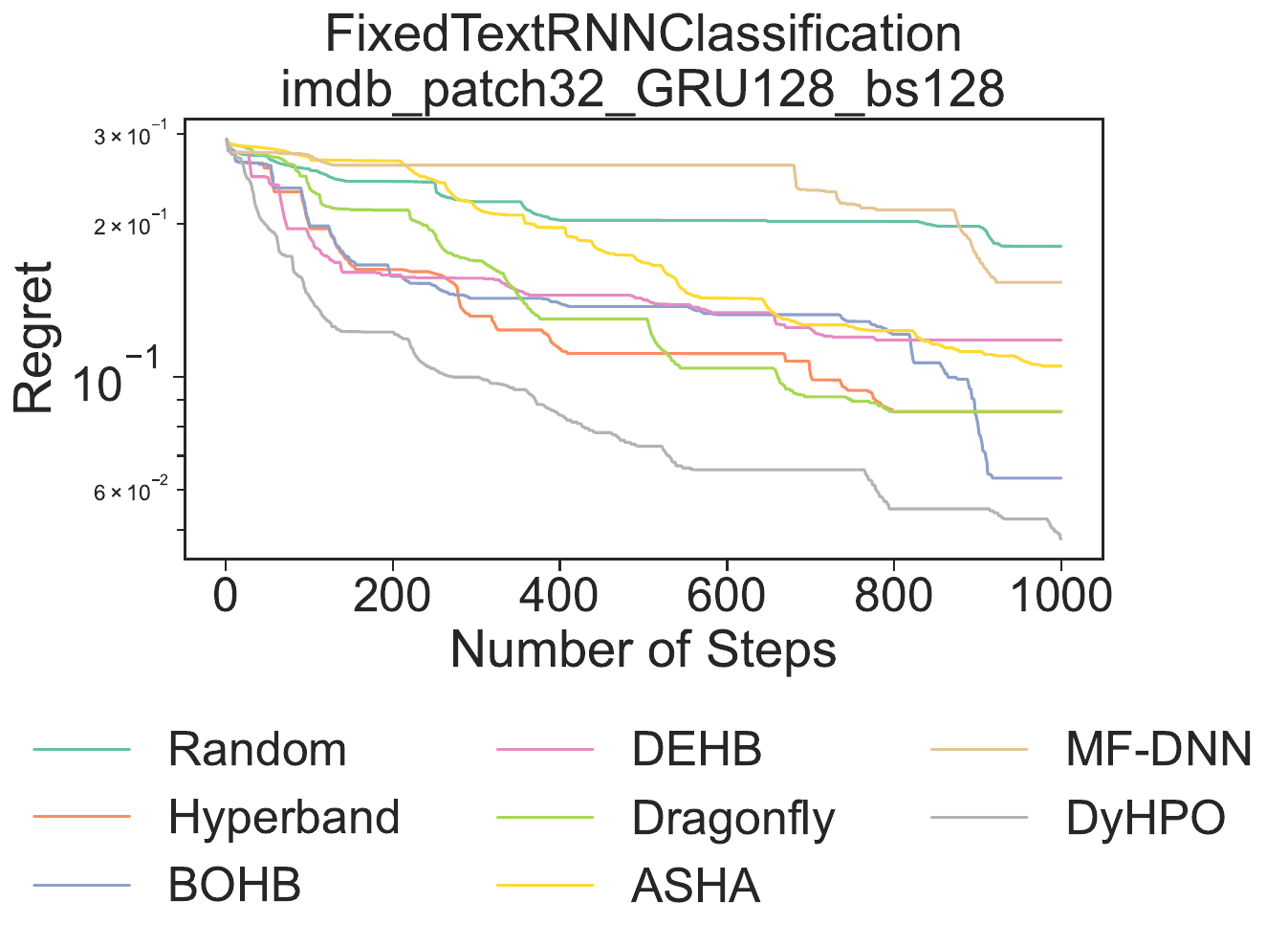}
    \includegraphics[width=0.32\textwidth]{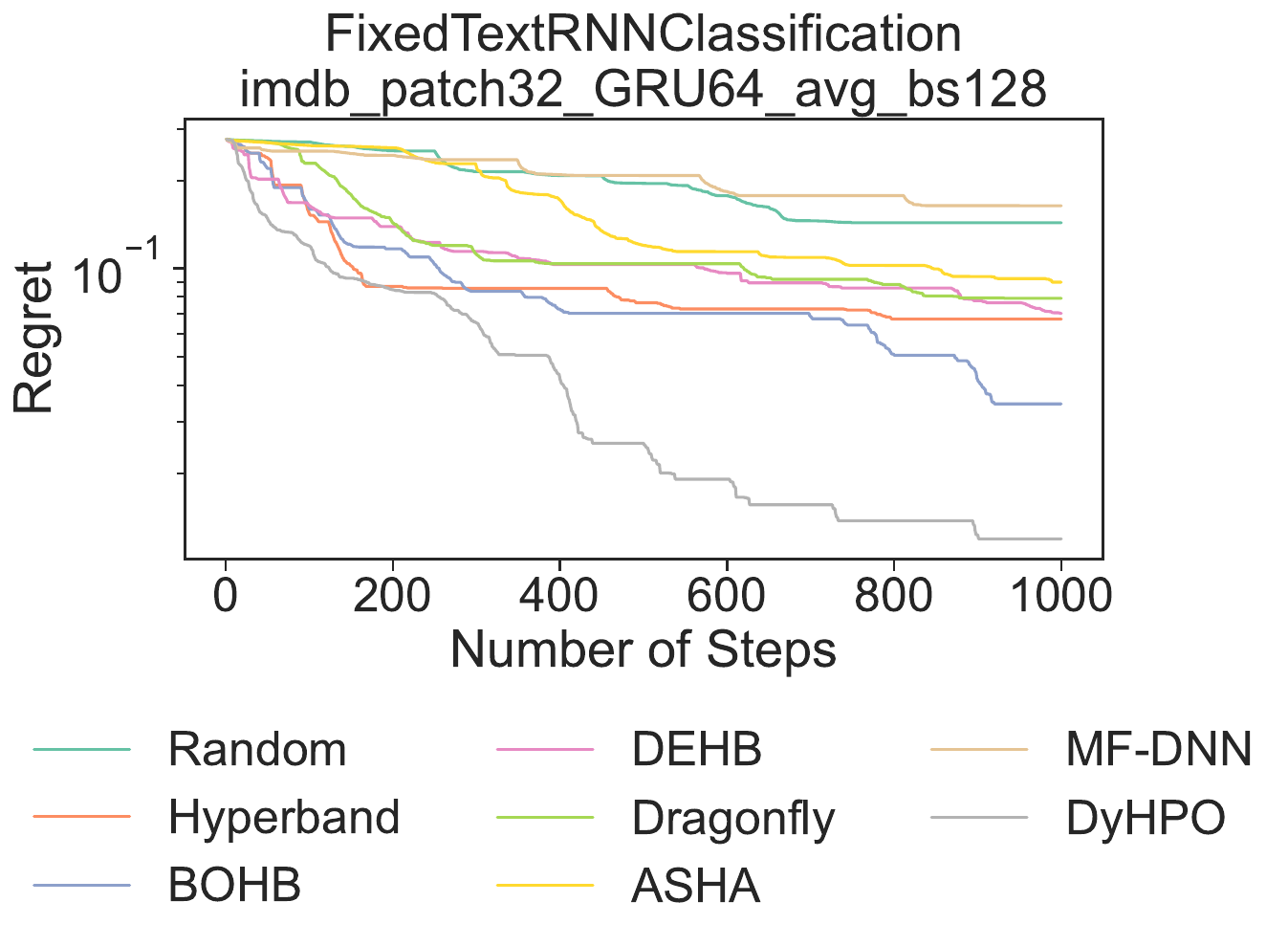}
    \includegraphics[width=0.32\textwidth]{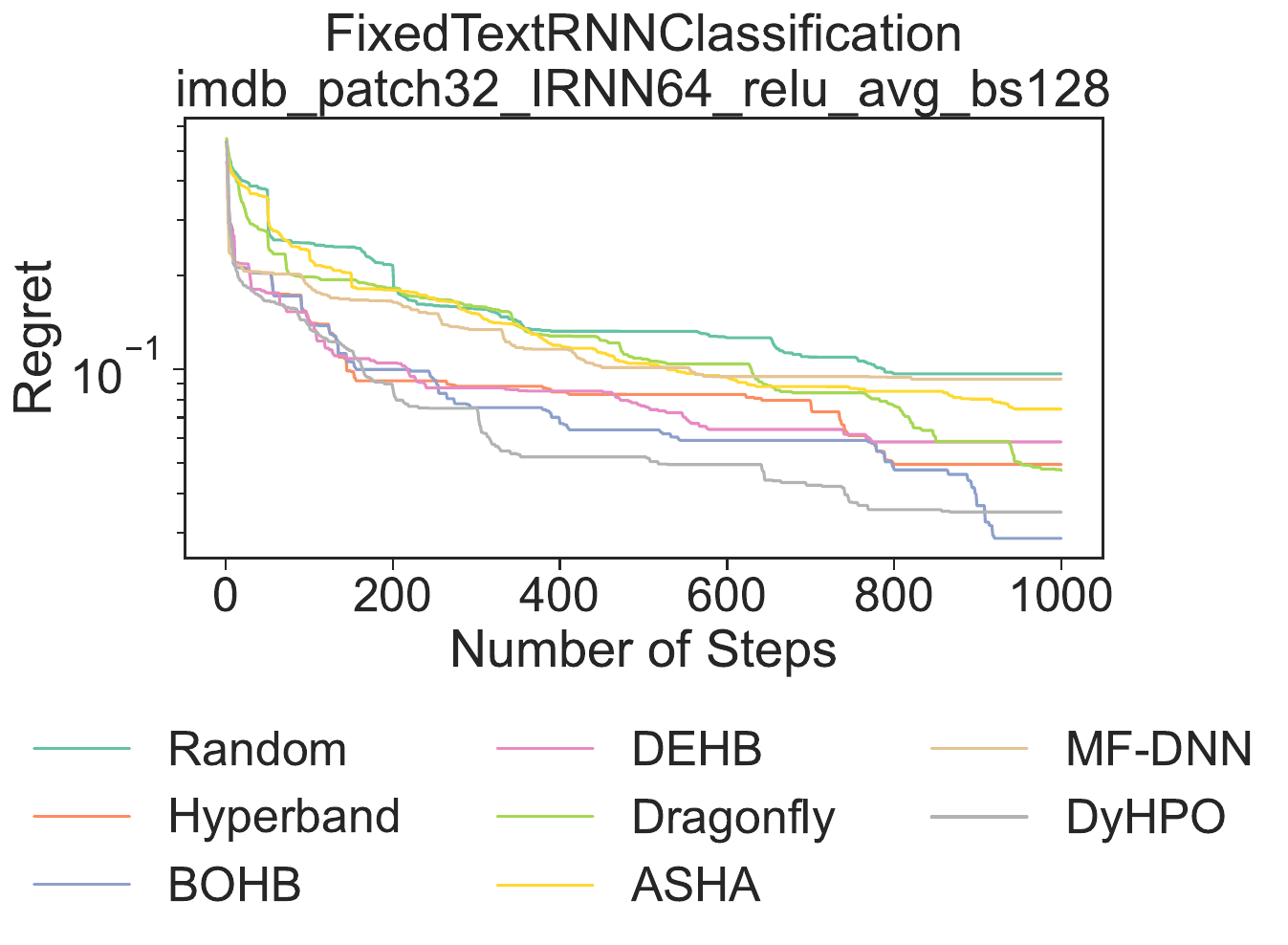}
    \includegraphics[width=0.32\textwidth]{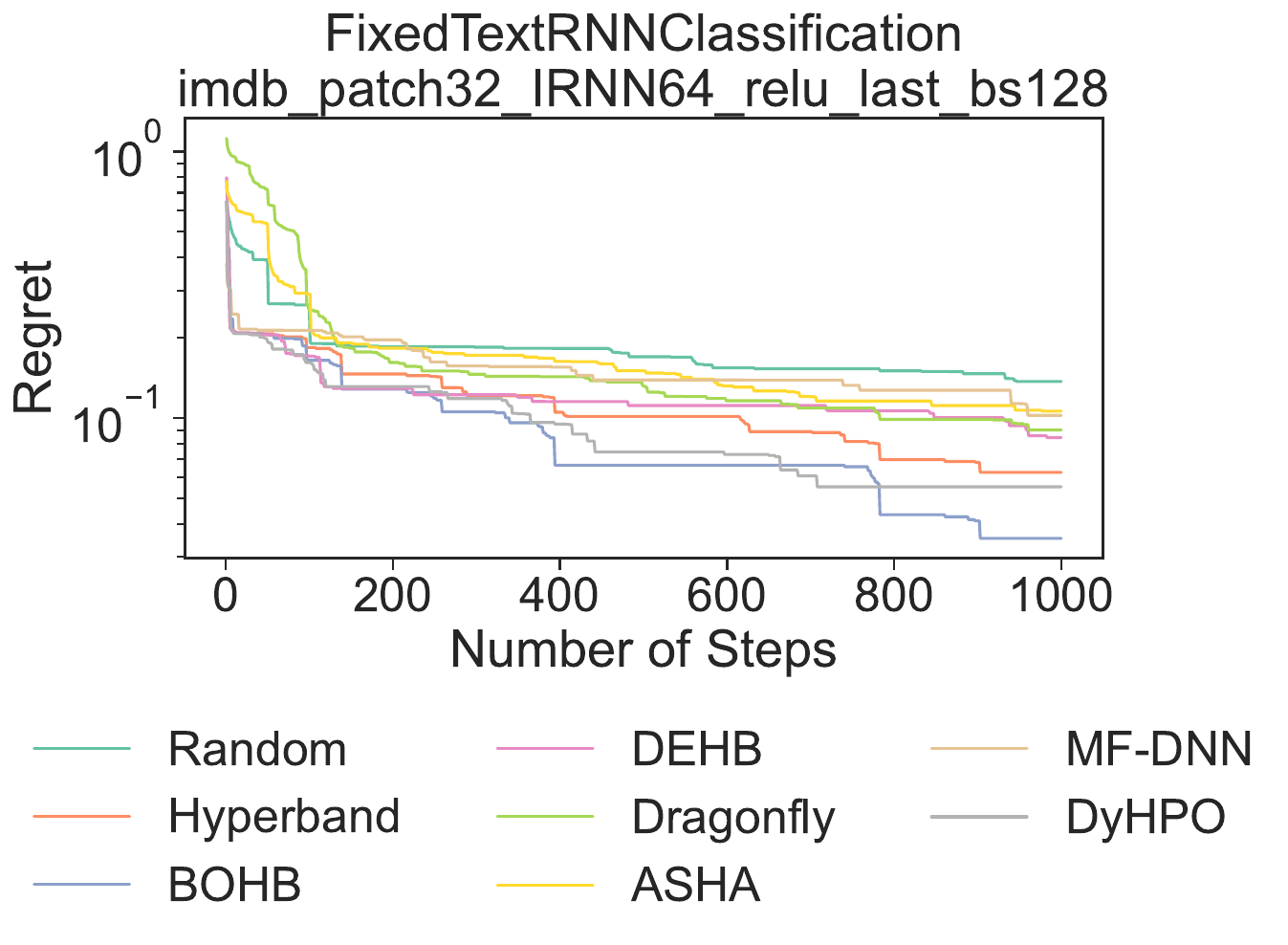}
    \includegraphics[width=0.32\textwidth]{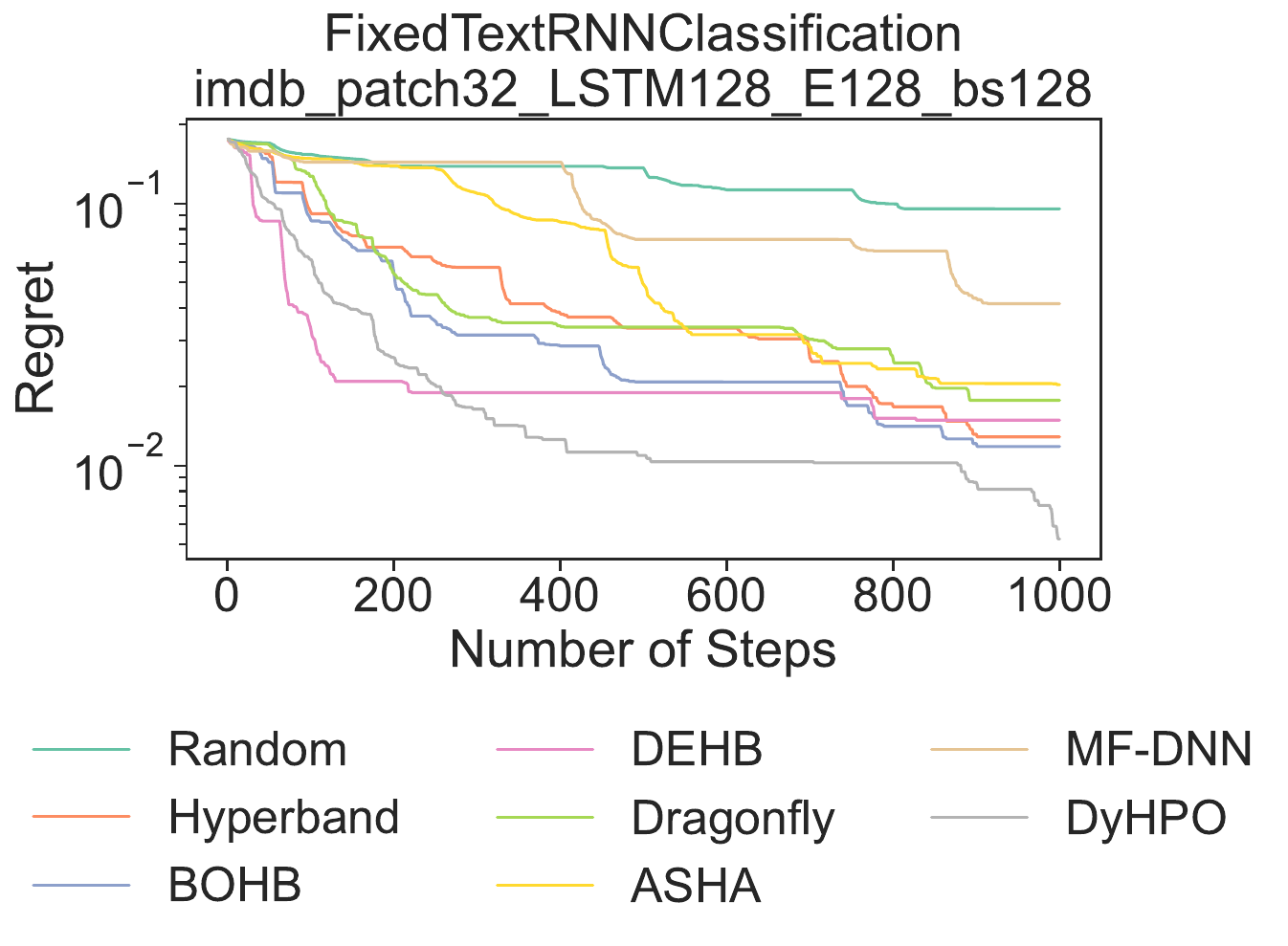}
    \includegraphics[width=0.32\textwidth]{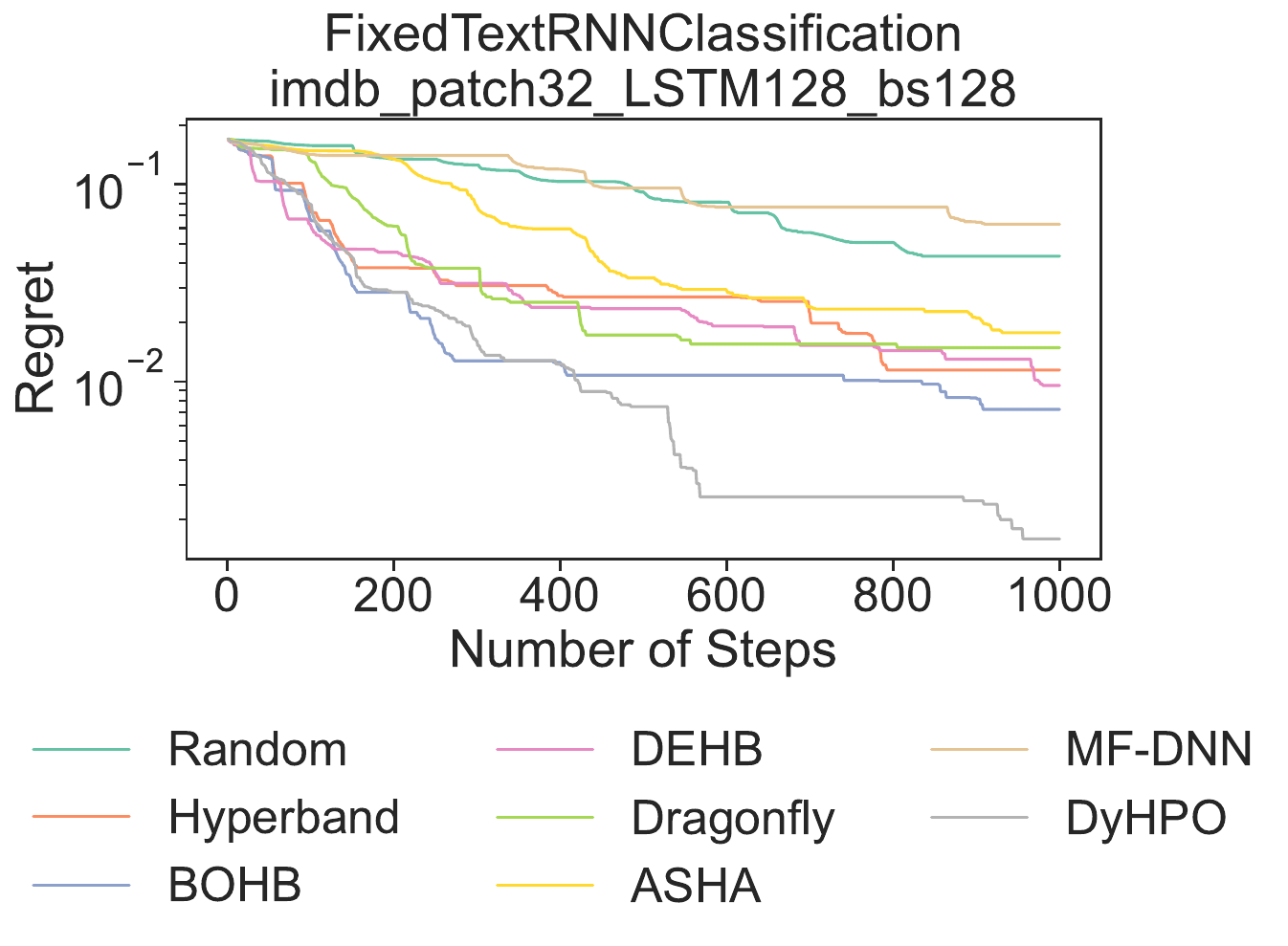}
    \includegraphics[width=0.32\textwidth]{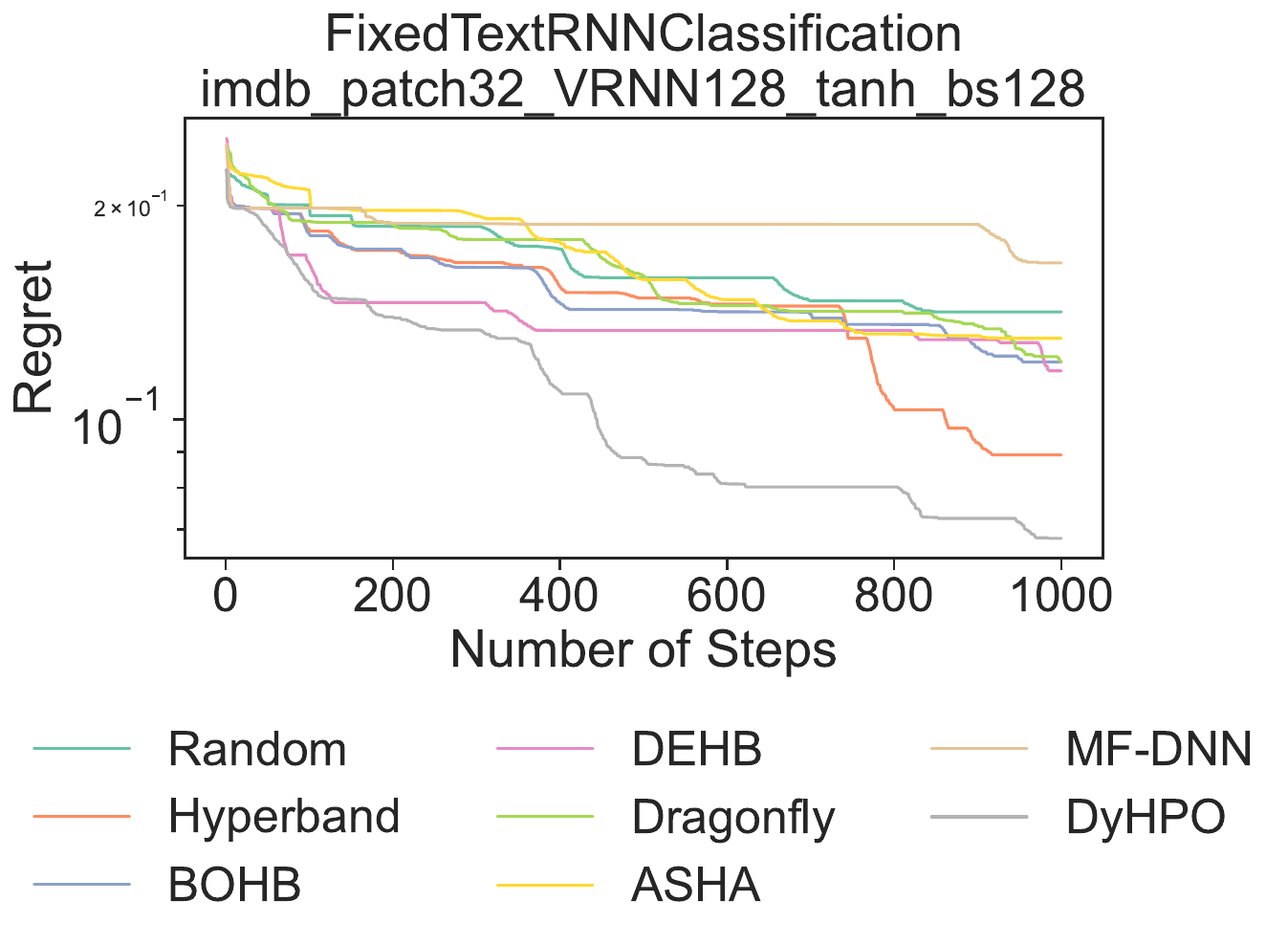}
    \includegraphics[width=0.32\textwidth]{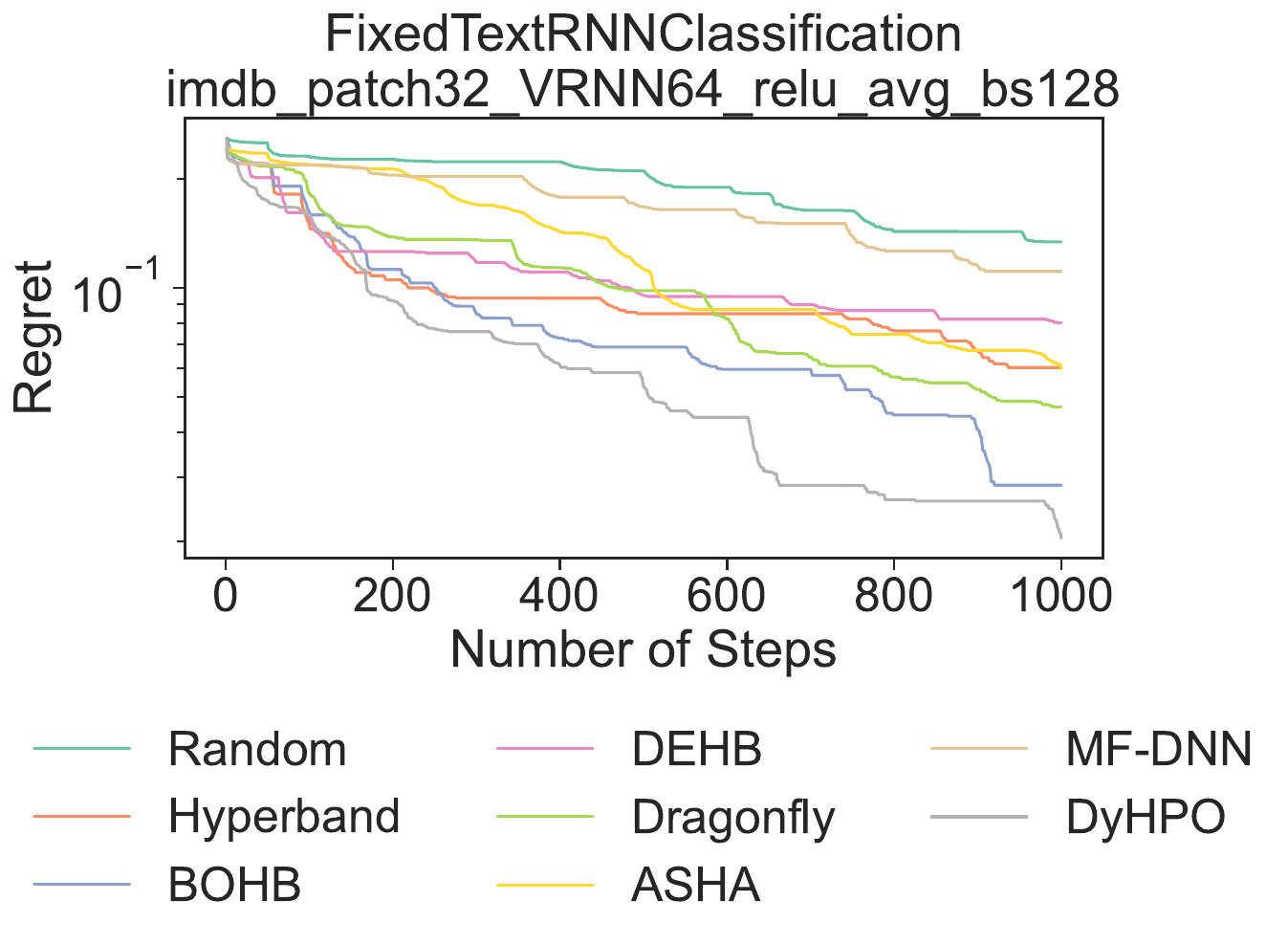}
    \includegraphics[width=0.32\textwidth]{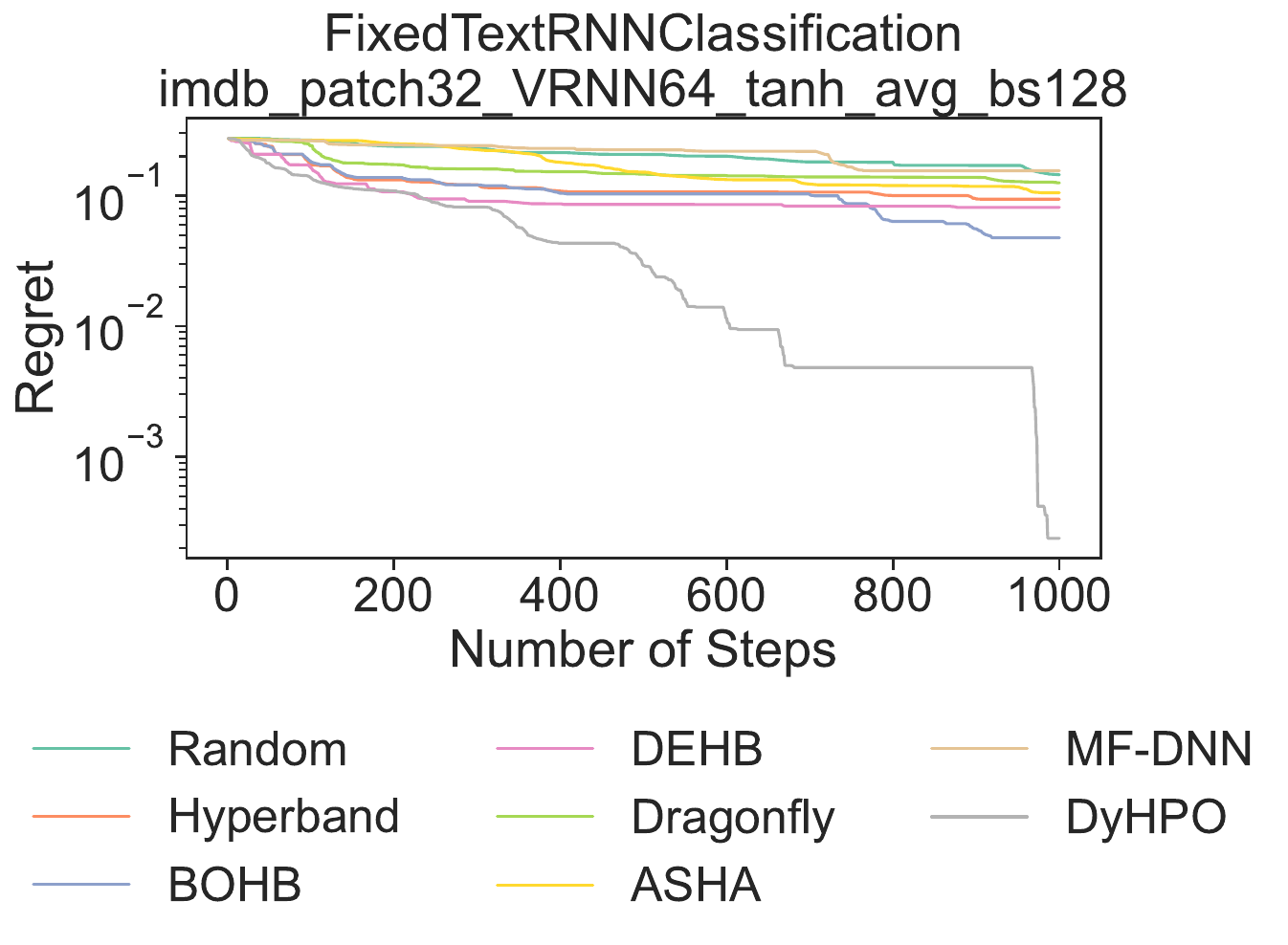}

  \caption{Performance comparison over the number of steps on a dataset level for TaskSet.}
  \label{fig:results_per_dataset_taskset}
\end{figure*}

Furthermore, Figure~\ref{fig:results_per_dataset_taskset} shows the performance comparison for the datasets chosen from TaskSet over the number of steps. Looking at the results, \algabbr{} is outperforming all methods convincingly on the majority of datasets by converging faster and with significant differences in the regret evaluation metric.

\begin{figure*}[htp]
  \centering
    \includegraphics[width=0.26\textwidth]{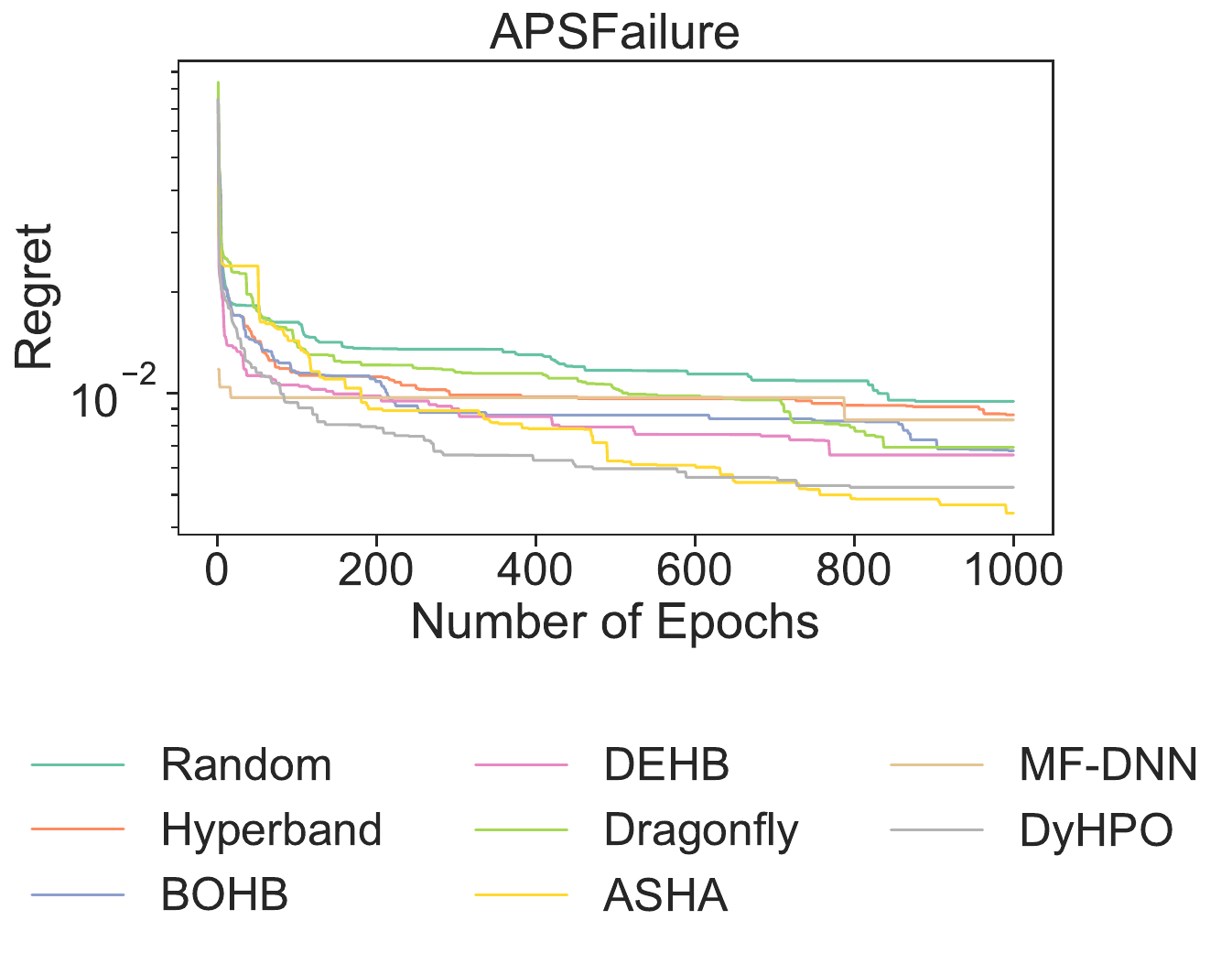}
    \includegraphics[width=0.26\textwidth]{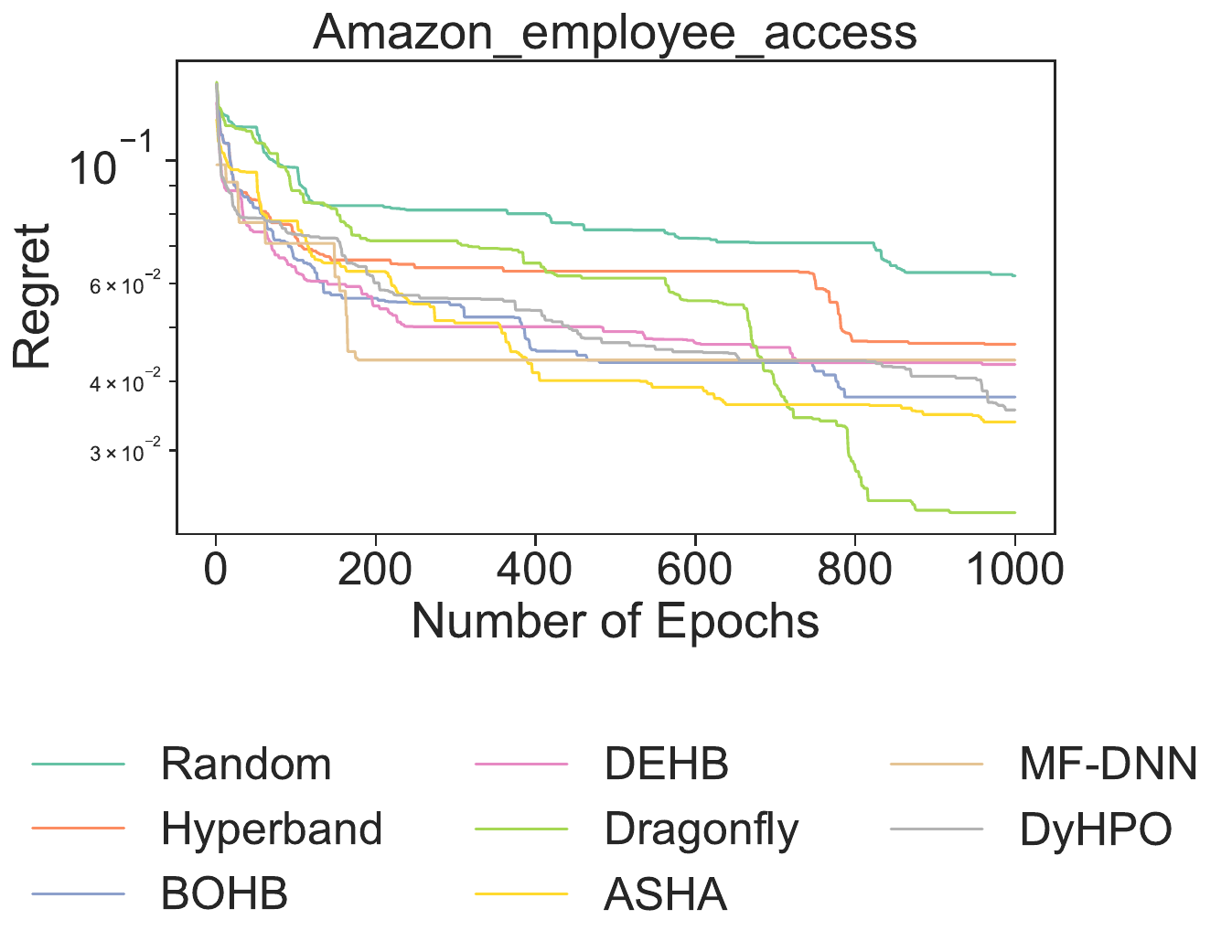}
    \includegraphics[width=0.26\textwidth]{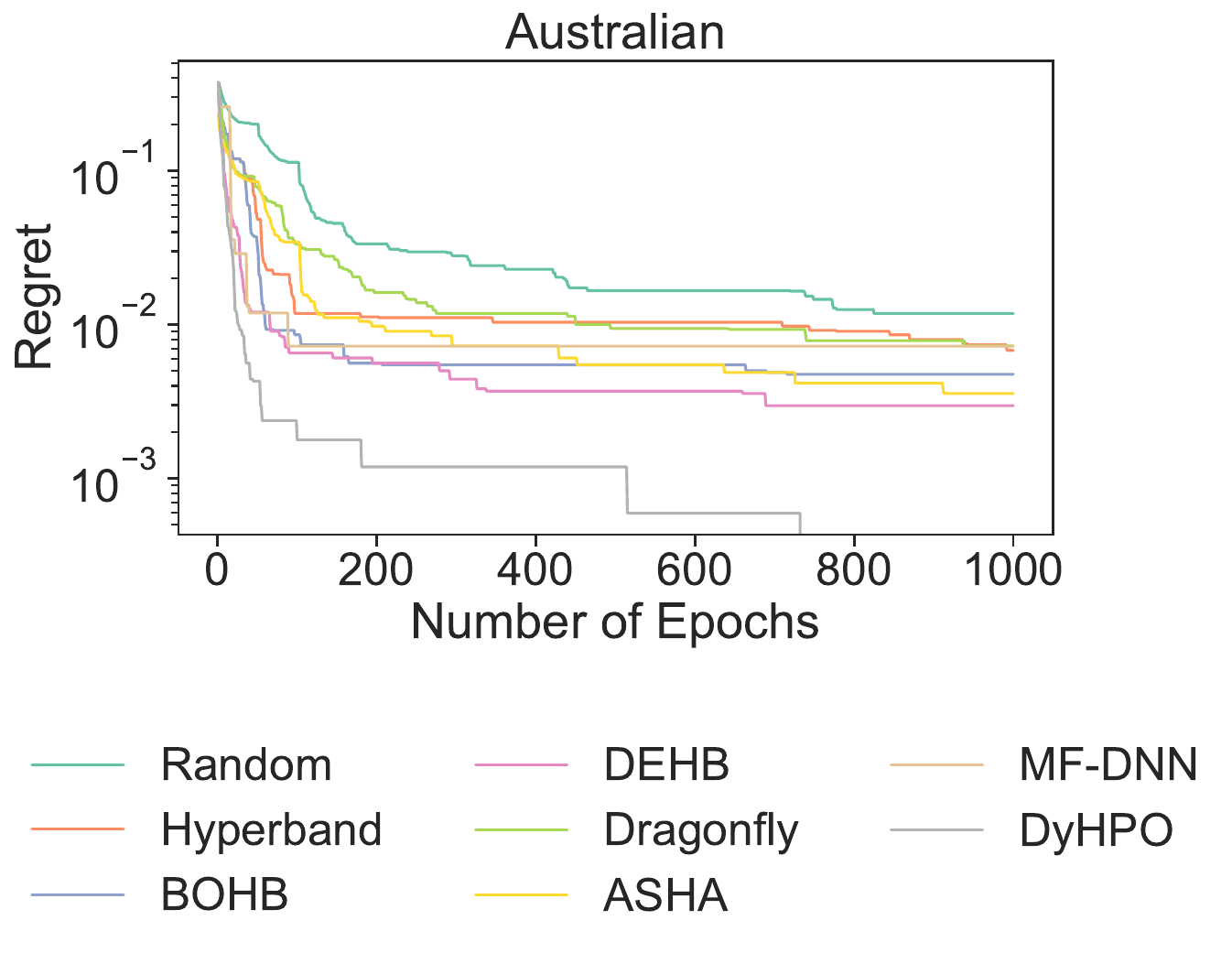}
    \includegraphics[width=0.26\textwidth]{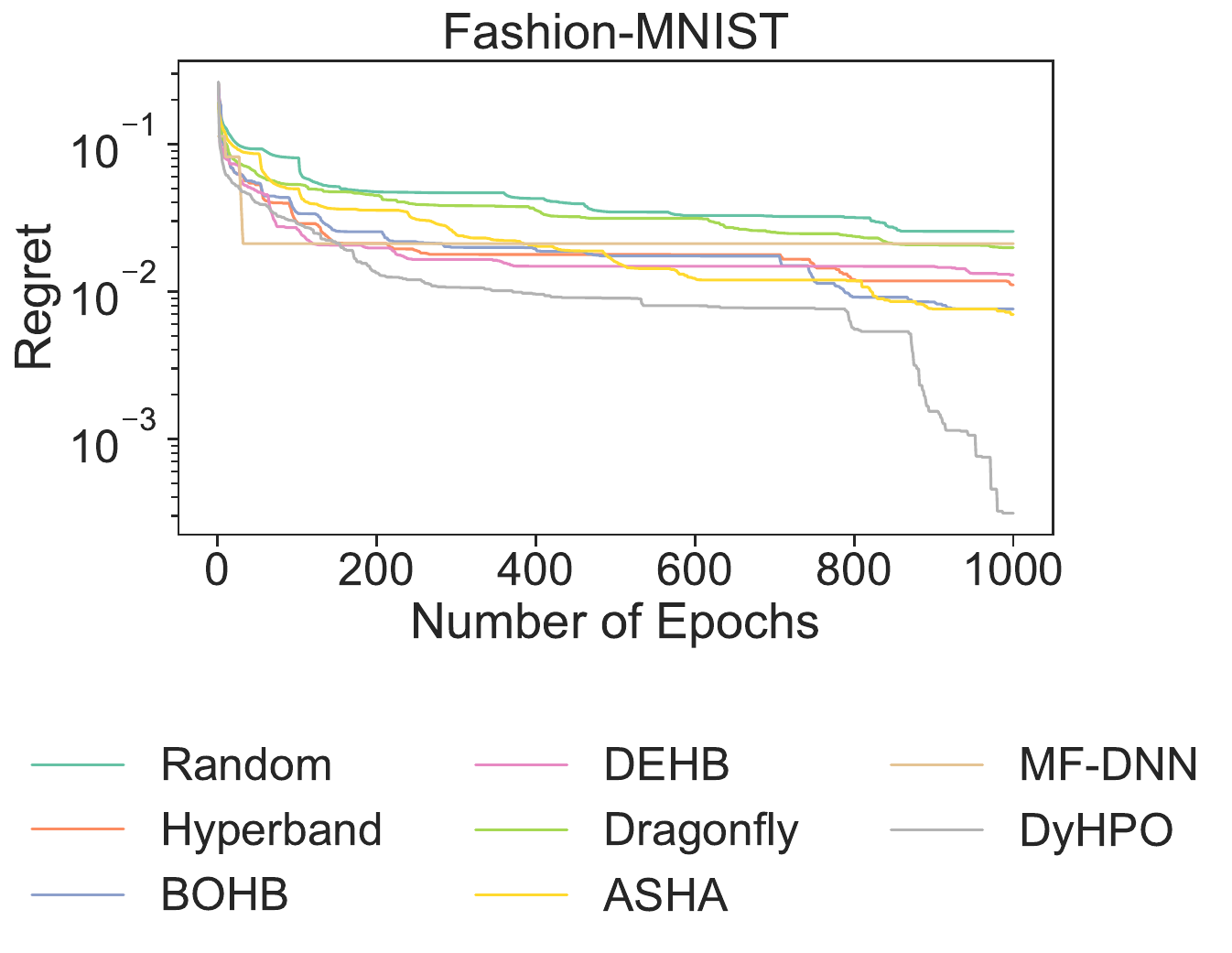}
    \includegraphics[width=0.26\textwidth]{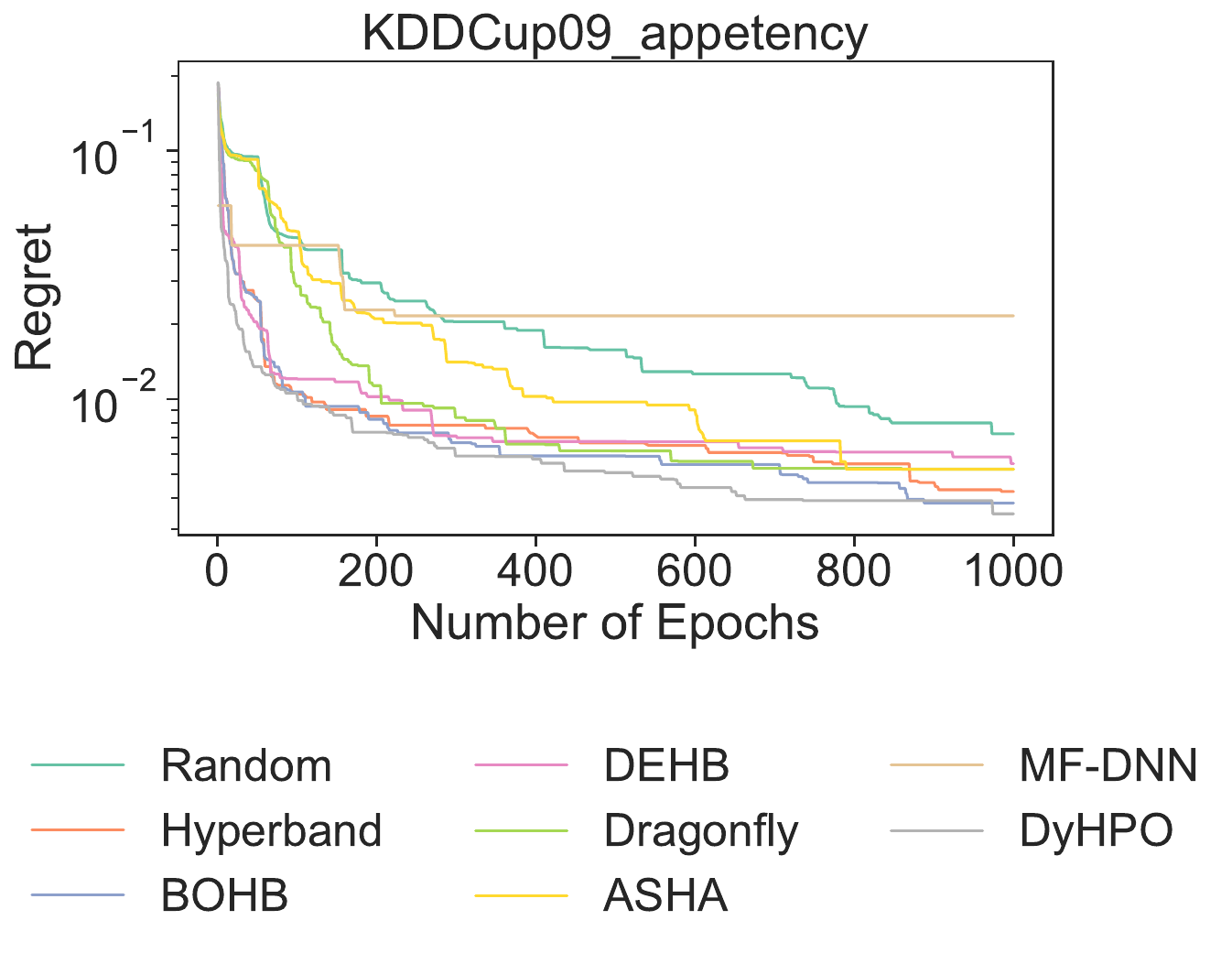}
    \includegraphics[width=0.26\textwidth]{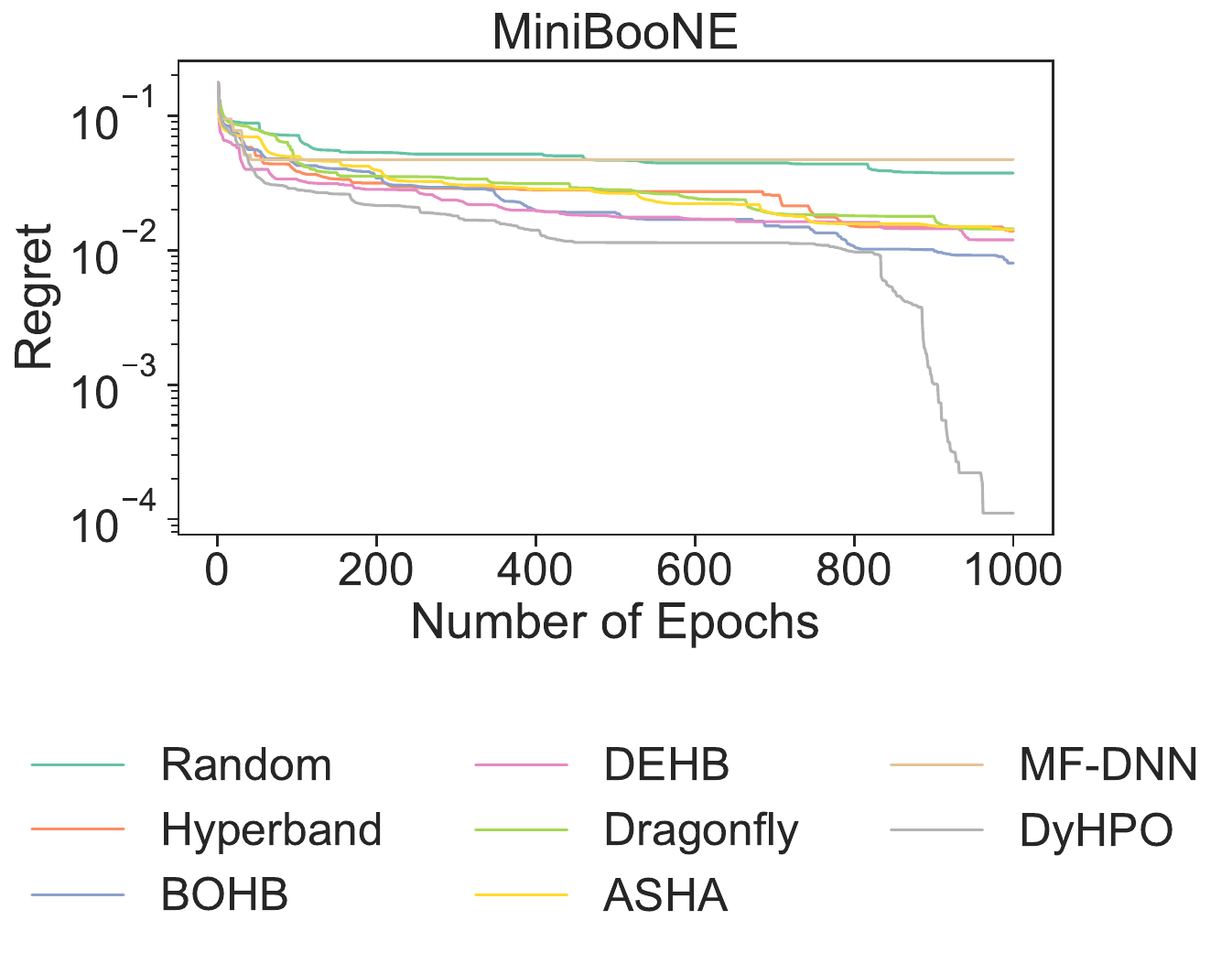}
    \includegraphics[width=0.26\textwidth]{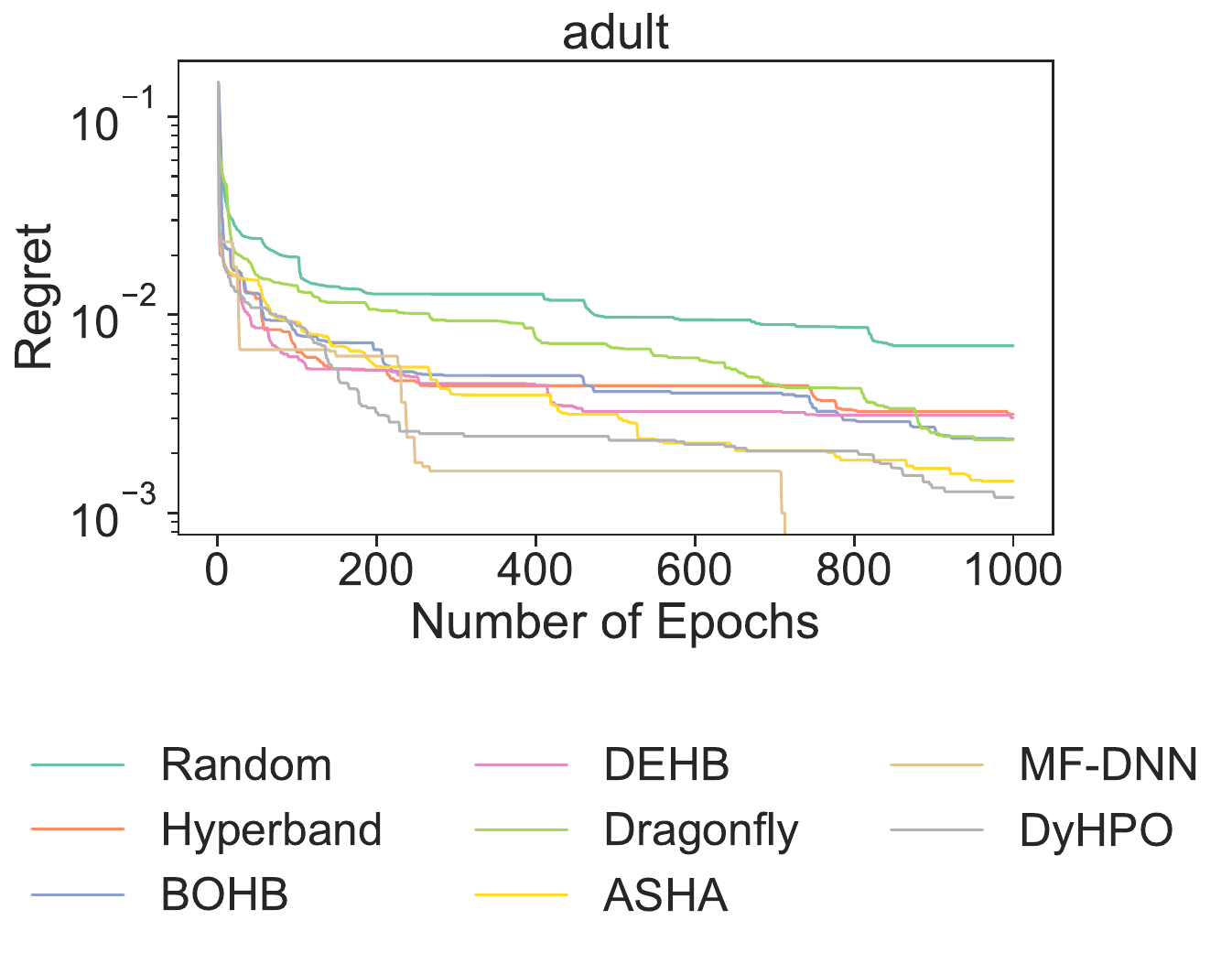}
    \includegraphics[width=0.26\textwidth]{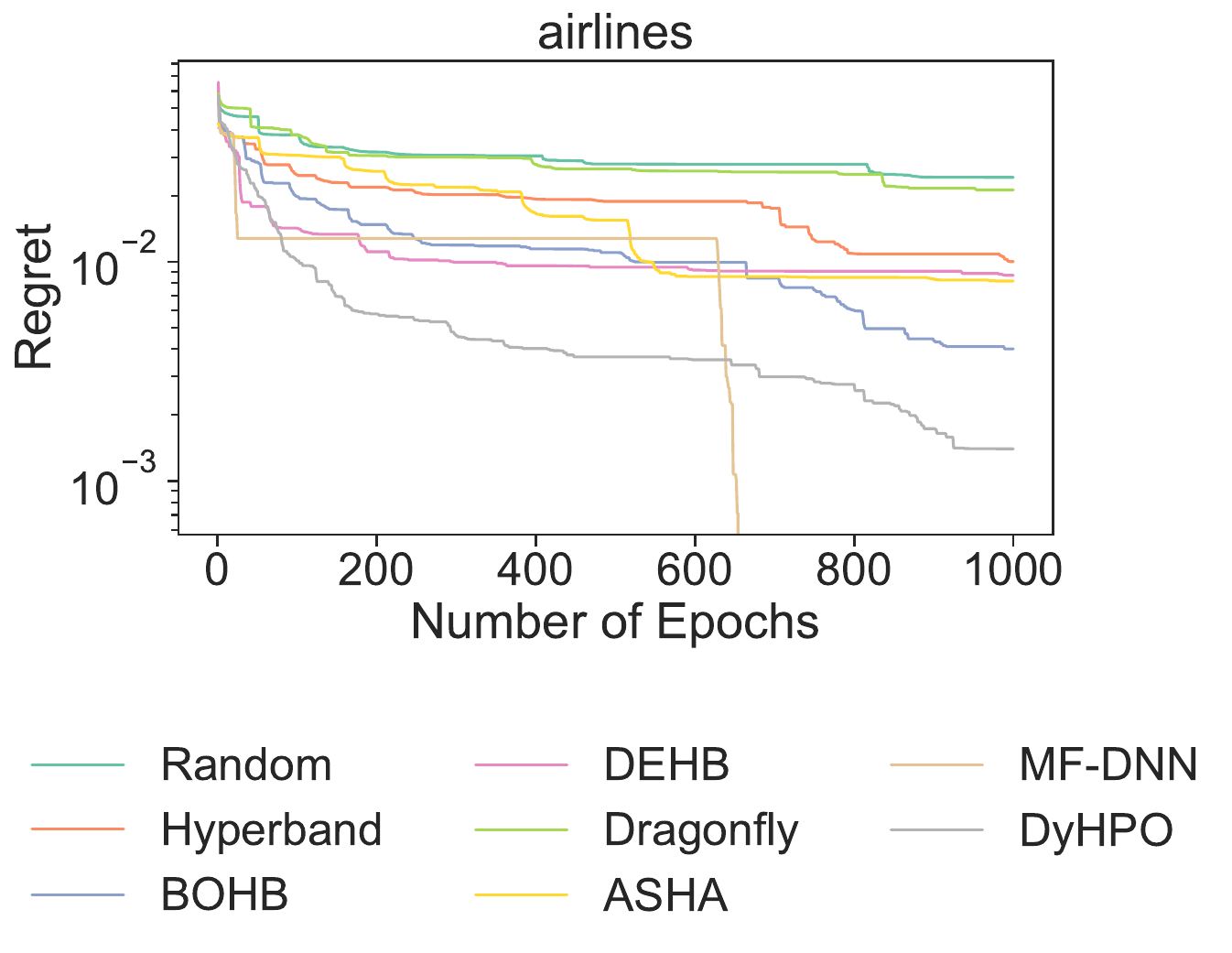}
    \includegraphics[width=0.26\textwidth]{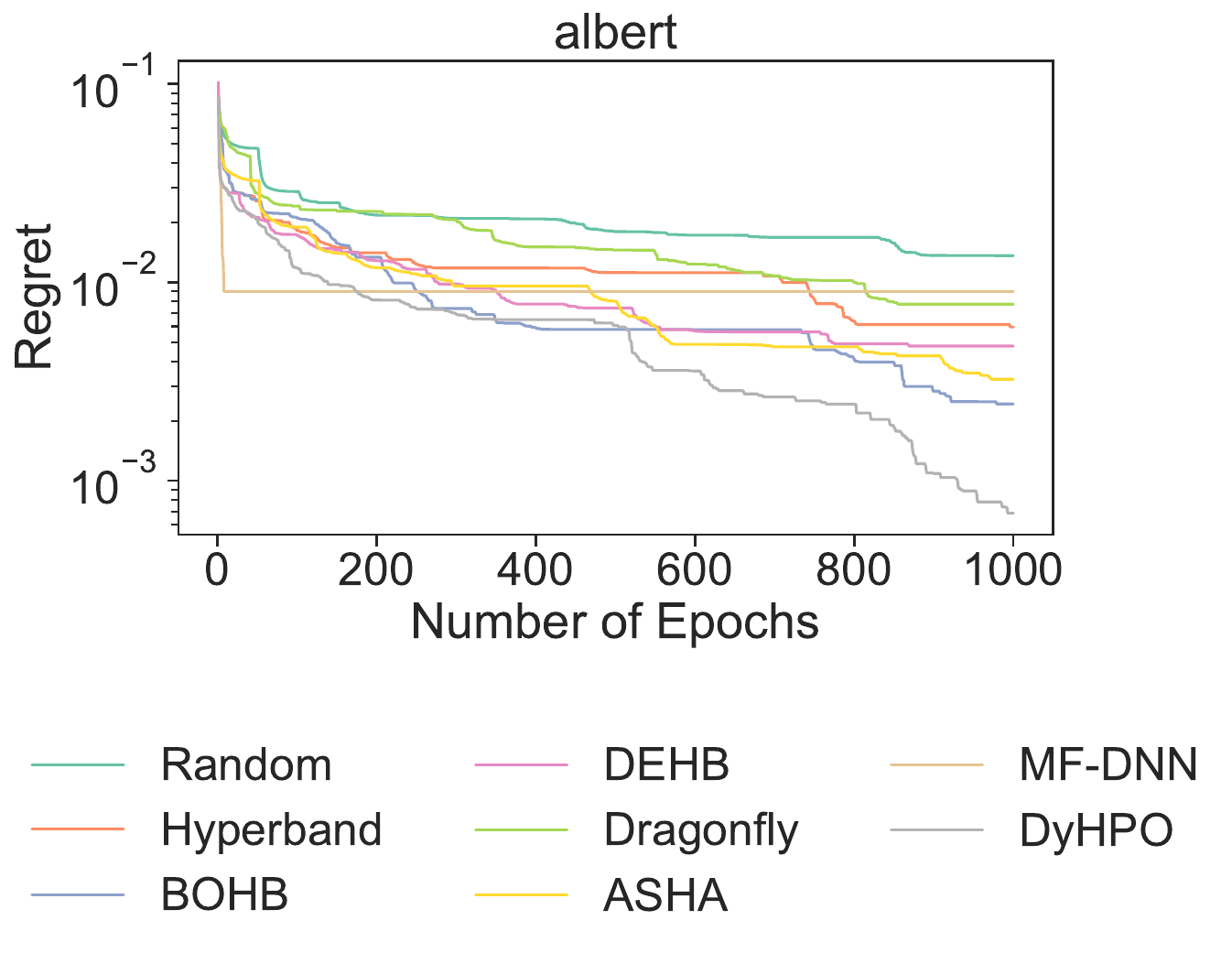}
    \includegraphics[width=0.26\textwidth]{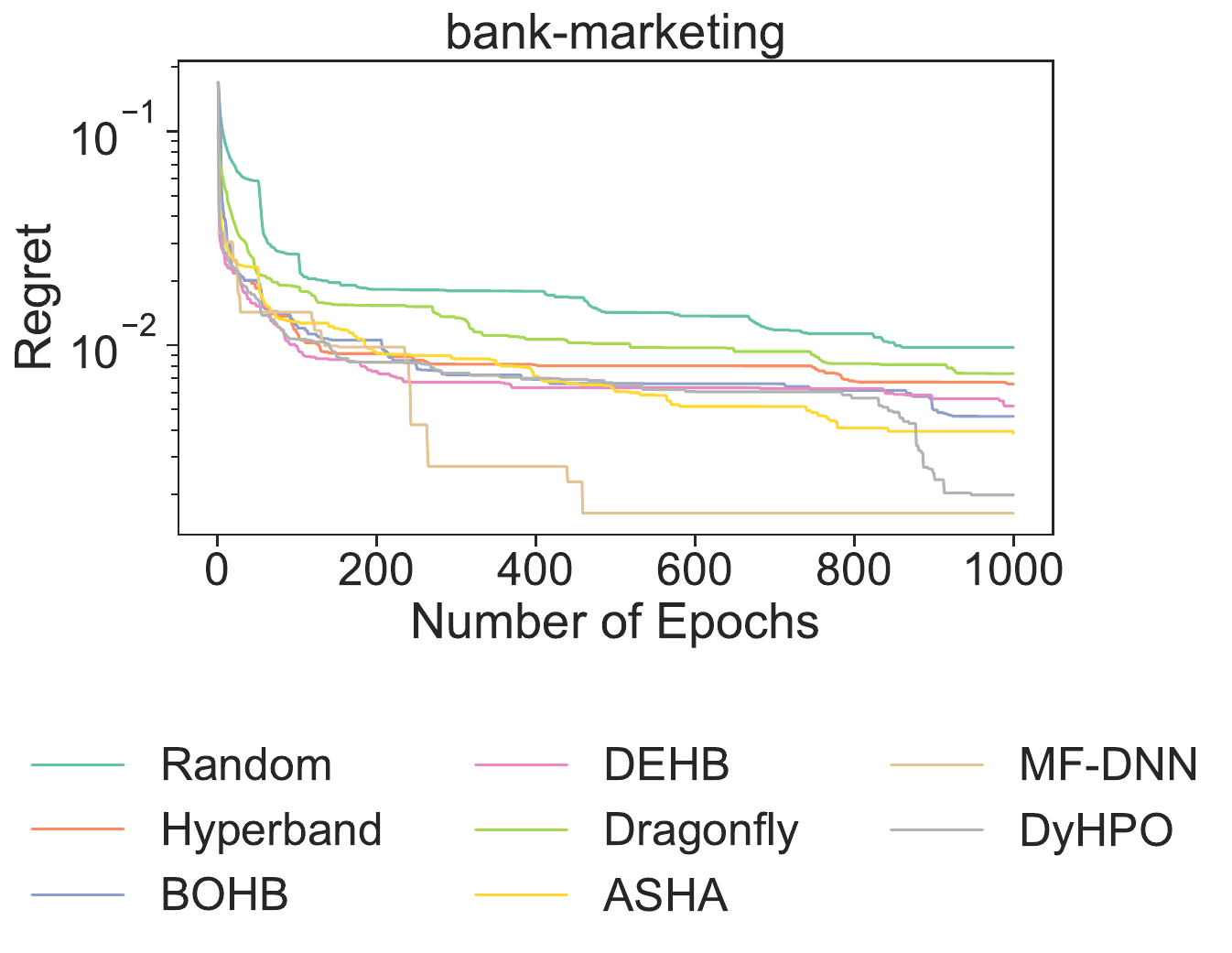}
    \includegraphics[width=0.26\textwidth]{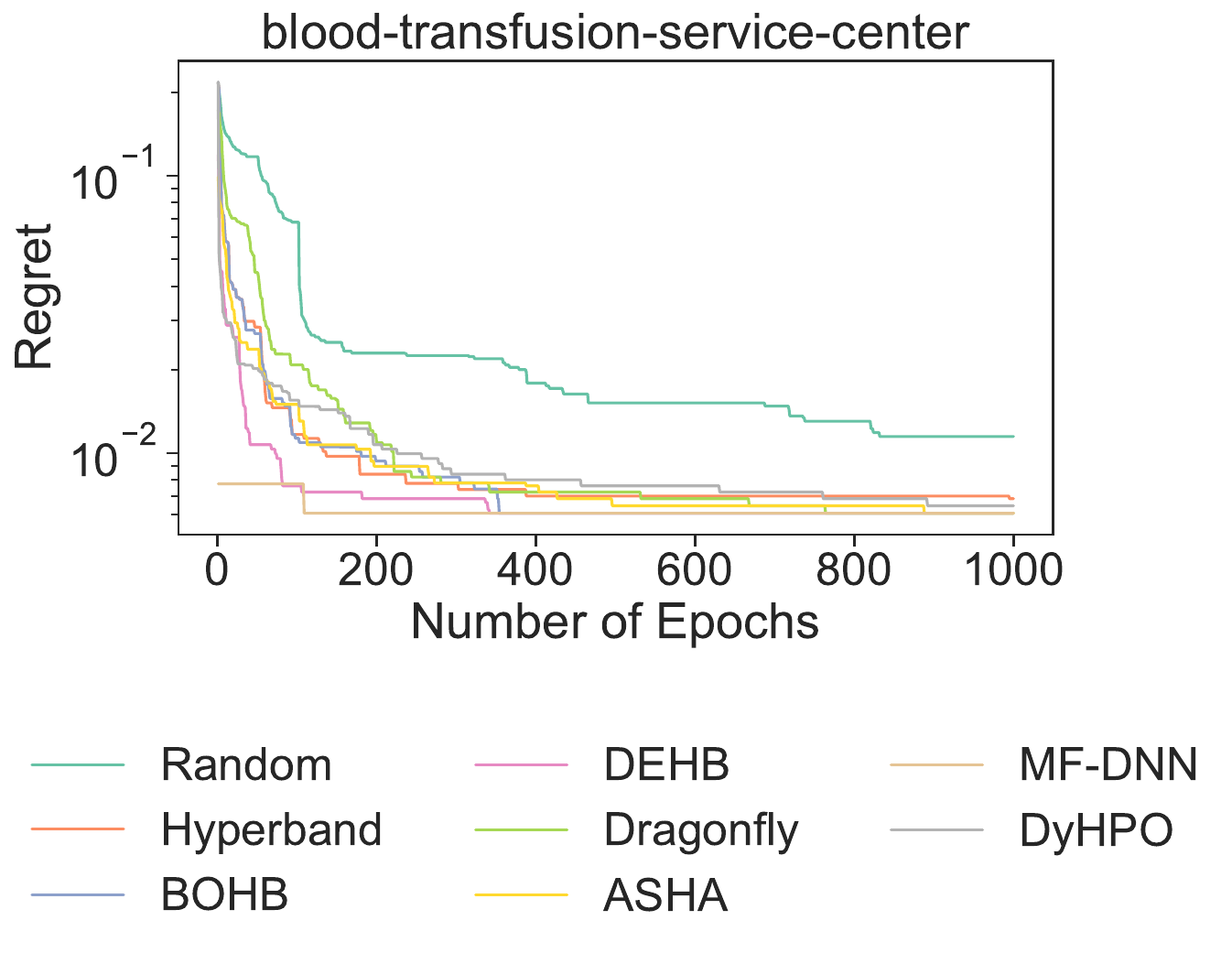}
    \includegraphics[width=0.26\textwidth]{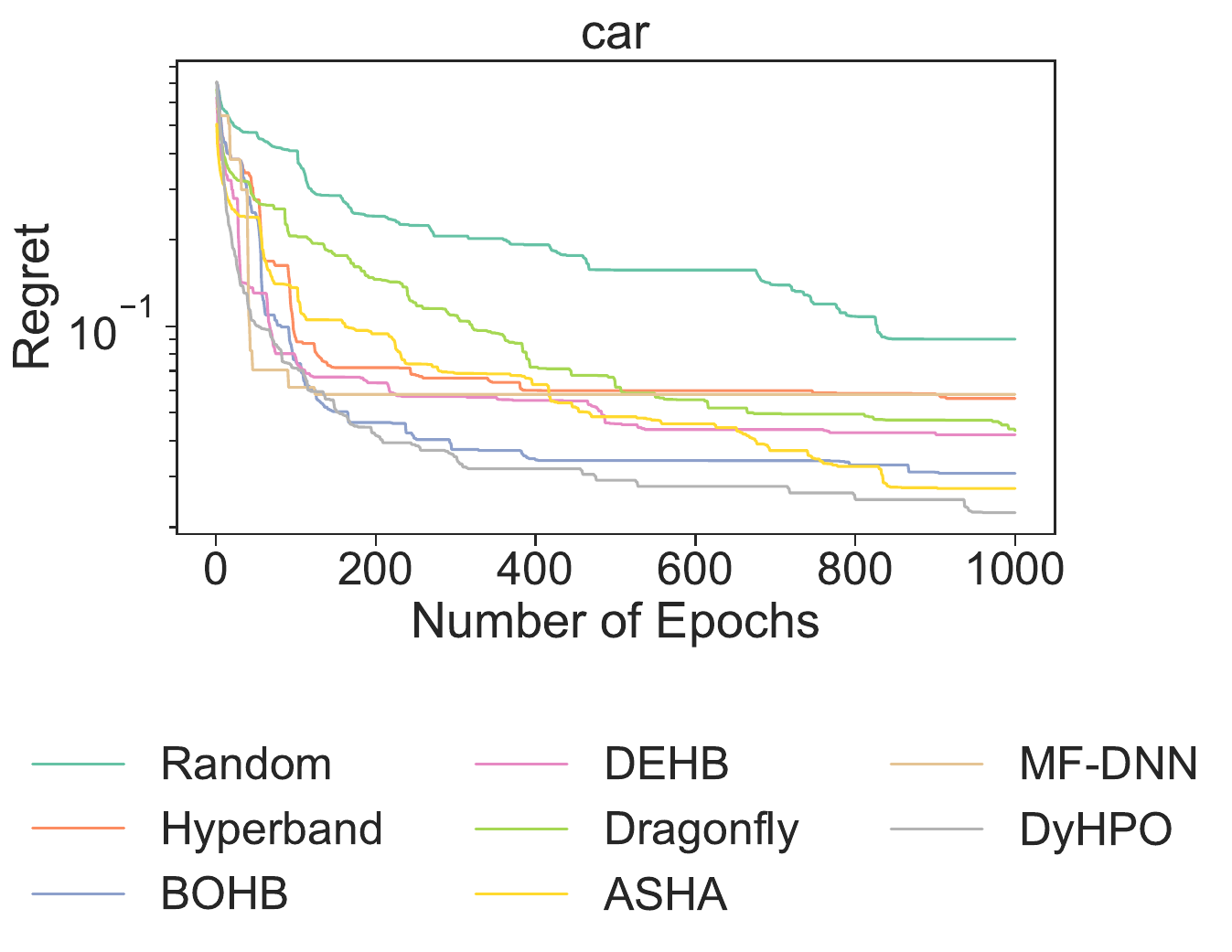}
    \includegraphics[width=0.26\textwidth]{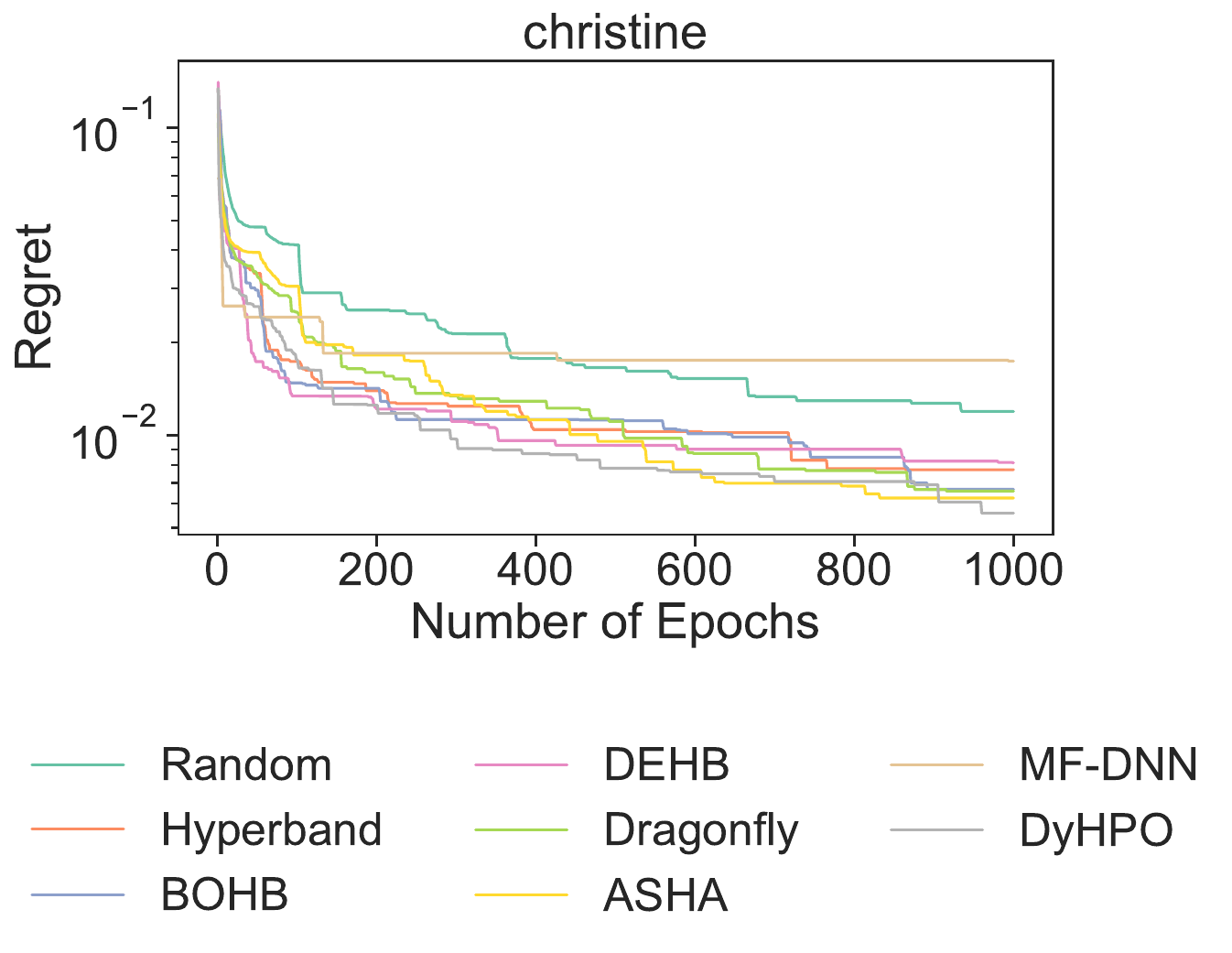}
    \includegraphics[width=0.26\textwidth]{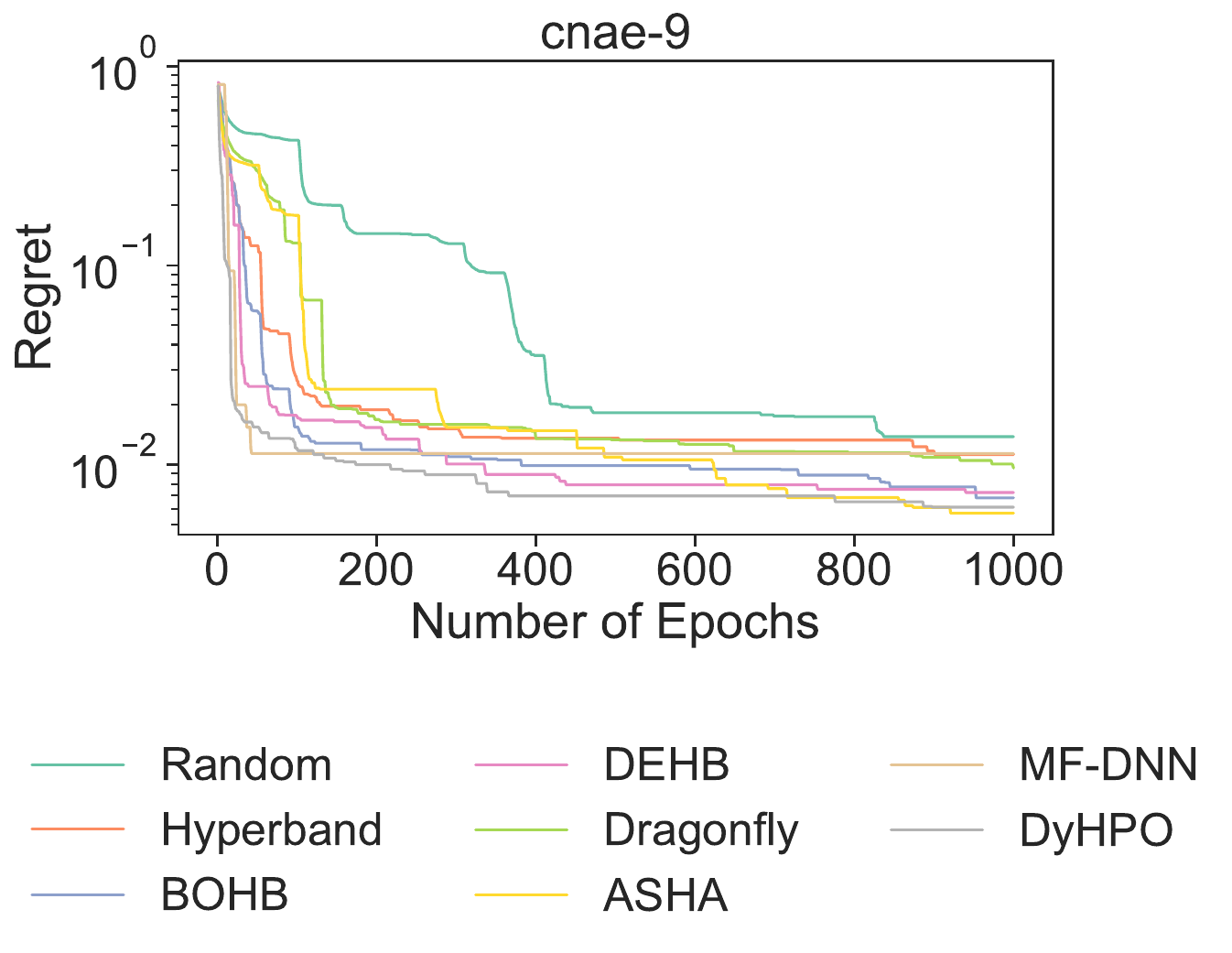}
    \includegraphics[width=0.26\textwidth]{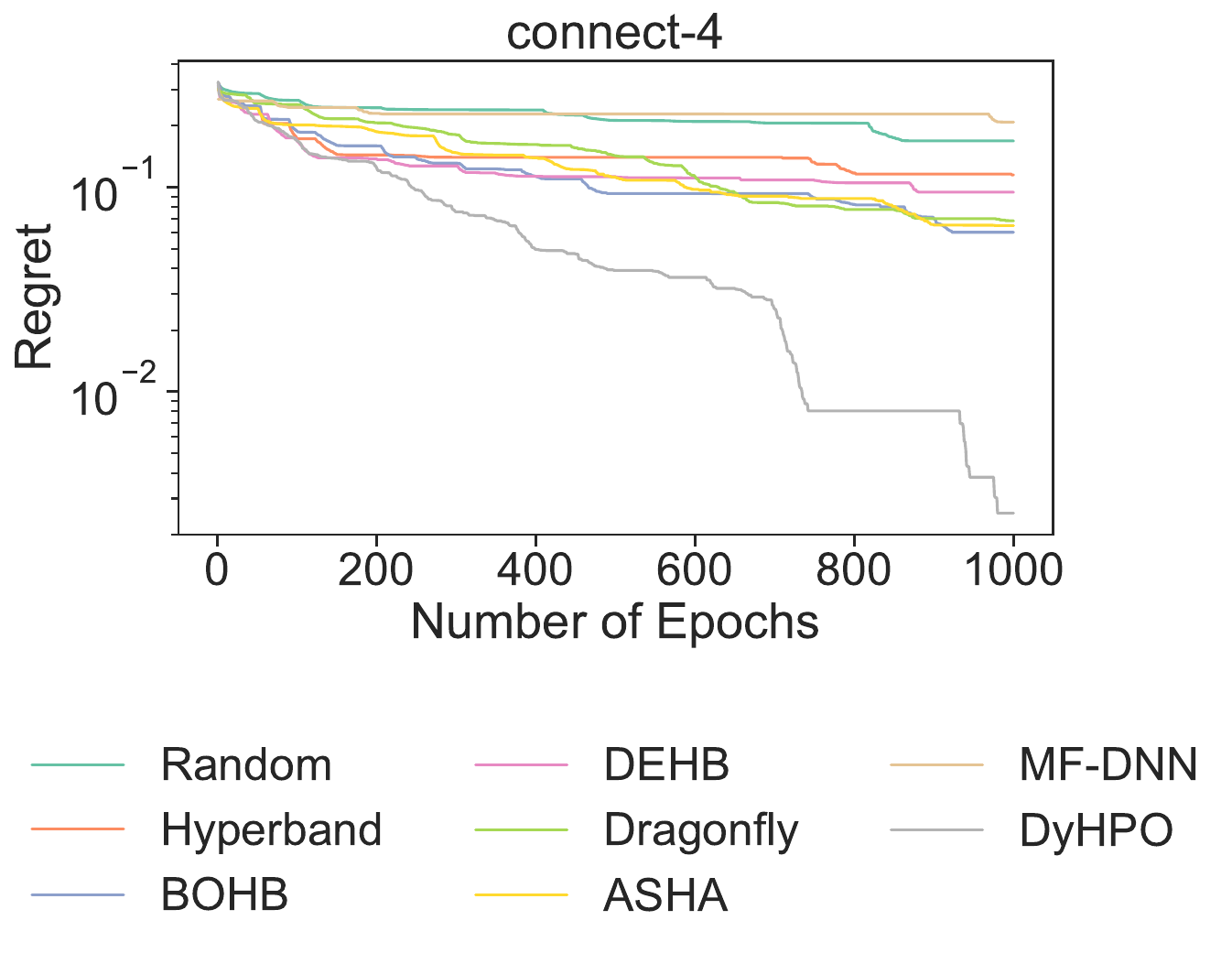}
    \includegraphics[width=0.26\textwidth]{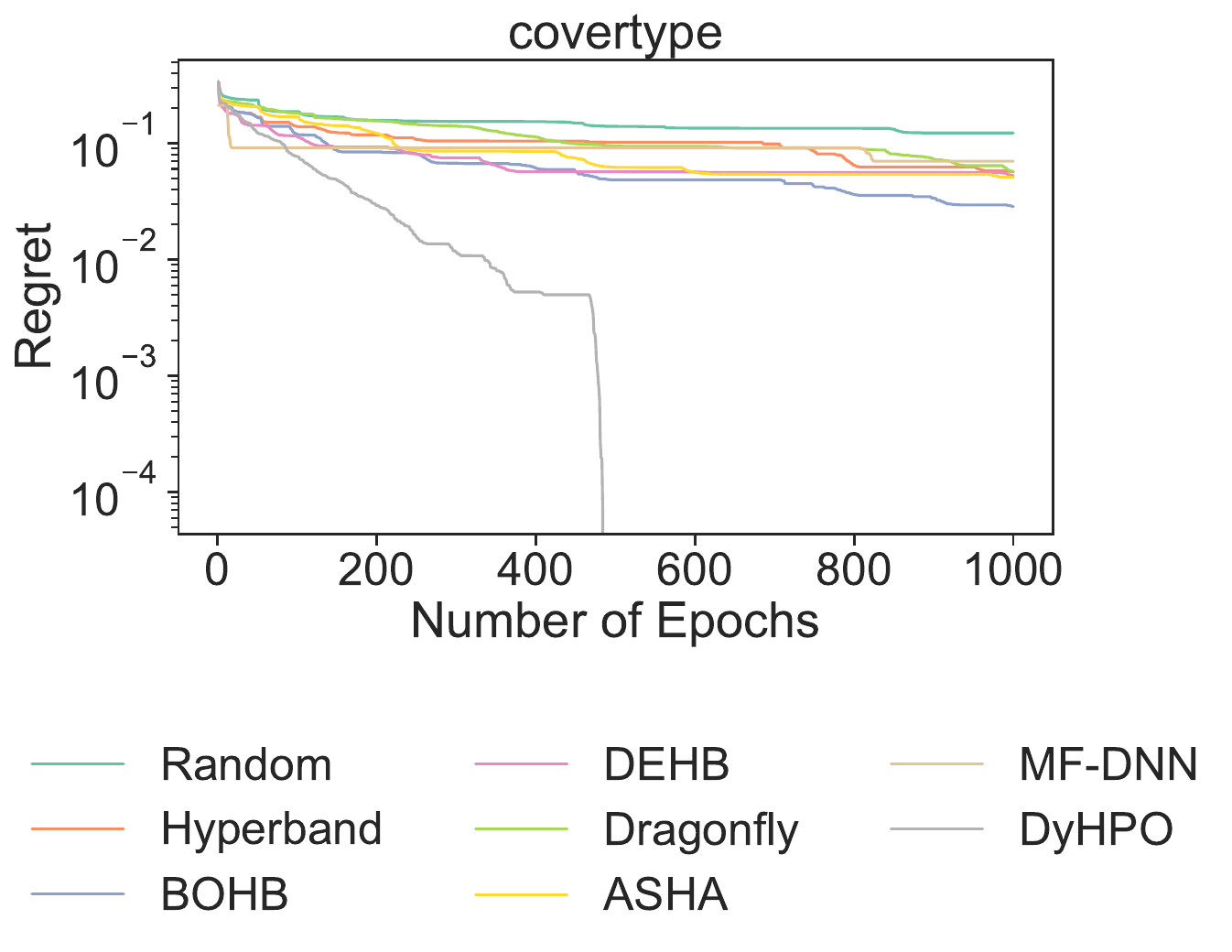}
    \includegraphics[width=0.26\textwidth]{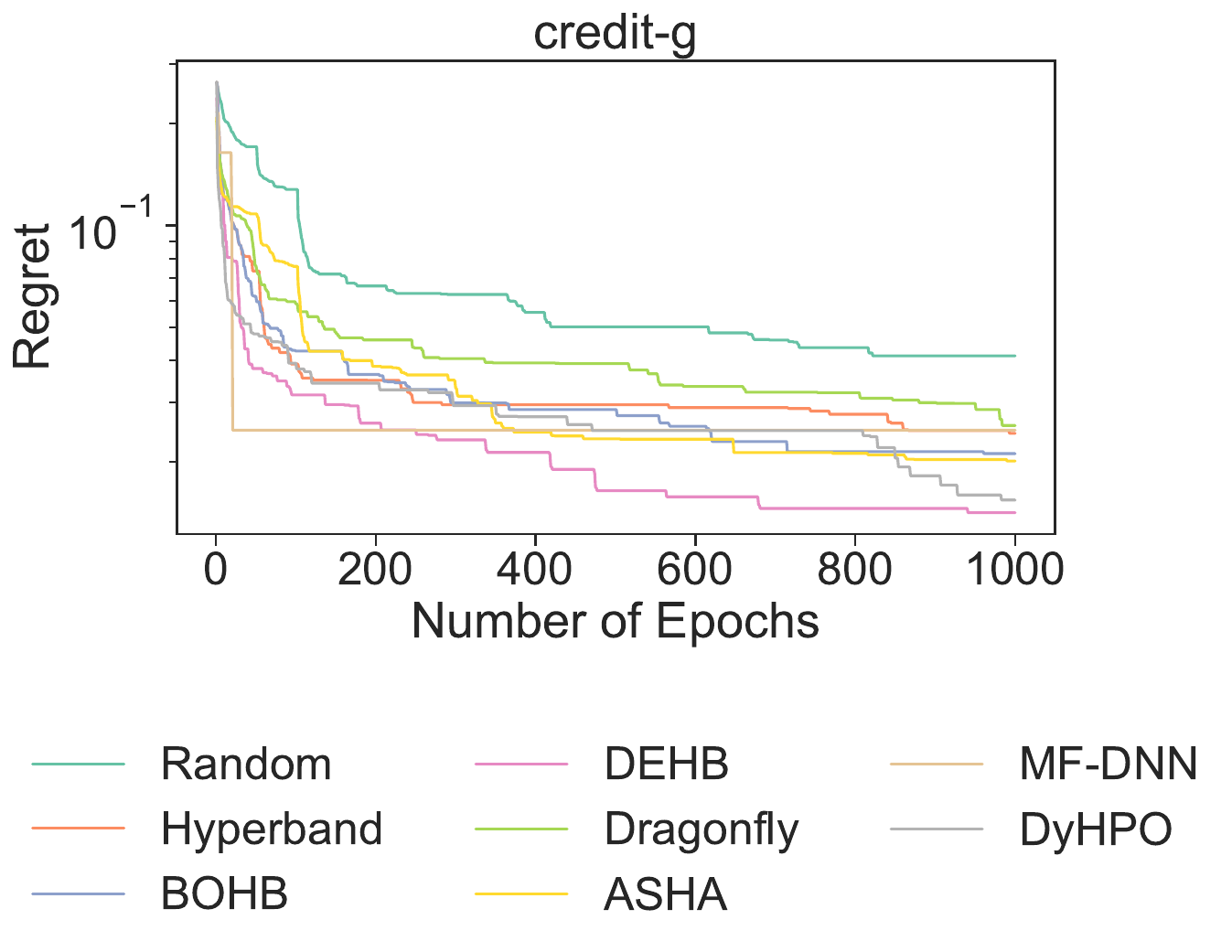}
    \includegraphics[width=0.26\textwidth]{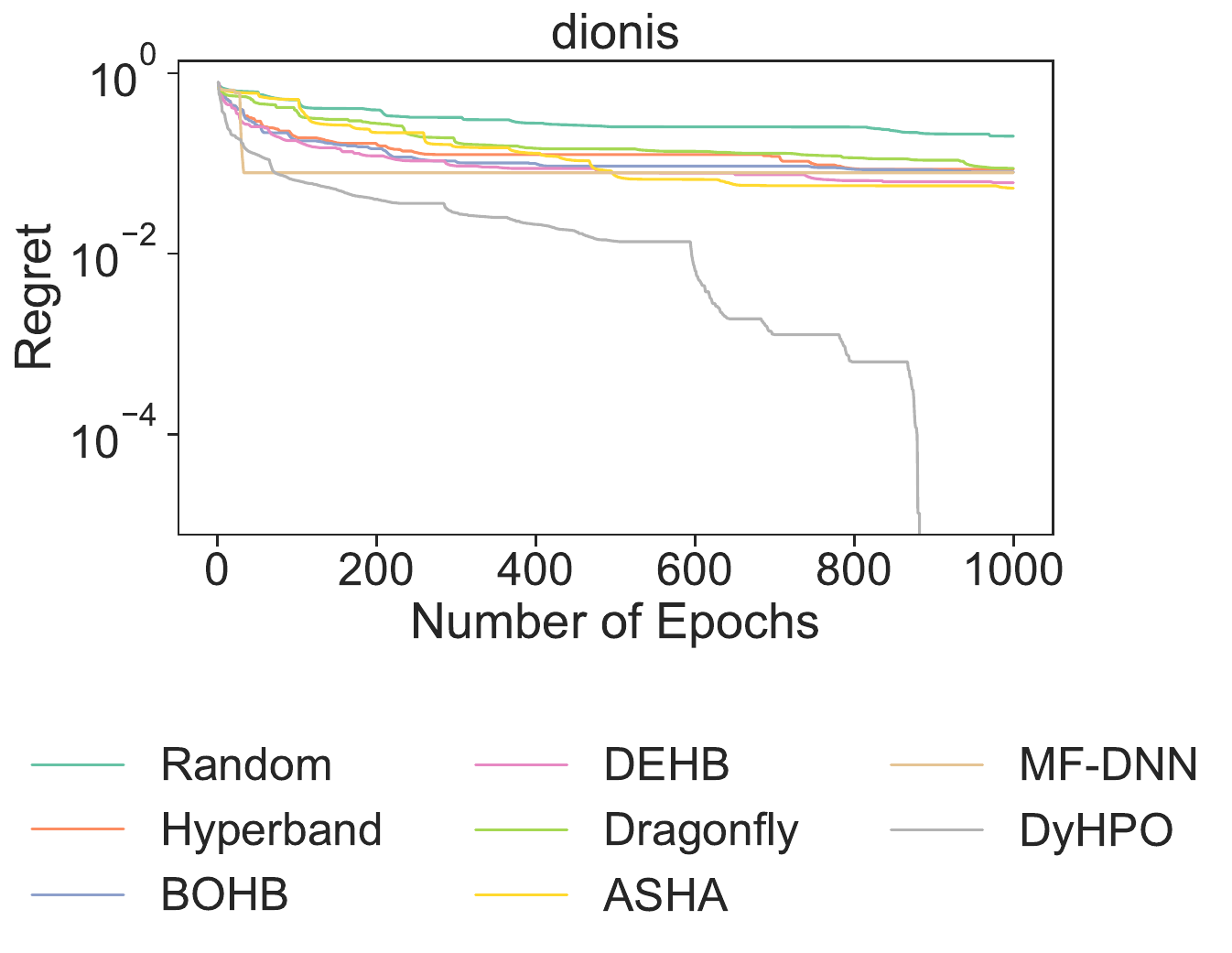}

  \caption{Performance comparison over the number of steps on a dataset level for LCBench.}
  \label{fig:results_per_dataset1}
\end{figure*}

\begin{figure*}[htp]
  \centering
    \includegraphics[width=0.26\textwidth]{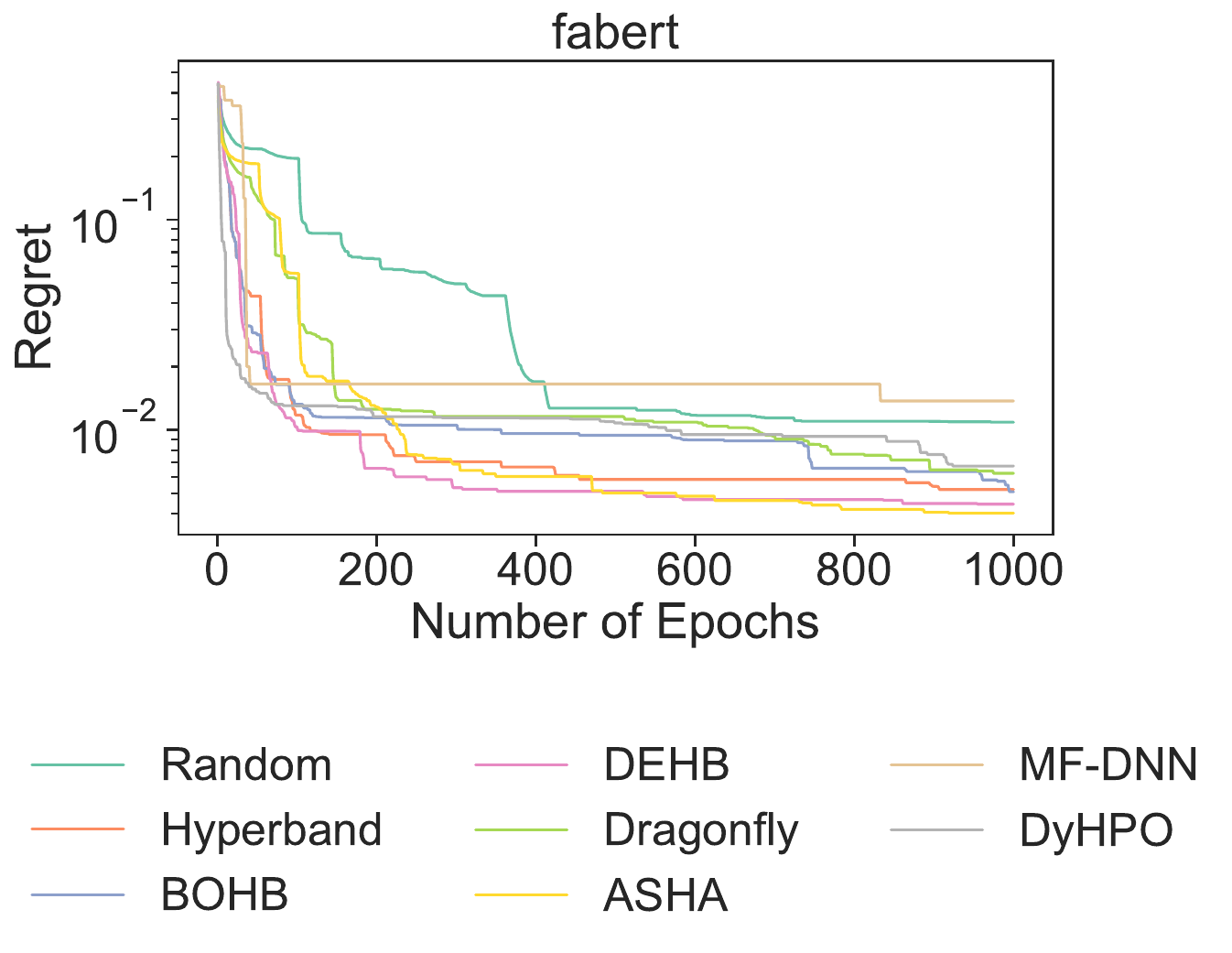}
    \includegraphics[width=0.26\textwidth]{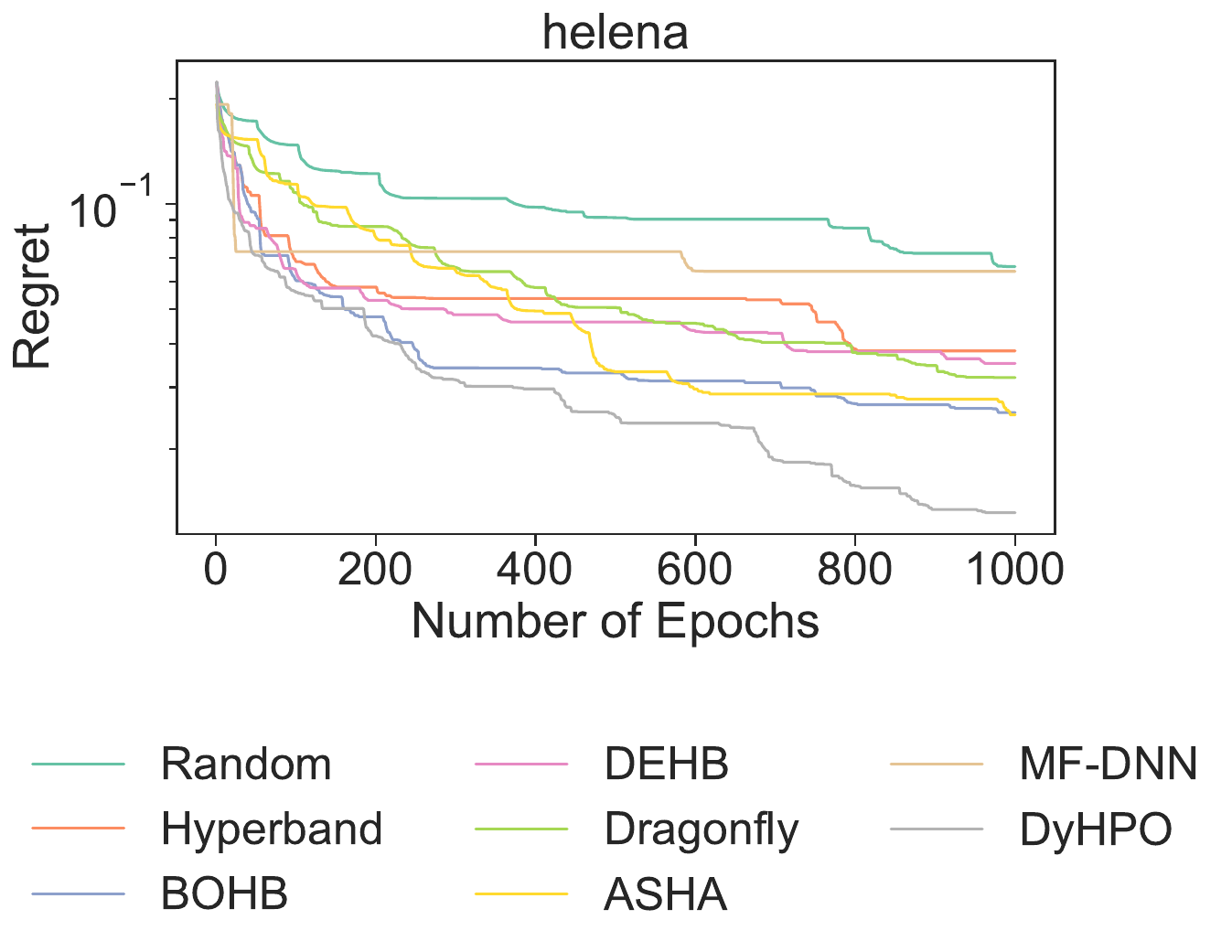}
    \includegraphics[width=0.26\textwidth]{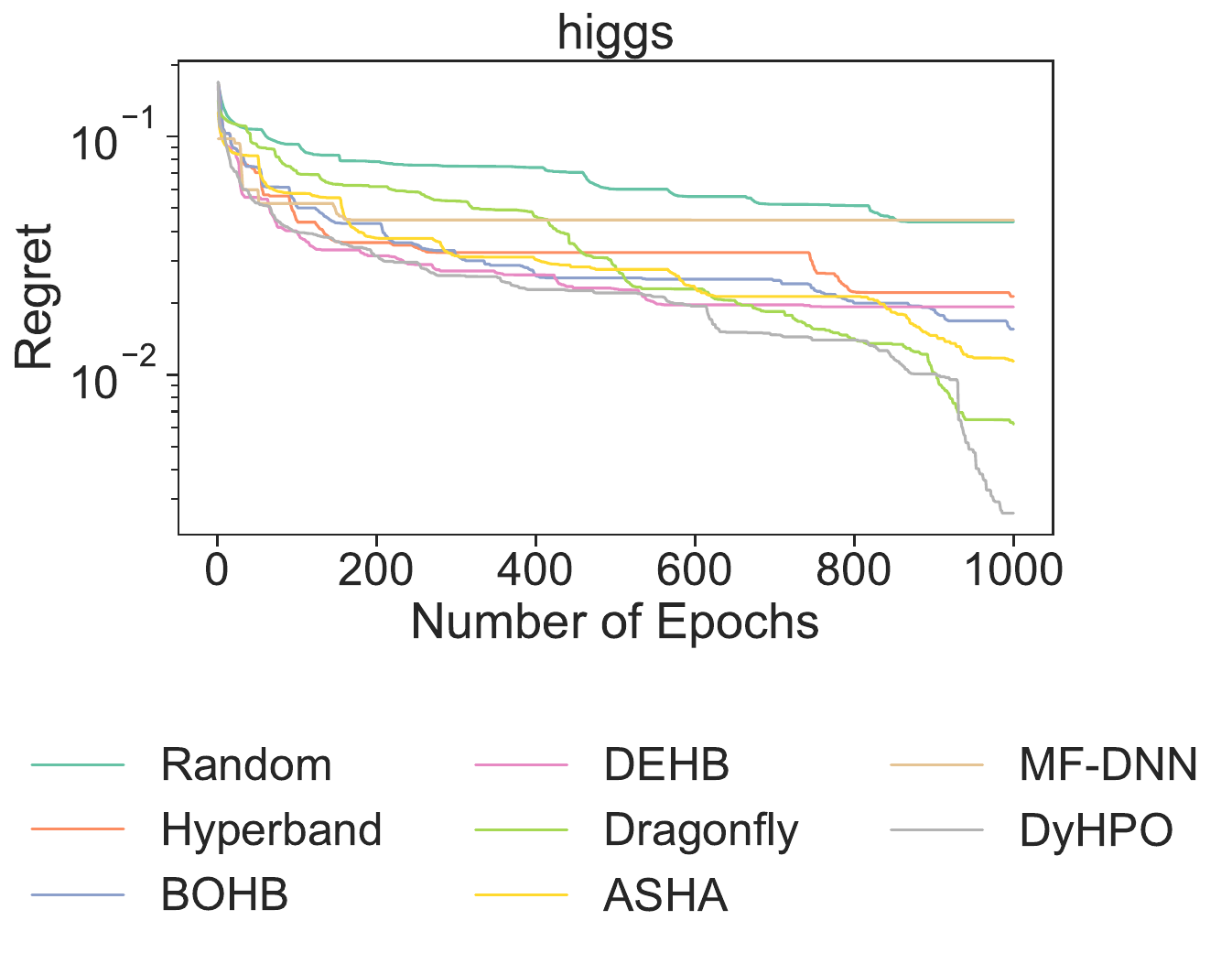}
    \includegraphics[width=0.26\textwidth]{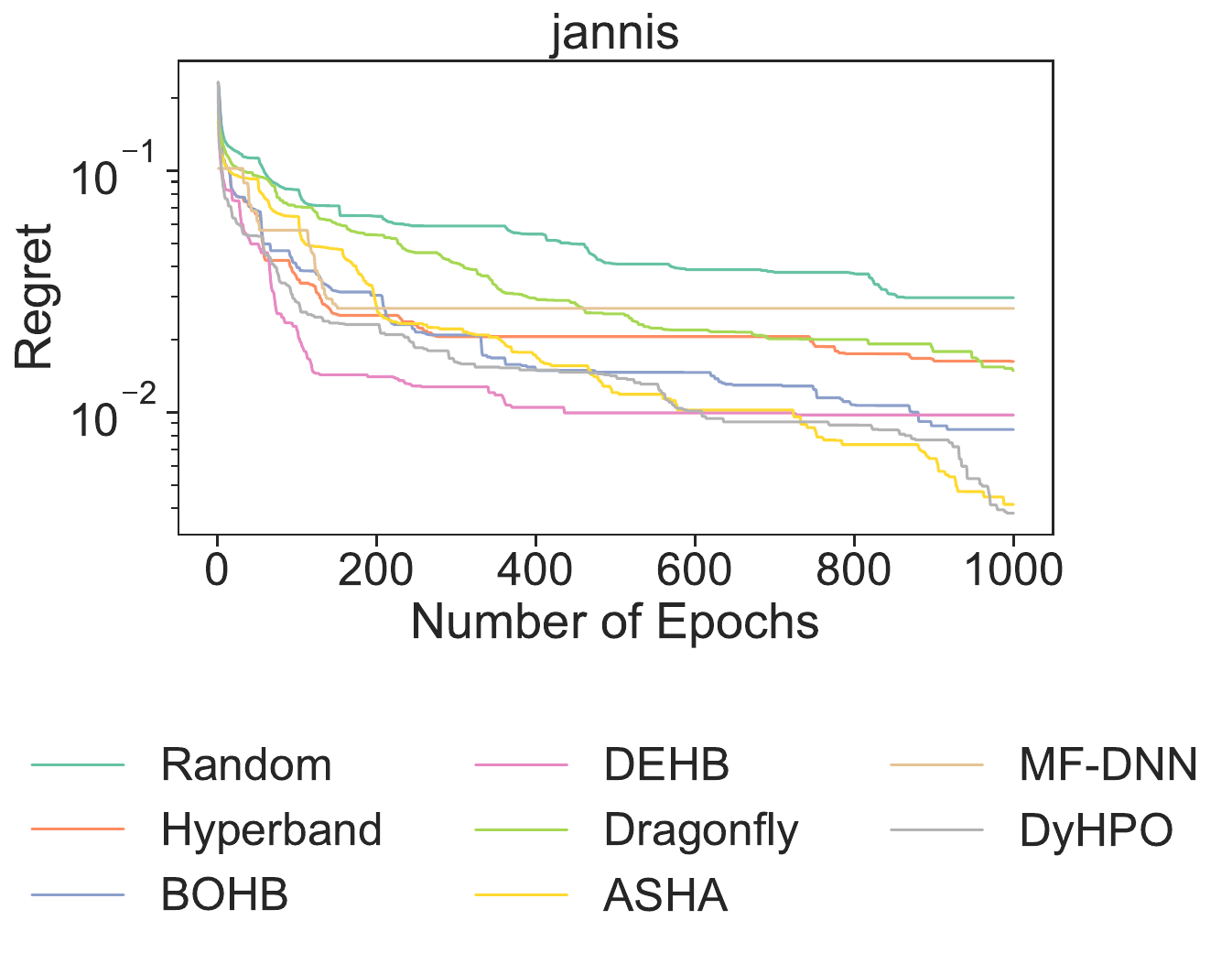}
    \includegraphics[width=0.26\textwidth]{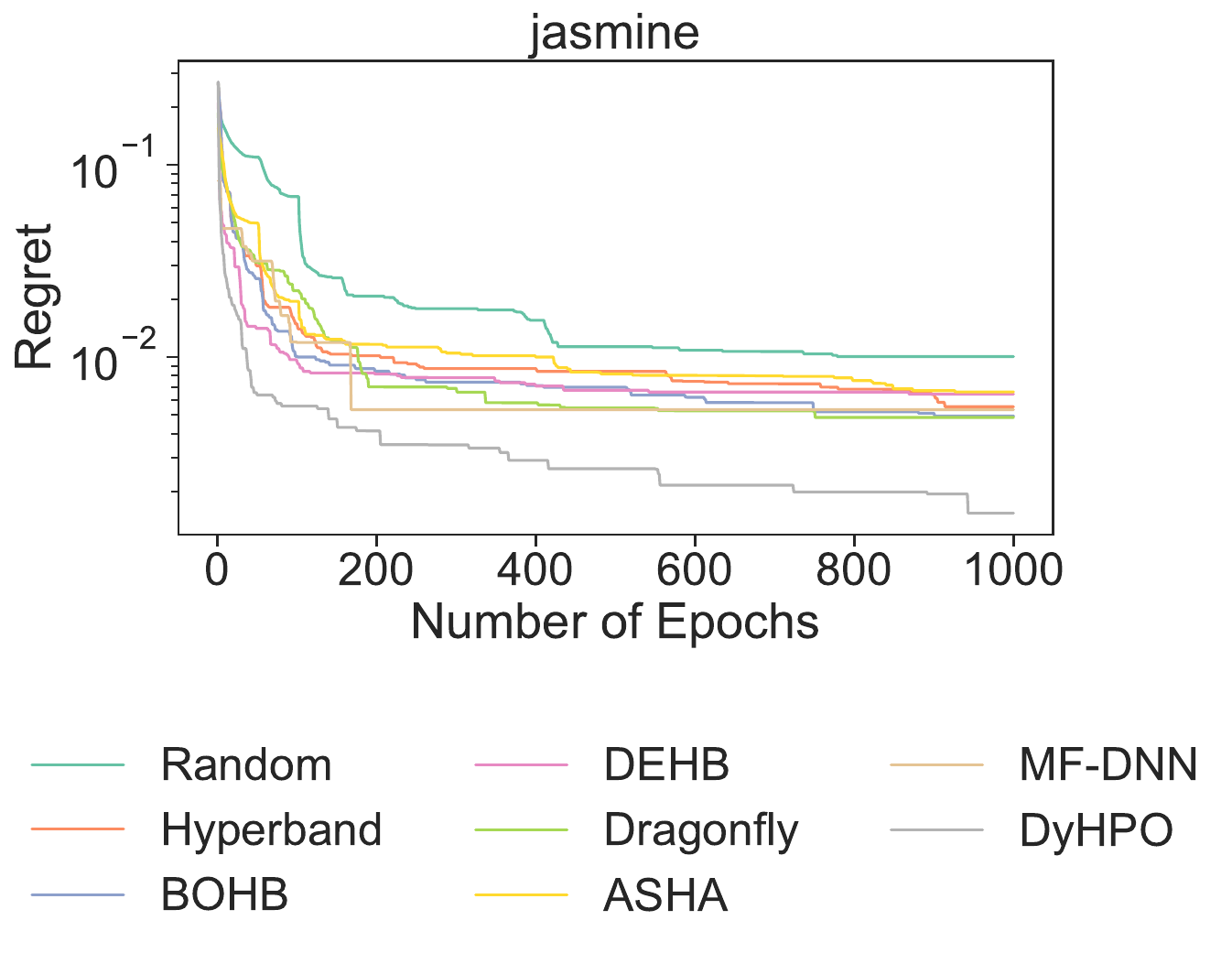}
    \includegraphics[width=0.26\textwidth]{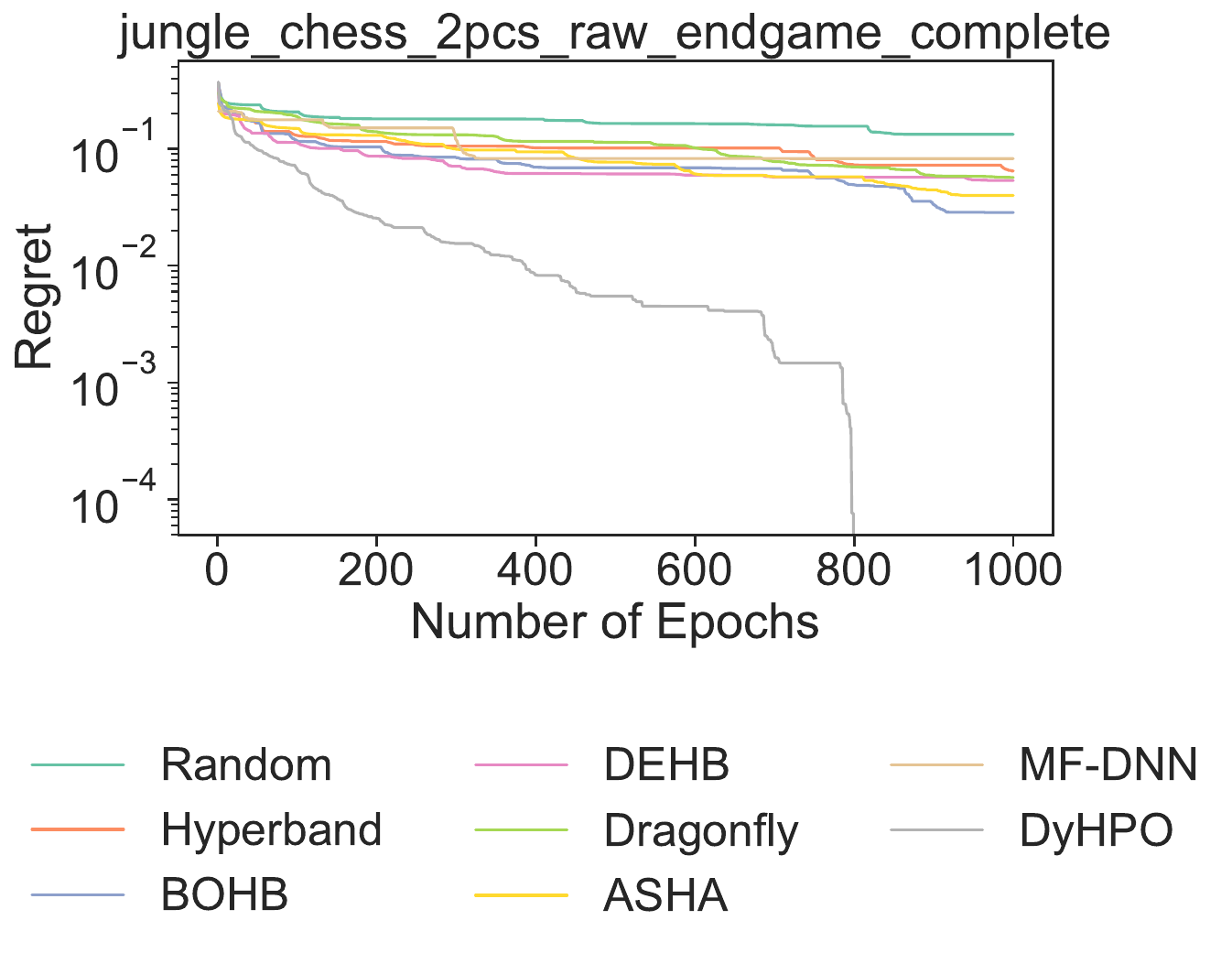}
    \includegraphics[width=0.26\textwidth]{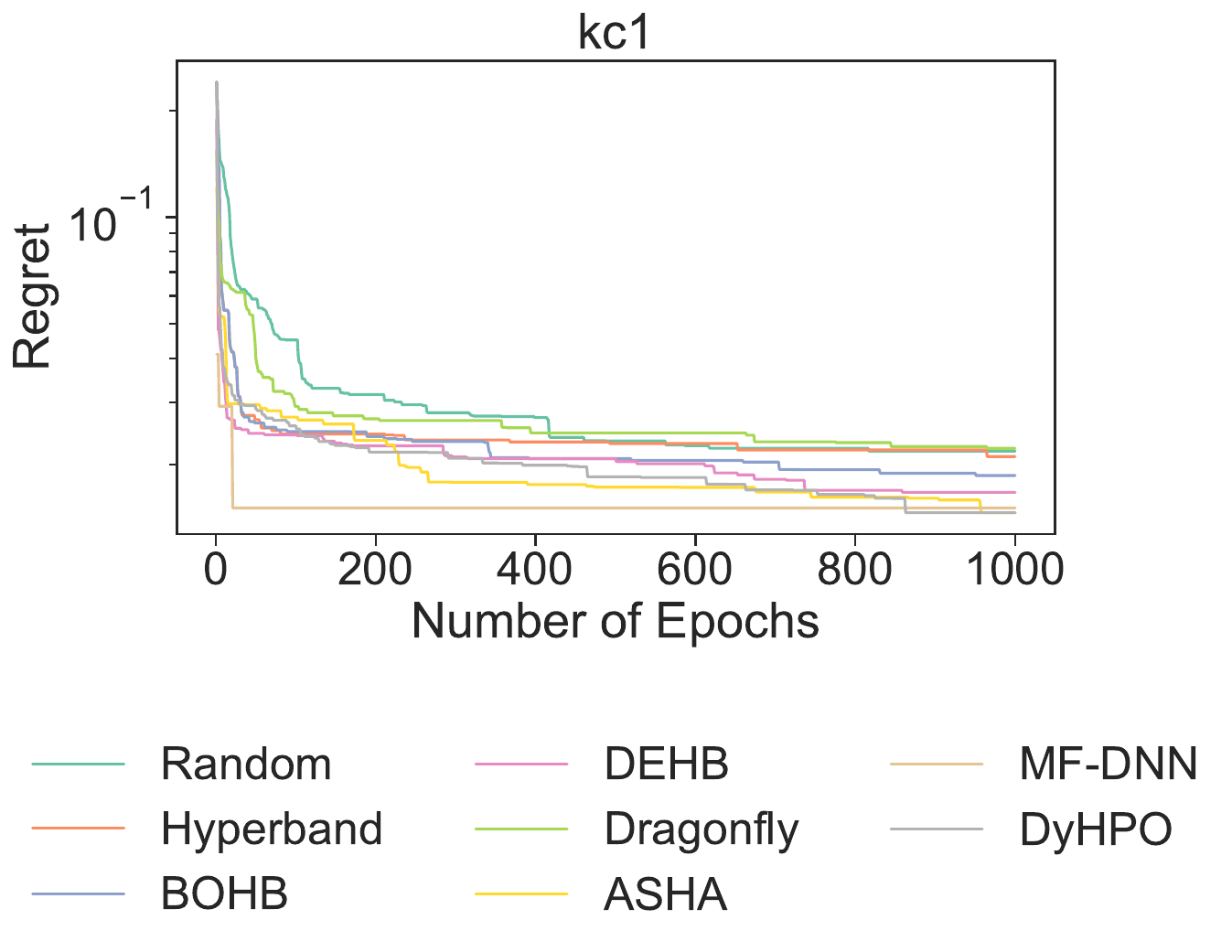}
    \includegraphics[width=0.26\textwidth]{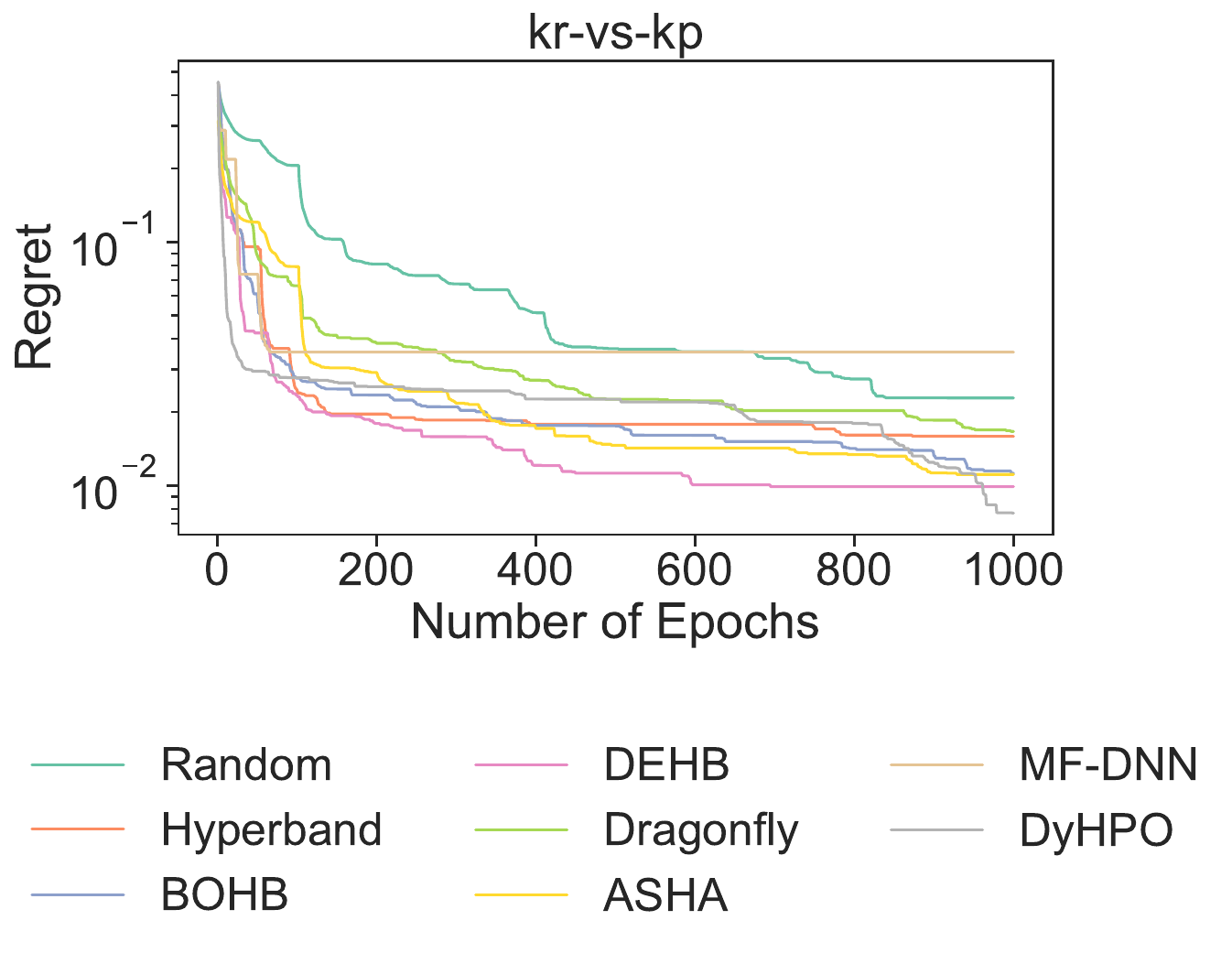}
    \includegraphics[width=0.26\textwidth]{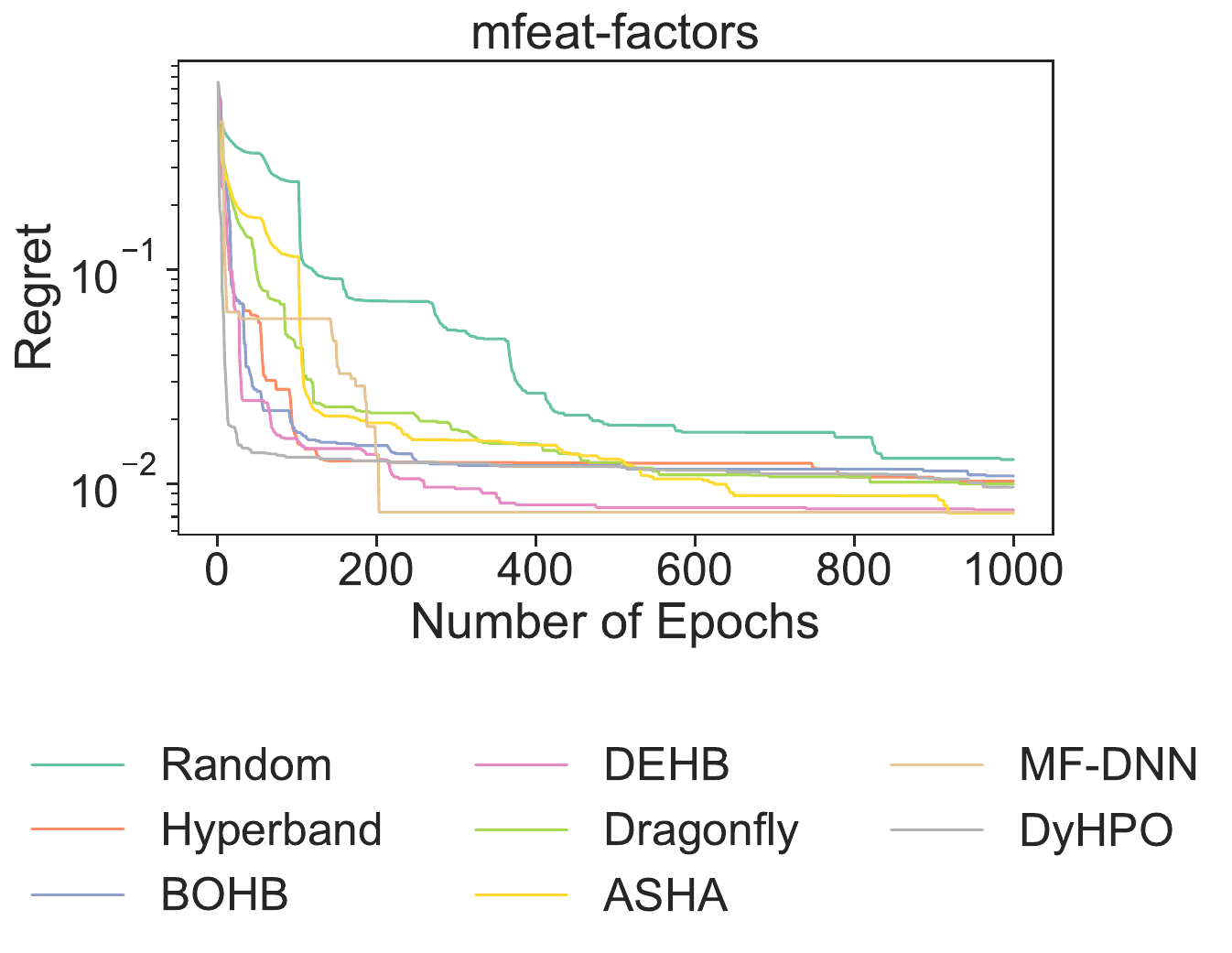}
    \includegraphics[width=0.26\textwidth]{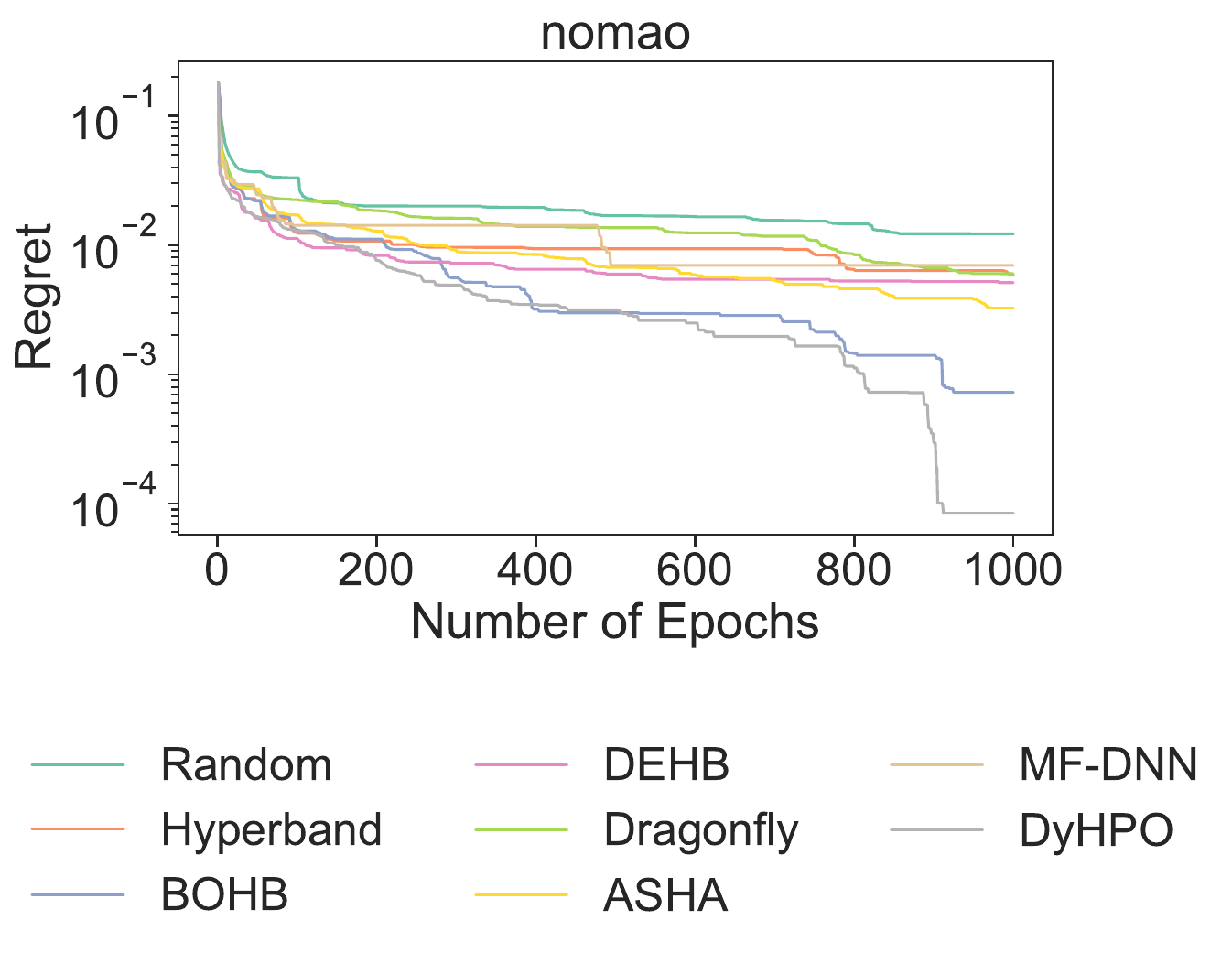}
    \includegraphics[width=0.26\textwidth]{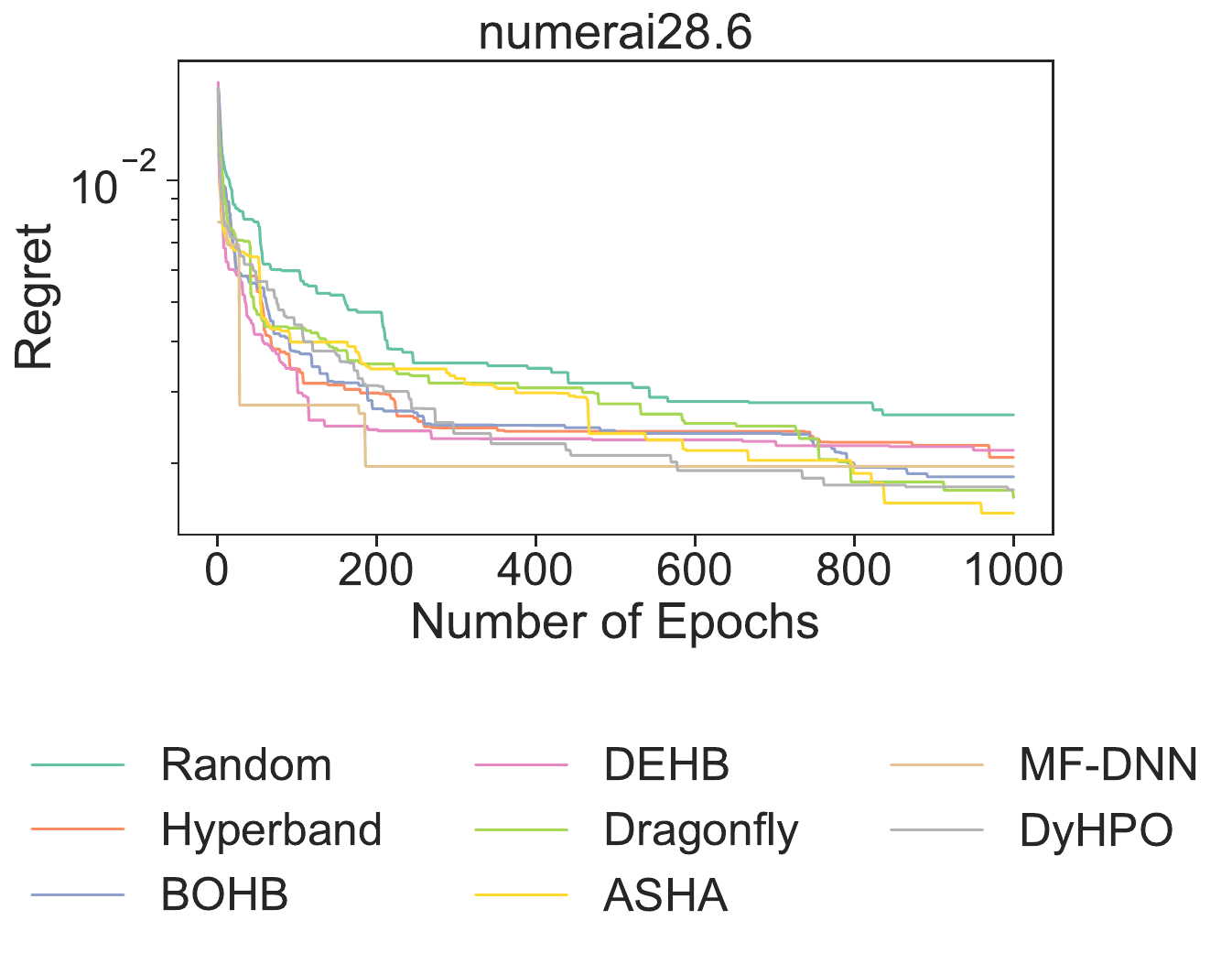}
    \includegraphics[width=0.26\textwidth]{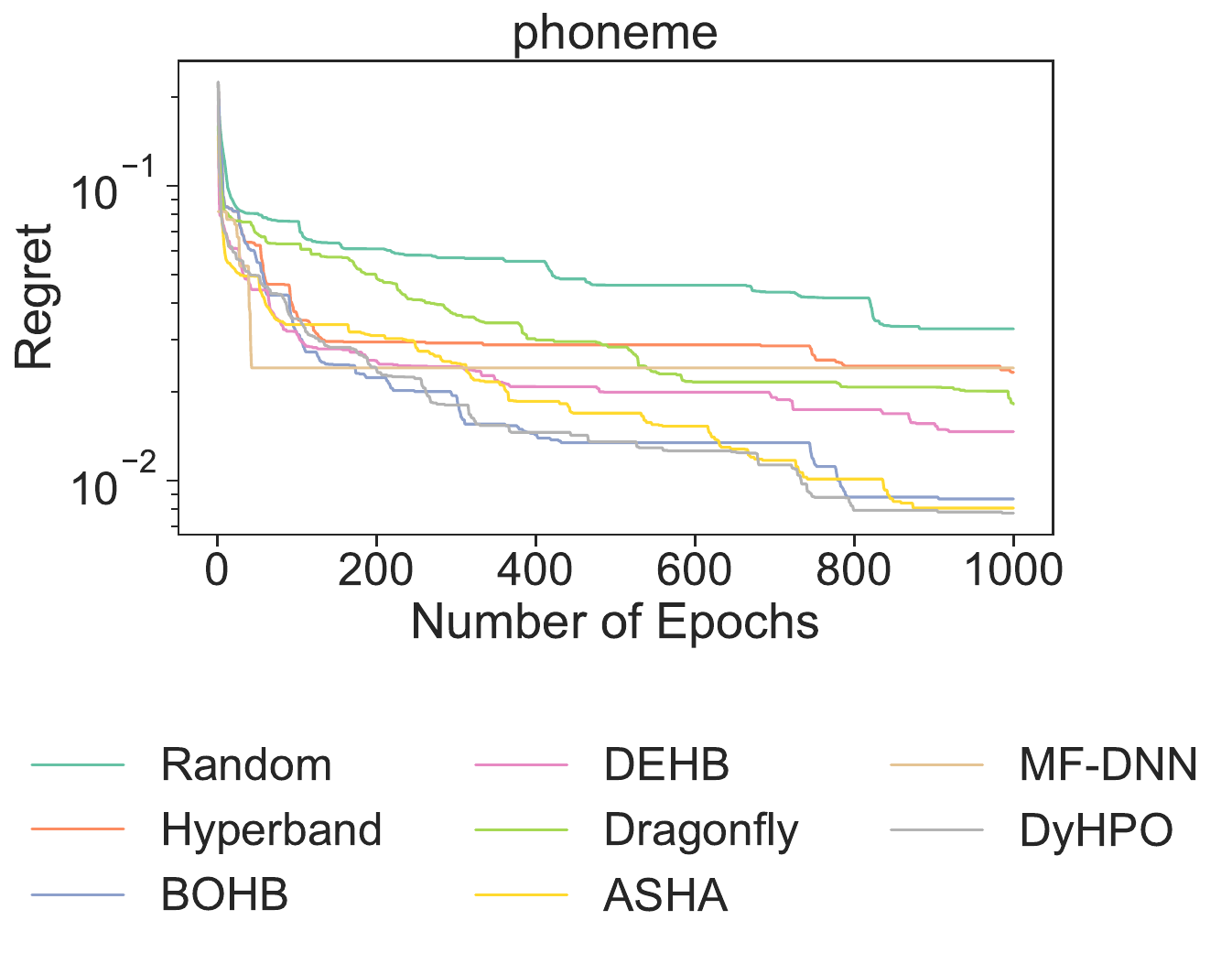}
    \includegraphics[width=0.26\textwidth]{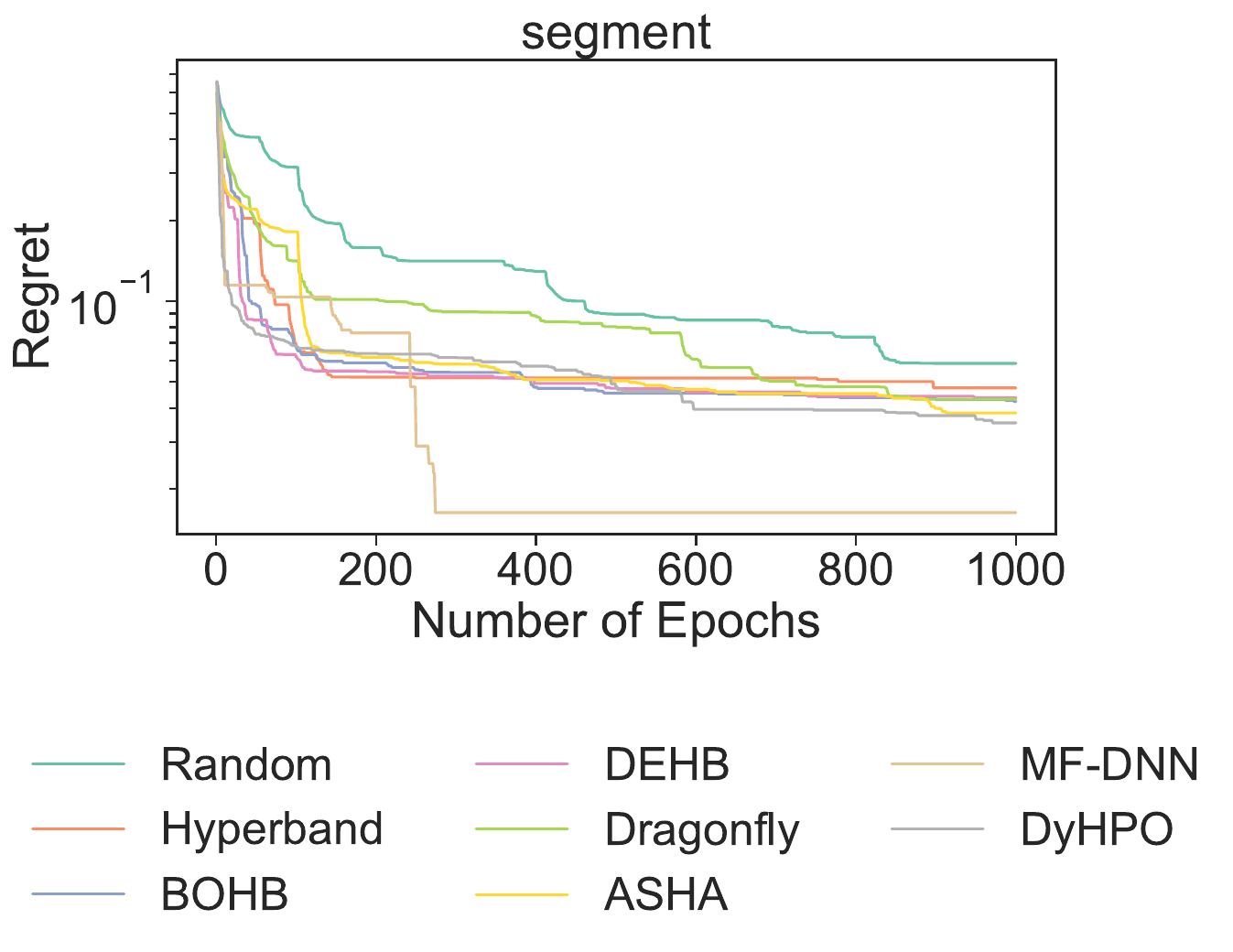}
    \includegraphics[width=0.26\textwidth]{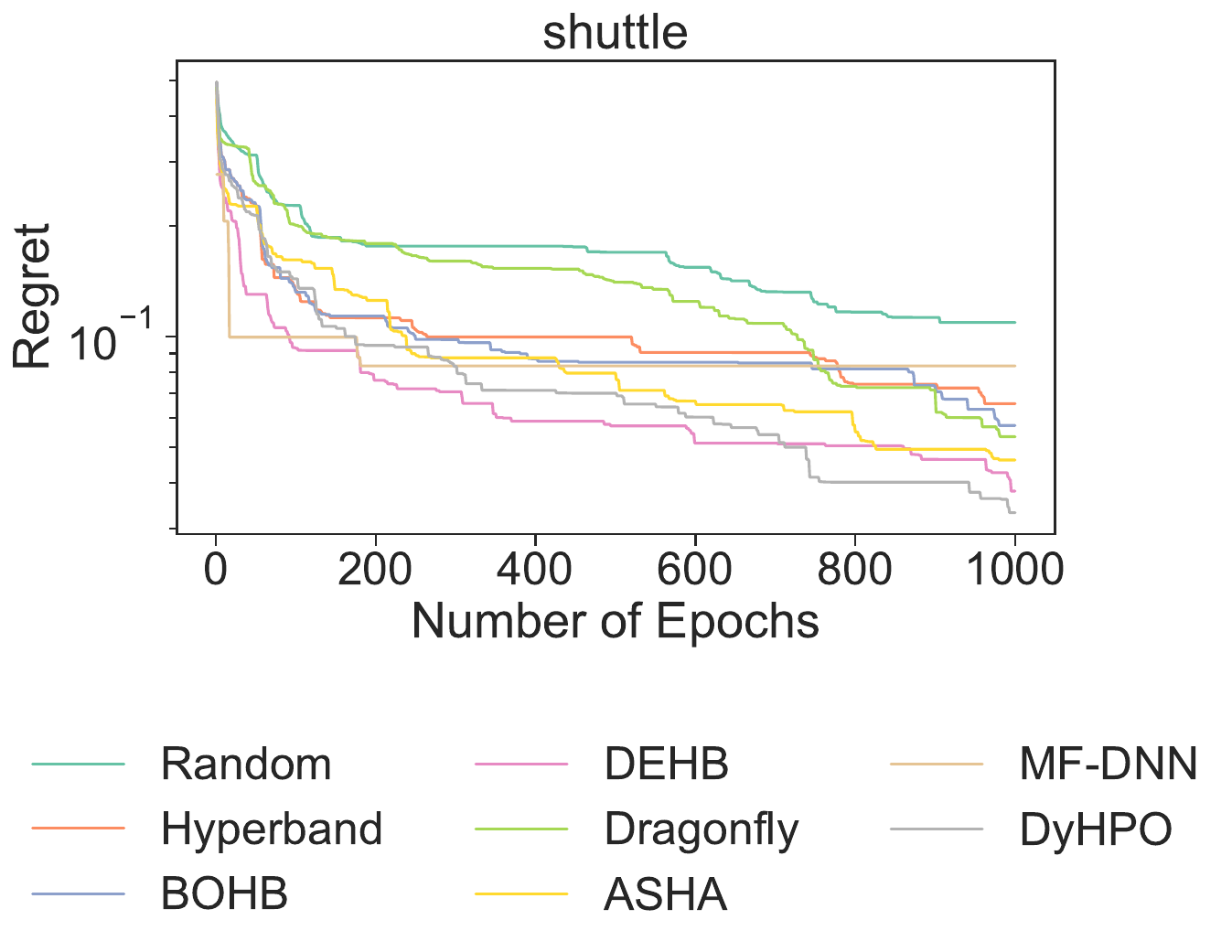}
    \includegraphics[width=0.26\textwidth]{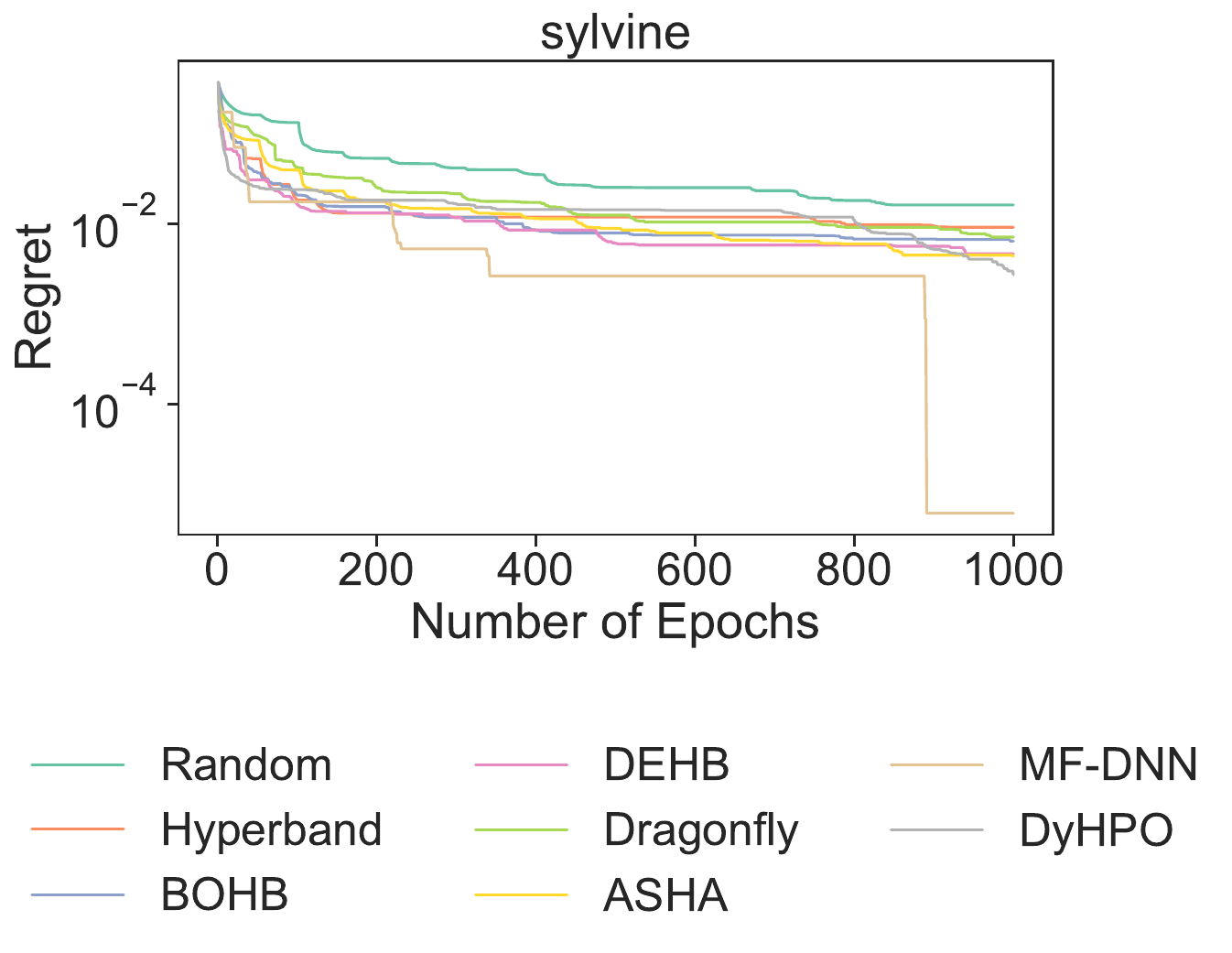}
    \includegraphics[width=0.26\textwidth]{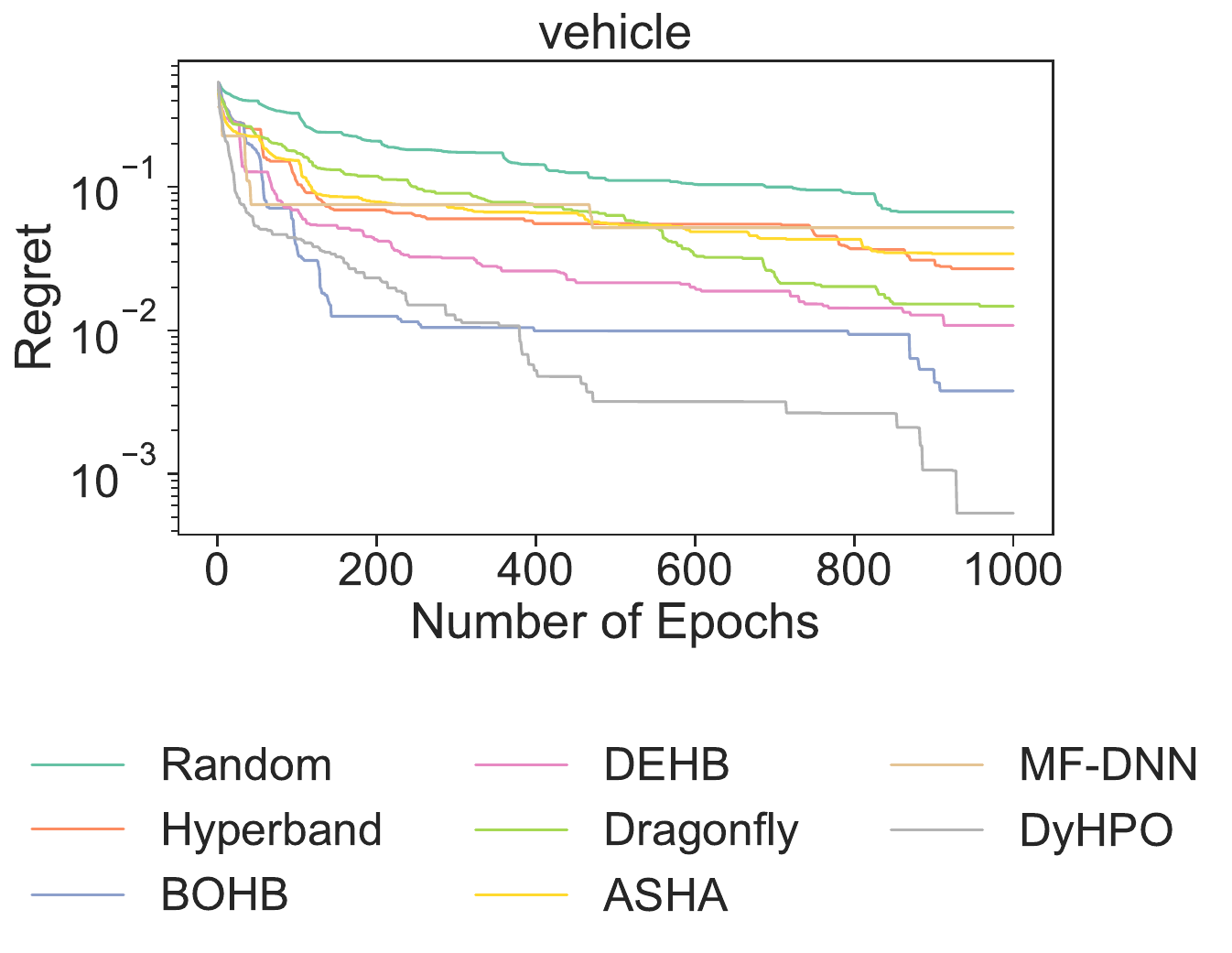}
    \includegraphics[width=0.26\textwidth]{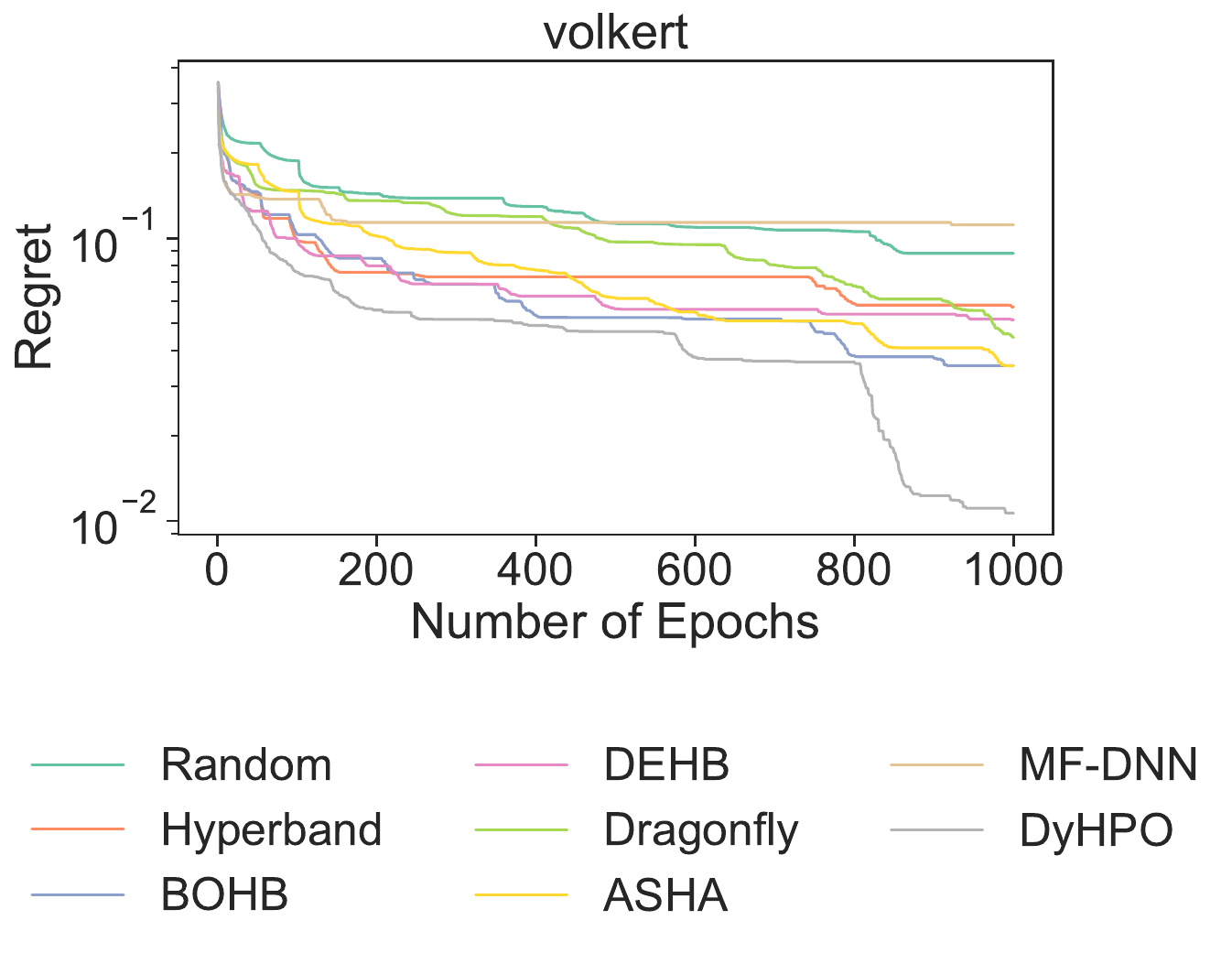}

  \caption{Performance comparison over the number of steps on a dataset level for LCBench (cont.).}
  \label{fig:results_per_dataset2}
\end{figure*}

\begin{figure*}[htp]
  \centering
    \includegraphics[width=0.28\textwidth]{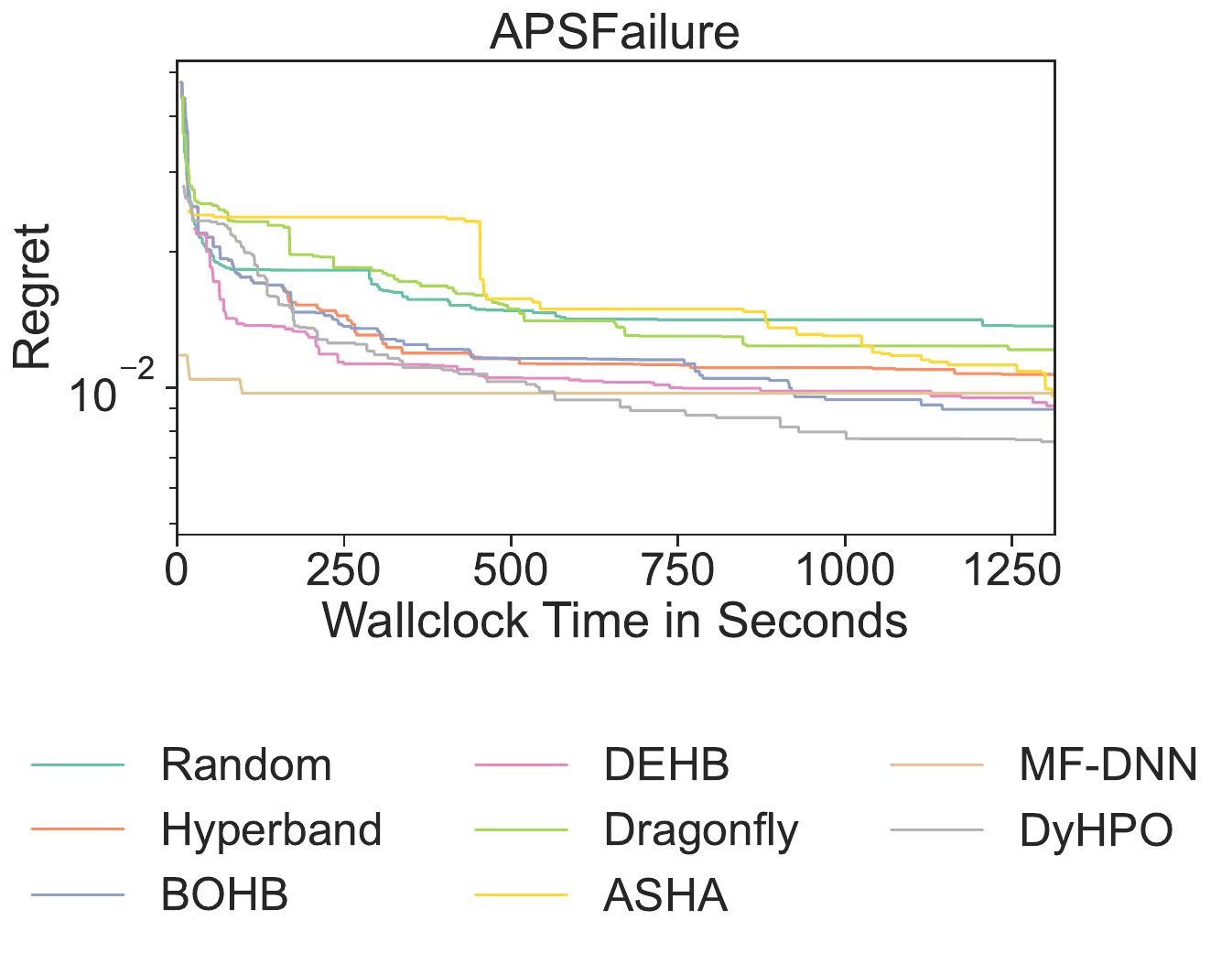}
    \includegraphics[width=0.28\textwidth]{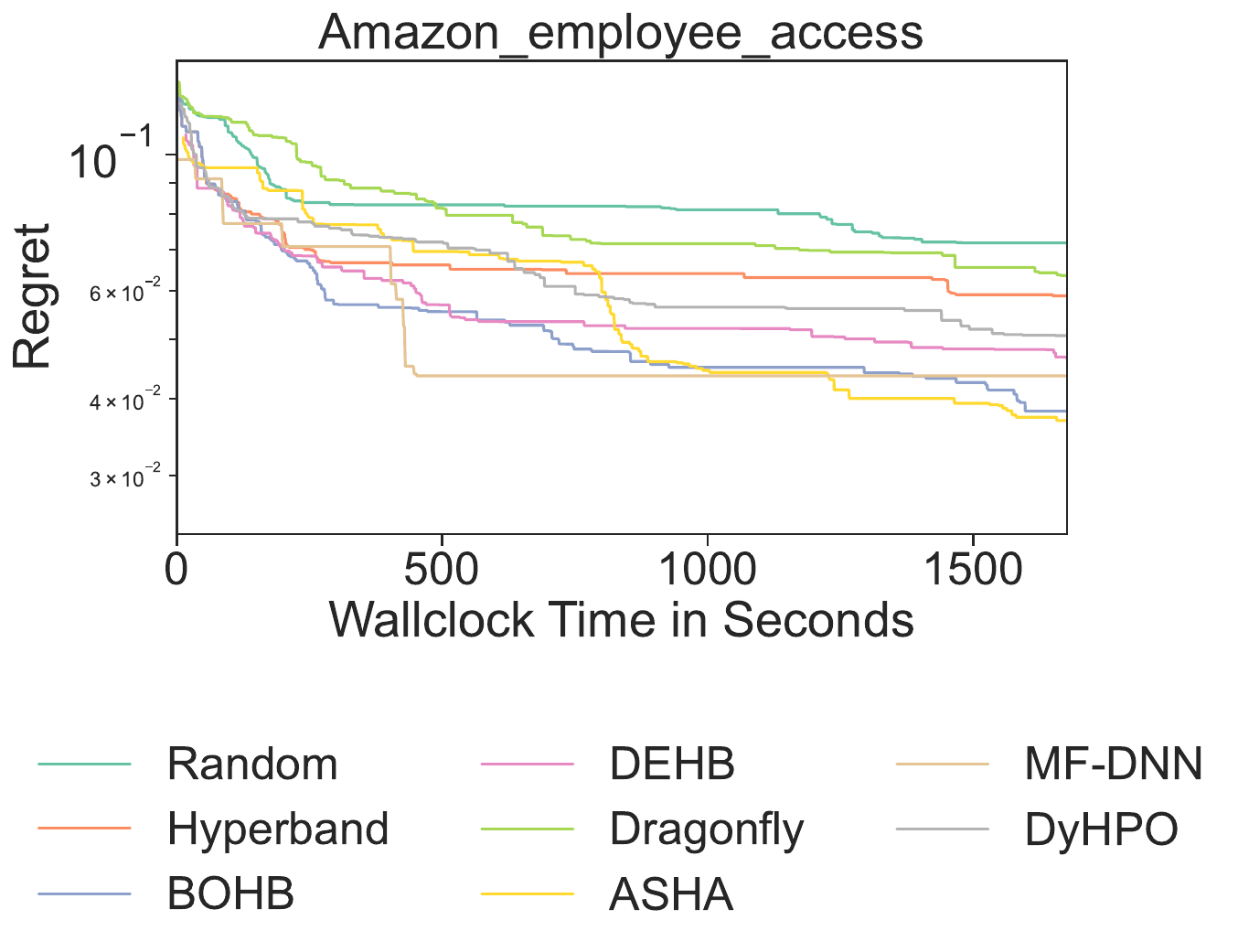}
    \includegraphics[width=0.28\textwidth]{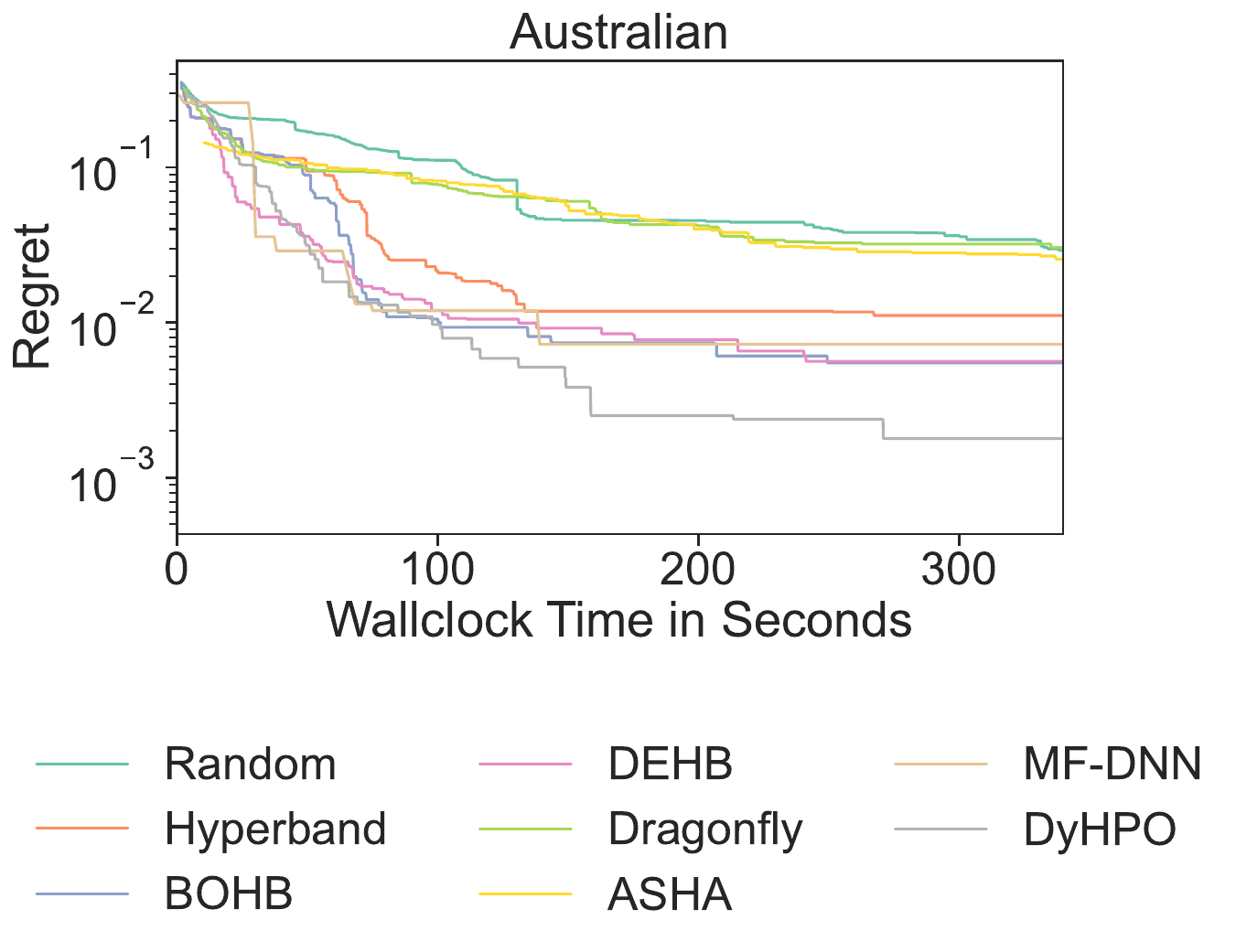}
    \includegraphics[width=0.28\textwidth]{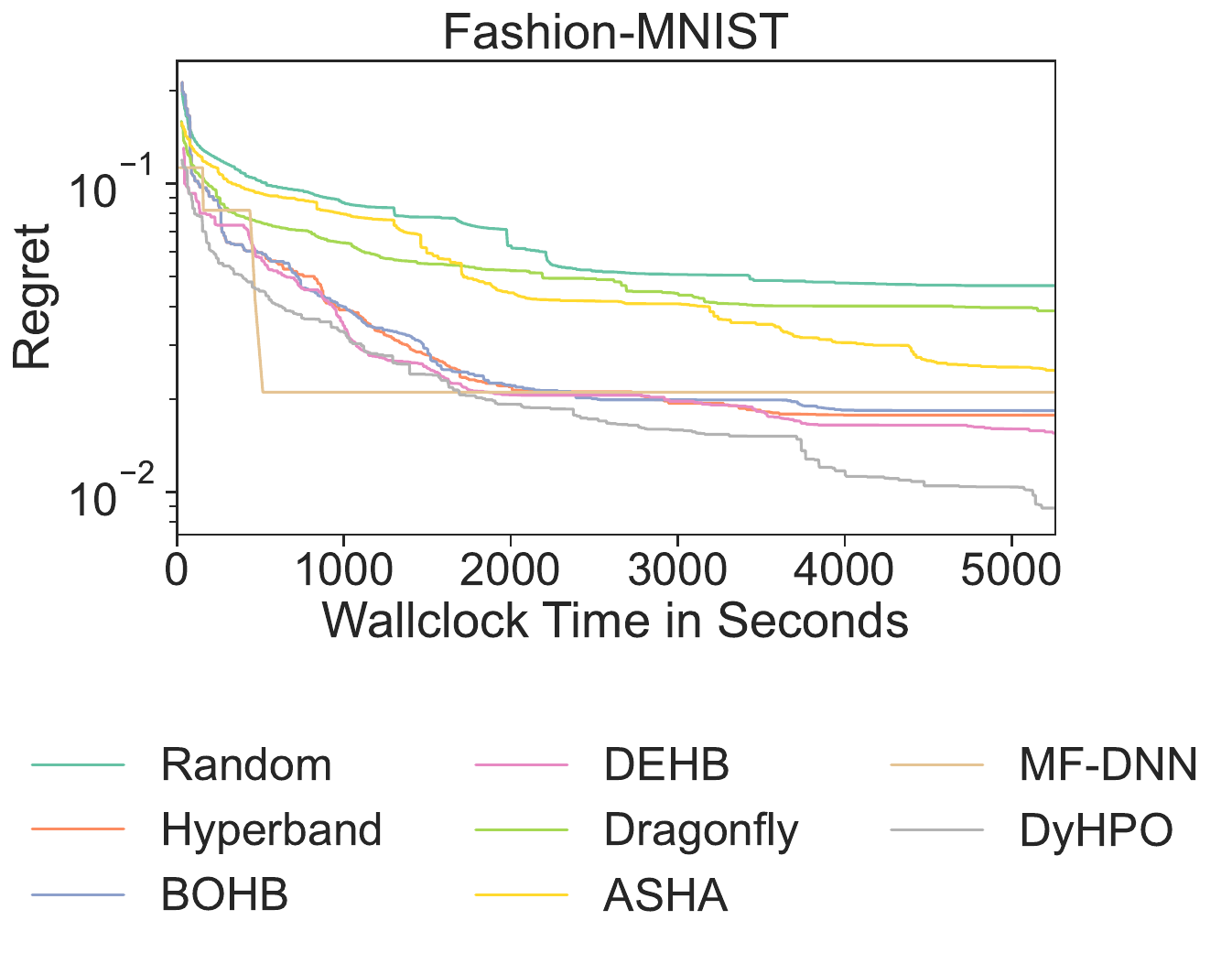}
    \includegraphics[width=0.28\textwidth]{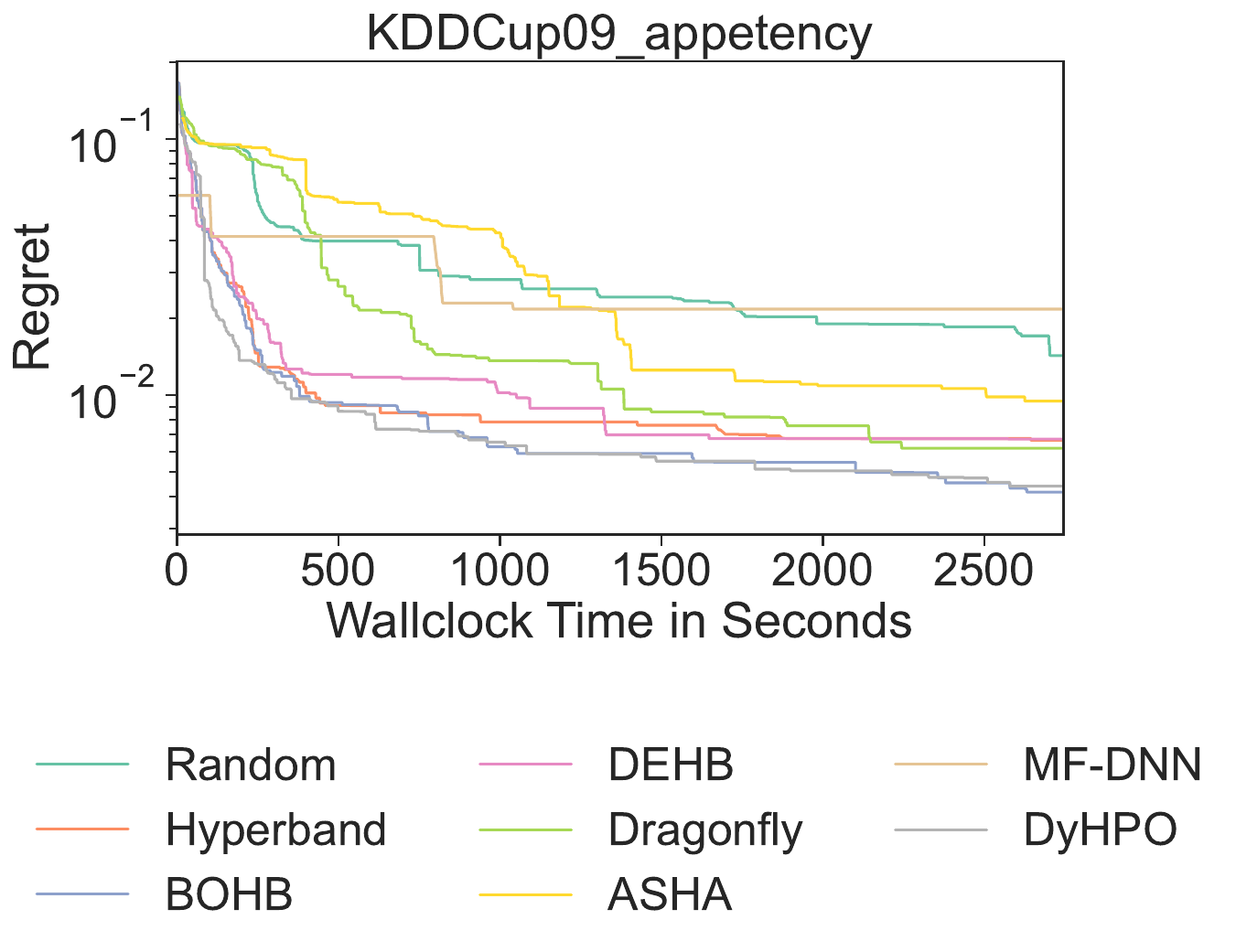}
    \includegraphics[width=0.28\textwidth]{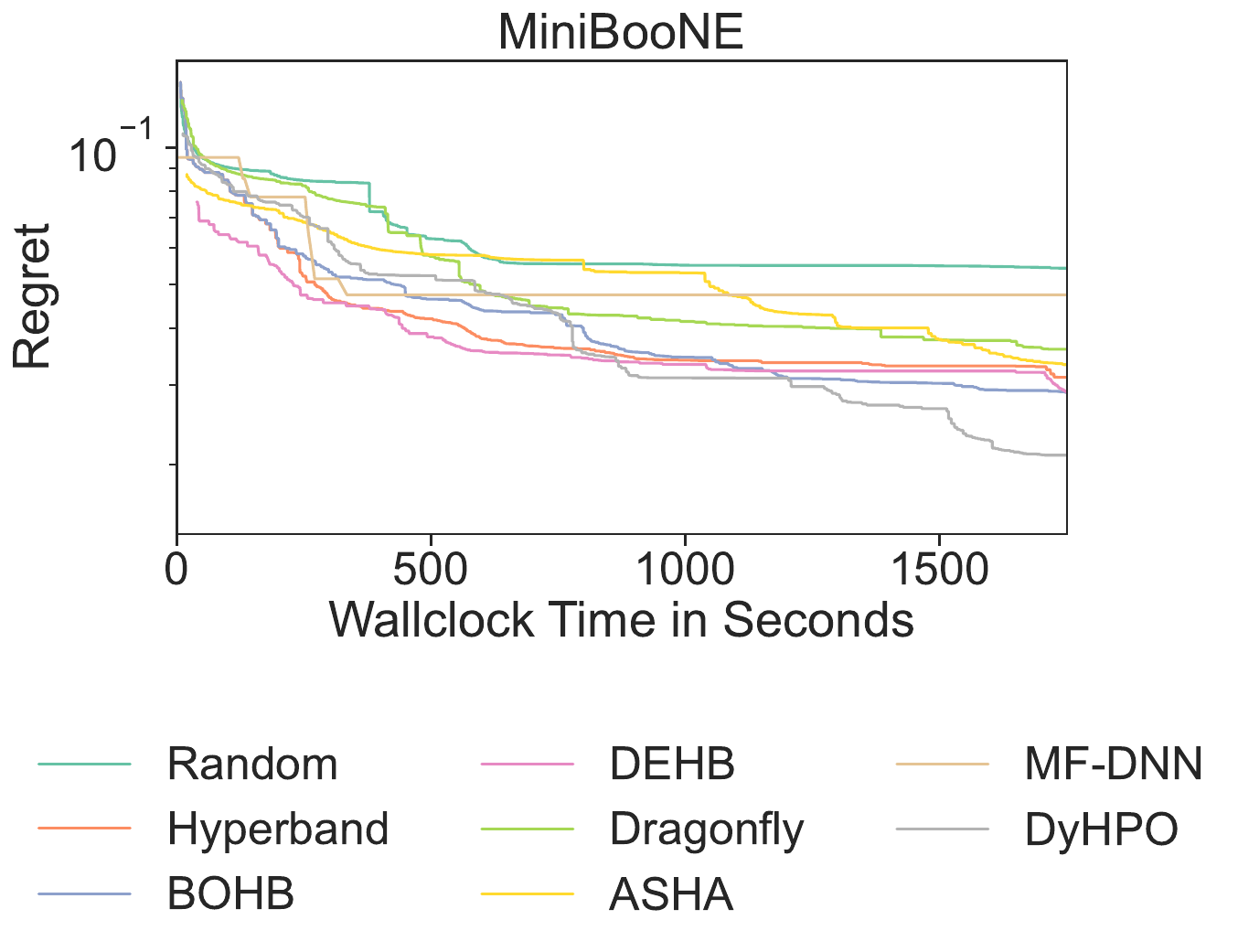}
    \includegraphics[width=0.28\textwidth]{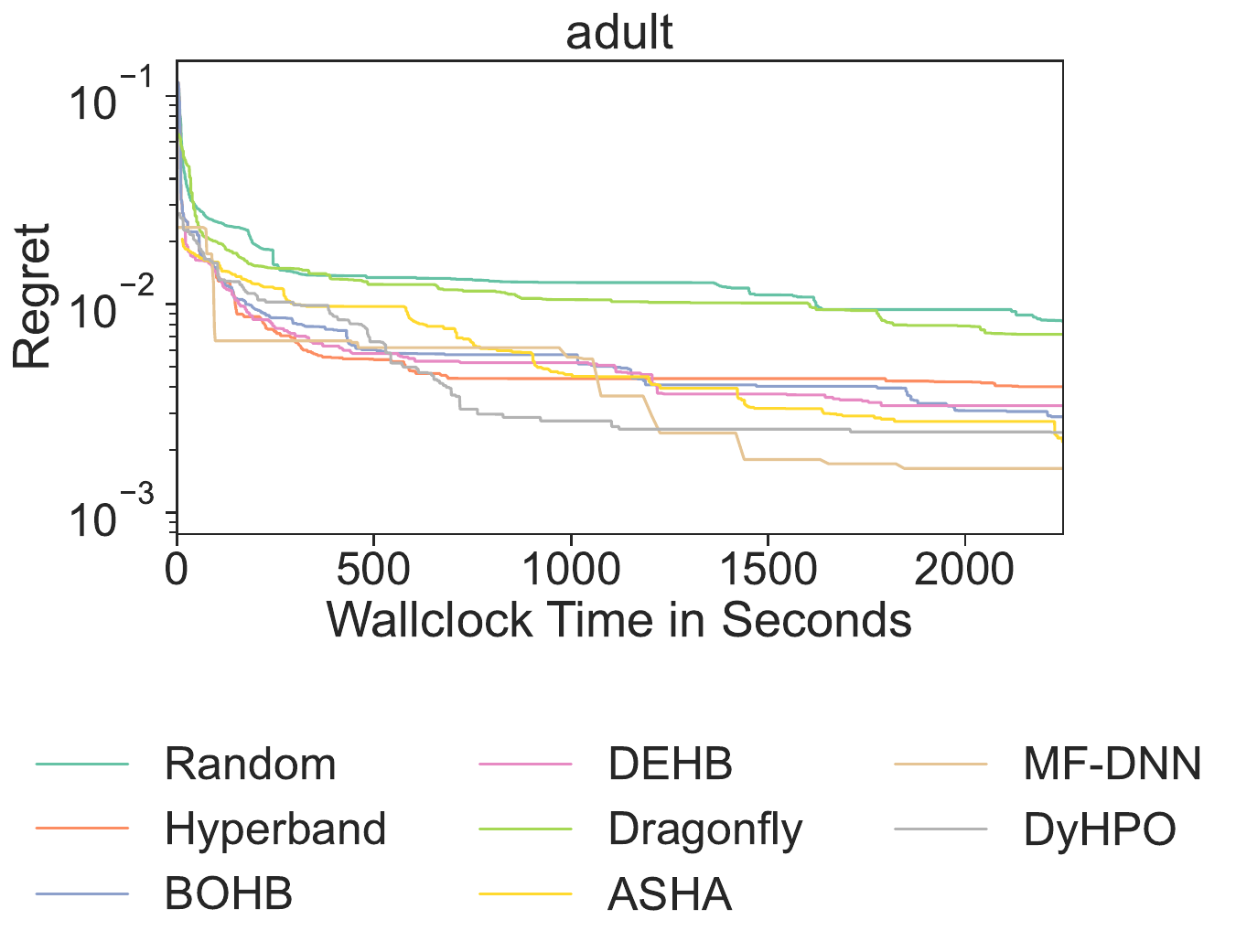}
    \includegraphics[width=0.28\textwidth]{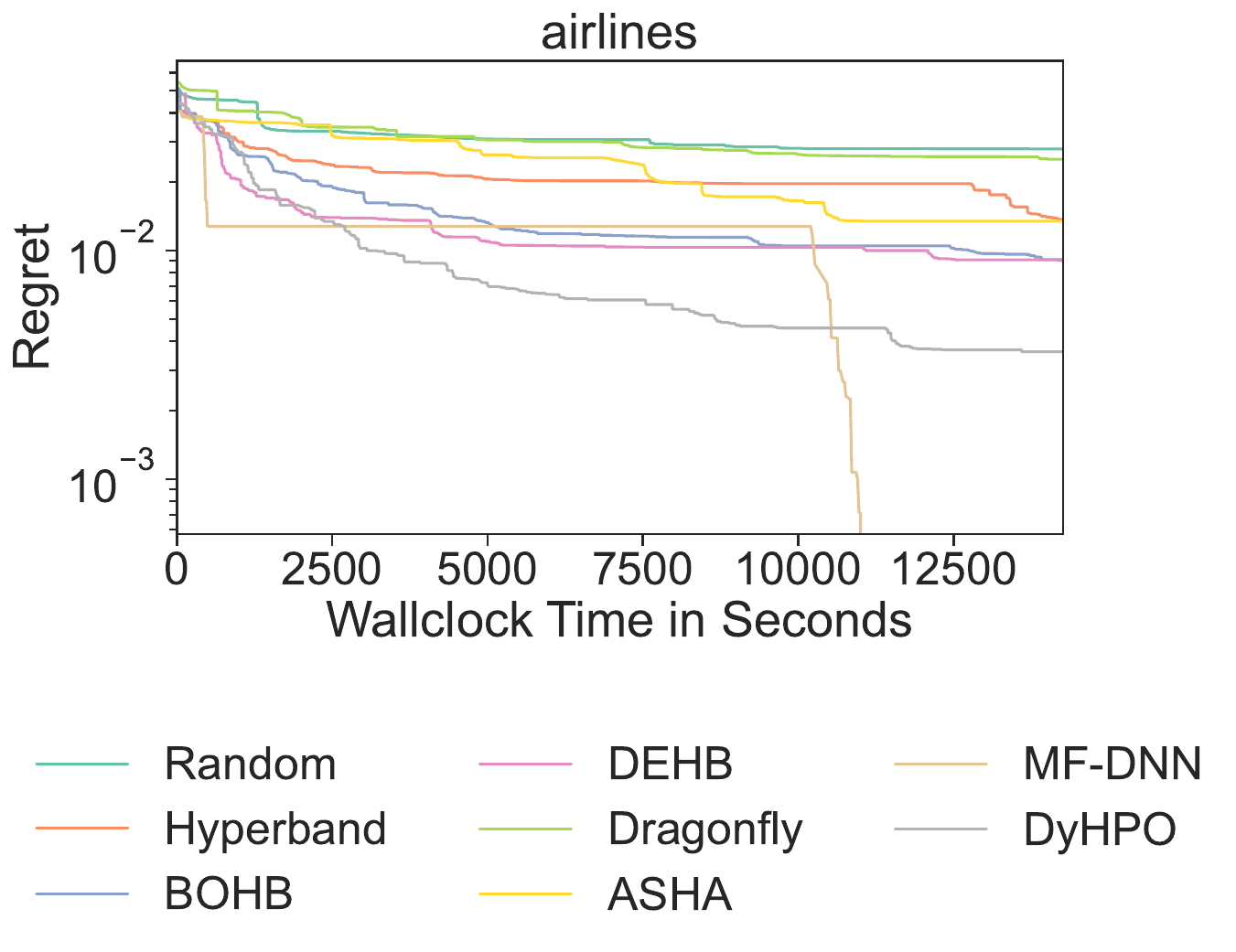}
    \includegraphics[width=0.28\textwidth]{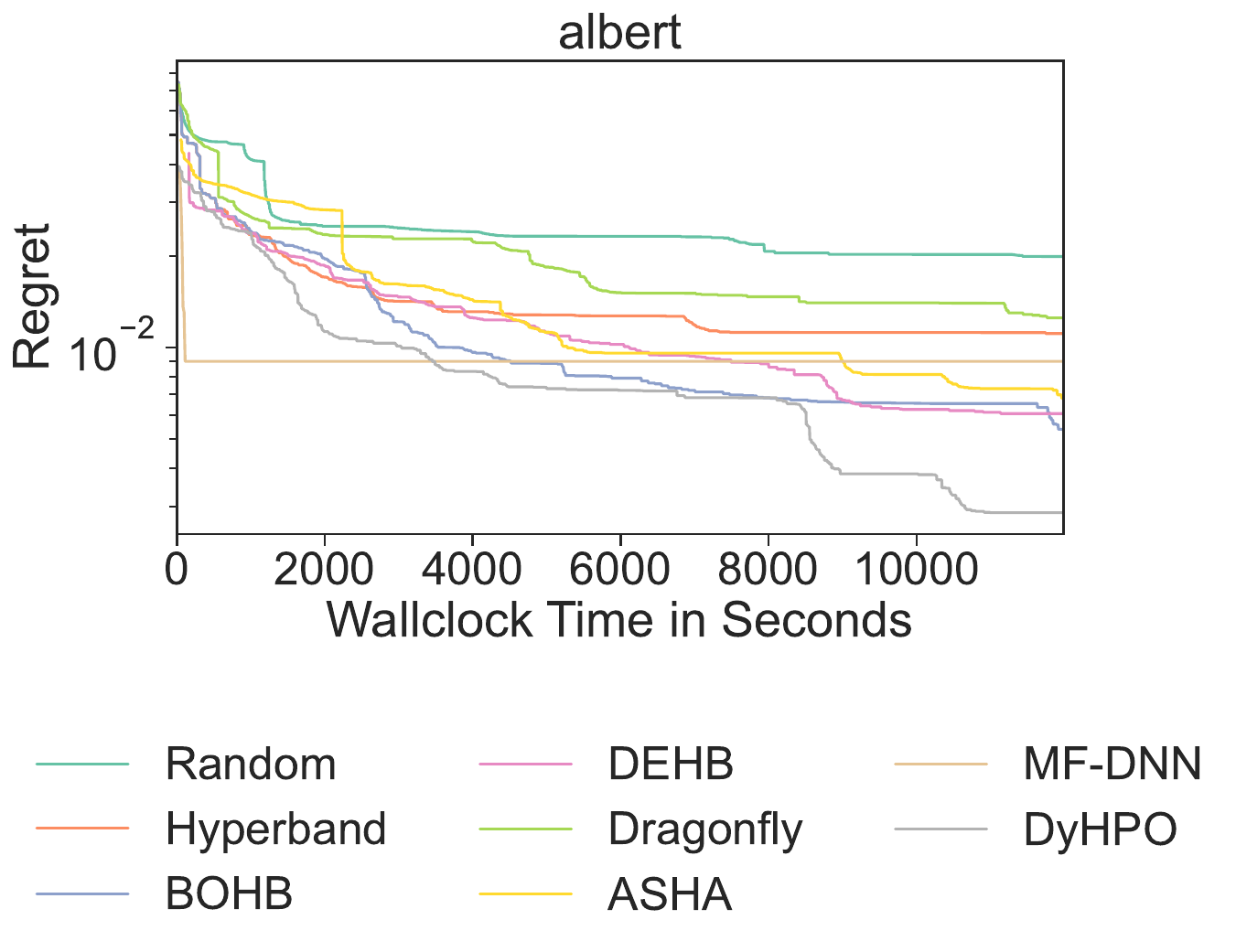}
    \includegraphics[width=0.28\textwidth]{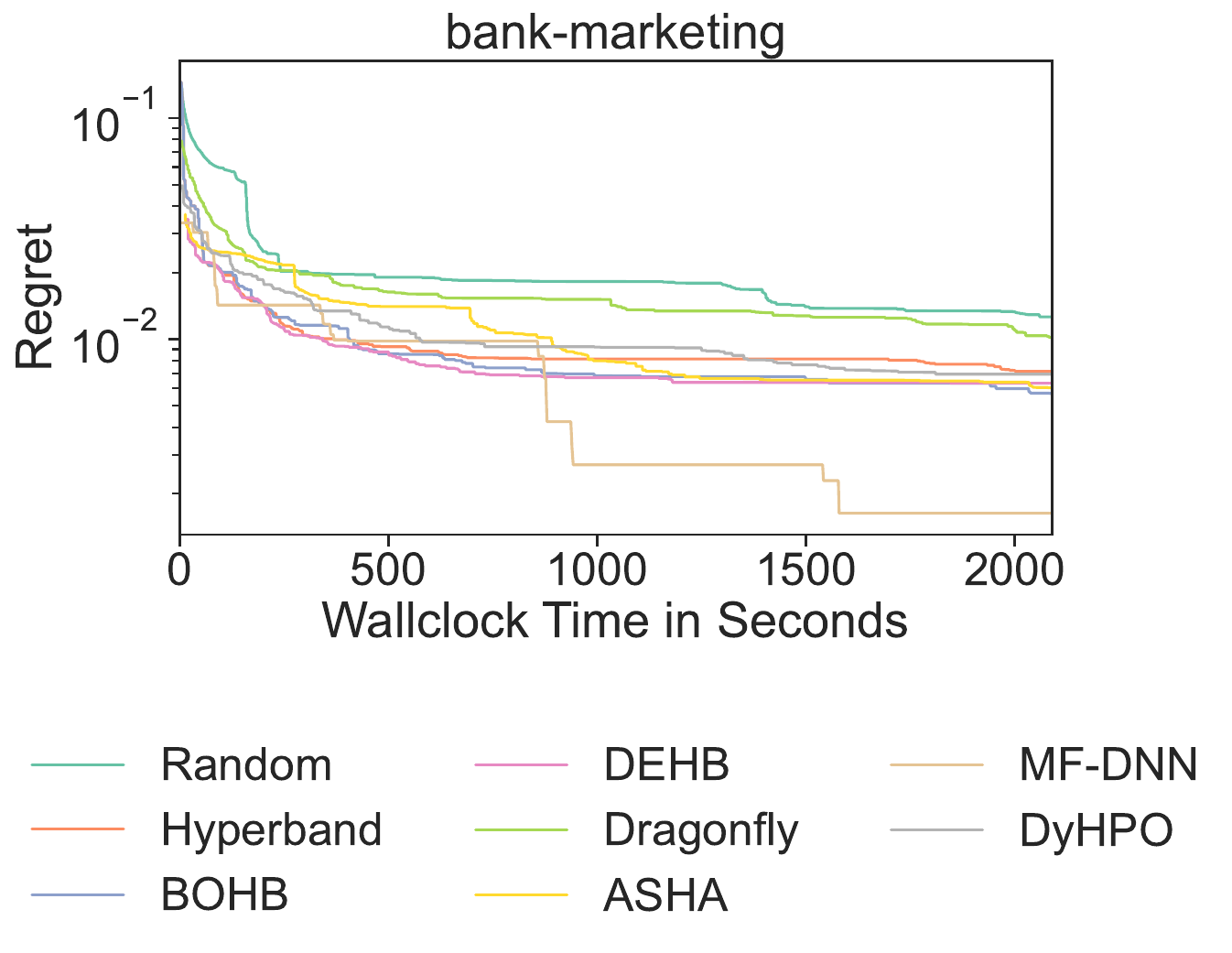}
    \includegraphics[width=0.28\textwidth]{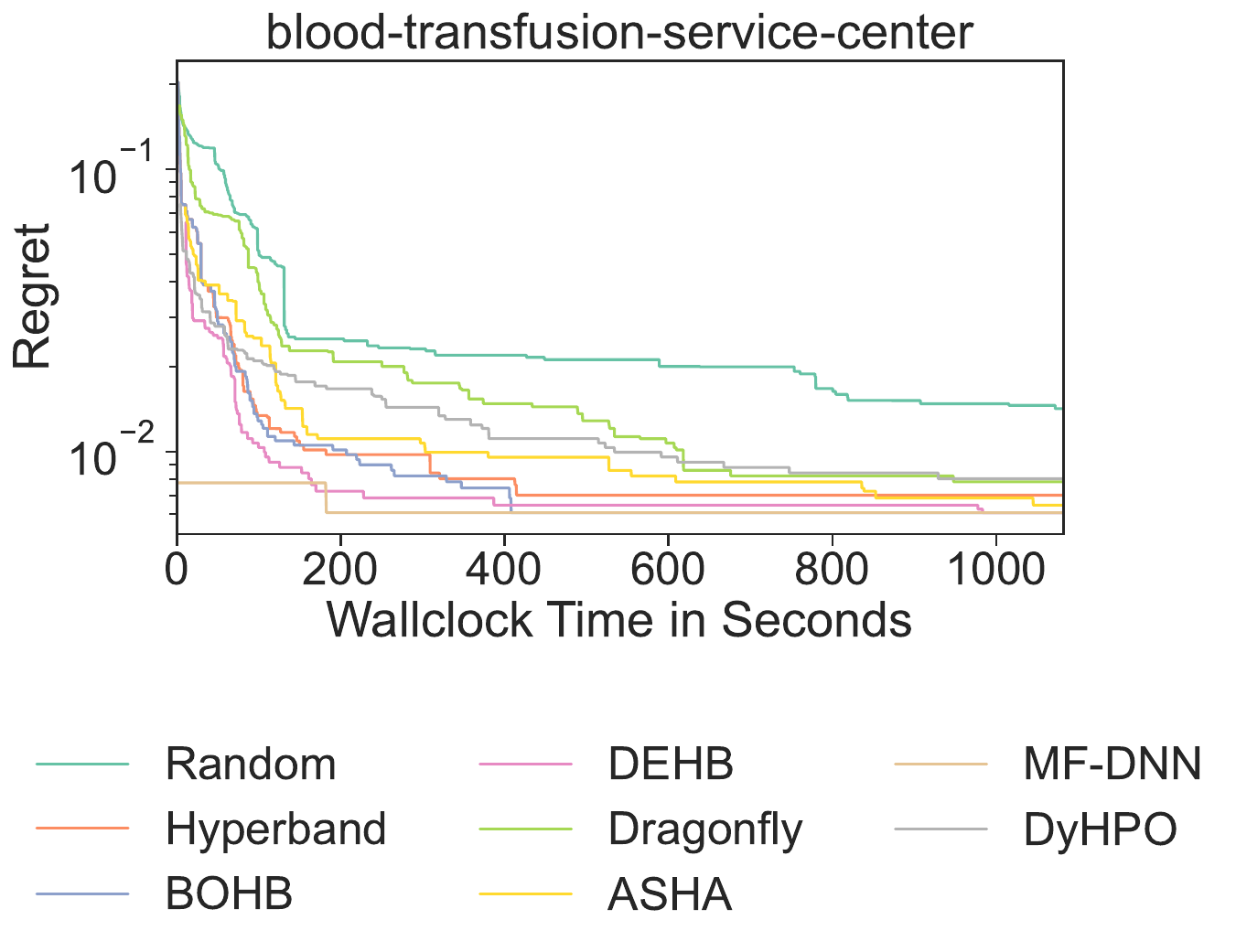}
    \includegraphics[width=0.28\textwidth]{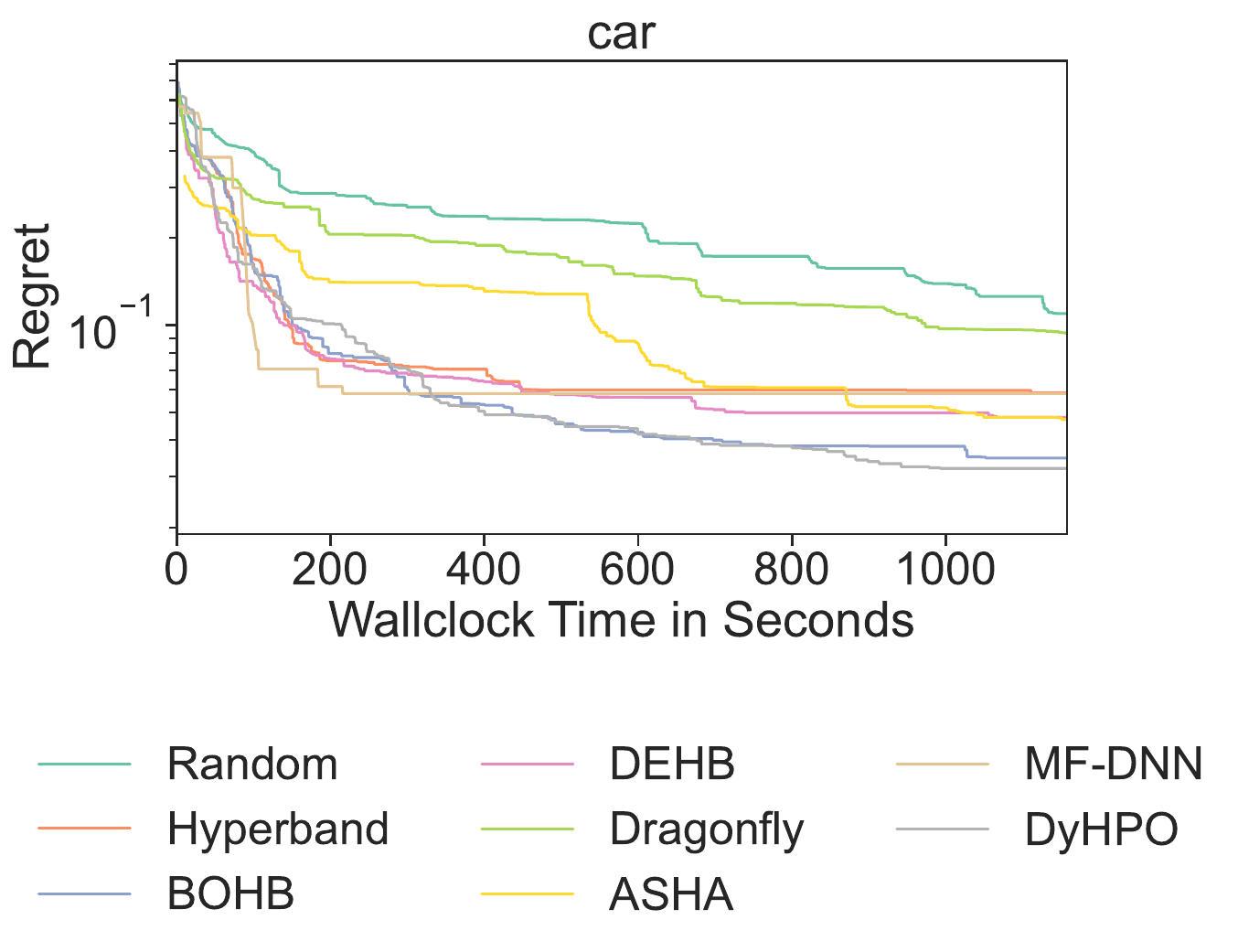}
    \includegraphics[width=0.28\textwidth]{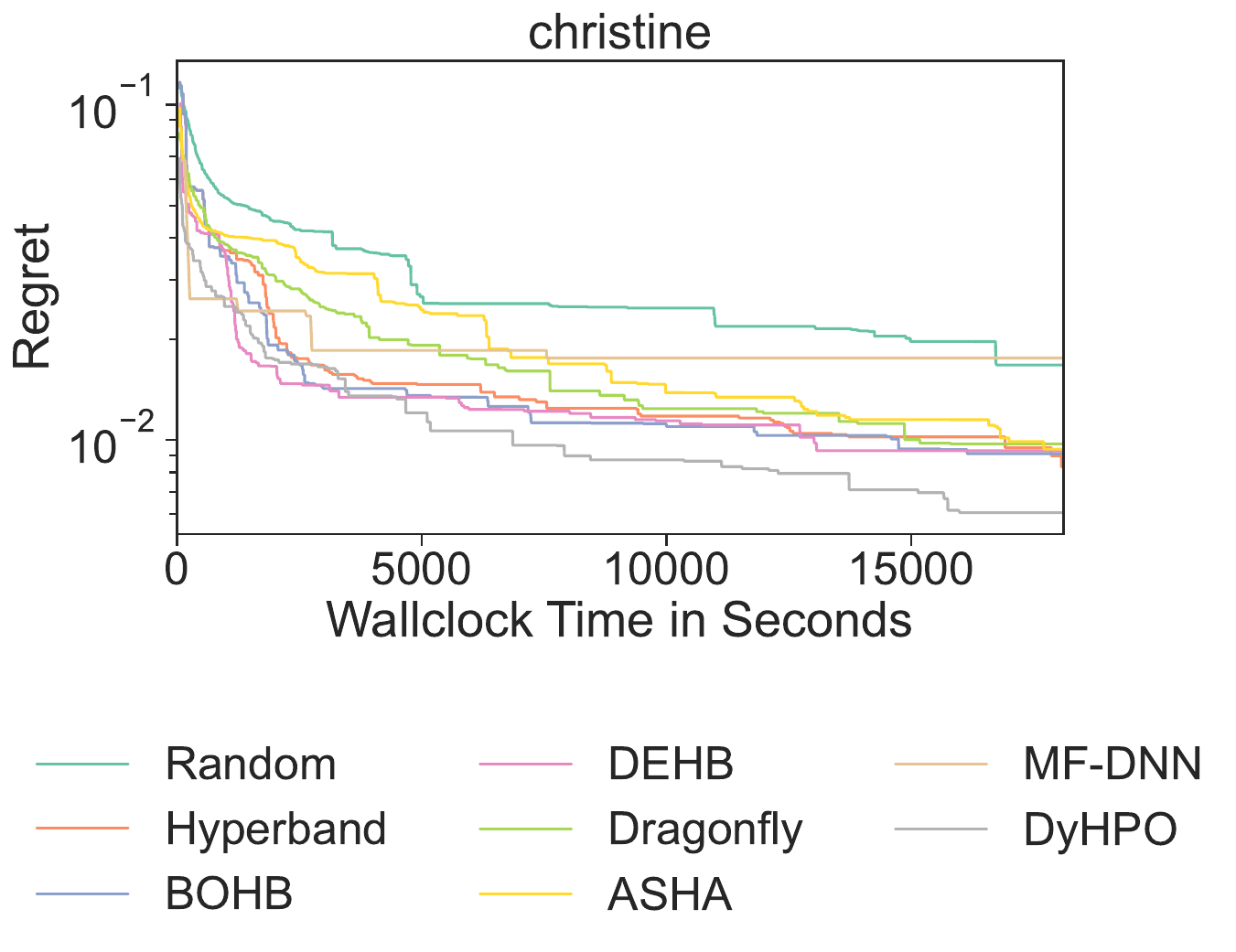}
    \includegraphics[width=0.28\textwidth]{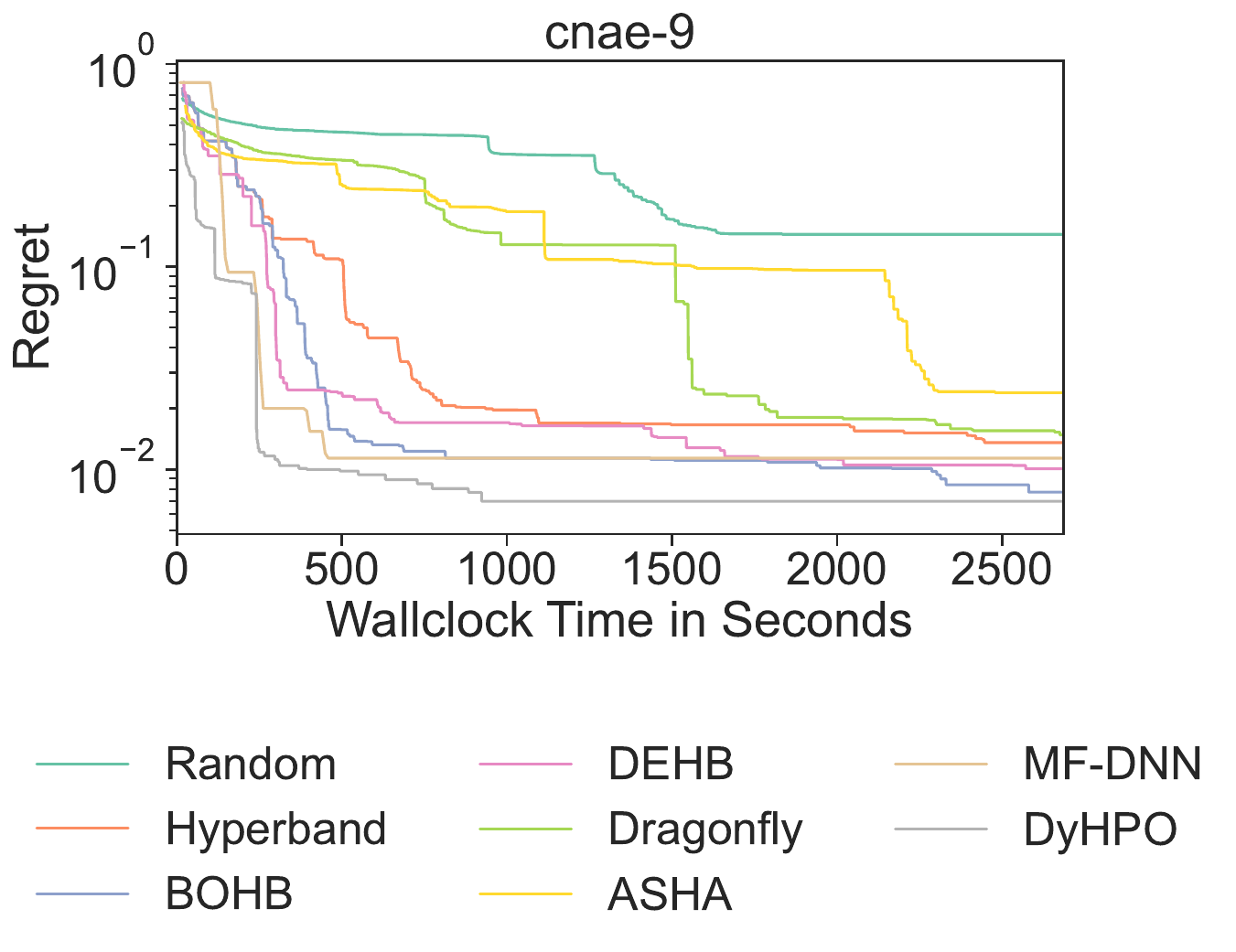}
    \includegraphics[width=0.28\textwidth]{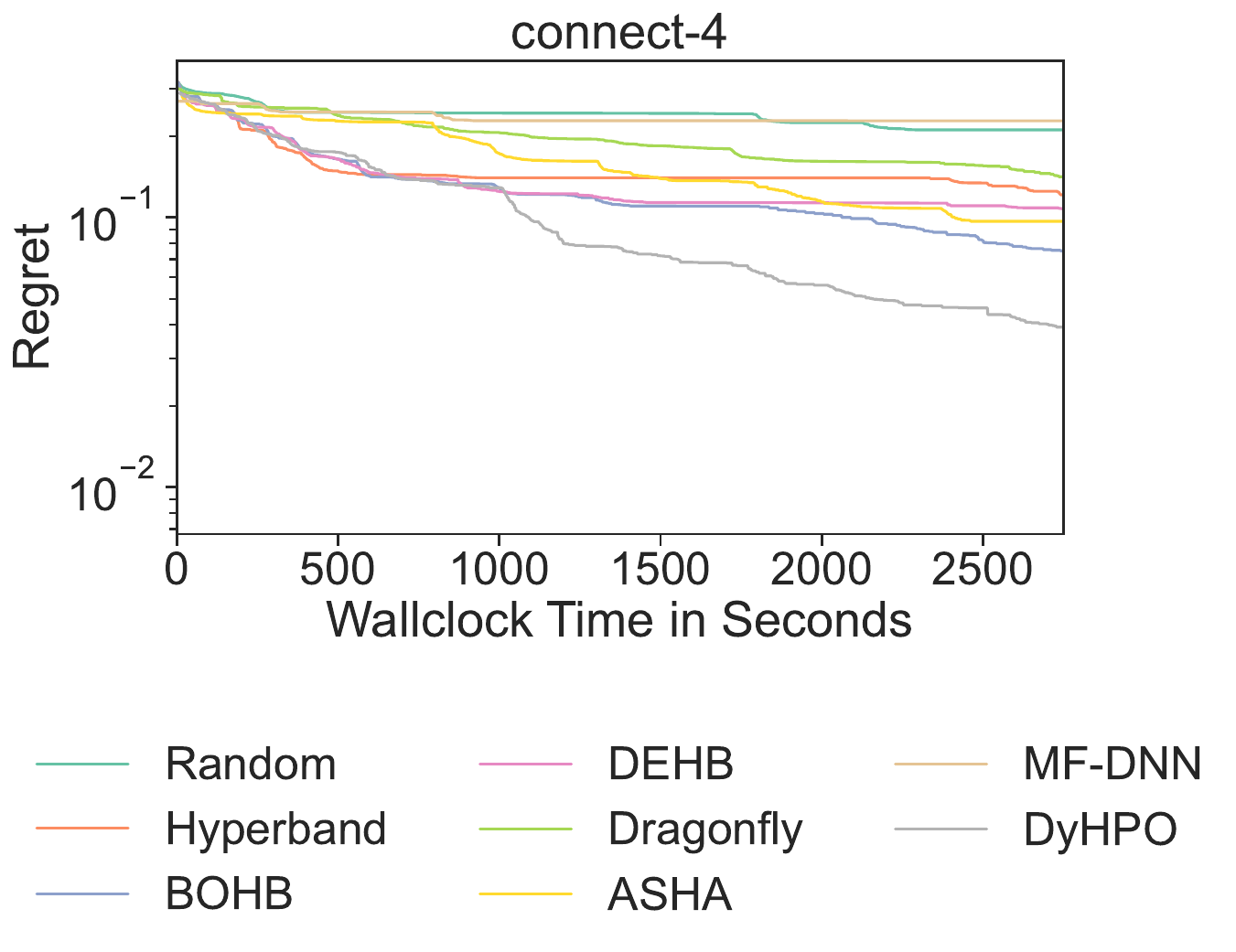}
    \includegraphics[width=0.28\textwidth]{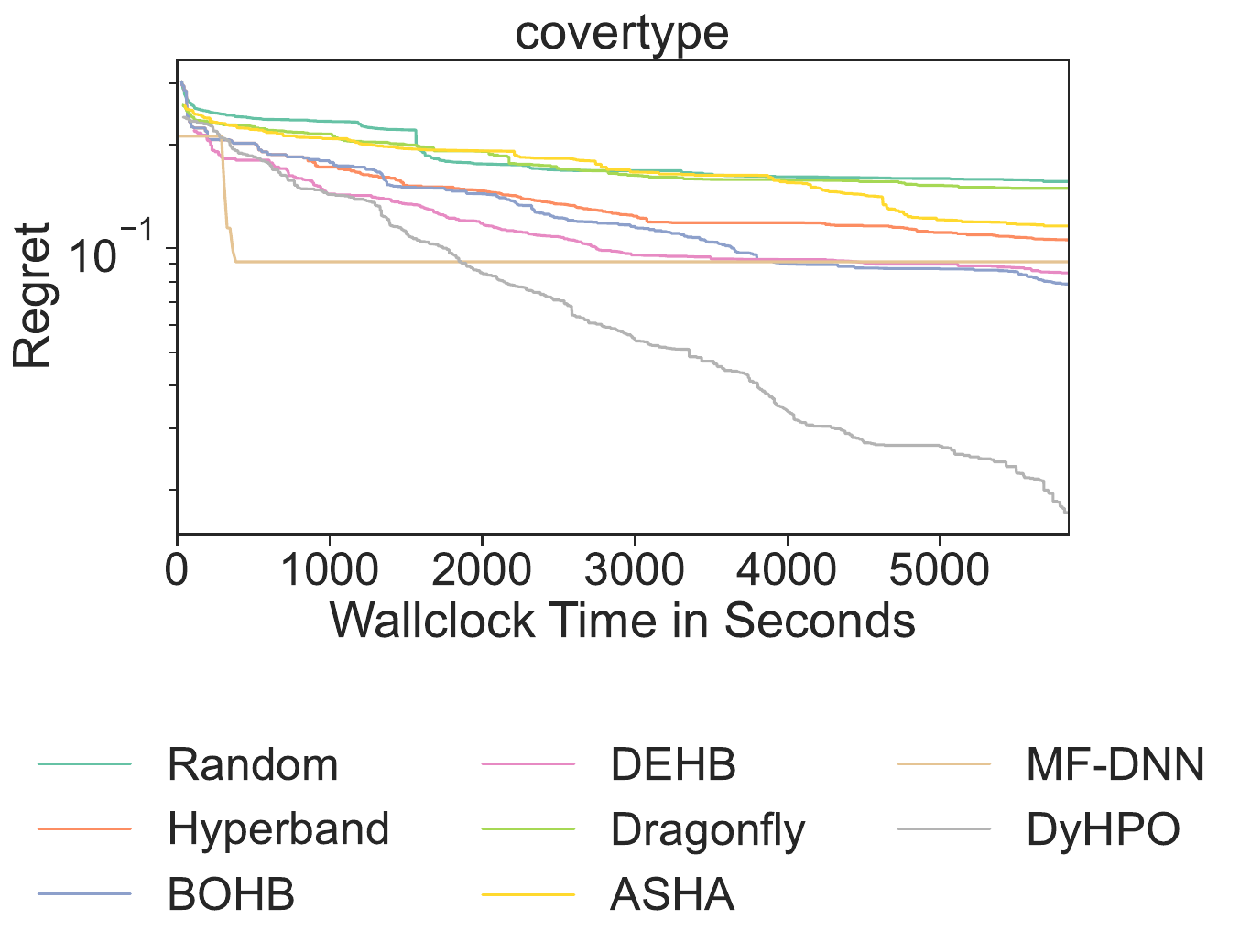}
    \includegraphics[width=0.28\textwidth]{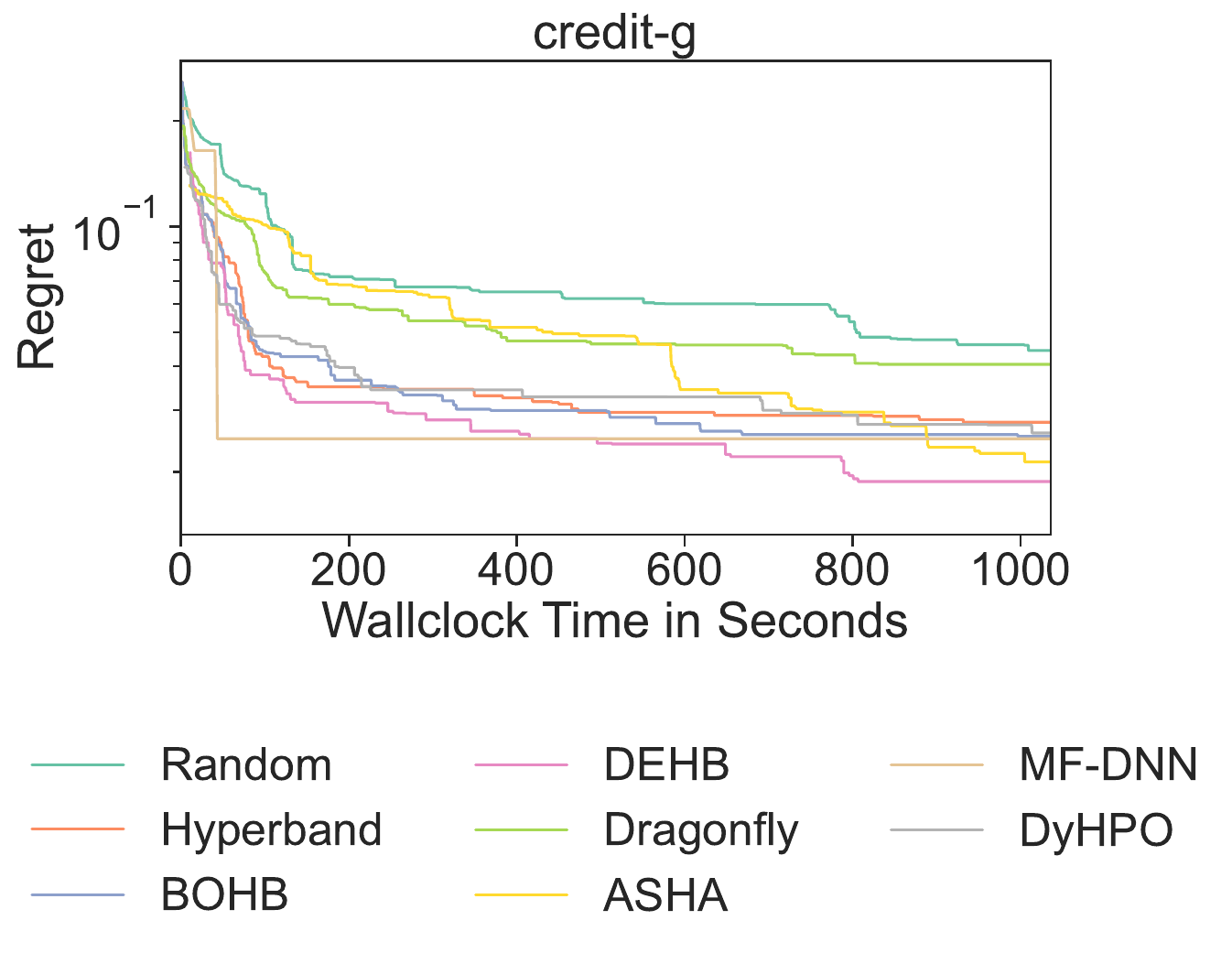}
    \includegraphics[width=0.28\textwidth]{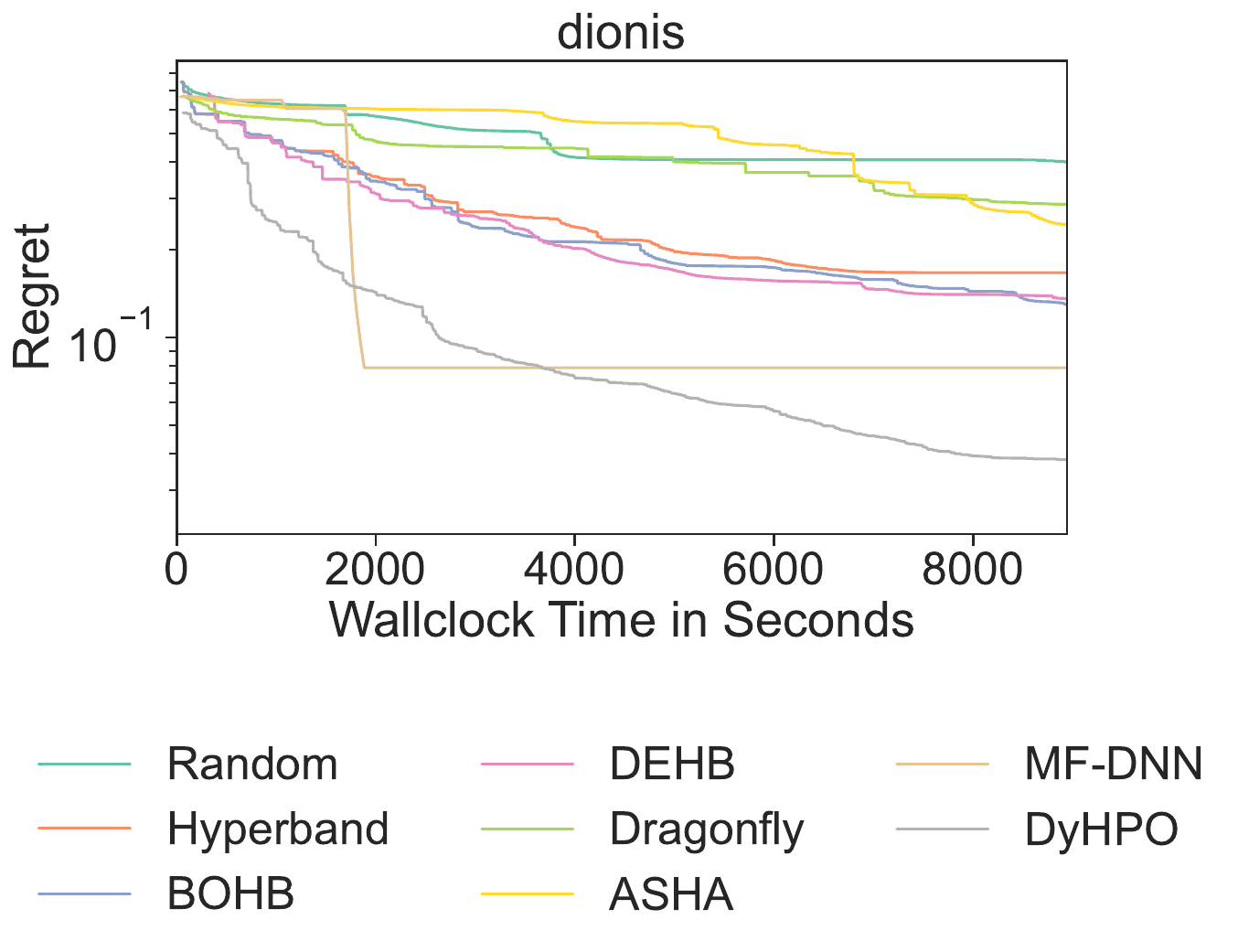}

  \caption{Performance comparison over time on a dataset level for LCBench with the overhead included.}
  \label{fig:results_per_dataset1_overhead}
\end{figure*}

\begin{figure*}[htp]
  \centering
    \includegraphics[width=0.28\textwidth]{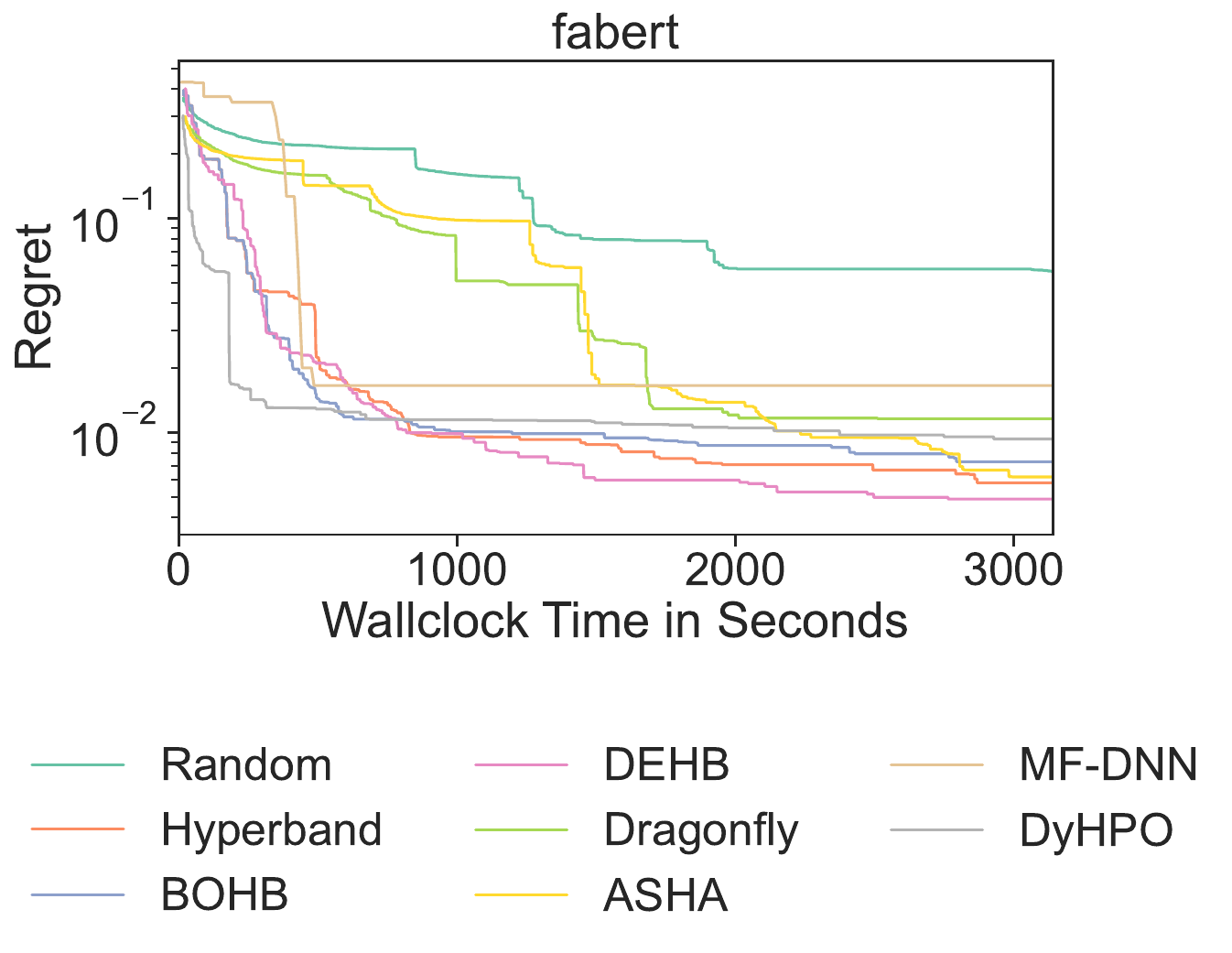}
    \includegraphics[width=0.28\textwidth]{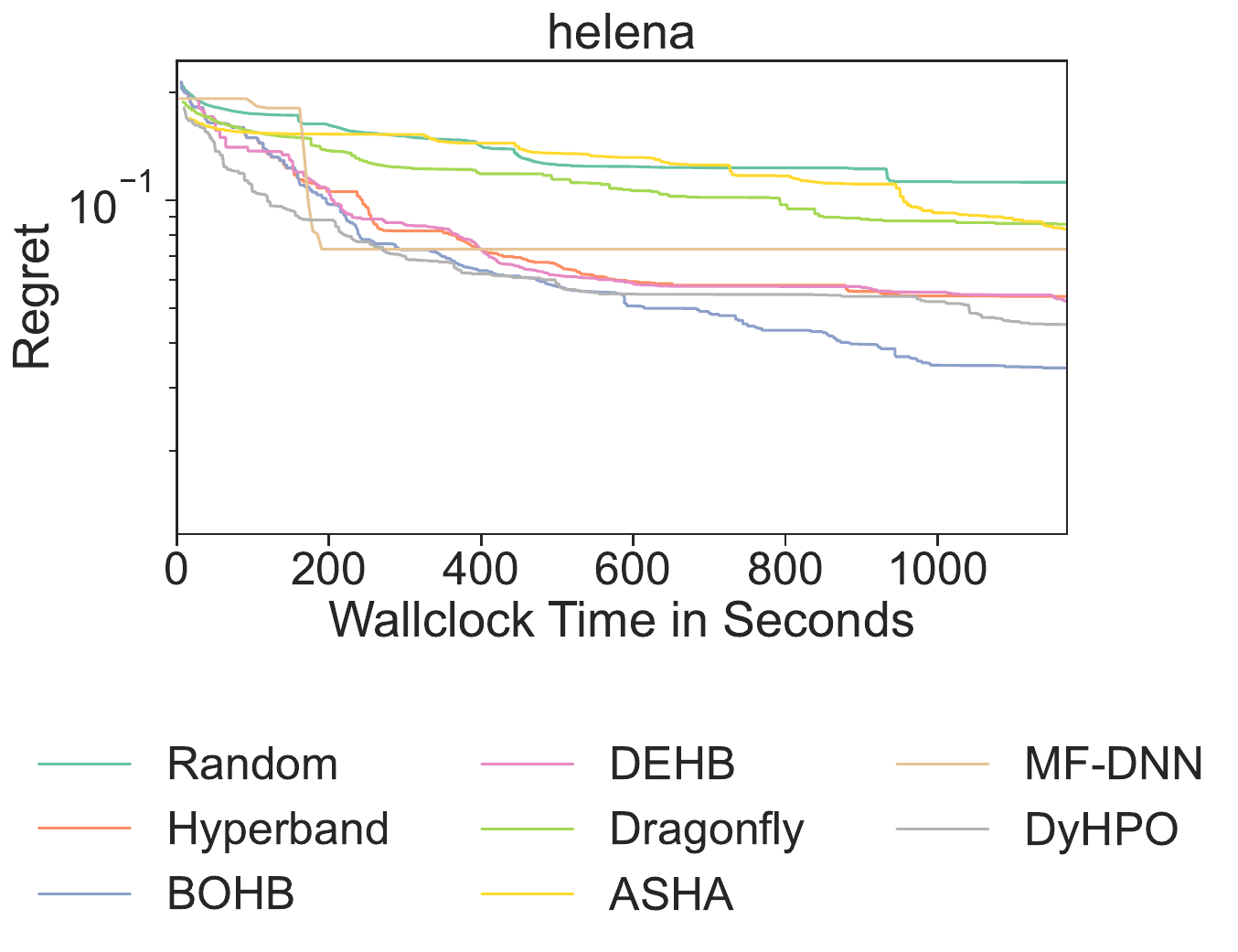}
    \includegraphics[width=0.28\textwidth]{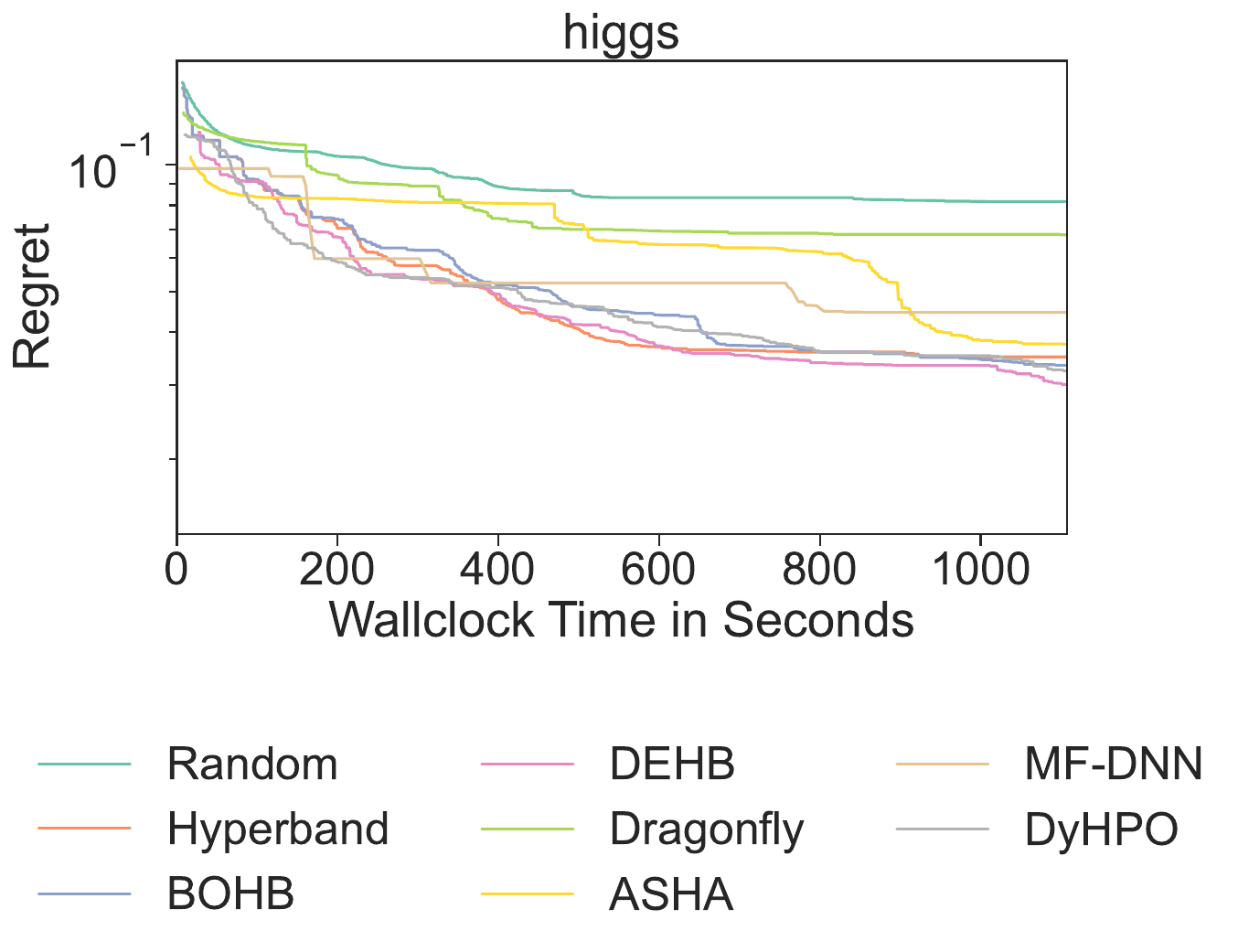}
    \includegraphics[width=0.28\textwidth]{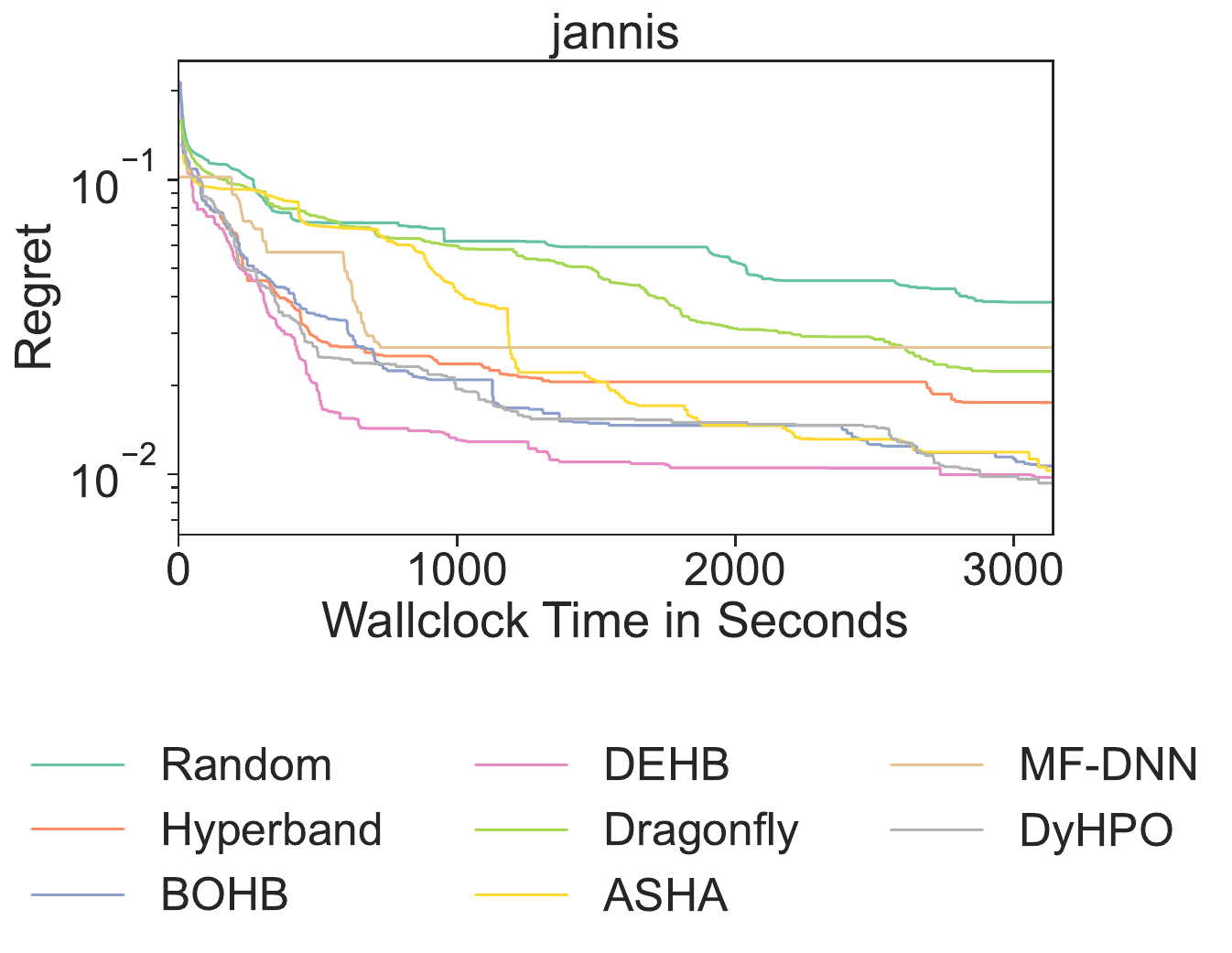}
    \includegraphics[width=0.28\textwidth]{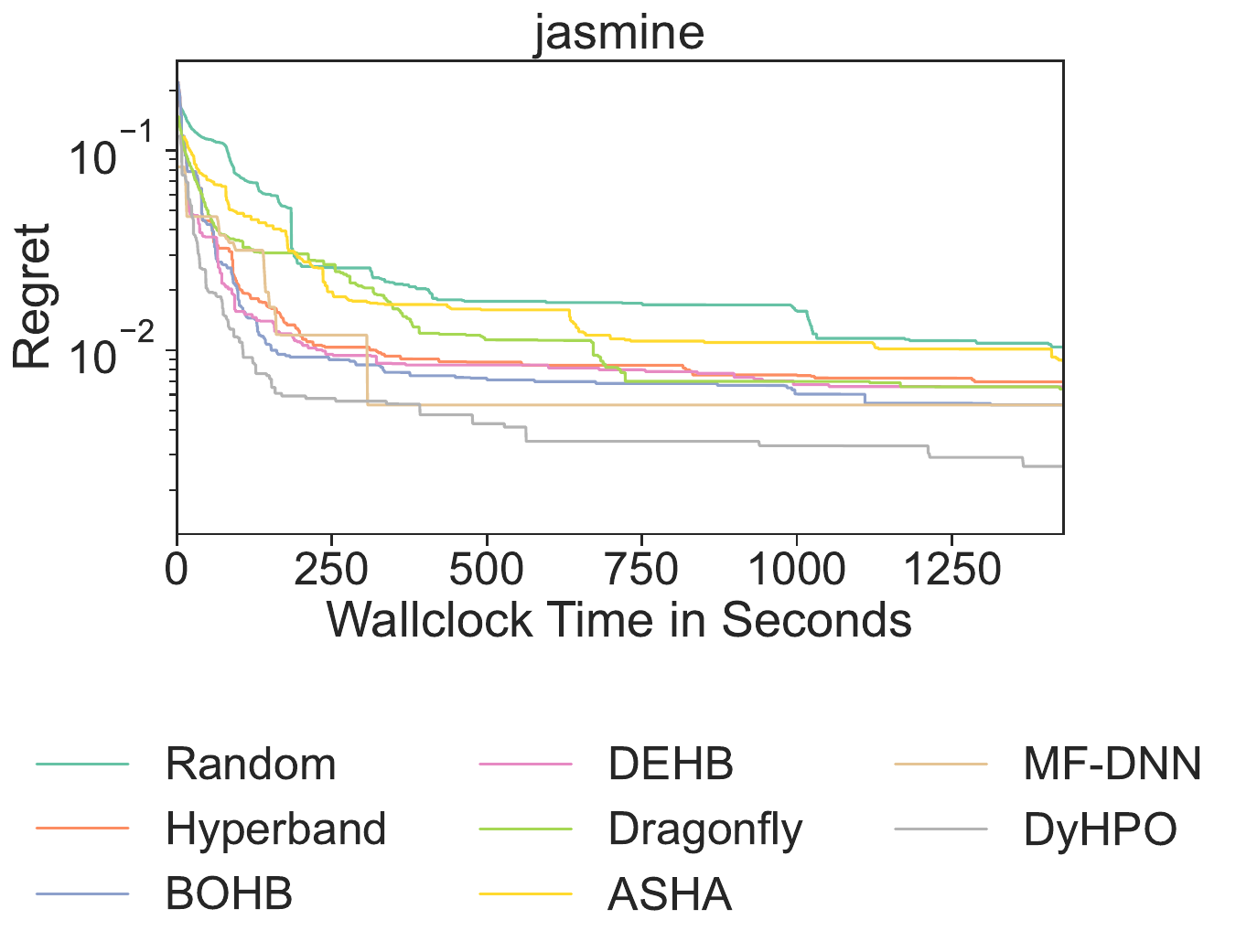}
    \includegraphics[width=0.28\textwidth]{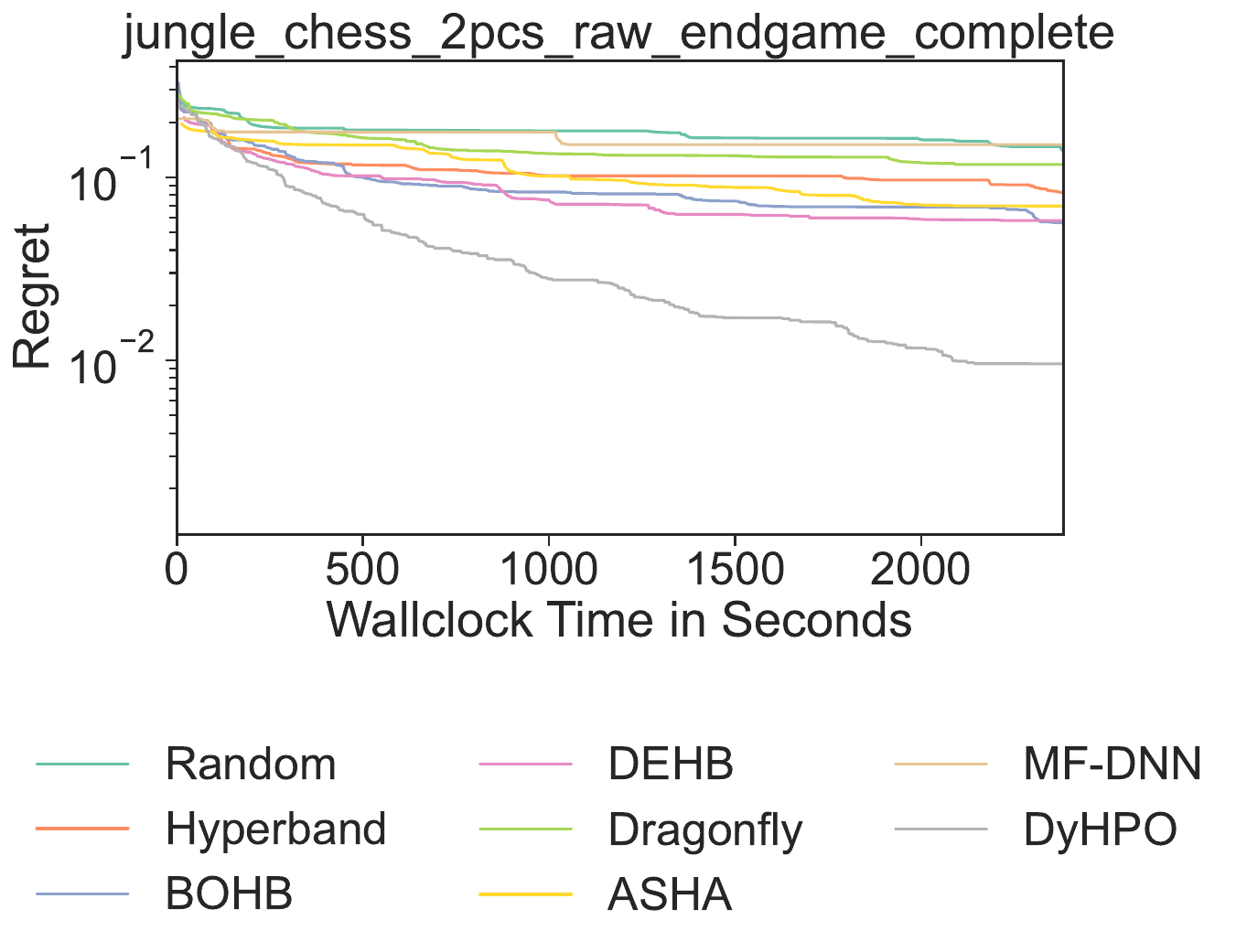}
    \includegraphics[width=0.28\textwidth]{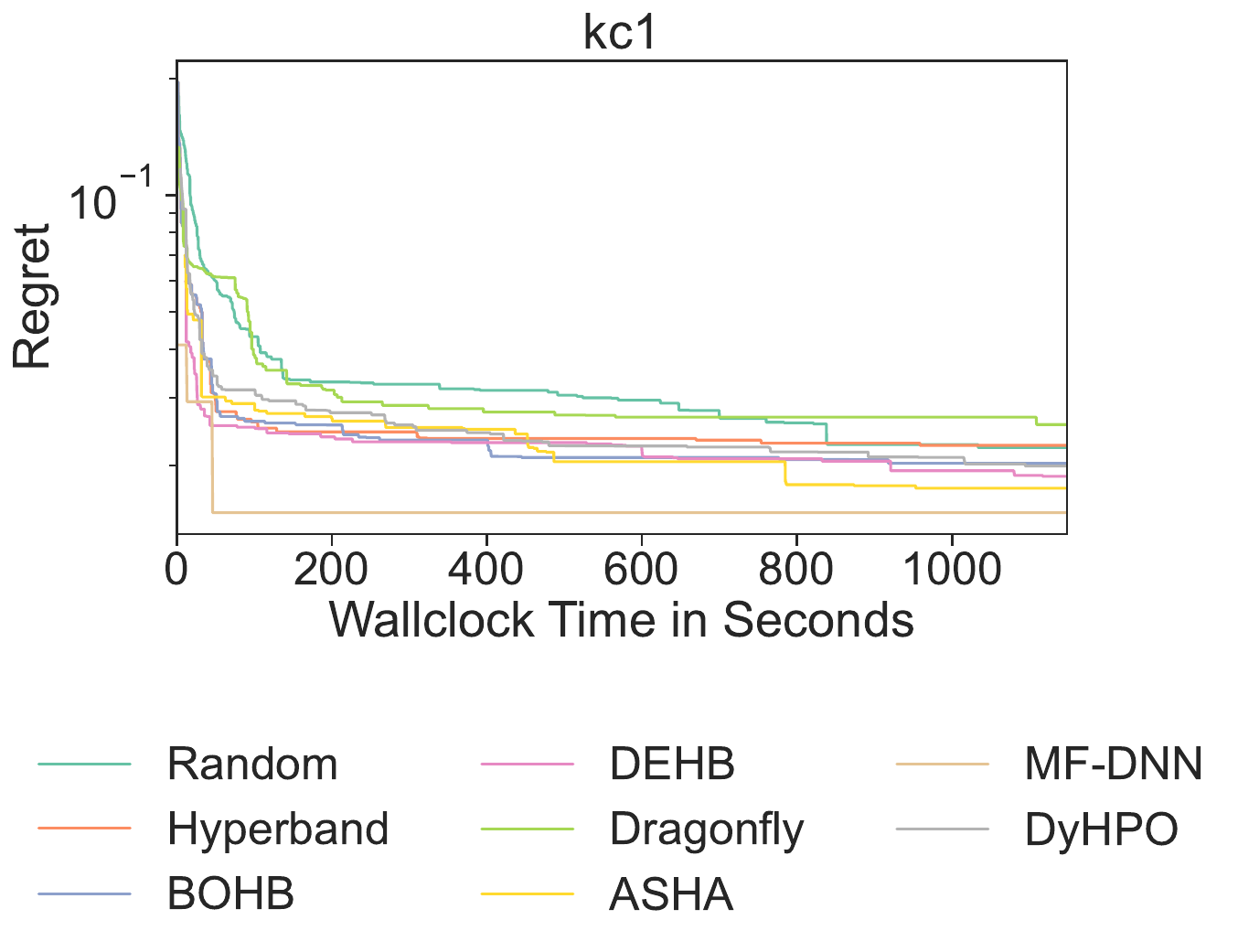}
    \includegraphics[width=0.28\textwidth]{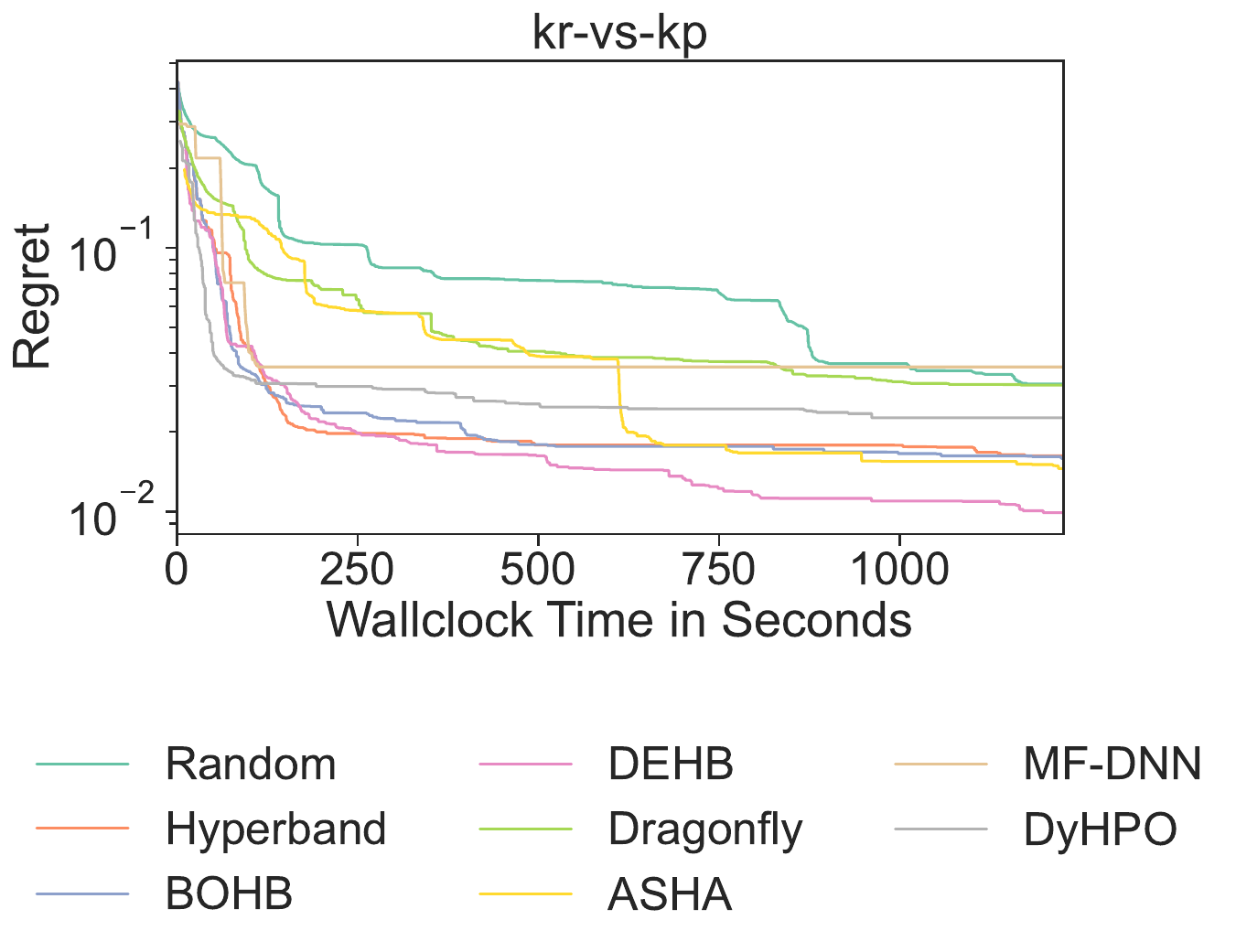}
    \includegraphics[width=0.28\textwidth]{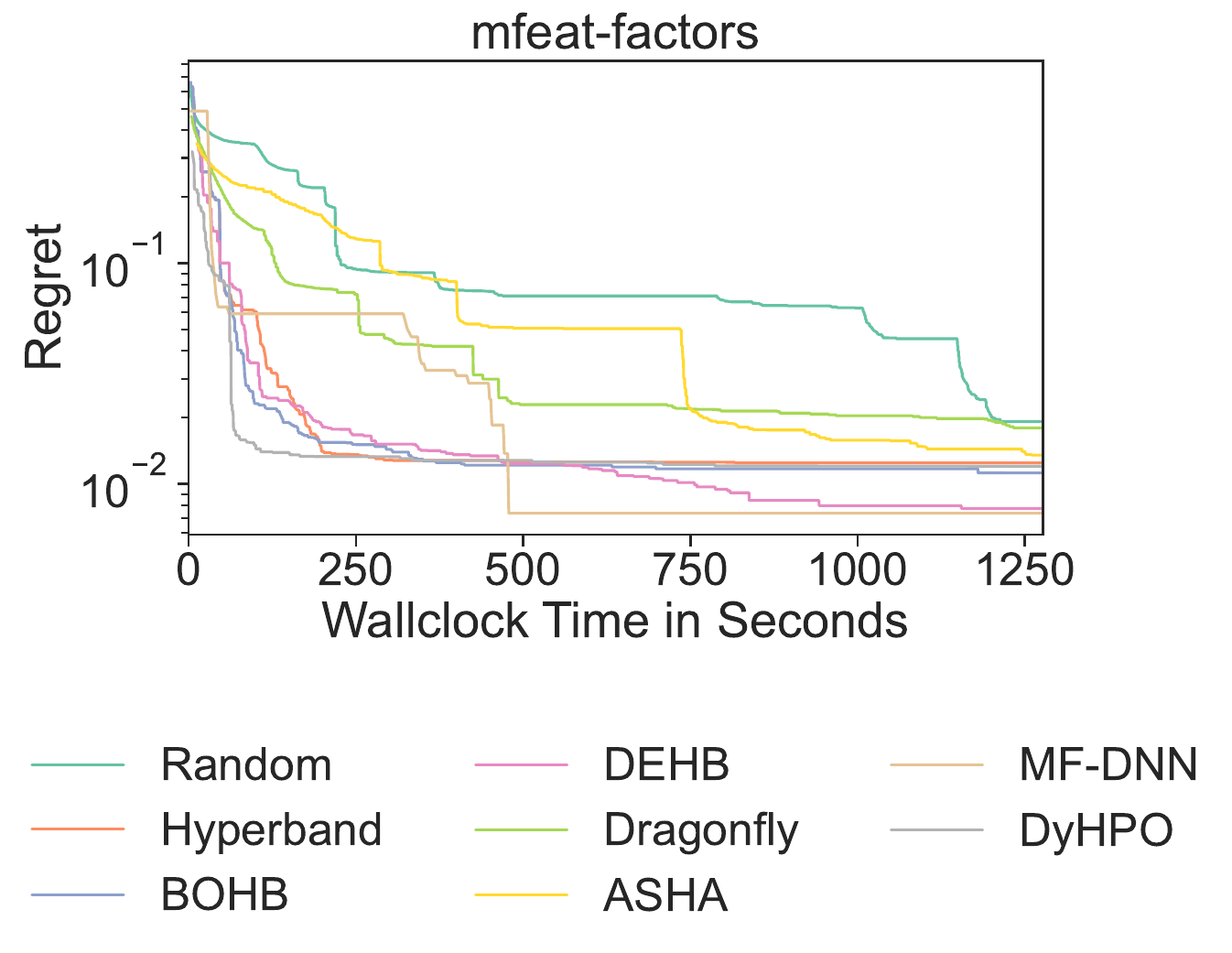}
    \includegraphics[width=0.28\textwidth]{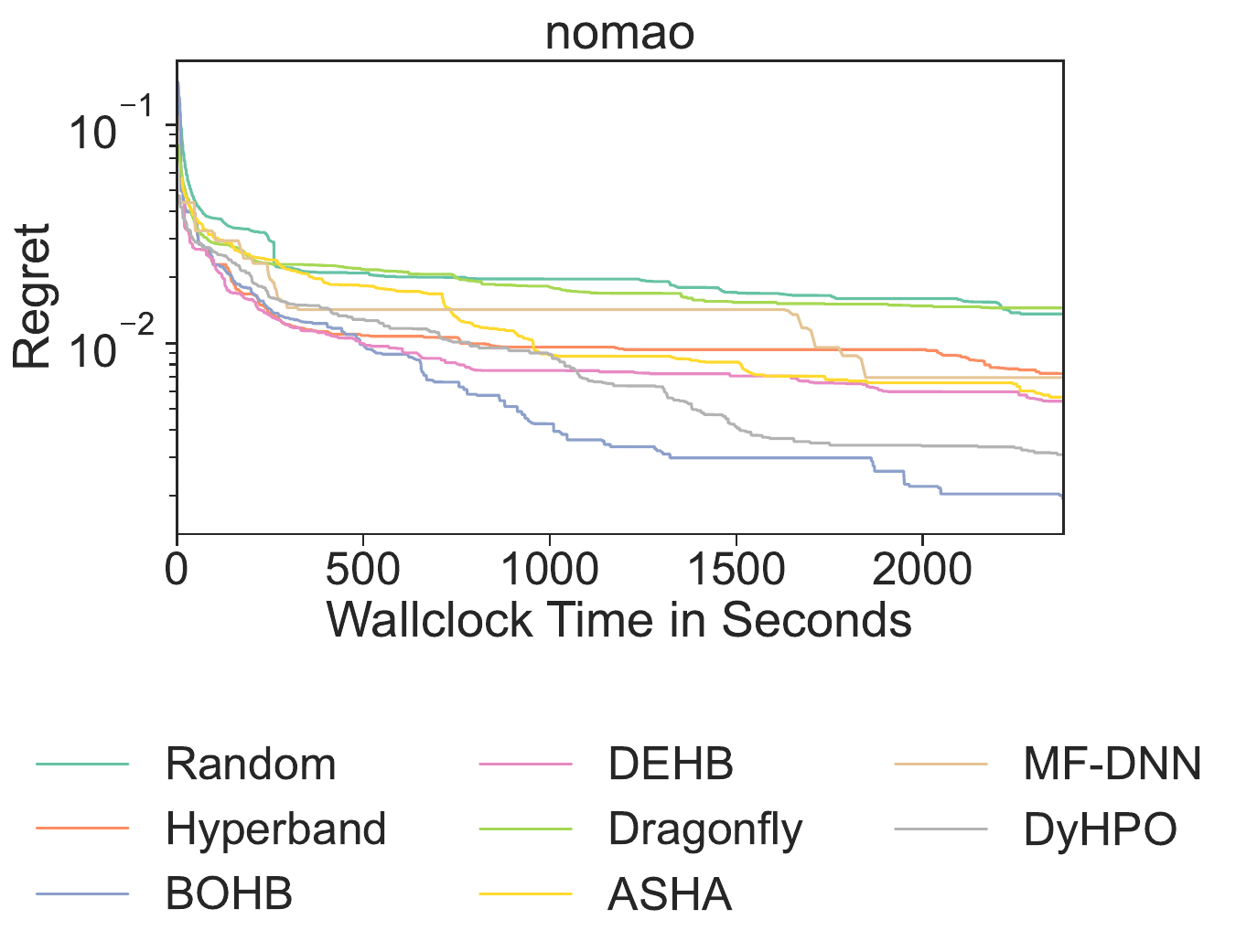}
    \includegraphics[width=0.28\textwidth]{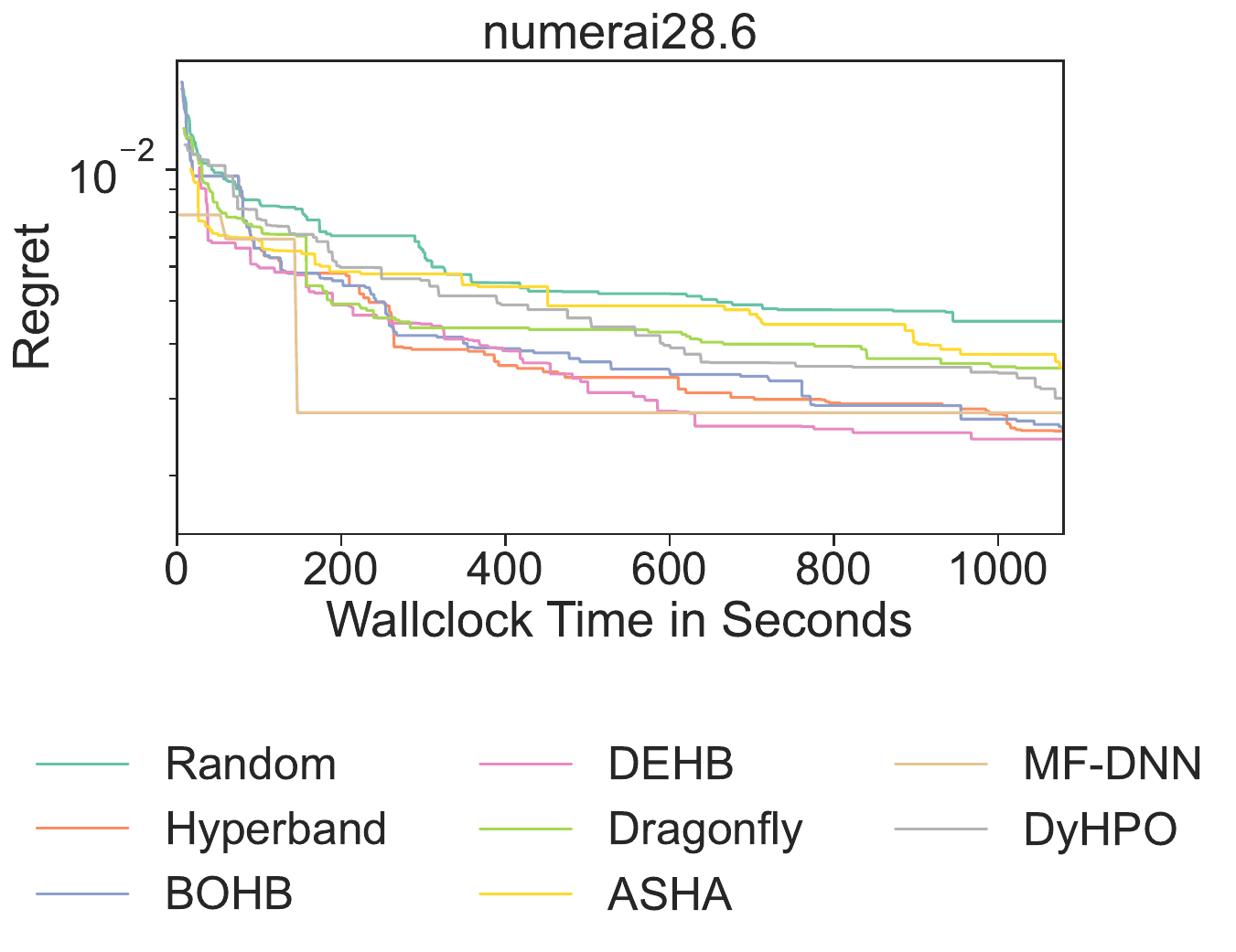}
    \includegraphics[width=0.28\textwidth]{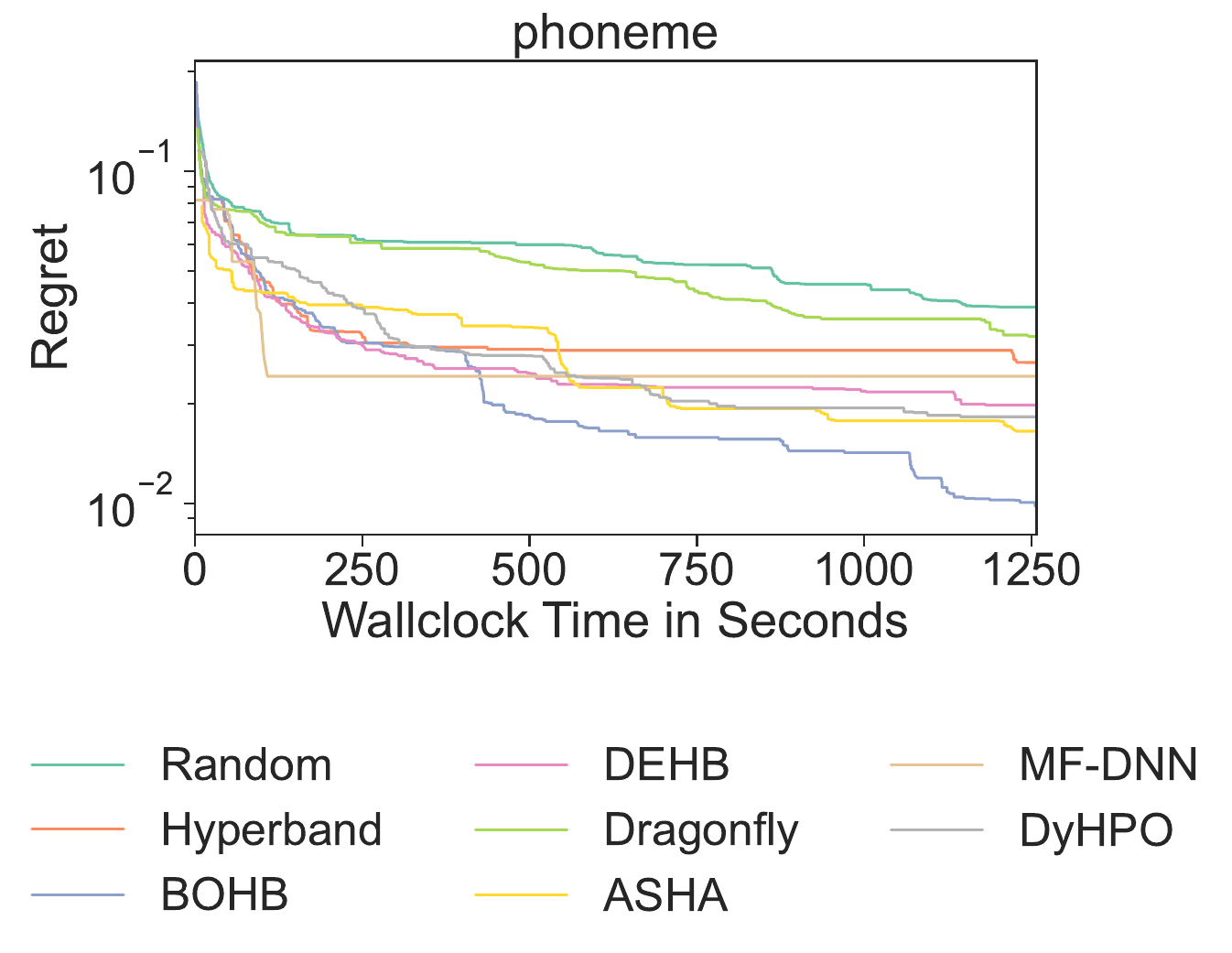}
    \includegraphics[width=0.28\textwidth]{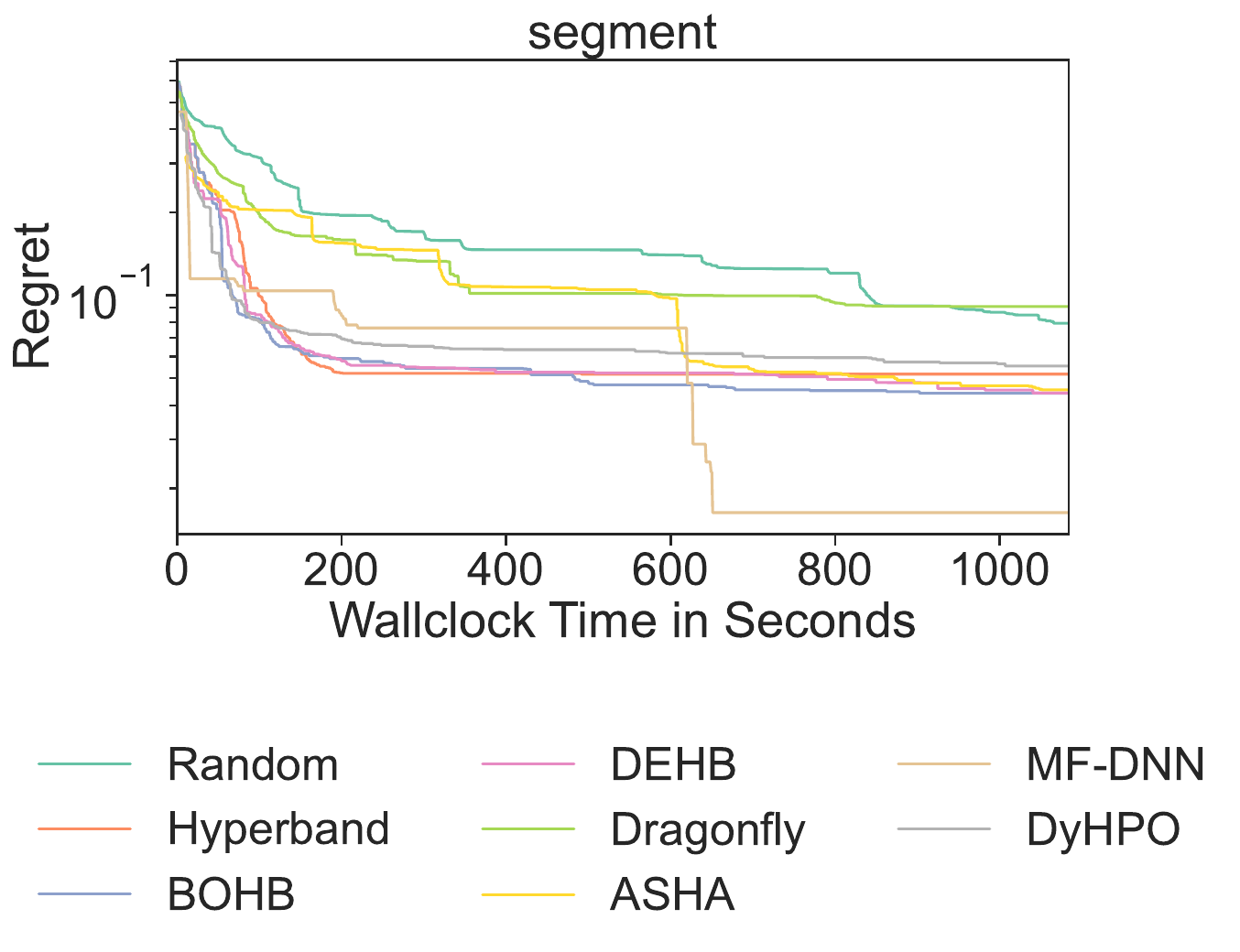}
    \includegraphics[width=0.28\textwidth]{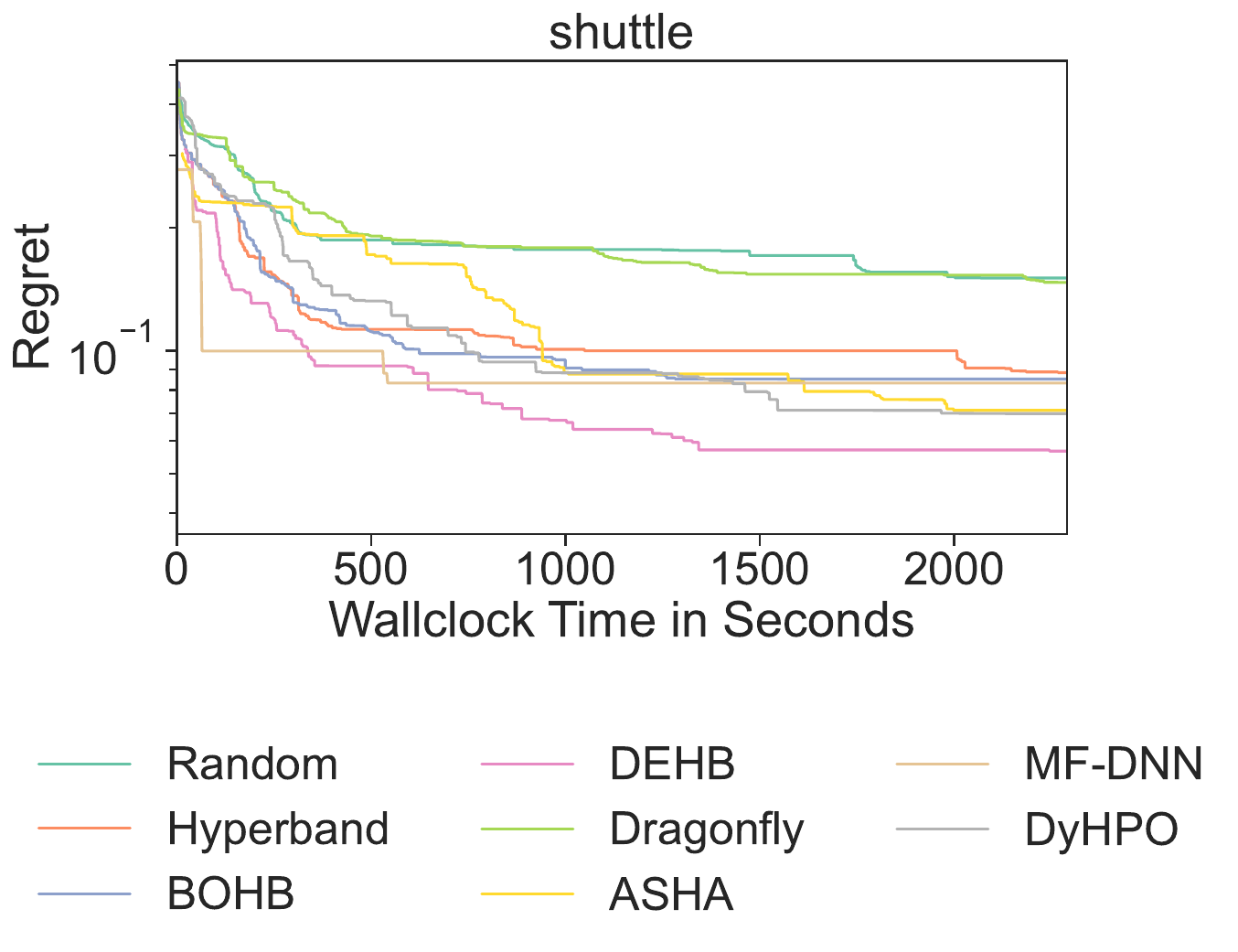}
    \includegraphics[width=0.28\textwidth]{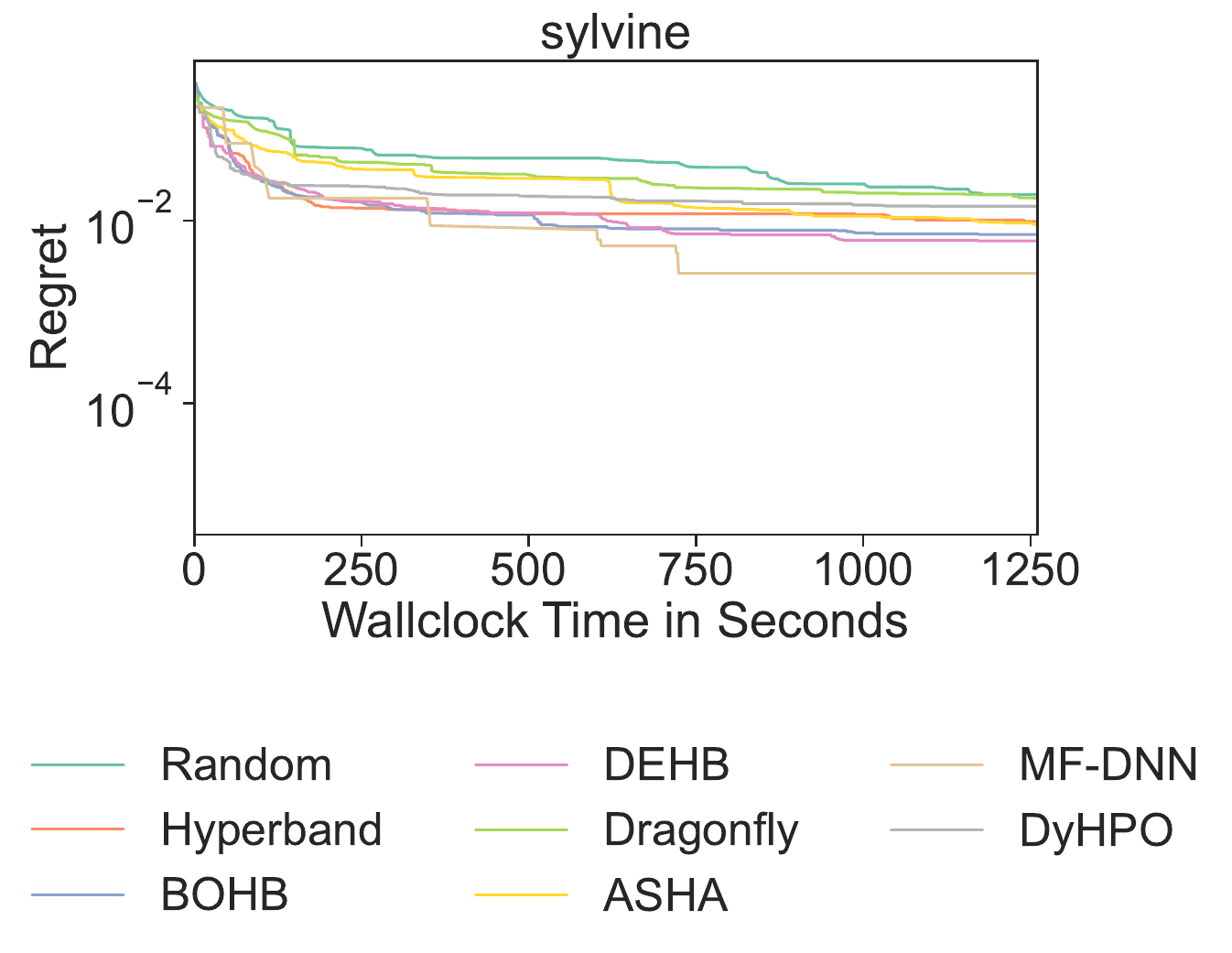}
    \includegraphics[width=0.28\textwidth]{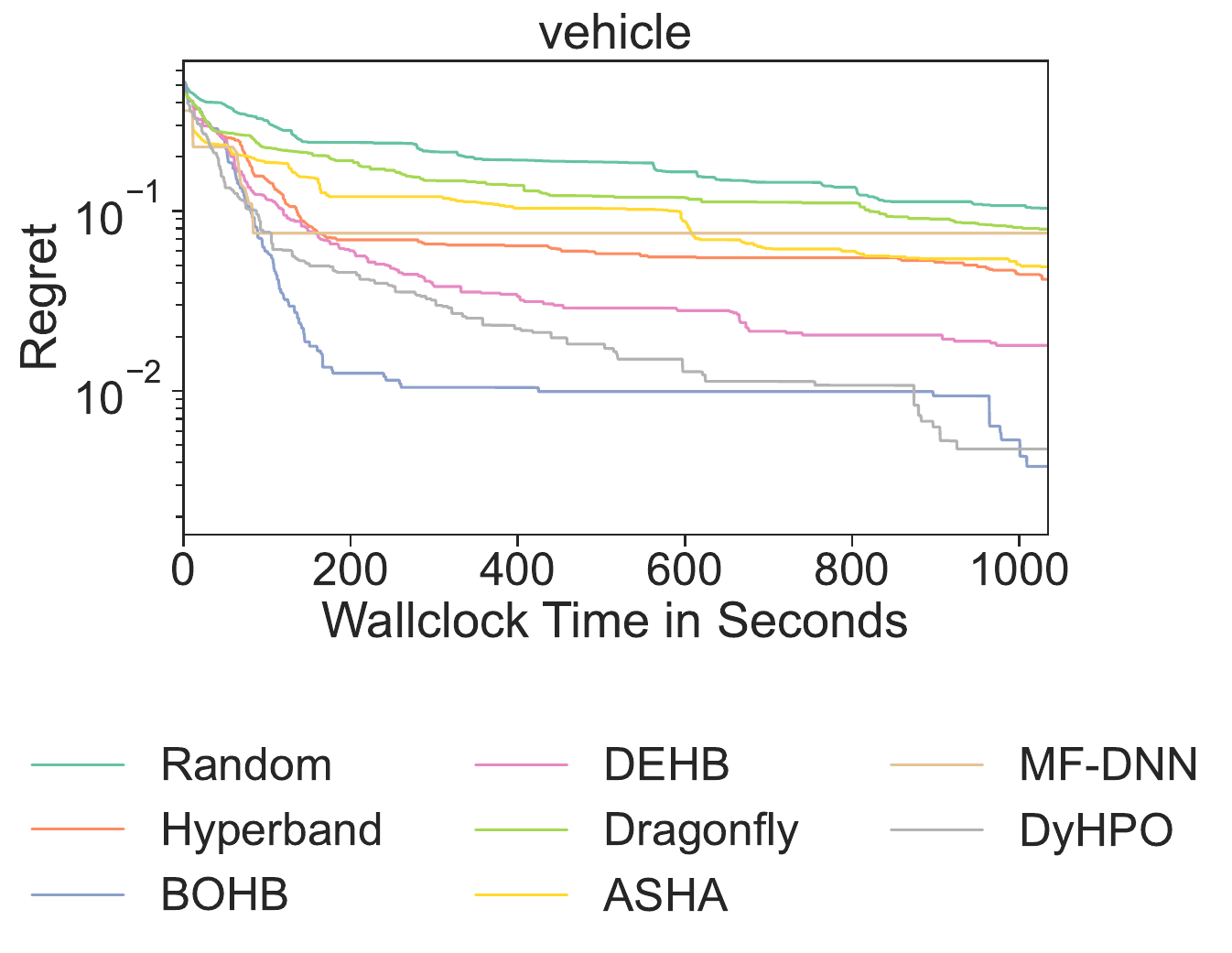}
    \includegraphics[width=0.28\textwidth]{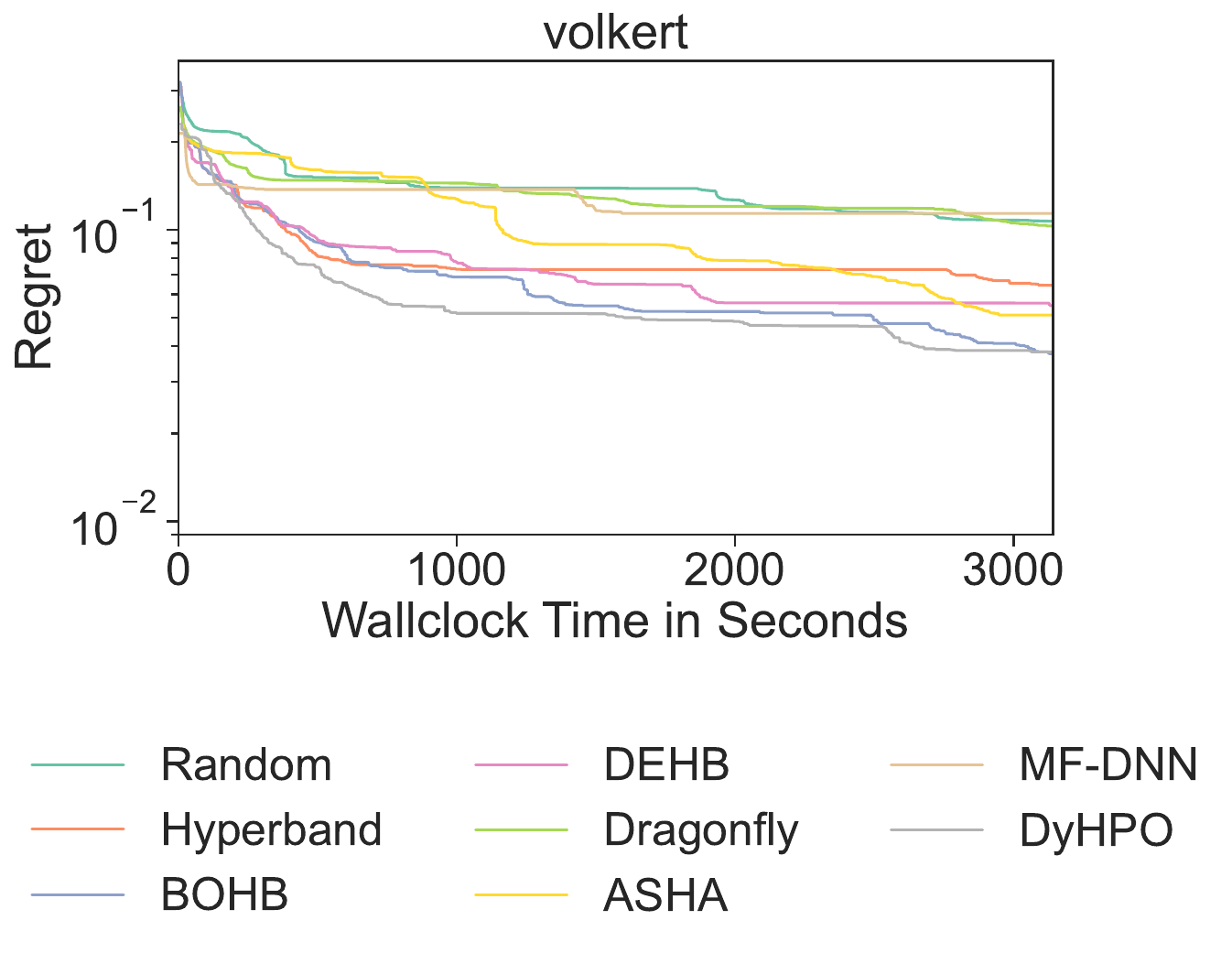}

  \caption{Performance comparison over time on a dataset level for LCBench with the overhead included. (cont.).}
  \label{fig:results_per_dataset2_overhead}
\end{figure*}

In Figure~\ref{fig:results_per_dataset1} and \ref{fig:results_per_dataset2}, we show the performance comparison for all the datasets from LCBench regarding regret over the number of epochs. Similarly, in Figure~\ref{fig:results_per_dataset1_overhead} and \ref{fig:results_per_dataset2_overhead}, we show the same performance comparison, however, over time. As can be seen, \algabbr{} manages to outperform the other competitors in the majority of the datasets, and in the datasets that it does not, it is always close to the top-performing method, and the difference between methods is marginal.

\begin{figure*}[t]
  \centering
    \includegraphics[width=0.32\textwidth]{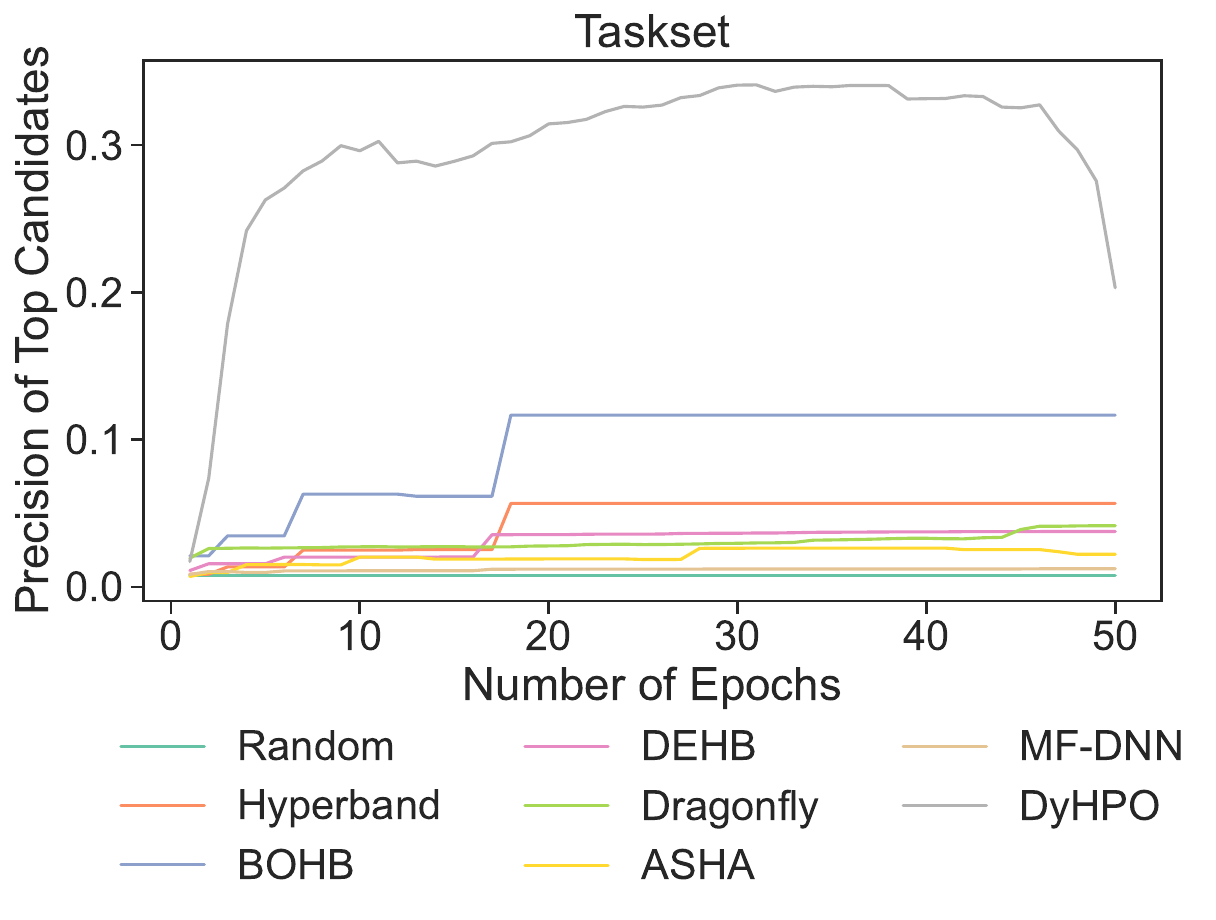}
    \includegraphics[width=0.32\textwidth]{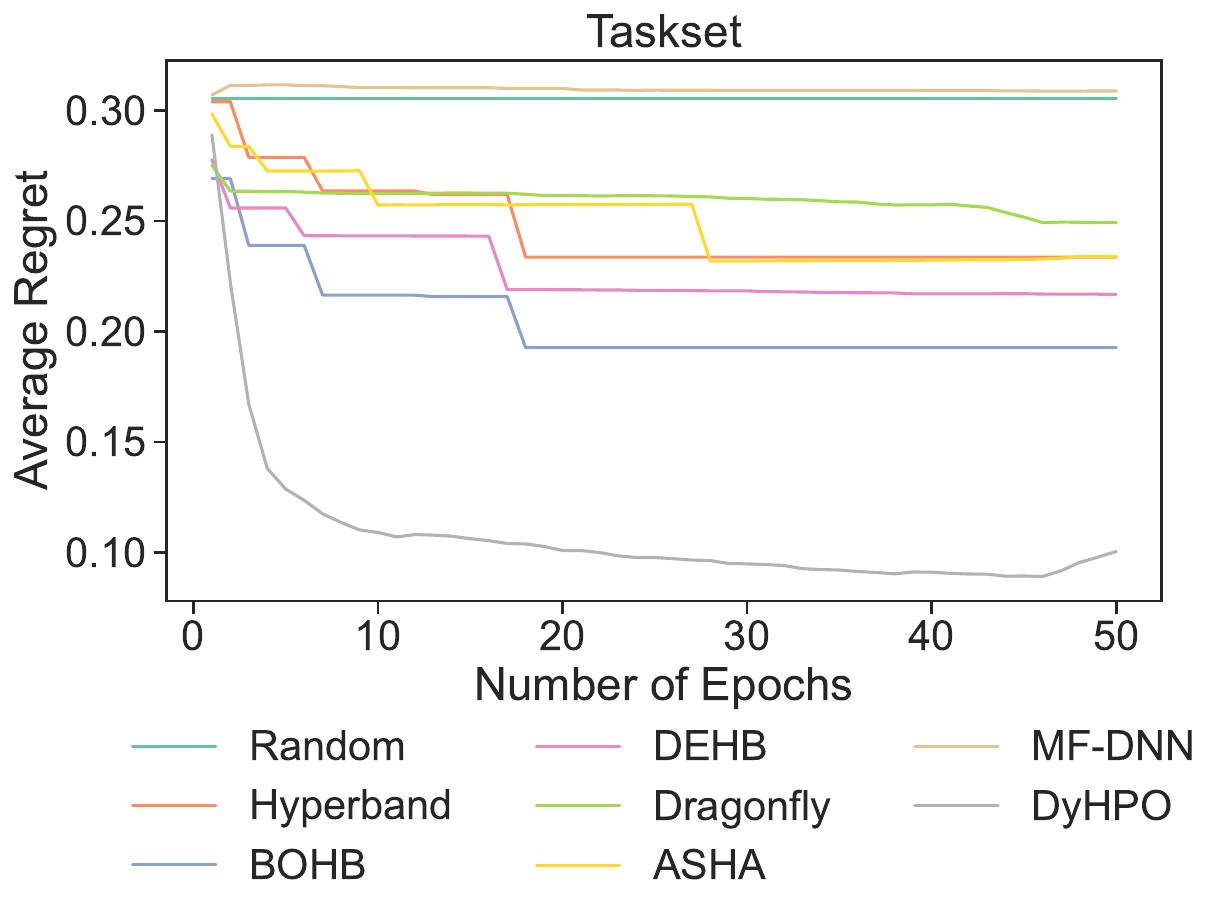}
    \includegraphics[width=0.32\textwidth]{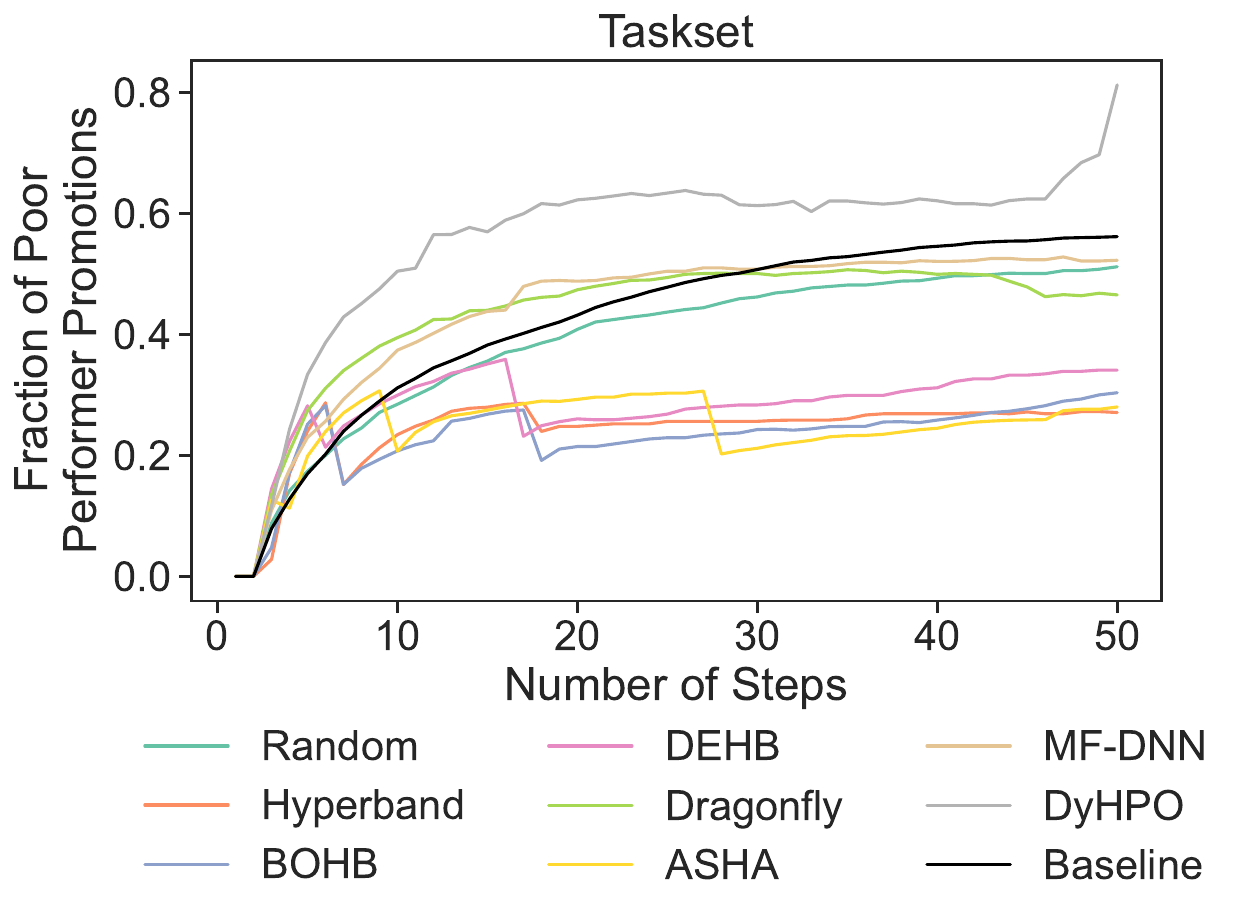}
  \caption{The efficiency of \algabbr{} as the optimization progresses. \textbf{Left:} The fraction of top-performing candidates from all candidates that were selected to be trained. \textbf{Middle:} The average regret for the configurations that were selected to be trained at a given budget. \textbf{Right:} The percentage of configurations that belong to the top 1/3 configurations at a given budget and that were in the top bottom 2/3 of the configurations at a previous budget. All of the results are from the Taskset benchmark.}
  \label{fig:efficiency_ts}
\end{figure*}

In Figure~\ref{fig:efficiency_ts} we provide the extended results of \textbf{Experiment 3} for TaskSet. We show the precision, average regret, and promotion percentage for poor-performing configurations for \algabbr{} and the other competitor methods.

\begin{figure*}[t]
  \centering
    \includegraphics[width=0.32\textwidth]{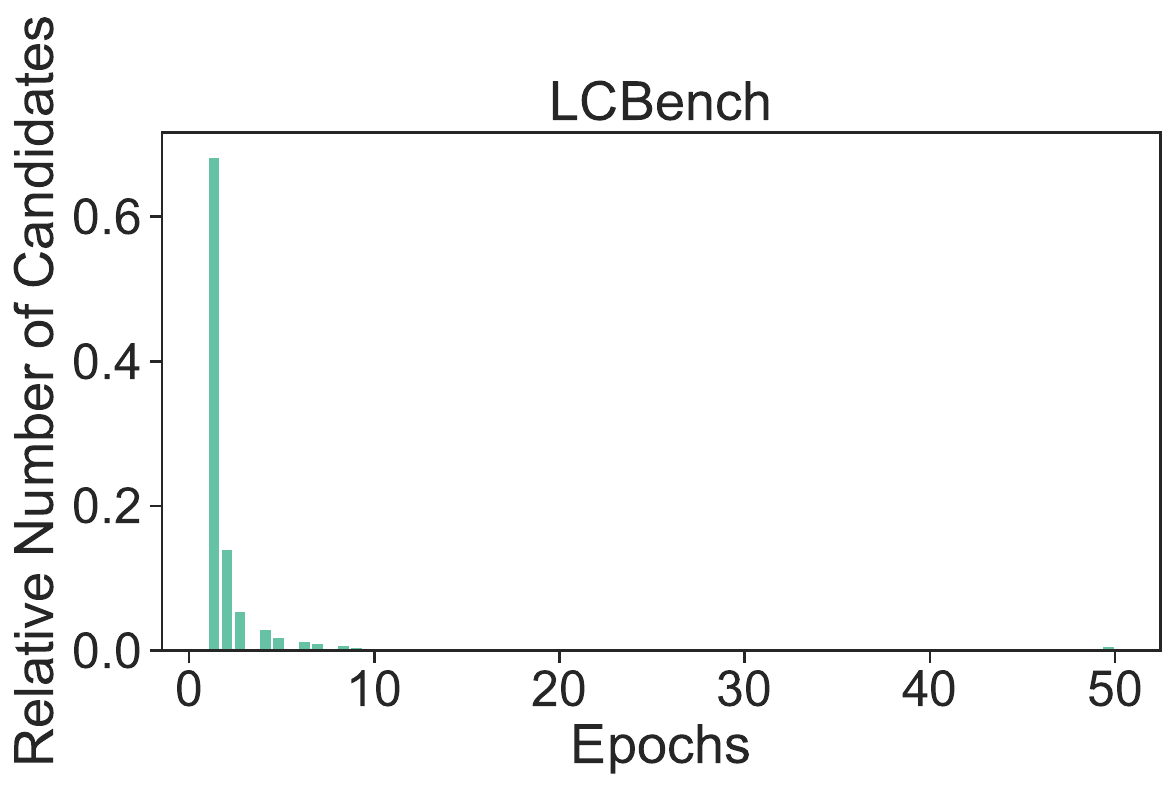}
    \includegraphics[width=0.32\textwidth]{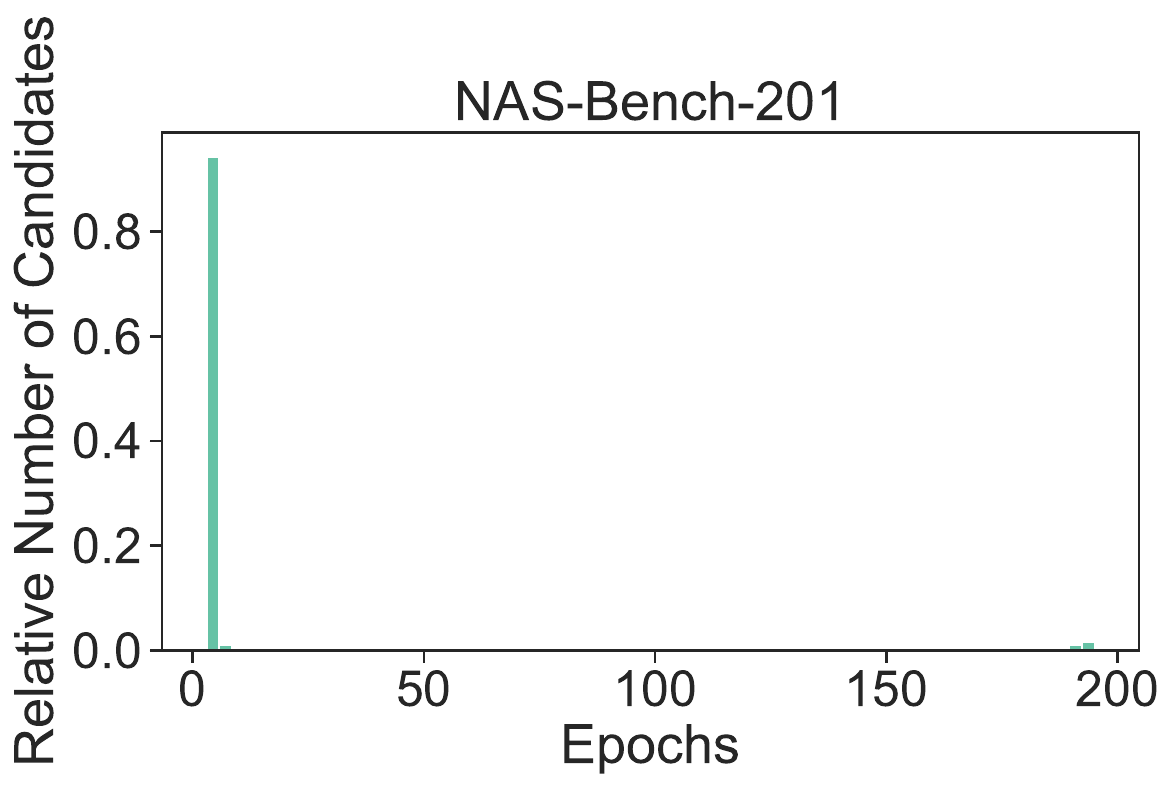}
    \includegraphics[width=0.32\textwidth]{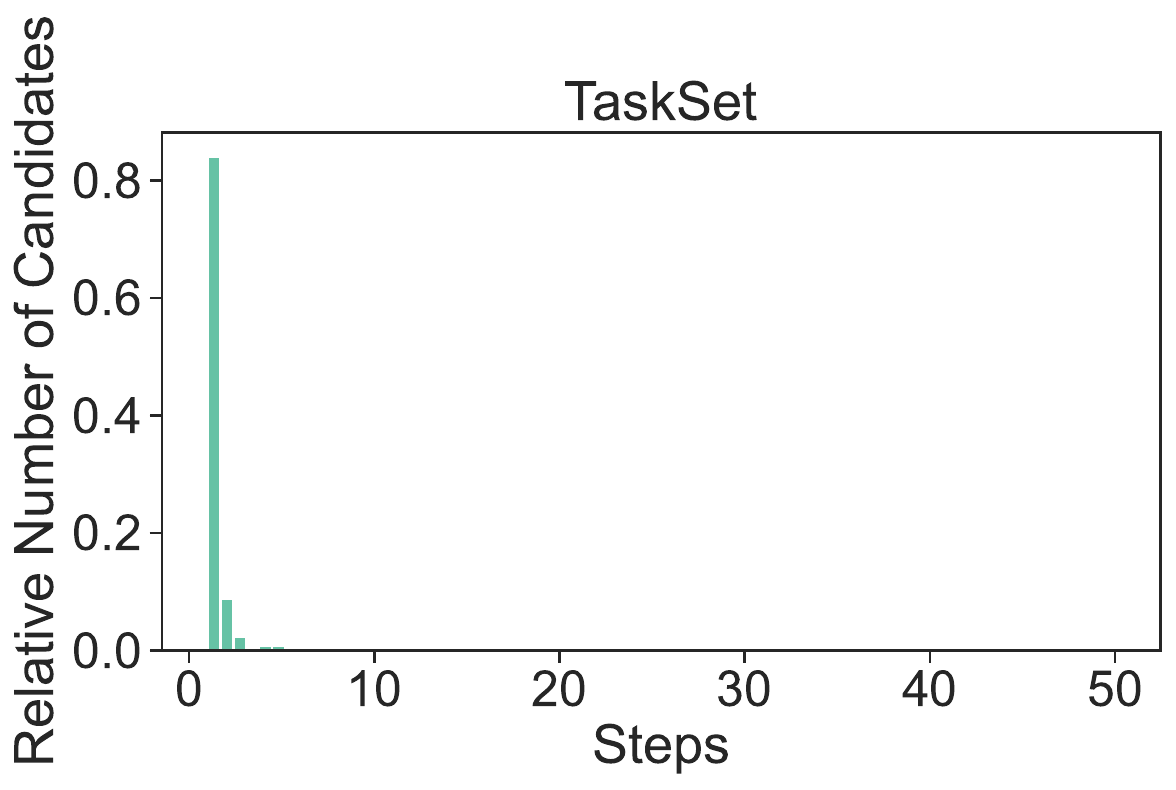}
  \caption{These plots shed light on how \algabbr{} behaves \emph{after} the configuration it finally returns as the best. The plots show how many epochs are spent per candidate. As we can see, for most candidates only a small budget was considered, indicating that \algabbr{} is mostly exploring at this point.}
  \label{fig:budget_invested}
\end{figure*}

Lastly, we explore the behavior of \algabbr{} after finding the configuration which is returned at the end of the optimization as the best configuration.
In Figure~\ref{fig:budget_invested}, we show how the budget is distributed on the configurations considered during that part of the optimization.
Clearly, \algabbr{} is spending very little budget on most configurations.
Furthermore, we investigated how many new configurations are considered during this phase.
For LCBench, 76.98\% of considered configurations are new demonstrating that \algabbr{} is investigating most of the budget into exploration.
These are even more extreme for TaskSet (93.16\% and NAS-Bench-201 (97.51\%).

\end{document}